\documentclass[11pt,a4paper,twoside,openright]{book}   	

\usepackage{geometry}                	
\let\origdoublepage\cleardoublepage
\newcommand{\clearemptydoublepage}{%
  \clearpage
  {\pagestyle{empty}\origdoublepage}%
}
\let\cleardoublepage\clearemptydoublepage

\newif\ifSOL
\SOLtrue      

\usepackage{chngcntr}  

\ifSOL
  \usepackage[sol, nogeom]{sty/ethuebungchaptersection}  
\else
  \usepackage[nogeom]{sty/ethuebungchaptersection}       
\fi
\UebungStyle{LargeSolutions}

\newcounter{mycounter}[chapter]
\counterwithin{mycounter}{chapter}
\counterwithin{equation}{chapter}
\counterwithin{uebeqloesung}{chapter}

\newcommand{\ex}[1]{\refstepcounter{mycounter}\exercise{#1}}
\newcommand{\exref}[1]{Exercise \ref{#1}}

\renewcommand{\theuebcounter}{\thechapter.\arabic{mycounter}}%

\renewcommand{\uebLoesungEqLabel}{S.\thechapter.\arabic{uebeqloesung}}

\UebungExTitleFont{\rmfamily }

\usepackage{nicefrac}
\usepackage{graphicx}
\usepackage{float}
\usepackage{amsmath,amssymb,amstext,amsthm,bbm,bm,mathtools}
\usepackage[utf8]{inputenc}
\usepackage{subfig}
\usepackage{booktabs}
\usepackage{xcolor}
\usepackage{url}
\usepackage{ccicons}

\usepackage[round, authoryear]{natbib}
\usepackage{minitoc}
\usepackage{cprotect}               
\usepackage{listings}

\usepackage{fancyhdr}
\usepackage{calc}
\newlength{\pageoffset}
\renewcommand{\sectionmark}[1]{}    
\fancypagestyle{plain}{%
    \fancyhead{}                       
    \fancyfoot{}                       

}

\usepackage{calligra}
\usepackage[T1]{fontenc}

\hypersetup{
    plainpages=false,
    colorlinks,
    linkcolor={red!50!black},
    citecolor={blue!50!black},
    urlcolor={blue!80!black}
}

\usepackage{scalerel}[2016/12/29] 

\usepackage{tikz}
\usetikzlibrary{arrows,shapes,backgrounds,through,shadows,positioning}
\usetikzlibrary{decorations.pathmorphing}
\usetikzlibrary{calc}
\usetikzlibrary{fit} %
\usetikzlibrary{graphs,graphs.standard}
\usepackage{pgfplots}
\usetikzlibrary{tikzmark,positioning}

\tikzstyle{neur}=[rectangle,draw=green!50,fill=green!50,minimum size=6mm,line width=2pt,>=stealth]  

\tikzstyle{fact}=[fill,minimum size=1.5mm,line width=2pt,>=stealth]

\tikzstyle{cont2}=[circle,draw=black!50,top color=white, 
bottom color=lilla!50!black!20,thick,minimum size=6mm,line width=1pt,>=stealth]  

\tikzstyle{contredb}=[circle,draw=red,top color=red, 
bottom color=red!50!black!20,thick,minimum size=8.1mm,line width=1pt,>=stealth]  
\tikzstyle{contyellowb}=[circle,draw=yellow,top color=yellow, 
bottom color=yellow!50!black!20,thick,minimum size=8.1mm,line width=1pt,>=stealth]  
\tikzstyle{contblueb}=[circle,draw=blue,top color=blue, 
bottom color=blue!50!black!20, thick,minimum size=8.1mm,line width=1pt,>=stealth]  

\tikzstyle{contred}=[circle,draw=red,top color=red, 
bottom color=white,thick,minimum size=6mm,line width=1pt,>=stealth]  
\tikzstyle{contyellow}=[circle,draw=yellow,top color=yellow, 
bottom color=yellow!50!black!20,thick,minimum size=6mm,line width=1pt,>=stealth]  
\tikzstyle{contblue}=[circle,draw=blue,top color=blue!80!black!50, 
bottom color=white, thick,minimum size=6mm,line width=1pt,>=stealth]  
\tikzstyle{contgreen}=[circle,draw=green,top color=green!80!black!50, 
bottom color=white, thick,minimum size=6mm,line width=1pt,>=stealth]  
\tikzstyle{contwhite}=[circle,draw=white,color=white, thick,minimum size=6mm,line width=1pt,>=stealth]  
\tikzstyle{contwhiteb}=[circle,draw=white,color=white, thick,minimum size=7.5mm,line width=1pt,>=stealth]  

\tikzstyle{cont}=[circle, draw,
thick,minimum size=6mm,line width=1pt,>=stealth]  

\tikzstyle{contb}=[circle,draw=black!50,top color=white, 
bottom color=darkgreen!50!black!20,thick,inner sep=0pt,minimum size=8.1mm,line width=1pt,>=stealth]  

\tikzstyle{ocont}=[ellipse,draw=blue!50,thick,minimum size=6mm,>=stealth]  
\tikzstyle{blackcont}=[circle,draw=black!50,thick,minimum size=6mm,line width=2pt,>=stealth]  
\tikzstyle{oval}=[ellipse,draw=blue!50,thick,minimum size=6mm,line width=1pt,>=stealth]  
\tikzstyle{ovalb}=[ellipse,draw=blue!50,thick,minimum size=7.5mm,line width=1pt,>=stealth]  
\tikzstyle{disc}=[rectangle,draw=blue!50,thick,line width=1pt,minimum size=6mm]  
\tikzstyle{obs}=[fill=blue!20,thick]  
\tikzstyle{opt}=[star,draw=red!50,thick,minimum size=6mm]  

\tikzstyle{fillred}=[fill=red!20,thick]  
\tikzstyle{fillgreen}=[fill=green!20,thick]  
\tikzstyle{purered}=[fill=red]  
\tikzstyle{state}=[rectangle,fill=red!20]  
\tikzstyle{sobs}=[fill=green!15,thick]  
\tikzstyle{fact}=[fill,minimum size=1.5mm,line width=2pt,>=stealth]
\tikzstyle{varfact}=[draw,minimum size=1.5mm,line width=2pt,>=stealth]
\tikzstyle{sep}=[rectangle,draw=magenta!50,thick,minimum size=6mm]  
\tikzstyle{det}=[fill=red!15,rectangle,draw=red!50,thick,minimum size=6mm]  

\tikzstyle{dethid}=[diamond,draw=red!50,thick,minimum size=6mm]  

\tikzstyle{lineball}=[fill,-*,draw=red!50,line width=1.5pt]
\tikzstyle{redball}=[mark=*,mark options={fill=red!50,draw=red},mark size=0.5pt]
\tikzstyle{greenball}=[mark=*,mark options={fill=green!50,draw=green},mark size=0.5pt]
\tikzstyle{hid}=[circle,draw,thick]  

\tikzstyle{dec}=[rectangle,draw=red!50,thick,minimum size=6mm]  
\tikzstyle{utility}=[diamond,draw=red!50,thick,minimum size=6mm]  
\tikzstyle{contdec}=[circle,draw=blue!50,thick,fill=blue!10,line width=2pt]  
\tikzstyle{decutility}=[diamond,draw=red!50,thick,minimum size=6mm]  

\tikzstyle{contobs}+=[cont]
\tikzstyle{contobs}+=[obs]
\tikzstyle{discobs}+=[disc]
\tikzstyle{discobs}+=[obs]

\tikzstyle{obsred}+=[obs]
\tikzstyle{obsred}+=[red]

\tikzstyle{background grid}=[draw, black!50,step=.1cm]
\tikzstyle{dgraph}=[->, line width=1.5pt]
\tikzstyle{ugraph}=[line width=1.5pt]

\newcommand{\fxmess}[3]{\scaleobj{#3}{\mu_{{#1\rightarrow #2}}}}
\newcommand{\xfmess}[3]{\scaleobj{#3}{\mu_{{#1\rightarrow #2}}}}

\newcommand{\fxmessb}[2]{\pmb{\mu_{{#1\rightarrow #2}}}}
\newcommand{\xfmessb}[2]{\pmb{\mu_{{#1\rightarrow #2}}}}

\newcommand{\mfxmess}[3]{\scaleobj{#3}{\gamma_{{#1\rightarrow #2}}}}
\newcommand{\mxfmess}[3]{\scaleobj{#3}{\gamma_{{#1\rightarrow #2}}}}

\definecolor{magenta}{cmyk}{0.1,1,1,0.5}
\definecolor{darkgreen}{cmyk}{0.6,0.1,0.6,0.6}
\definecolor{pink}{cmyk}{0.1,1,1,0.1}
\definecolor{azzurro}{cmyk}{0.9333, 0.2471, 0.5569, 0.102}
\definecolor{lilla}{rgb}{0.5, 0, 0.5}
\definecolor{mygreen}{rgb}{0, 0.5, 0}
\definecolor{darkorange}{rgb}{1, 0.4, 0} 
\definecolor{darkred}{rgb}{0.8, 0, 0}

\newcommand{\MB}{\textrm{MB}}
\newcommand{\nondesc}{\textrm{nondesc}}
\newcommand{\desc}{\textrm{desc}}

\newcommand{\data}{\mathcal{D}}
\newcommand{\normal}{\mathcal{N}}

\newcommand{\var}{\mathbb{V}}
\newcommand{\Var}{\var}
\newcommand{\E}{\mathbb{E}}
\renewcommand{\Pr}{\mathbb{P}}

\usepackage{centernot}
\newcommand{\independent}{\mathrel{\perp\mspace{-10mu}\perp}}
\newcommand{\notind}{\centernot{\independent}}

\newcommand{\pre}{\mathrm{pre}}
\newcommand{\pa}{\mathrm{pa}}
\newcommand{\pap}{\pi}
\renewcommand{\ne}{\mathrm{ne}}

\newcommand{\Gauss}{\mathcal{N}}
\newcommand{\BetaDist}{\mathcal{B}}

\DeclareMathOperator*{\argmax}{\mathrm{argmax}}
\DeclareMathOperator*{\argmin}{\mathrm{argmin}}

\DeclareMathOperator{\diag}{diag}
\newcommand{\zerob}{\boldsymbol{0}}
\DeclareMathOperator{\tr}{tr}

\newcommand{\ud}{\mathrm{d}}

\renewcommand{\a}{\mathbf{a}}
\renewcommand{\b}{\mathbf{b}}
\renewcommand{\c}{\mathbf{c}}
\newcommand{\e}{\mathbf{e}}

\newcommand{\g}{\mathbf{g}}
\newcommand{\h}{\mathbf{h}}

\newcommand{\m}{\mathbf{m}}

\newcommand{\p}{\mathbf{p}}

\makeatletter
\@ifundefined{r}{%
\newcommand{\r}{\mathbf{r}}}{}
\makeatother
\renewcommand{\r}{\mathbf{r}}

\renewcommand{\u}{\mathbf{u}}

\renewcommand{\v}{\mathbf{v}}

\newcommand{\w}{\mathbf{w}}
\newcommand{\x}{\mathbf{x}}
\newcommand{\y}{\mathbf{y}}
\newcommand{\z}{\mathbf{z}}

\newcommand{\mytheta}{\boldsymbol{\theta}}

\newcommand{\thetab}{\mytheta}

\newcommand{\epsilonb}{\boldsymbol{\epsilon}}
\newcommand{\myalpha}{\boldsymbol{\alpha}}
\newcommand{\alphab}{\myalpha}
\newcommand{\mybeta}{\boldsymbol{\beta}}
\newcommand{\betab}{\mybeta}

\newcommand{\mub}{\pmb{\mu}}

\newcommand{\lambdab}{\boldsymbol{\lambda}}

\newcommand{\Ind}{\mathcal{I}}

\newcommand{\A}{\mathbf{A}}
\newcommand{\B}{\mathbf{B}}
\makeatletter
\@ifundefined{C}{%
\newcommand{\C}{\mathbf{C}}
}{}
\makeatother
\renewcommand{\C}{\mathbf{C}}

\newcommand{\F}{\mathbf{F}}
\renewcommand{\H}{\mathbf{H}}
\newcommand{\I}{\mathbf{I}}
\newcommand{\K}{\mathbf{K}}
\newcommand{\M}{\mathbf{M}}

\newcommand{\PP}{\mathbf{P}}
\newcommand{\R}{\mathbf{R}}

\makeatletter
\@ifundefined{U}{%
\newcommand{\U}{\mathbf{U}}
}{}
\makeatother
\renewcommand{\U}{\mathbf{U}}
\newcommand{\V}{\mathbf{V}}
\newcommand{\W}{\mathbf{W}}

\newcommand{\EE}{\mathbf{E}}
\newcommand{\Lambdab}{\mathbf{\Lambda}}
\newcommand{\Sigmab}{\pmb{\Sigma}}
\newcommand{\Psib}{\pmb{\Psi}}

\newcommand{\red}[1]{\textcolor{red}{#1}}
\newcommand{\blue}[1]{\textcolor{blue}{#1}}

\newcommand{\approxpropto}{\mathrel{\vcenter{
  \offinterlineskip\halign{\hfil$##$\cr
    \propto\cr\noalign{\kern2pt}\sim\cr\noalign{\kern-2pt}}}}}

\newcommand{\ind}{\mathbbm{1}}

\newcommand{\xBar}{\bar{x}}

\newcommand{\KL}{\text{KL}}

\newcommand{\ELBO}{\mathcal{L}}

\newcommand{\ELBOx}{\ELBO_{\x}}

\definecolor{dkblue}{rgb}{0,0.1,0.5}
\definecolor{dkgreen}{rgb}{0,0.4,0}
\definecolor{dkred}{rgb}{0.6,0,0}
\definecolor{midred}{RGB}{224, 100, 89}
\definecolor{verylightgrey}{RGB}{245, 245, 245}

\usepackage{tikz}
\usepackage[%
framemethod=tikz,
skipbelow=\topskip,
skipabove=\topskip
]{mdframed}
\mdfsetup{%
        leftmargin=6pt,
        rightmargin=6pt,
        innertopmargin=6pt,
        innerbottommargin=4pt,
        backgroundcolor=verylightgrey,
        middlelinecolor=black,
        roundcorner=4
}

\BeforeBeginEnvironment{lstlisting}{\begin{mdframed}\vspace{-0.7em}}
        \AfterEndEnvironment{lstlisting}{\vspace{-0.5em}\end{mdframed}}  

\graphicspath{
  {./figs/}
  {./figs/linear-algebra/}
  {./figs/directed-graphical-models/}
  {./figs/inference-for-hidden-markov-models/}
  {./figs/model-based-learning/}
  {./figs/sampling-and-monte-carlo-integration/}
  {./figs/variational-inference/}
}

\begin{document}

\frontmatter

%
%
%
%

\begin{titlepage} 
	
	\raggedleft 
	
	\rule{1pt}{\textheight} 
	\hspace{0.05\textwidth} 
	\parbox[b]{0.75\textwidth}{ 

    \resizebox{8cm}{!}{\Huge \hspace{-1.5ex} \bfseries \calligra Pen \& Paper }\\[\baselineskip]
    {\huge\bfseries Exercises in Machine Learning}\\[4\baselineskip] 
		{\Large \bfseries Michael U. Gutmann}\\[\baselineskip]
		
		\vspace{0.5\textheight} 

    {\noindent University of Edinburgh}\\[\baselineskip]
	}

\end{titlepage}


\newpage
\thispagestyle{empty}

\rule{0pt}{0.95\textheight}
\noindent This work is licensed under the
\href{https://creativecommons.org/licenses/by/4.0/}{Creative Commons Attribution
4.0 International License} \ccby. To view a copy of this license, visit
\url{http://creativecommons.org/licenses/by/4.0/}.

\addtocontents{toc}{\protect{\pdfbookmark[0]{\contentsname}{toc}}}
\dominitoc
\setcounter{tocdepth}{1}    
\setcounter{secnumdepth}{1} 
\tableofcontents
\cleardoublepage

\chapter{Preface}
\label{ch:preface}
We may have all heard the saying ``use it or lose it''. We experience it when we
feel rusty in a foreign language or sports that we have not practised in a
while. Practice is important to maintain skills but it is also key when learning
new ones. This is a reason why many textbooks and courses feature exercises.
However, the solutions to the exercises feel often overly brief, or are
sometimes not available at all. Rather than an opportunity to practice the new
skills, the exercises then become a source of frustration and are ignored.

\hspace{2ex} This book contains a collection of exercises with \emph{detailed}
solutions. The level of detail is, hopefully, sufficient for the reader to
follow the solutions and understand the techniques used. The exercises, however,
are not a replacement of a textbook or course on machine learning. I assume that
the reader has already seen the relevant theory and concepts and would now like
to deepen their understanding through solving exercises.

\hspace{2ex} While coding and computer simulations are extremely important in
machine learning, the exercises in the book can (mostly) be solved with pen and
paper. The focus on pen-and-paper exercises reduced length and simplified the
presentation. Moreover, it allows the reader to strengthen their mathematical
skills. However, the exercises are ideally paired with computer exercises to
further deepen the understanding.

\hspace{2ex} The exercises collected here are mostly a union of exercises that I
developed for the courses ``Unsupervised Machine Learning'' at the University of
Helsinki and ``Probabilistic Modelling and Reasoning'' at the University of
Edinburgh. The exercises do not comprehensively cover all of machine learning
but focus strongly on unsupervised methods, inference and learning.

\hspace{2ex} I am grateful to my students for providing feedback and asking
questions. Both helped to improve the quality of the exercises and solutions. I
am further grateful to both universities for providing the research and teaching
environment.

\hspace{2ex} My hope is that the collection of exercises will grow with time. I
intend to add new exercises in the future and welcome contributions from the
community. Latex source code is available at
\url{https://github.com/michaelgutmann/ml-pen-and-paper-exercises}. Please use
GitHub's issues to report mistakes or typos, and please get in touch if you
would like to make larger contributions.

\begin{flushright}
  Michael Gutmann\\
  Edinburgh, June 2022\\
  \url{https://michaelgutmann.github.io}
\end{flushright}

\mainmatter

\chapter{Linear Algebra}
\minitoc

\ex{Gram--Schmidt orthogonalisation}
\label{ex:gram-schmidt} 

\begin{exenumerate}
\item Given two vectors $\a_1$ and $\a_2$ in $\mathbb{R}^n$, show that 
  \begin{align}
    \u_1 &= \a_1 \\
    \u_2 &= \a_2 - \frac{\u_1^\top \a_2}{\u_1^\top \u_1}\u_1
  \end{align}
  are orthogonal to each other.
  \begin{solution}
    Two vectors $\u_1$ and $\u_2$ of $\mathbb{R}^n$ are orthogonal
    if their inner product equals zero. Computing the inner product $\u_1^\top \u_2$ gives
    \begin{align}
      \u_1^{\top}\u_2 &= \u_1^\top(\a_2 - \frac{\u_1^\top\a_2}{\u_1^\top\u_1}\u_1) \\
                      &= \u_1^\top\a_2 - \frac{\u_1^\top\a_2}{\u_1^\top\u_1}\u_1^\top\u_1 \\
                      &= \u_1^\top\a_2 - \u_1^\top\a_2 \\
                      &= 0.
    \end{align}
    Hence the vectors $\u_1$ and $\u_2$ are orthogonal.
    
    If $\a_2$ is a multiple of $\a_1$, the orthogonalisation procedure produces
    a zero vector for $\u_2$. To see this, let $\a_2 = \alpha \a_1$ for some
    real number $\alpha$. We then obtain
    \begin{align} \u_2 &= \a_2 - \frac{\u_1^\top \a_2}{\u_1^\top \u_1}\u_1\\
                       &= \alpha \u_1 - \frac{\alpha \u_1^\top \u_1}{\u_1^\top \u_1}\u_1\\
                       &= \alpha \u_1 - \alpha \u_1\\ &=\mathbf{0}.
    \end{align}
  \end{solution}
  
\item Show that any linear combination of (linearly independent) $\a_1$ and $\a_2$ can be written in terms of $\u_1$ and $\u_2$. 
  \begin{solution}
    Let $\v$ be a linear combination of $\a_1$ and $\a_2$, i.e. $\v = \alpha{\a_1} + \beta{\a_2}$ for some real numbers $\alpha$ and $\beta$.
    Expressing $\u_1$ and $\u_2$ in term of $\a_1$ and $\a_2$, we can write $\v$ as
    \begin{align}
      \v  &= \alpha \a_1 + \beta\a_2 \\
          &= \alpha\u_1 + \beta(\u_2 + \frac{\u_1^\top\a_2}{\u_1^\top\u_1}\u_1) \\
          &= \alpha\u_1 + \beta\u_2 + \beta \frac{\u_1^\top\a_2}{\u_1^\top\u_1}\u_1 \\
          &= (\alpha + \beta\frac{\u_1^\top\a_2}{\u_1^\top\u_1})\u_1 + \beta\u_2,
    \end{align}
    Since $\alpha + \beta((\u_1^\top\a_2)/(\u_1^\top\u_1))$ and $\beta$ are real
    numbers, we can write $\v$ as a linear combination of $\u_1$ and
    $\u_2$. Overall, this means that any vector in the span of $\{\a_1, \a_2\}$
    can be expressed in the orthogonal basis $\{\u_1, \u_2\}$.
    
  \end{solution}

\item Show by induction that for any $k \le n$ linearly independent vectors
  $\mathbf{a}_1,\ldots,\mathbf{a}_k$, the vectors $\mathbf{u}_i$, $i=1, \ldots
  k$, are orthogonal, where
  \begin{align}
    \u_i &= \a_i - \sum_{j=1}^{i-1} \frac{\u_j^\top \a_i}{\u_j^\top\u_j} \u_j. \label{eq:Gram-Schmidt-basis-def}
  \end{align}
  The calculation of the vectors $\u_i$ is called Gram–Schmidt orthogonalisation.
  \begin{solution}
   We have shown above that the claim holds for two vectors. This is the base
   case for the proof by induction. Assume now that the claim holds for $k$
   vectors. The induction step in the proof by induction then consists of
   showing that the claim also holds for $k+1$ vectors.
   
   Assume that $\u_1, \u_2, \ldots, \u_k$ are orthogonal vectors. The linear independence assumption ensures that none of the
   $\u_i$ is a zero vector. We then have for $\u_{k+1}$
    \begin{align}
      \u_{k+1} &= \a_{k+1} - \frac{\u_1^\top\a_{k+1}}{\u_1^\top\u_1}\u_1 - \frac{\u_2^\top\a_{k+1}}{\u_2^\top\u_2}\u_2
                 - \ldots - \frac{\u_k^\top\a_{k+1}}{\u_k^\top\u_k}\u_k,
                 \label{eq:u_k+1}
    \end{align}
    and for all $i = 1, 2, \ldots, k$
    \begin{align}
      \u_i^\top\u_{k+1} &= \u_i^\top\a_{k+1} - \frac{\u_1^\top\a_{k+1}}{\u_1^\top\u_1}\u_i^\top\u_1 - 
                          \ldots - \frac{\u_k^\top\a_{k+1}}{\u_k^\top\u_k}\u_i^\top\u_k.
    \end{align}
    By assumption $\u_i^\top\u_j = 0$ if $i \neq j$, so that
    \begin{align}
      \u_i^\top\u_{k+1} &= \u_i^\top\a_{k+1} - 0 - \ldots - \frac{\u_i^\top\a_{k+1}}{\u_i^\top\u_i}\u_i^\top\u_i - 0 \ldots - 0 \\
                        &= \u_i^\top\a_{k+1} - \u_i^\top\a_{k+1} \\
                        &= 0,
    \end{align}
    which means that $\u_{k+1}$ is orthogonal to $\u_1, \ldots, \u_k$.
    
  \end{solution}

\item Show by induction that any linear combination of (linear independent)
  $\a_1, \a_2, \ldots, \a_k$ can be written in terms of $\u_1, \u_2, \ldots,
  \u_k$.
  
  \begin{solution} The base case of two vectors was proved above. Using
    induction, we assume that the claim holds for $k$ vectors and we will prove that
    it then also holds for $k + 1$ vectors: Let $\v$ be a linear combination of
    $\a_1, \a_2, \ldots,
    \a_{k+1}$, i.e. $\v = \alpha_1\a_1 + \alpha_2\a_2 + \ldots + \alpha_k\a_k +
    \alpha_{k+1}\a_{k+1}$ for some real numbers $\alpha_1, \alpha_2, \ldots,
    \alpha_{k+1}$. Using the induction assumption, $\v$ can be written as
    \begin{equation}
      \v = \beta_1\u_1 + \beta_2\u_2 + \ldots + \beta_k\u_k + \alpha_{k+1}\a_{k+1},
    \end{equation}
    for some real numbers $\beta_1, \beta_2, \ldots, \beta_k$ Furthermore, using equation \eqref{eq:u_k+1}, $\v$ can be written as
    \begin{align}
      \v &= \beta_1\u_1  + \ldots + \beta_k\u_k + \alpha_{k+1}\u_{k+1} + \alpha_{k+1}\frac{\u_1^\top\a_{k+1}}{\u_1^\top\u_1}\u_1 \\ 
         &\phantom{=}+ \ldots + \alpha_{k+1}\frac{\u_k^\top\a_{k+1}}{\u_k^\top\u_k}\u_k.
    \end{align}
    With $\gamma_i = \beta_i + \alpha_{k+1}(\u_i^\top\a_{k+1})/(\u_i^\top\u_i)$, $\v$ can thus be written as
    \begin{align}
      \v &= \gamma_1\u_1 + \gamma_2\u_2 + \ldots + \gamma_k\u_k + \alpha_{k+1}\u_{k+1},
    \end{align}
    which completes the proof. Overall, this means that the $\u_1, \u_2, \ldots,
    \u_k$ form an orthogonal basis for $\text{span}(\a_1, \ldots, \a_k)$, i.e.\
    the set of all vectors that can be obtained by linearly combining the
    $\a_i$.

  \end{solution}
  
\item Consider the case where $\a_1, \a_2, \ldots, \a_k$ are linearly
  independent and $\a_{k+1}$ is a linear combination of $\a_1, \a_2, \ldots,
  \a_k$. Show that $\u_{k+1}$, computed according to
  \eqref{eq:Gram-Schmidt-basis-def}, is zero.

  \begin{solution}
    Starting with \eqref{eq:Gram-Schmidt-basis-def}, we have
    \begin{equation}
      \u_{k+1} = \a_{k+1} - \sum_{j=1}^{k} \frac{\u_j^\top \a_{k+1}}{\u_j^\top\u_j} \u_j.
    \end{equation}
    By assumption, $\a_{k+1}$ is a linear combination of $\a_1, \a_2, \ldots,
    \a_k$. By the previous question, it can thus also be written as a linear
    combination of the $\u_1, \ldots, \u_k$. This means that there are some
    $\beta_i$ so that
    \begin{equation}
      \a_{k+1} = \sum_{i=1}^k \beta_i \u_i
    \end{equation}
    holds. Inserting this expansion into the equation above gives
    \begin{align}
      \u_{k+1}  &=  \sum_{i=1}^k \beta_i \u_i - \sum_{j=1}^k \sum_{i=1}^k \beta_i \frac{\u_j^\top\u_i}{\u_j^\top\u_j} \u_j\\
                &= \sum_{i=1}^k \beta_i \u_i - \sum_{i=1}^k \beta_i  \frac{\u_i^\top\u_i}{\u_i^\top\u_i} \u_i
    \end{align}
    because $\u_j^\top \u_i =0$ if $i \neq j$. We thus obtain the desired result:
    \begin{align}
      \u_{k+1} & =  \sum_{i=1}^k \beta_i \u_i - \sum_{i=1}^k \beta_i \u_i\\
               &= 0
    \end{align}
    This property of the Gram-Schmidt process in
    \eqref{eq:Gram-Schmidt-basis-def} can be used to check whether a list of
    vectors $\a_1, \a_2, \ldots, \a_d$ is linearly independent or not. If, for
    example, $\u_{k+1}$ is zero, $\a_{k+1}$ is a linear combination of the $\a_1,
    \ldots, \a_k$. Moreover, the result can be used to extract a sublist of linearly
    independent vectors: We would remove $\a_{k+1}$ from the list and restart
    the procedure in \eqref{eq:Gram-Schmidt-basis-def} with $\a_{k+2}$ taking
    the place of $\a_{k+1}$. Continuing in this way constructs a list of
    linearly independent $\a_j$ and orthogonal $\u_j$, $j=1, \ldots, r$, where
    $r$ is the number of linearly independent vectors among the $\a_1, \a_2,
    \ldots, \a_d$.

  \end{solution}
\end{exenumerate}

\ex{Linear transforms}
\label{ex:linear-transforms}

\begin{exenumerate}
\item Assume two vectors $ \a_1$ and $\a_2$ are in $\mathbb{R}^2$. Together,
  they span a parallelogram. Use \exref{ex:gram-schmidt} to show that the
  squared area $S^2$ of the parallelogram is given by
  \begin{equation}
    S^2 = (\a_2^T\a_2)(\a_1^T\a_1)-(\a_2^T\a_1)^2
  \end{equation}
  
  \begin{solution}
    Let $\a_1$ and $\a_2$ be the vectors that span the parallelogram. From geometry we
    know that the area of parallelogram is base times height, which is
    equivalent to the length of the base vector times the length of the height vector.
    Denote this by $S^2 = ||\a_1||^2||\u_2||^2$, where is $\a_1$ is the base
    vector and $\u_2$ is the height vector which is orthogonal to the base
    vector. Using the Gram--Schmidt process for the vectors $\a_1$ and $\a_2$ in that
    order, we obtain the vector $\u_2$ as the second  output.
    \begin{figure}[h!]
      \centering
      \scalebox{0.75}{
        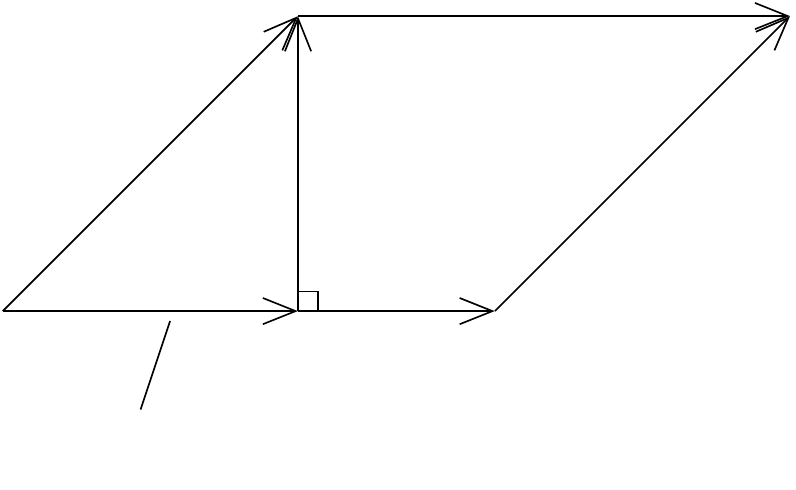 }
    \end{figure}
    
    Therefore $||\u_2||^2$ equals
    \begin{align}
      ||\u_2||^2 &= \u_2^\top\u_2 \\
                 &= \left(\a_2 - \frac{\a_1^\top\a_2}{\a_1^\top\a_1}\a_1\right)^\top\left(\a_2 - \frac{\a_1^\top\a_2}{\a_1^\top\a_1}\a_1\right) \\
                 &= \a_2^\top\a_2 - \frac{(\a_1^\top\a_2)^2}{\a_1^\top\a_1} - \frac{(\a_1^\top\a_2)^2}{\a_1^\top\a_1} + \left(\frac{\a_1^\top\a_2}{\a_1^\top\a_1}\right)^2\a_1^\top\a_1 \\
                 &= \a_2^\top\a_2 - \frac{(\a_1^\top\a_2)^2}{\a_1^\top\a_1}.
    \end{align}
    Thus, $S^2$ is:
    \begin{align}
      S^2 &= ||\a_1||^2||\u_2||^2 \\
          &= (\a_1^\top\a_1)(\u_2^\top\u_2) \\
          &= (\a_1^\top\a_1)\left(\a_2^\top\a_2 - \frac{(\a_1^\top\a_2)^2}{\a_1^\top\a_1}\right) \\
          &= (\a_2^\top\a_2)(\a_1^\top\a_1) - (\a_1^\top\a_2)^2.
    \end{align}

  \end{solution}
  
\item Form the matrix $\A=(\a_1 \; \a_2)$ where $\a_1$ and $\a_2$ are
  the first and second column vector, respectively. Show that
  \begin{equation}
    S^2 = (\det \A)^2.
  \end{equation}

  \begin{solution}
    We form the matrix $\A$,
    \begin{equation}
      \A = \begin{pmatrix} \a_1 & \a_2
           \end{pmatrix}  =
           \begin{pmatrix} a_{11} & a_{12} \\ a_{21} & a_{22}
           \end{pmatrix}.
    \end{equation}
    The determinant of $\A$ is $\det \A = a_{11}a_{22} - a_{12}a_{21}$. By multiplying out $(\a_2^\top\a_2)$, $(\a_1^\top\a_1)$ and 
    $(\a_1^\top\a_2)^2$, we get
    \begin{align}
      \a_2^\top\a_2 	&= a_{12}^2 + a_{22}^2 \\ 
      \a_1^\top\a_1 	&= a_{11}^2 + a_{21}^2 \\
      (\a_1^\top\a_2)^2 	&= (a_{11}a_{12} + a_{21}a_{22})^2 = a_{11}^2a_{12}^2 + a_{21}^2a_{22}^2 + 2a_{11}a_{12}a_{21}a_{22}.
    \end{align}
    Therefore the area equals
    \begin{align}
      S^2 &= (a_{12}^2 + a_{22}^2)(a_{11}^2 + a_{21}^2) - (\a_1^\top\a_2)^2 \\
          &= a_{12}^2a_{11}^2 + a_{12}^2a_{21}^2 + a_{22}^2a_{11}^2 + a_{22}^2a_{21}^2 \nonumber\\
          &\phantom{=}- (a_{12}^2a_{11}^2 + a_{21}^2a_{22}^2 + 2a_{11}a_{12}a_{21}a_{22}) \\
          &= a_{12}^2a_{21}^2 + a_{22}^2a_{11}^2 - 2a_{11}a_{12}a_{21}a_{22} \\
          &= (a_{11}a_{22} - a_{12}a_{21})^2,
    \end{align}
    which equals $(\det \A)^2$.
  \end{solution}
\item Consider the linear transform $\y=\A\x$ where $\A$
  is a $2 \times 2$ matrix. Denote the image of the rectangle $U_x=[x_1 \;
  x_1+\triangle_1]\times [x_2 \; x_2+\triangle_2]$ under the transform
  $\A$ by $U_y$. What is $U_y$? What is the area of $U_y$?
  
  \begin{solution} $U_y$ is parallelogram that is spanned by the column
    vectors $\a_1$ and $\a_2$ of $\A$, when $\A = (\a_1 \; \a_2)$.
    
    A rectangle with the same area as $U_x$ is spanned by vectors $(\Delta_1,
    0)$ and $(0, \Delta_2)$. Under the linear transform $\A$ these spanning
    vectors become $\Delta_1\a_1$ and $\Delta_2\a_2$. Therefore a parallelogram
    with the same area as $U_y$ is spanned by $\Delta_1\a_1$ and
    $\Delta_2\a_2$ as shown in the following figure.
    \begin{figure}[htp] \centering \scalebox{0.4}{
        \begin{picture}(0,0)%
          \includegraphics{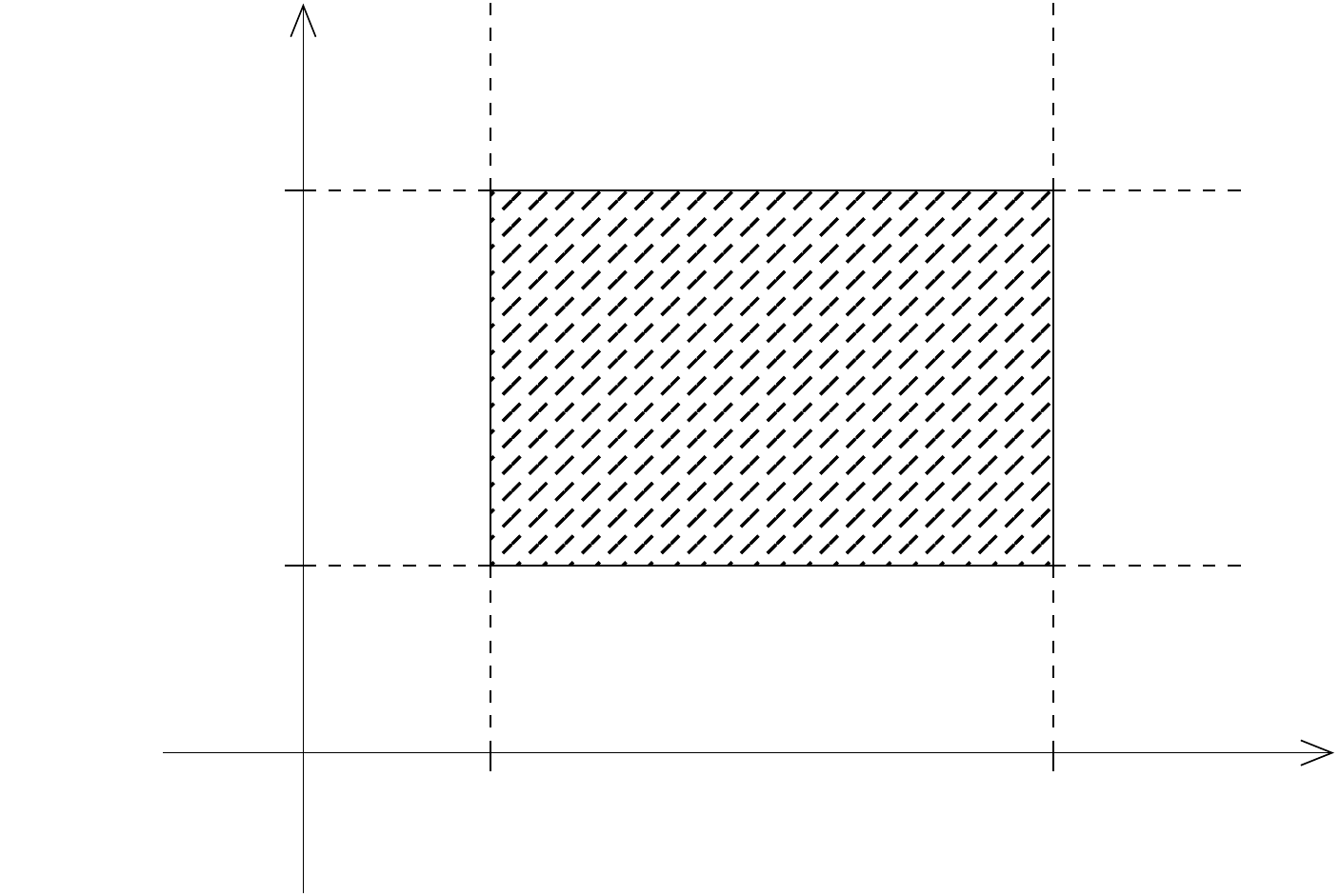}%
        \end{picture}%
        \setlength{\unitlength}{4144sp}%
        \begingroup\makeatletter\ifx\SetFigFont\undefined%
          \gdef\SetFigFont#1#2#3#4#5{%
            \reset@font\fontsize{#1}{#2pt}%
            \fontfamily{#3}\fontseries{#4}\fontshape{#5}%
            \selectfont}%
        \fi\endgroup%
        \begin{picture}(6417,4299)(-1454,152)
          \put(2026,3809){\makebox(0,0)[lb]{\smash{{\SetFigFont{25}{30.0}{\familydefault}{\mddefault}{\updefault}{\color[rgb]{0,0,0}$U_x$}%
                }}}}
          \put(-629,1649){\makebox(0,0)[lb]{\smash{{\SetFigFont{25}{30.0}{\familydefault}{\mddefault}{\updefault}{\color[rgb]{0,0,0}$x_2$}%
                }}}}
          \put(3016,434){\makebox(0,0)[lb]{\smash{{\SetFigFont{25}{30.0}{\familydefault}{\mddefault}{\updefault}{\color[rgb]{0,0,0}$x_1 + \Delta_1$}%
                }}}}
          \put(-1439,3449){\makebox(0,0)[lb]{\smash{{\SetFigFont{25}{30.0}{\familydefault}{\mddefault}{\updefault}{\color[rgb]{0,0,0}$x_2 + \Delta_2$}%
                }}}}
          \put(721,434){\makebox(0,0)[lb]{\smash{{\SetFigFont{25}{30.0}{\familydefault}{\mddefault}{\updefault}{\color[rgb]{0,0,0}$x_1$}%
                }}}}
        \end{picture}%
      }
      \scalebox{0.4}{ 
        \begin{picture}(0,0)%
          \includegraphics{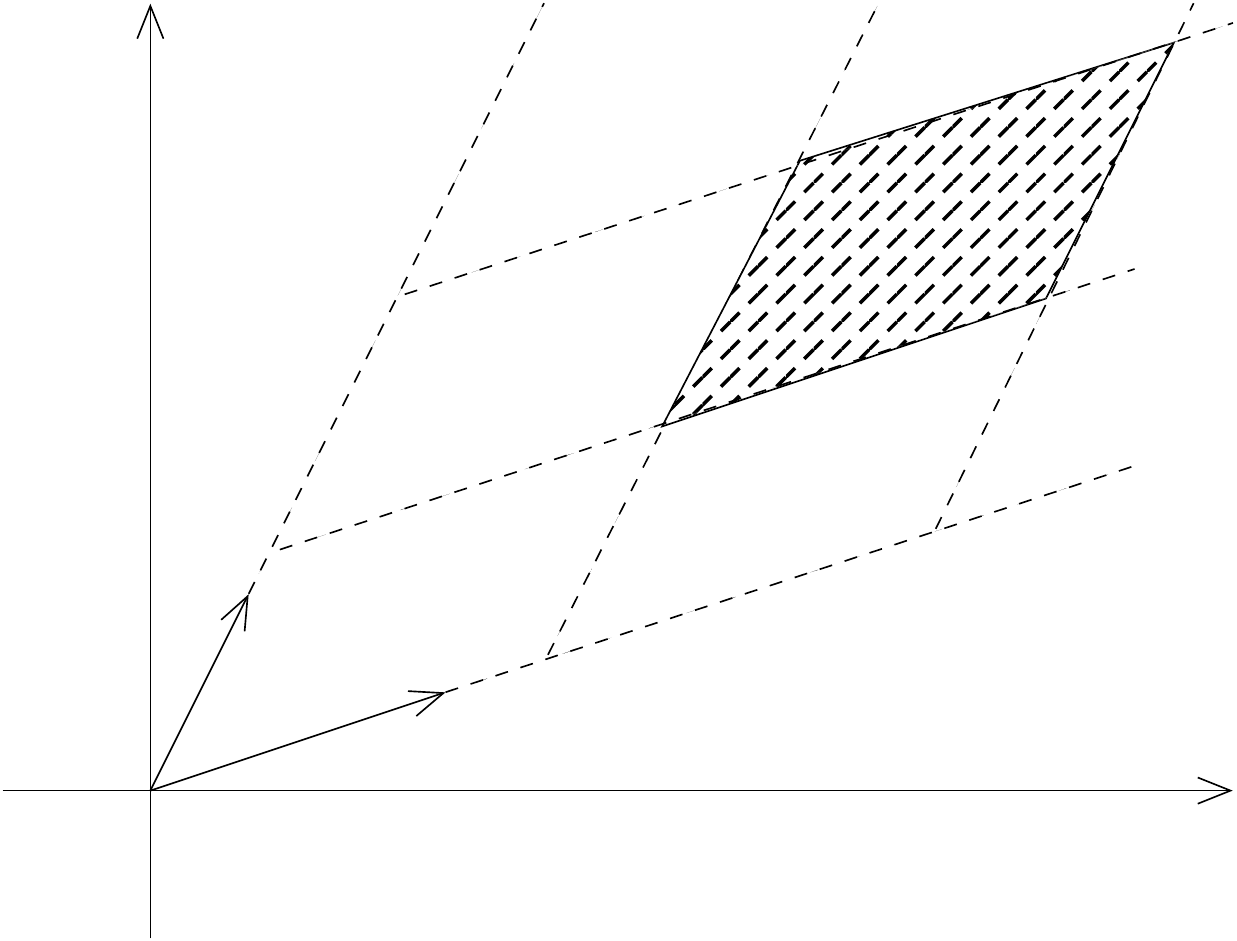}%
        \end{picture}%
        \setlength{\unitlength}{4144sp}%
        \begingroup\makeatletter\ifx\SetFigFont\undefined%
          \gdef\SetFigFont#1#2#3#4#5{%
            \reset@font\fontsize{#1}{#2pt}%
            \fontfamily{#3}\fontseries{#4}\fontshape{#5}%
            \selectfont}%
        \fi\endgroup%
        \begin{picture}(5649,4299)(-686,152)
          \put(1711,1199){\makebox(0,0)[lb]{\smash{{\SetFigFont{17}{20.4}{\familydefault}{\mddefault}{\updefault}{\color[rgb]{0,0,0}$x_1$}%
                }}}}
          \put(3241,1694){\makebox(0,0)[lb]{\smash{{\SetFigFont{17}{20.4}{\familydefault}{\mddefault}{\updefault}{\color[rgb]{0,0,0}$x_1 + \Delta_1$}%
                }}}}
          \put( 91,3044){\makebox(0,0)[lb]{\smash{{\SetFigFont{17}{20.4}{\familydefault}{\mddefault}{\updefault}{\color[rgb]{0,0,0}$x_2 + \Delta_2$}%
                }}}}
          \put(181,1874){\makebox(0,0)[lb]{\smash{{\SetFigFont{17}{20.4}{\familydefault}{\mddefault}{\updefault}{\color[rgb]{0,0,0}$x_2$}%
                }}}}
          \put(991,929){\makebox(0,0)[lb]{\smash{{\SetFigFont{20}{24.0}{\familydefault}{\mddefault}{\updefault}{\color[rgb]{0,0,0}$\mathbf{a}_1$}%
                }}}} 
          \put(406,1289){\makebox(0,0)[lb]{\smash{{\SetFigFont{20}{24.0}{\familydefault}{\mddefault}{\updefault}{\color[rgb]{0,0,0}$\mathbf{a}_2$}%
                }}}}
          \put(3691,4214){\makebox(0,0)[lb]{\smash{{\SetFigFont{25}{30.0}{\familydefault}{\mddefault}{\updefault}{\color[rgb]{0,0,0}$U_y$}%
                }}}}
        \end{picture}%
      }
    \end{figure}
         
    From the previous question, the $A_{U_y}$ of $U_y$ equals the absolute value of the determinant of the matrix $(\Delta_1\a_1 \; \Delta_2\a_2)$:
    \begin{align}
      A_{U_y} &= |\det \begin{pmatrix} \Delta_1a_{11} & \Delta_2a_{12} \\ \Delta_1a_{21} & \Delta_2a_{22} \end{pmatrix} |\\
              &= |\Delta_1\Delta_2a_{11}a_{22} - \Delta_1\Delta_2a_{12}a_{21}| \\ 
              &= |\Delta_1\Delta_2(a_{11}a_{22} - a_{12}a_{21})| \\
              &= \Delta_1\Delta_2 |\det \A|
    \end{align}
    Therefore the area of $U_y$ is the area of $U_x$ times $|\det \A$|.
  \end{solution}
  
\item Give an intuitive explanation why we have equality in the change of variables formula 
  \begin{equation} \int_{U_y} f(\y)d\y = \int_{U_x} f(\A \x) |\det \A| d\x.
  \end{equation} where $\A$ is such that $U_x$ is an axis-aligned (hyper-)
  rectangle as in the previous question.
  
  \begin{solution} We can think that, loosely speaking, the two integrals
    are limits of the following two sums
    \begin{equation} \sum_{\y_i \in U_y}
      f(\y_i) \textrm{vol}(\Delta_{\y_i}) \quad \quad \quad \sum_{\x_i \in U_x}
      f(\A \x_i) |\det \A| \textrm{vol}(\Delta_{\x_i})
    \end{equation}
    where $\x_i = \A^{-1} \y_i$, which means that $\x$ and $\y$ are related
    by $\y = \A \x$. The set of function values $f(\y_i)$ and $f(\A \x_i)$
    that enter the two sums are exactly the same. The volume
    $\textrm{vol}(\Delta_{\x_i})$ of a small axis-aligned hypercube (in $d$
    dimensions) equals $\prod_{i=1}^d \Delta_i$. The image of this small
    axis-aligned hypercube under $\A$ is a parallelogram $\Delta_{\y_i}$
    with volume $\textrm{vol}(\Delta_{\y_i})= |\det \A|
    \textrm{vol}(\Delta_{\x_i})$. Hence
    \begin{equation}
      \sum_{\y_i \in U_y} f(\y_i) \textrm{vol}(\Delta_{\y_i}) = \sum_{\x_i \in U_x} f(\A \x_i) |\det \A| \textrm{vol}(\Delta_{\x_i}).
    \end{equation}
    We must have the term $|\det \A|$ to compensate for the fact that the volume of $U_x$ and $U_y$ are not the same. For example, let $\A$ be a diagonal matrix $\diag(10, 100)$ so that $U_x$ is much smaller than $U_y$. The determinant $\det \A = 1000$ then compensates for the fact that the $\x_i$ values are more condensed than the $\y_i$.
  \end{solution}
  
\end{exenumerate}


\ex{Eigenvalue decomposition}
\label{ex:eigenvalue-decomposition}
For a square matrix $\A$ of size $n \times n$, a vector $\u_i
\neq 0$ which satisfies
\begin{equation}
  \A\u_i = \lambda_i \u_i
  \label{eq:eigenvector}
\end{equation}
is called a eigenvector of $\A$, and $\lambda_i$ is the corresponding
eigenvalue. For a matrix of size $n \times n$, there are $n$
eigenvalues $\lambda_i$ (which are not necessarily distinct).
\begin{exenumerate}
\item Show that if $\u_1$ and $\u_2$ are eigenvectors with $\lambda_1=\lambda_2$, then
  $\u = \alpha \u_1 +\beta \u_2$ is also an eigenvector with the same eigenvalue.
  \begin{solution}
    We compute
    \begin{align}
      \A\u &= \alpha \A\u_1 + \beta \A\u_2 \\
           &= \alpha \lambda \u_1 + \beta \lambda \u_2 \\
           &= \lambda(\alpha\u_1 + \beta\u_2) \\
           &= \lambda \u,
    \end{align}
    so $\u$ is an eigenvector of $\A$ with the same eigenvalue as $\u_1$ and $\u_2$.
  \end{solution}

\item Assume that none of the eigenvalues of $\A$ is zero. Denote by $\U$
  the matrix where the column vectors are linearly independent eigenvectors
  $\u_i$ of $\A$. Verify that \eqref{eq:eigenvector} can be written in matrix form as $\A\U=\U\Lambdab$, where $\Lambdab$ is a
  diagonal matrix with the eigenvalues $\lambda_i$ as diagonal elements.
 
  \begin{solution}
    By basic properties of matrix multiplication, we have
    \begin{align}
      \A\U &= (\A\u_1 \; \A\u_2 \;\ldots \; \A\u_n)
    \end{align}
    With $\A\u_i =\lambda_i \u_i$ for all $i = 1, 2, \ldots, n$, we thus obtain
    \begin{align}
      \A\U & = (\lambda_1\u_1 \; \lambda_2\u_2 \;\ldots \; \lambda_n\u_n) \\
           &= \U\Lambdab.
    \end{align}
  \end{solution}
  
\item Show that we can write, with $\V^T=\U^{-1}$,
  \begin{align}
    \A &= \U \Lambdab \V^\top,&     \A &= \sum_{i=1}^n \lambda_i \u_i\v_i^T,\\
    \A^{-1} &= \U \Lambdab^{-1} \V^\top, & \A^{-1} &= \sum_{i=1}^n \frac{1}{\lambda_i} \u_i\v_i^\top,
  \end{align}
  where $\v_i$ is the $i$-th column of $\V$.
  \begin{solution}
    \begin{enumerate}
    \item[(i)] Since the columns of $\U$ are linearly independent,
      $\U$ is invertible. Because $\A\U = \U\Lambdab$, multiplying from
      the right with the inverse of $\U$ gives $\A = \U\Lambdab \U^{-1} =
      \U\Lambda \V^\top$.
    \item[(ii)] Denote by $\u^{[i]}$ the $i$th row of $\U$,
      $\v^{(j)}$ the $j$th column of $\V^\top$ and $\v^{[j]}$
      the $j$th row of $\V$ and denote $\B = \sum_{i=1}^n
      \lambda_i\u_i\v_i^\top$. Let $\e^{[i]}$ be a row vector
      with 1 in the $i$th place and 0 elsewhere and $\e^{(j)}$ be a
      column vector with 1 in the $j$th place and 0
      elsewhere. Notice that because $\A = \U\Lambdab \V^\top$,
      the element in the $i$th row and $j$th column is
      \begin{align}
        A_{ij}  &= \u^{[i]}\Lambdab \v^{(j)} \\
                &= \u^{[i]}\Lambdab {\v^{[j]}}^\top \\
                &= \u^{[i]} \left( \begin{matrix} \lambda_1V_{j1} \\ \vdots \\ \lambda_nV_{jn} \end{matrix} \right) \\
                &= \sum_{k=1}^n \lambda_k V_{jk}U_{ik}.
      \end{align}
      On the other hand, for matrix $\B$ the element in the $i$th row and $j$th column is
      \begin{align}
        B_{ij}  &= \sum_{k=1}^n \lambda_k\e^{[i]}\u_k\v_k^\top \e^{(j)} \\
                &= \sum_{k=1}^n \lambda_kU_{ik}V_{jk},
      \end{align}
      which is the same as $A_{ij}$. Therefore $\A = \B$.
      
    \item[(iii)] Since $\Lambdab$ is a diagonal matrix with no zeros as diagonal
      elements, it is invertible. We have thus
      \begin{align}
        \A^{-1} &= (\U\Lambdab{\U^{-1}})^{-1} \\
                & = (\Lambdab \U^{-1})^{-1}\U^{-1} \\
                &= \U \Lambda^{-1} \U^{-1} \\
                &= \U\Lambda^{-1}\V^\top.
      \end{align}

    \item[(iv)] This follows from $\A = \U \Lambdab \V^\top = \sum_i \u_i \lambda_i \v_i^\top$, when $\lambda_i$ is replaced with $1/\lambda_i$.
    \end{enumerate}
  \end{solution}  
\end{exenumerate}


\ex{Trace, determinants and eigenvalues}
\label{ex:trace-determinants-eigenvalues}
\begin{exenumerate}
\item Use \exref{ex:eigenvalue-decomposition} to show that $\tr(\A)=\sum_i A_{ii} = \sum_i\lambda_i$. (You can use $\tr(\A\B)=\tr(\B\A)$.)
  \begin{solution}
    Since $\tr(\A\B) = \tr(\B\A$ and $\A = \U\Lambdab \U^{-1}$
    \begin{align}
      \tr(\A)  &= \tr(\U\Lambdab \U^{-1}) \\
               &= \tr(\Lambdab \U^{-1}\U) \\
               &= \tr(\Lambda)\\
               &= \sum_i \lambda_i.
    \end{align}
  \end{solution}

\item Use \exref{ex:eigenvalue-decomposition} to show that $\det \A = \prod_i \lambda_i$. (Use $\det \A^{-1}=1/(\det \A)$ and $\det(\A\B)=\det(\A)\det(\B)$ for any $\A$ and $\B$.)
  \begin{solution}
    We use the eigenvalue decomposition of $\A$ to obtain
    \begin{align}
      \det(\A) &= \det(\U\Lambdab \U^{-1})\\
               &= \det(\U)\det(\Lambdab)\det(\U^{-1}) \\
               &= \frac{\det(\U)\det(\Lambdab)}{\det(\U)}\\
               &= \det(\Lambda)\\
               &= \prod_i \lambda_i,
    \end{align}
    where, in the last line, we have used that the determinant of a diagonal matrix is the product of its elements.

  \end{solution}
\end{exenumerate}


\ex{Eigenvalue decomposition for symmetric matrices}
\label{ex:eigenvalue-decomposition-symmetric-matrices}
\begin{exenumerate}
\item Assume that a matrix $\A$ is symmetric, i.e. $\A^\top=\A$. Let $\u_1$ and
  $\u_2$ be two eigenvectors of $\A$ with corresponding eigenvalues $\lambda_1$
  and $\lambda_2$, with $\lambda_1 \neq \lambda_2$. Show that the two vectors
  are orthogonal to each other.

  \begin{solution}
    Since $\A\u_2 = \lambda_2\u_2$, we have
    \begin{equation}
      \u_1^\top \A\u_2 = \lambda_2\u_1^\top\u_2.
    \end{equation}
    Taking the transpose of $\u_1^\top \A\u_2$ gives
    \begin{align}
      (\u_1^\top \A\u_2)^\top &= (\A\u_2)^\top(\u_1^\top)^\top = \u_2^\top \A^\top \u_1 = \u_2^\top \A\u_1 \\
                              & = \lambda_1\u_2^\top\u_1
    \end{align}
    because $\A$ is symmetric and $\A \u_1 = \lambda_1 \u_1$. On the other hand, the same operation gives
    \begin{align}
      (\u_1^\top \A\u_2)^\top &= (\lambda_2\u_1^\top\u_2)^\top = \lambda_2\u_2^\top\u_1
    \end{align}
    Therefore $\lambda_1\u_2^\top\u_1 = \lambda_2\u_2^\top\u_1$, which is equivalent to $\u_2^\top\u_1(\lambda_1 - \lambda_2) = 0$. Because
    $\lambda_1 \neq \lambda_2$, the only possibility is that
    $\u_2^\top\u_1 = 0$. Therefore $\u_1$ and $\u_2$ are orthogonal to each other.

    The result implies that the eigenvectors of a symmetric matrix $\A$ with
    distinct eigenvalues $\lambda_i$ forms an orthogonal basis. The result
    extends to the case where some of the eigenvalues are the same (not proven).

  \end{solution}

\item A symmetric matrix $\A$ is said to be positive definite if $\v^T\A\v>0$
  for all non-zero vectors $\v$. Show that positive definiteness implies that
  $\lambda_i>0$, $i=1,\ldots, M$. Show that, vice versa, $\lambda_i>0$,
  $i=1 \ldots M$ implies that the matrix $\A$ is positive definite. Conclude
  that a positive definite matrix is invertible.

  \begin{solution}

    Assume that $\v^\top \A\v > 0$ for all $\v \neq 0$. Since eigenvectors are not
    zero vectors, the assumption holds also for eigenvector $\u_k$ with
    corresponding eigenvalue $\lambda_k$. Now
    \begin{align}
      \u_k^\top \A\u_k &= \u_k^\top\lambda_k\u_k = \lambda_k(\u_k^\top\u_k) = \lambda_k||\u_k|| > 0
    \end{align}
    and because $||\u_k|| > 0$, we obtain $\lambda_k > 0$.

    Assume now that all the eigenvalues of $\A$, $\lambda_1, \lambda_2, \ldots, \lambda_n$, are positive and nonzero. We have shown above that there exists an orthogonal basis consisting of eigenvectors $\u_1, \u_2, \ldots, \u_n$ and therefore every vector $\v$ can be written as a
    linear combination of those vectors (we have only shown it for the case of distinct eigenvalues but it holds more generally). Hence for a nonzero vector $\v$ and for some real numbers $\alpha_1, \alpha_2, \ldots, \alpha_n$, we have
    \begin{align}
      \v^\top\A\v &= (\alpha_1\u_1 + + \ldots + \alpha_n\u_n)^\top \A(\alpha_1\u_1 + \ldots + \alpha_n\u_n) \\
                  &= (\alpha_1\u_1 + \ldots + \alpha_n\u_n)^\top(\alpha_1\A\u_1 + \ldots + \alpha_n\A\u_n) \\
                  &= (\alpha_1\u_1 + \ldots + \alpha_n\u_n)^\top(\alpha_1\lambda_1\u_1 + \ldots + \alpha_n\lambda_n\u_n) \\
                  &= \sum_{i,j} \alpha_i\u_i^\top\alpha_j\lambda_j\u_j \\
                  &= \sum_i \alpha_i\alpha_i\lambda_i\u_i^\top\u_i \\
                  &= \sum_i (\alpha_i)^2||\u_i||^2\lambda_i,
    \end{align}
    where we have used that $\u_i^T\u_j = 0$ if $i \neq j$, due to orthogonality of the basis. Since $(\alpha_i)^2 > 0$, $||\u_i||^2 > 0$ and $\lambda_i > 0$ for all $i$, we find that $\v^\top\A\v > 0.$

    Since every eigenvalue of $\A$ is nonzero, we can use \exref{ex:eigenvalue-decomposition} to conclude that inverse of $\A$ exists and equals $\sum_i 1/\lambda_i \u_i \u_i^\top$.

  \end{solution}

\end{exenumerate}


\ex{Power method}
\label{ex:power-method}
We here analyse an algorithm called the ``power method''. The power method takes
as input a positive definite symmetric matrix $\Sigmab$ and calculates the
eigenvector that has the largest eigenvalue (the ``first eigenvector''). For
example, in case of principal component analysis, $\Sigmab$ is the covariance
matrix of the observed data and the first eigenvector is the first principal
component direction.

The power method consists in iterating the update equations
\begin{align}
  \v_{k+1} &= \Sigmab \w_{k}, & \w_{k+1} &= \frac{\v_{k+1}}{||\v_{k+1}||_2},
\end{align}
where $||\v_{k+1}||_2$ denotes the Euclidean norm.

\begin{exenumerate}
\item Let $\U$ the matrix with the (orthonormal) eigenvectors $\u_i$ of $\Sigmab$ as columns. What is the eigenvalue decomposition of the covariance matrix $\Sigmab$?
  
  \begin{solution}
    Since the columns of $\U$ are orthonormal (eigen)vectors, $\U$ is orthogonal, i.e.
    $\U^{-1} = \U^\top$. With \exref{ex:eigenvalue-decomposition} and \exref{ex:eigenvalue-decomposition-symmetric-matrices}, we obtain
    \begin{align}
      \Sigmab &= \U\Lambdab \U^\top,
    \end{align}
    where $\Lambdab$ is the diagonal matrix with eigenvalues $\lambda_i$ of $\Sigmab$ as diagonal elements. Let the eigenvalues be ordered $\lambda_1 > \lambda_2 > \ldots > \lambda_n>0$ (and, as additional assumption, all distinct).
  \end{solution}

\item Let $\tilde{\v}_{k}= \U^T \v_k$ and $\tilde{\w}_{k}= \U^T \w_k$. Write the update equations of the power method in terms of $\tilde{\v}_{k}$ and $\tilde{\w}_{k}$. This means that we are making a change of basis to represent the vectors $\w_k$ and $\v_k$ in the basis given by the eigenvectors of $\Sigmab$.
  
  \begin{solution}
    With
    \begin{align}
      \v_{k+1} &= \Sigmab \w_k \\
               & = \U\Lambdab \U^\top\w_k
    \end{align}
    we obtain
    \begin{align}
      \U^\top\v_{k+1} = \Lambdab \U^\top\w_k.
    \end{align}
    Hence $\tilde{\v}_{k+1} = \Lambdab\tilde{\w}_k$. The norm of $\tilde{\v}_{k+1}$ is the same as the norm of $\v_{k+1}$:
    \begin{align}
      ||\tilde{\v}_{k+1}||_2 &= ||\U^\top\v_{k+1}||_2 \\
                             &= \sqrt{(\U^\top\v_{k+1})^\top(\U^\top\v_{k+1})} \\
                             &= \sqrt{\v_{k+1}^\top \U\U^\top\v_{k+1}} \\
                             &= \sqrt{\v_{k+1}^\top\v_{k+1}} \\
                             &= ||\v_{k+1}||_2.
    \end{align}
    Hence, the update equation, in terms of $\tilde{\v}_{k}$ and $\tilde{\w}_k$, is
    \begin{align}
      \tilde{\v}_{k+1} &= \Lambdab \tilde{\w}_k, & \tilde{\w}_{k+1} &= \frac{\tilde{\v}_{k+1}}{||\tilde{\v}_{k+1}||}.
    \end{align}
    
  \end{solution}

\item Assume you start the iteration with $\tilde{\w}_{0}$. To which vector $\tilde{\w}^{\ast}$ does the iteration converge to? 
  
  \begin{solution}
    Let $\tilde{\w}_0 = \begin{pmatrix} \alpha_1 & \alpha_2 & \ldots & \alpha_n \end{pmatrix}^\top$. Since $\Lambdab$ is a diagonal matrix, we obtain
    \begin{align}
      \tilde{\v}_1 &=  
                     \begin{pmatrix} 
                       \lambda_1\alpha_1 \\ 
                       \lambda_2\alpha_2 \\ 
                       \vdots \\ 
                       \lambda_n\alpha_n 
                     \end{pmatrix} 
                     = 
                     \lambda_1\alpha_1
                     \begin{pmatrix} 
                       1 \\ 
                       \frac{\alpha_2}{\alpha_1}\frac{\lambda_2}{\lambda_1} \\ 
                       \vdots \\ 
                       \frac{\alpha_n}{\alpha_1}\frac{\lambda_n}{\lambda_1}
                     \end{pmatrix}
    \end{align}
    and therefore 
    \begin{align}
      \tilde{\w}_1 &= \frac{\lambda_1\alpha_1}{c_1}
                     \begin{pmatrix} 1 \\
                       \frac{\alpha_2}{\alpha_1}\frac{\lambda_2}{\lambda_1} \\ 
                       \vdots \\ 
                       \frac{\alpha_n}{\alpha_1}\frac{\lambda_n}{\lambda_1}
                     \end{pmatrix},
    \end{align}
    where $c_1$ is a normalisation constant such that $\|\tilde{\w}_1\| = 1$ (i.e. $c_1 = \|\tilde{\v}_1\|$). Hence, for $\tilde{\w}_k$ it holds that
    \begin{align}
      \tilde{\w}_k &= \tilde{c}_k 
                     \begin{pmatrix}
                       1 \\
                       \frac{\alpha_2}{\alpha_1}\left(\frac{\lambda_2}{\lambda_1}\right)^k \\
                       \vdots \\
                       \frac{\alpha_n}{\alpha_1}\left(\frac{\lambda_n}{\lambda_1}\right)^k \\
                     \end{pmatrix},
    \end{align}
    where $\tilde{c}_k$ is again a normalisation constant such that $||\tilde{\w}_k|| = 1$.
    
    As $\lambda_1$ is the dominant eigenvalue, $|\lambda_j/\lambda_1| < 1$ for $j = 2, 3, \ldots, n$, so that
    \begin{equation}
      \lim_{k \rightarrow \infty} \left(\frac{\lambda_j}{\lambda_1} \right)^k = 0, \quad j=2, 3, \ldots, n,
    \end{equation}
    and hence
    \begin{equation}
      \lim_{k\to \infty}
      \begin{pmatrix}
        1 \\
        \frac{\alpha_2}{\alpha_1}\left(\frac{\lambda_2}{\lambda_1}\right)^k \\
        \vdots \\
        \frac{\alpha_n}{\alpha_1}\left(\frac{\lambda_n}{\lambda_1}\right)^k 
      \end{pmatrix}
      = \begin{pmatrix}
          1 \\
          0\\
          \vdots\\
          0
        \end{pmatrix}.
    \end{equation}

    For the normalisation constant $\tilde{c}_k$, we obtain
    \begin{equation}
      \tilde{c}_k = \frac{1}{\sqrt{1 + \sum_{i=2}^n \left(\frac{\alpha_i}{\alpha_1}\right)^2 \left(\frac{\lambda_i}{\lambda_1}\right)^{2k}}},
    \end{equation}
    and therefore
    \begin{align}
      \lim_{k \rightarrow \infty} \tilde{c}_k &= \frac{1}{\sqrt{1 + \sum_{i=2}^n \left(\frac{\alpha_i}{\alpha_1}\right)^2 \underset{k \rightarrow \infty}
                                                {\lim}\left(\frac{\lambda_i}{\lambda_1}\right)^{2k}}} \\
                                              &= \frac{1}{\sqrt{1 + \sum_{i=2}^n \left(\frac{\alpha_i}{\alpha_1}\right)^2 \cdot 0}} \\
                                              &= 1.
    \end{align}
    The limit of the product of two convergent sequences is the product of the limits so that 
    \begin{align}
      \lim_{k \rightarrow \infty} \tilde{\w}_k &= \lim_{k \rightarrow \infty} \tilde{c}_k 
                                                 \lim_{k \rightarrow \infty} \begin{pmatrix}
                                                                               1 \\
                                                                               \frac{\alpha_2}{\alpha_1}\left(\frac{\lambda_2}{\lambda_1}\right)^k \\
                                                                               \vdots \\
                                                                               \frac{\alpha_n}{\alpha_1}\left(\frac{\lambda_n}{\lambda_1}\right)^k \\
                                                                             \end{pmatrix}
                                                 = 
                                                 \begin{pmatrix} 
                                                   1 \\ 
                                                   0 \\ 
                                                   \vdots \\ 
                                                   0 
                                                 \end{pmatrix}.
    \end{align}
    
  \end{solution}
\item Conclude that the power method finds the first eigenvector.
  
  \begin{solution}
Since $\w_k = \U\tilde{\w}_k$, we obtain
\begin{align}
  \lim_{k \rightarrow \infty} \w_k &= \U 
                                     \begin{pmatrix}
                                       1 \\
                                       0 \\
                                       \vdots \\
                                       0
                                     \end{pmatrix}
                                     = \u_1,
\end{align}
which is the eigenvector with the largest eigenvalue, i.e. the ``first'' or ``dominant'' eigenvector. 
\end{solution}

\end{exenumerate}

\chapter{Optimisation} 
\minitoc

\ex{Gradient of vector-valued functions}
\label{ex:grad-vector}
For a function $J$ that maps a column vector $\w \in \mathbb{R}^n$ to $\mathbb{R}$, the gradient is defined as
\begin{equation}
  \nabla J(\w) = \left(
    \begin{array}{c}
      \frac{\partial J(\w)}{\partial w_1} \\ %
      \vdots \\
      \frac{\partial J(\w)}{\partial w_n}
    \end{array} \right),
\end{equation}
where $\partial J(\w)/\partial w_i$ are the partial derivatives of $J(\w)$ with
respect to the $i$-th element of the vector $\w = (w_1, \ldots, w_n)^\top$ (in
the standard basis). Alternatively, it is defined to be the column vector $\nabla J(\w)$ such that
\begin{align}
  J(\w+\epsilon \h) &=  J(\w) + \epsilon \left( \nabla J(\w)\right)^\top \h+O(\epsilon^2) \label{eq:grad-vector}
\end{align}
for an arbitrary perturbation $\epsilon \h$. This phrases the derivative in
terms of a first-order, or affine, approximation to the perturbed function
$J(\w+\epsilon \h)$. The derivative $\nabla J$ is a linear transformation that
maps $\h \in \mathbb{R}^n$ to $\mathbb{R}$ \citep[see e.g.][Chapter 9, for a
formal treatment of derivatives]{Rudin1976}.

Use either definition to determine $\nabla J(\w)$ for the following functions where $\a \in \mathbb{R}^n$, $\A \in \mathbb{R}^{n \times n}$ and
$f: \mathbb{R} \rightarrow \mathbb{R}$ is a differentiable function.
\begin{exenumerate}
\item $J(\w)=\a^\top \w$.

  \begin{solution}
    First method:
    \begin{align}
      J(\w) &= \a^\top\w = \sum_{k=1}^n a_kw_k& \implies&& \frac{\partial{J(\w)}}{\partial{w_i}} &= a_i
    \end{align}
    Hence
    \begin{equation}
      \nabla J(\w) = \begin{pmatrix} a_1 \\ a_2 \\ \vdots \\ a_n \end{pmatrix} = \a.
    \end{equation}
    Second method:
    \begin{align}
      J(\w + \epsilon\h) &= \a^\top(\w + \epsilon\h) = \underbrace{\a^\top\w}_{J(\w)} + \epsilon \underbrace{\a^\top \h}_{\nabla J^\top\h}
    \end{align}
    Hence we find again $\nabla J(\w) = \a$.
  \end{solution}

\item $J(\w)=\w^\top \A\w$.

  \begin{solution}
    First method: We start with
    \begin{align}
      J(\w) &= \w^\top\A\w = \sum_{i=1}^n \sum_{j=1}^n w_iA_{ij}w_j
    \end{align}
    Hence,
    \begin{align}
      \frac{\partial{J(\w)}}{\partial{w_k}} &= \sum_{j=1}^n A_{kj}w_j + \sum_{i=1}^n w_i A_{ik} \\
                                            &= \sum_{j=1}^n A_{kj}w_j + \sum_{i=1}^n w_i (\A^\top)_{ki}\\
                                            &= \sum_{j=1}^m \left(A_{kj}+(\A^\top)_{kj}\right)w_j
    \end{align}
    where we have used that the entry in row $i$ and column $k$ of the matrix $\A$ equals the entry in row $k$ and column $i$ of its transpose $\A^\top$. It follows that
    \begin{align}
      \nabla J(\w) &=
                     \begin{pmatrix}
                       \sum_{j=1}^n  \left(A_{1j}+(\A^\top)_{1j}\right)w_j\\
                       \vdots \\
                       \sum_{j=1}^n  \left(A_{nj}+(\A^\top)_{nj}\right)w_j
                     \end{pmatrix} \\
                   &= (\A+\A^\top)\w,
    \end{align}
    where we have used that sums like $\sum_j B_{ij} w_j$ are equal to the $i$-th element of the matrix-vector product $\B \w$.

    Second method:
    \begin{align}
      J(\w + \epsilon\h) &= (\w + \epsilon\h)^\top \A(\w + \epsilon\h) \\
                         &= \w^\top\A\w + \w^\top\A(\epsilon\h) + \epsilon\h^\top\A\w +
                           \underbrace{\epsilon\h^\top \A\epsilon\h}_{O(\epsilon^2)} \\
                         &= \w^\top\A\w + \epsilon(\w^\top\A\h + \w^\top\A^\top\h) + O(\epsilon^2) \\
                         &= \underbrace{\w^\top\A\w}_{J(\w)} +\epsilon(\underbrace{\w^\top\A + \w^\top\A^\top}_{\nabla J(\w)^\top})\h + O(\epsilon^2)
    \end{align}
    where we have used that $\h^\top\A\w$ is a scalar so that $\h^\top\A\w = (\h^\top\A\w)^\top = \w^\top \A^\top \h$. Hence
    \begin{equation}
      \nabla J(\w)^\top = \w^\top\A + \w^\top\A^\top = \w^\top(\A+\A^\top)
    \end{equation}
    and
    \begin{equation}
      \nabla J(\w) = (\A + \A^\top)\w.
    \end{equation}
  \end{solution}

\item $J(\w)=\w^\top \w$.

  \begin{solution}
    The easiest way to calculate the gradient of $J(\w) = \w^\top\w$ is to use the previous question with $\A = \I$ (the identity matrix).
    Therefore
    \begin{equation}
      \nabla J(\w) = \I\w + \I^\top\w = \w + \w = 2\w.
    \end{equation}

  \end{solution}

\item $J(\w)=||\w||_2$.

  \begin{solution}
    Note that $||\w||_2 = \sqrt{\w^\top \w}$.

    First method: We use the chain rule
    \begin{equation}
      \frac{\partial{J(\w)}}{\partial{w_k}} = \frac{\partial \sqrt{\w^\top \w}}{\partial \w^\top \w}\frac{\partial \w^\top \w}{\partial w_k}
    \end{equation}
    and that
    \begin{equation}
      \frac{\partial \sqrt{\w^\top \w}}{\partial \w^\top \w} = \frac{1}{2\sqrt{\w^\top \w}}
    \end{equation}
    The derivatives $\partial \w^\top \w/\partial w_k$ were calculated in the question above so that
    \begin{equation}
      \nabla J(\w) = \frac{1}{2\sqrt{\w^\top \w}} 2 \w = \frac{\w}{||\w||_2}
    \end{equation}
    Second method: Let $f(\w) = \w^\top \w$. From the previous question, we know that
    \begin{align}
      f(\w + \epsilon \h) & = f(\w) + \epsilon 2\w^\top \h + O(\epsilon^2).
    \end{align}
    Moreover,
    \begin{align}
      \sqrt{z + \epsilon u + O(\epsilon^2)} & = \sqrt{z} + \frac{1}{2\sqrt{z}}(\epsilon u + O(\epsilon^2)) + O(\epsilon^2)\\
                                            & = \sqrt{z} + \epsilon \frac{1}{2\sqrt{z}} u + O(\epsilon^2)
    \end{align}
    With $z=f(\w)$ and $u = 2 \w^\top \h$, we thus obtain
    \begin{align}
      J(\w + \epsilon \h) &= \sqrt{f(\w+\epsilon \h)} \\
                          &= \sqrt{f(\w)} + \epsilon \frac{1}{2\sqrt{f(\w)}} 2\w^\top \h + O(\epsilon^2)\\
                          &= \sqrt{f(\w)} + \epsilon \frac{\w^\top}{\sqrt{f(\w)}}\h + O(\epsilon^2)\\
                          &= J(\w) +  \epsilon \frac{\w^\top}{\sqrt{||\w||_2}}\h + O(\epsilon^2)
    \end{align}
    so that
    \begin{equation}
      \nabla J(\w) = \frac{\w}{||\w||_2}.
    \end{equation}

  \end{solution}

\item $J(\w)=f(||\w||_2)$.
  \begin{solution}
    Either the chain rule or the approach with the Taylor expansion can be used
    to deal with the outer function $f$. In any case:
    \begin{align}
      \nabla J(\w) &= f'(||\w||_2) \nabla ||\w||_2 = f'(||\w||_2) \frac{\w}{||\w||_2},
    \end{align}
    where $f'$ is the derivative of the function $f$.
  \end{solution}

\item $J(\w)=f(\w^\top\a)$.

  \begin{solution}
    We have seen that $\nabla_\w \a^\top\w = \a$. Using the chain rule then yields
    \begin{align}
      \nabla J(\w)  &= f'(\w^\top\a)\nabla (\w^\top\a) \\
      =& f'(\w^\top\a) \a
    \end{align}

  \end{solution}

\end{exenumerate}


\ex{Newton's method}
\label{ex:newtons_method}
Assume that in the neighbourhood of $\w_0$, a function $J(\w)$ can be described by the quadratic approximation
\begin{equation}
  f(\w)=c+ \g^\top(\w-\w_0)+\frac{1}{2}(\w-\w_0)^\top \H(\w-\w_0),
\end{equation}
where $c=J(\w_0)$, $\g$ is the gradient of $J$ with respect to $\w$, and $\H$ a
symmetric positive definite matrix (e.g.\ the Hessian matrix for $J(\w)$ at
$\w_0$ if positive definite).

\begin{exenumerate}
\item Use \exref{ex:grad-vector} to determine $\nabla f(\w)$.
  
  \begin{solution}
    We first write $f$ as
    \begin{align}
      f(\w) &= c + \g^\top(\w-\w_0)+\frac{1}{2}(\w-\w_0)^T\H(\w-\w_0) \\
            &= c - \g^\top \w_0 + \frac{1}{2} \w_0^\top\H \w_0 + \nonumber \\
            &\phantom{=} \g^\top \w + \frac{1}{2}\w^\top\H\w - \frac{1}{2} \w_0^\top \H \w - \frac{1}{2} \w^\top \H \w_0
    \end{align}
	  Using now that $\w^\top \H \w_0$ is a scalar and that $\H$ is symmetric, we have
    \begin{equation}
      \w^\top \H \w_0= (\w^\top \H \w_0)^\top = \w_0^\top \H^\top \w = \w_0^\top \H \w
    \end{equation}
    and hence
    \begin{equation}
      f(\w) = \text{const} + (\g^\top - \w_0^\top \H) \w +  \frac{1}{2}\w^\top\H\w
    \end{equation}
    With the results from \exref{ex:grad-vector} and the fact that $\H$ is symmetric, we thus obtain
    \begin{align}
      \nabla f(\w) &= \g-\H^\top \w_0 + \frac{1}{2}(\H^\top\w + \H\w) \\
                   & = \g-\H \w_0 + \H\w
    \end{align}
    
    The expansion of $f(\w)$ due to the $\w-\w_0$ terms is a bit
    tedious. It is simpler to note that gradients define a linear approximation
    of the function. We can more efficiently deal with $\w-\w_0$ by changing the
    coordinates and determine the linear approximation of $f$ as a function of
    $\v = \w-\w_0$, i.e.\ locally around the point $\w_0$. We then have
    \begin{align}
      \tilde{f}(\v) & = f(\v+\w_0)\\
                    & = c + \g^\top \v + \frac{1}{2} \v^\top \H \v
    \end{align}
    With \exref{ex:grad-vector}, the derivative is
    \begin{equation}
      \nabla_{\v} \tilde{f}(\v) = \g + \H \v
    \end{equation}
    and the linear approximation becomes
    \begin{equation}
      \tilde{f}(\v+\epsilon \h) =  c + \epsilon (\g + \H \v)^\top \h + O(\epsilon^2)
    \end{equation}
    The linear approximation for $\tilde{f}$ determines a linear approximation of $f$ around $\w_0$, i.e.\
    \begin{equation}
      f(\w + \epsilon \h) =  \tilde{f}(\w-\w_0+\epsilon \h) =  c + \epsilon (\g + \H (\w-\w_0))^\top \h + O(\epsilon^2)
    \end{equation}
    so that the derivative for $f$ is
    \begin{equation}
      \nabla_{\w}f(\w) = \g + \H (\w-\w_0) = \g - \H \w_0 + \H \w,
    \end{equation}
    which is the same result as before.
  \end{solution}
  
\item A necessary condition for $\w$ being optimal (leading either to a maximum, minimum or a saddle point) is $\nabla f(\w)=0$. 
  Determine $\w^\ast$ such that $\nabla f(\w)\big|_{\w=\w^\ast}=0$. Provide arguments why $\w^\ast$ is a minimiser of $f(\w)$.
  
  \begin{solution}
    We set the gradient to zero and solve for $\w$:
    \begin{equation}
        \g + \H(\w -\w_0) = 0 \quad \leftrightarrow \quad \w-\w_0 = - \H^{-1} \g
      \end{equation}
      so that
      \begin{equation}
        \w^\ast = \w_0 - \H^{-1} \g.
      \end{equation}
    As we assumed that $\H$ is positive definite, the inverse $\H$ exists (and
    is positive definite too).
    
    Let us consider $f$ as a function of $\v$ around $\w^\ast$, i.e.\ $\w = \w^\ast+\v$. With $\w^\ast+\v-\w_0= -\H^{-1}\g + \v$, we have
    \begin{align}
      f(\w^\ast+\v) & = c + \g^\top(- \H^{-1} \g+\v) + \frac{1}{2}(-\H^{-1}\g + \v)^\top \H (-\H^{-1}\g + \v)
    \end{align}
    Since $\H$ is positive definite, we have that $(-\H^{-1}\g + \v)^\top \H
    (-\H^{-1}\g + \v)>0$ for all $\v$. Hence, as we move away from $\w^\ast$,
    the function increases quadratically, so that $\w^\ast$ minimises $f(\w)$.
    
  \end{solution}

\item In terms of Newton's method to minimise $J(\w)$, what do $\w_0$ and $\w^\ast$ stand for?
  
  \begin{solution}
    The equation
    \begin{equation}
      \w^\ast = \w_0 - \H^{-1} \g.
    \end{equation}
    corresponds to one update step in Newton's method where $\w_0$ is the
    current value of $\w$ in the optimisation of $J(\w)$ and $\w^\ast$ is the
    updated value. In practice rather than determining the inverse $\H^{-1}$, we solve
    \begin{equation}
      \H \p = \g
    \end{equation}
    for $\p$ and then set $\w^\ast = \w_0 - \p$. The vector $\p$ is the
    search direction, and it is possible include a step-length $\alpha$ so that
    the update becomes $\w^\ast = \w_0 - \alpha \p$. The value of $\alpha$ may
    be set by hand or can be determined via line-search methods \citep[see
    e.g.][]{Nocedal1999}.
\end{solution}
\end{exenumerate}


\ex{Gradient of matrix-valued functions}
\label{ex:grad-matrix}
For functions $J$ that map a matrix $\W \in \mathbb{R}^{n\times m}$ to $\mathbb{R}$, the gradient is defined as 
\begin{equation}
  \nabla J(\W) = 
  \begin{pmatrix}
    \frac{\partial J(\W)}{\partial W_{11}} & \ldots &  \frac{\partial J(\W)}{\partial W_{1m}} \\
    \vdots & \vdots & \vdots\\
    \frac{\partial J(\W)}{\partial W_{n1}} & \ldots &  \frac{\partial J(\W)}{\partial W_{nm}}
  \end{pmatrix}.
\end{equation}

Alternatively, it is defined to be the matrix $\nabla J$ such that
\begin{align}
  J(\W+\epsilon \H) &= J(\w) + \epsilon \tr(\nabla J^\top \H) + O(\epsilon^2) \label{eq:grad-matrix}\\
                    &= J(\w) + \epsilon \tr(\nabla J \H^\top) + O(\epsilon^2)
\end{align}
This definition is analogue to the one for vector-valued functions
in \eqref{eq:grad-vector}. It phrases the derivative in terms of a linear
approximation to the perturbed objective $J(\W+\epsilon \H)$ and, more formally,
$\tr \nabla J^\top$ is a linear transformation that maps
$\H \in \mathbb{R}^{n \times m}$ to $\mathbb{R}$ \citep[see e.g.][Chapter 9, for a formal treatment of derivatives]{Rudin1976}.

Let $\e^{(i)}$ be \emph{column} vector which is everywhere zero but in slot $i$
where it is 1. Moreover let $\e^{[j]}$ be a \emph{row} vector which is
everywhere zero but in slot $j$ where it is 1. The outer product
$\e^{(i)}\e^{[j]}$ is then a matrix that is everywhere zero but in row $i$ and
column $j$ where it is one. For $\H = \e^{(i)} \e^{[j]}$, we obtain
\begin{align}
  J(\W+\epsilon \e^{(i)}\e^{[j]}) &= J(\W) + \epsilon \tr((\nabla J)^\top \e^{(i)} \e^{[j]}) + O(\epsilon^2)\\
                                  &= J(\W) + \epsilon  \e^{[j]} (\nabla J)^\top \e^{(i)} + O(\epsilon^2)\\
                                  &= J(\W) + \epsilon \e^{[i]} \nabla J  \e^{(j)}+O(\epsilon^2)
\end{align}
Note that $\e^{[i]} \nabla J \e^{(j)}$ picks the element of the
matrix $\nabla J$ that is in row $i$ and column $j$, i.e.\ $\e^{[i]} \nabla J
\e^{(j)}=\partial J/\partial W_{ij}$.

Use either of the two definitions to find $\nabla
J(\W)$ for the functions below, where
$\u \in \mathbb{R}^n, \v \in \mathbb{R}^m, \A \in \mathbb{R}^{n \times m}$, and
$f: \mathbb{R} \rightarrow \mathbb{R}$ is differentiable.

\begin{exenumerate}
\item $J(\W) = \u^\top\W\v$.
  
\begin{solution}
  First method: With $J(\W) = \sum_{i=1}^n \sum_{j=1}^m u_iW_{ij}v_j$ we have
  \begin{align}
    \frac{\partial J(\W)}{W_{kl}} &= u_kv_l = (\u\v^\top)_{kl}
  \end{align}
  and hence
\begin{equation}
  \nabla J(\W) = \u\v^\top
\end{equation}

Second method: 
\begin{align}
  J(\W + \epsilon \H) & = \u^\top (\W+\epsilon \H)\v\\
                      & = J(\W) + \epsilon \u^\top \H \v\\
                      & = J(\W) + \epsilon \tr(\u^\top \H \v)\\
                      & = J(\W) + \epsilon \tr(\v \u^\top \H)
\end{align}
Hence:
\begin{align}
  \nabla J(\W) & = \u\v^\top
\end{align}

\end{solution}

\item $J(\W) = \u^\top(\W+\A)\v$.
\begin{solution}
  Expanding the objective function gives $J(\W) = \u^\top \W \v
  + \u^\top \A \v$. The second term does not depend on $\W$. With the previous
  question, the derivative thus is
  \begin{equation}
    \nabla J(\W) = \u \v^\top
  \end{equation}
  
\end{solution}
\item $J(\W) = \sum_n f(\w_n^\top\v)$, where $\w_n^\top$ are the rows of the matrix $\W$.
  
  \begin{solution}
    First method:
    \begin{align}
      \frac{\partial J(\W)}{\partial W_{ij}} &= \sum_{k=1}^n \frac{\partial}{\partial W_{ij}} f(\w_k^\top\v) \\
                                             &= f'(\w_i^\top\v) \frac{\partial}{\partial W_{ij}} \underbrace{\w_i^\top\v}_\text{$\sum_{j=1}^m W_{ij}v_j$} \\
                                             &= f'(\w_i^\top\v)v_j
    \end{align}
    Hence
    \begin{equation}
      \nabla J(\W) = f'(\W\v)\v^\top,
    \end{equation}
    where $f'$ operates element-wise on the vector $\W\v$.
    
    Second method:
    \begin{align}
      J(\W) &= \sum_{k=1}^n f(\w_k^\top\v)\\
            &= \sum_{k=1}^n f(\e^{[k]}\W\v),
    \end{align}
    where $\e^{[k]}$ is the unit row vector that is zero everywhere but for element $k$ which equals one. We now perform a perturbation of $\W$ by $\epsilon \H$.
    \begin{align}
J(\W + \epsilon \H) &= \sum_{k=1}^n f(\e^{[k]}(\W + \epsilon \H)\v) \\
                    &= \sum_{k=1}^n f(\e^{[k]}\W\v + \epsilon\e^{[k]}\H\v) \\
         					  &= \sum_{k=1}^n (f(\e^{[k]}\W\v) + \epsilon f'(\e^{[k]}\W\v)\e^{[k]}\H\v + O(\epsilon^2)\\
                    &= J(\W) + \epsilon \left( \sum_{k=1}^n  f'(\e^{[k]}\W\v)\e^{[k]}\right) \H \v + O(\epsilon^2)
    \end{align}
    The term $f'(\e^{[k]}\W\v) \e^{[k]}$ is a row vector that equals $(0, \ldots, 0,f'(\e^{[k]}\W\v), 0, \ldots, 0)$. Hence, we have
    \begin{equation}
      \sum_{k=1}^n  f'(\e^{[k]}\W\v)\e^{[k]}) = f'(\W\v)^\top
    \end{equation}
    where $f'$ operates element-wise on the column vector $\W\v$. The perturbed objective function thus is
    \begin{align}
      J(\W + \epsilon \H) & = J(\W) + \epsilon f'(\W\v)^\top \H \v  + O(\epsilon^2)\\
                          & =  J(\W) + \epsilon \tr\left(f'(\W\v)^\top \H \v\right)  + O(\epsilon^2)\\
                          & =  J(\W) + \epsilon \tr\left(\v f'(\W\v)^\top \H \right)  + O(\epsilon^2)
    \end{align}
    Hence, the gradient is the transpose of $\v f'(\W\v)^\top$, i.e.\
    \begin{equation}
      \nabla J(\W) = f'(\W\v) \v^\top
\end{equation}
\end{solution}

\item $J(\W) = \u^\top \W^{-1}\v$ (\emph{Hint: $(\W+\epsilon \H)^{-1} = \W^{-1} -\epsilon \W^{-1}\H\W^{-1}+O(\epsilon^2)$.})
  \begin{solution}
    We first verify the hint:
    \begin{align}
      \left(\W^{-1} -\epsilon \W^{-1}\H\W^{-1}+O(\epsilon^2)\right) (\W+\epsilon \H)&=  \I + \epsilon \W^{-1} \H - \epsilon \W^{-1}\H + O(\epsilon^2)\\
                                                                                    &= \I + O(\epsilon^2)
    \end{align}
    Hence the identity holds up to terms smaller than $\epsilon^2$, which is
    sufficient we do not care about terms of order $\epsilon^2$ and smaller 
    in the definition of the gradient in \eqref{eq:grad-matrix}.
    
    Let us thus make a first-order approximation of the perturbed objective $J(\W + \epsilon \H)$: 
    \begin{align}
      J(\W + \epsilon\H) &= \u^\top(\W + \epsilon\H\v)^{-1}\v \\
                         &\overset{\textrm{hint}}{=} \u^\top(\W^{-1} - \epsilon \W^{-1}\H \W^{-1} + O(\epsilon^2))\v \\
                         &= \u^\top \W^{-1}\v - \epsilon \u^\top\W^{-1}\H\W^{-1}\v + O(\epsilon^2) \\
                         &= J(\W) -  \epsilon \tr\left(\u^\top\W^{- 1}\H\W^{-1}\v\right) + O(\epsilon^2) \\
                         &= J(\W) - \epsilon \tr\left(\W^{-1}\v\u^\top\W^{-1}\H\right) + O(\epsilon^2)
    \end{align}
    Comparison with \eqref{eq:grad-matrix} gives
    \begin{equation}
      \nabla J^\top = -\W^{-1}\v\u^\top\W^{-1}
    \end{equation}
    and hence
    \begin{equation}
      \nabla J = -\W^{-\top}\u\v^\top\W^{-\top},
    \end{equation}
where $\W^{-\top}$ is the transpose of the inverse of $\W$.

\end{solution}
\end{exenumerate}


\ex{Gradient of the log-determinant}
\label{ex:grad-log-det}
The goal of this exercise is to determine the gradient of
\begin{equation}
  J(\W) = \log | \det(\W)|.
\end{equation}

\begin{exenumerate}
\item Show that the $n$-th eigenvalue $\lambda_n$ can be written as
  \begin{equation}
    \lambda_n=\v_n^\top \W \u_n,
  \end{equation}
  where $\u_n$ is the $n$th eigenvector and $\v_n$ the $n$th column vector of $\U^{-1}$, with $\U$ being the matrix with the eigenvectors $\u_n$ as columns.
  
  \begin{solution}
    As in \exref{ex:eigenvalue-decomposition}, let $\U\Lambdab \V^\top$ be the
    eigenvalue decomposition of $\W$ (with $\V^\top = \U^{-1}$). Then $\Lambdab = \V^\top
    \W\U$ and
    \begin{align}
      \lambda_n &= \e^{[n]}\Lambdab\e^{(n)} \\
                &= \e^{[n]}\V^\top \W \U\e^{(n)} \\
                &= (\V\e^{(n)})^\top \W\U\e^{(n)} \\
                &= \v_n^\top \W\u_n,
    \end{align}
    where $\e^{(n)}$ is the standard basis (unit) vector with a 1 in the $n$-th
    slot and zeros elsewhere, and $\e^{[n]}$ is the corresponding row vector.
  \end{solution}
  
\item Calculate the gradient of $\lambda_n$ with respect to $\W$, i.e. $\nabla \lambda_n(\W)$.

  \begin{solution}
    With \exref{ex:grad-matrix}, we have
    \begin{align}
      \nabla_{\W} \lambda_n(\W) &= \nabla_{\W} \v_n^\top \W \u_n = \v_n \u_n^\top.
    \end{align}
  \end{solution}
  
\item Write $J(\W)$ in terms of the eigenvalues $\lambda_n$ and calculate $\nabla J(\mathbf{\W})$.
  \begin{solution}
    In \exref{ex:trace-determinants-eigenvalues}, we have shown that $\det(\W) = \prod_i \lambda_i$ and hence $|\textrm{det}(W)| = \prod_i |\lambda_i|.$
    \begin{enumerate}
    \item[(i)] If $\W$ is positive definite, its eigenvalues are positive and
      we can drop the absolute values so that $|\det(W)| = \prod_i \lambda_i$.
    \item[(ii)] If $\W$ is a matrix with real entries, then $\W\u = \lambda\u$
        implies $\W\bar{\u} = \bar{\lambda}\bar{\u}$, i.e. if $\lambda$ is a
        complex eigenvalue, then $\bar{\lambda}$ (the complex conjugate of
        $\lambda$) is also an eigenvalue. Since $|\lambda|^2 =
        \lambda\bar{\lambda}$,
        \begin{align}
          |\det(\W)| = \left(\prod_{\lambda_i \in \mathbb{C}} \lambda_i \right) \left(\prod_{\lambda_j \in \mathbb{R}} |\lambda_j| \right).
        \end{align}
      \end{enumerate}
      Now we can write $J(\W)$ in terms of the eigenvalues:
      \begin{align}
        J(\W)  &= \log |\det(\W)| \\
               &= \log\left(\prod_{\lambda_i \in \mathbb{C}} \lambda_i \right) \left(\prod_{\lambda_j \in \mathbb{R}} |\lambda_j|\right) \\
               &= \log\left(\prod_{\lambda_i \in \mathbb{C}} \lambda_i \right) + \log\left(\prod_{\lambda_j \in \mathbb{R}} |\lambda_j| \right) \\
               &= \sum_{\lambda_i \in \mathbb{C}} \log \lambda_i + \sum_{\lambda_j \in \mathbb{R}} \log |\lambda_j|.
      \end{align}
      Assume that the real-valued $\lambda_j$ are non-zero so that
      \begin{align}
        \nabla_{\W} \log |\lambda_j| & = \frac{1}{|\lambda_j|} \nabla_{\W} |\lambda_j|\\
                                     & =  \frac{1}{|\lambda_j|} \textrm{sign}(\lambda_j)\nabla_{\W}\lambda_j
      \end{align}
      Hence
      \begin{align}
        \nabla J(\W) &= \sum_{\lambda_i \in \mathbb{C}}  \nabla_{\W}\log \lambda_i + \sum_{\lambda_j \in \mathbb{R}}  \nabla_{\W} \log |\lambda_j|\\ 
                     &= \sum_{\lambda_i \in \mathbb{C}} \frac{1}{\lambda_i} \nabla_{\W} \lambda_i + 
                       \sum_{\lambda_i \in \mathbb{R}} \frac{1}{|\lambda_i|} \textrm{sign}(\lambda_i)\nabla_{\W}\lambda_i\\
                     &= \sum_{\lambda_i \in \mathbb{C}} \frac{\v_i\u_i^\top}{\lambda_i} + 
                       \sum_{\lambda_i \in \mathbb{R}} \frac{\textrm{sign}(\lambda_i)\v_i\u_i^\top}{|\lambda_i|} \\
                     &= \sum_{\lambda_i \in \mathbb{C}} \frac{\v_i\u_i^\top}{\lambda_i} + 
                       \sum_{\lambda_i \in \mathbb{R}} \frac{\v_i\u_i^\top}{\lambda_i} \\
                     &= \sum_i  \frac{\v_i\u_i^\top}{\lambda_i}.
      \end{align}
      
    \end{solution}
  \item Show that
    \begin{equation}
      \nabla J(\W) = (\W^{-1})^\top.
    \end{equation}
    
    \begin{solution}
    This follows from \exref{ex:eigenvalue-decomposition} where we have found
    that
    \begin{equation}
      \W^{-1}= \sum_i \frac{1}{\lambda_i}\u_i \v_i^\top.
    \end{equation}
    Indeed:
    \begin{align}
      \nabla J(\W) = \sum_i  \frac{\v_i\u_i^\top}{\lambda_i} = \sum_i \frac{1}{\lambda_i} (\u_i\v_i^\top)^\top =  \left(\W^{-1}\right)^\top.
    \end{align}
    
  \end{solution}
\end{exenumerate}


\ex{Descent directions for matrix-valued functions}
\label{ex:descent-directions-for-matrix-valued-functions}
Assume we would like to minimise a matrix valued function $J(\W)$ by gradient
descent, i.e. the update equation is
\begin{equation}
  \W \leftarrow \W - \epsilon \nabla J(\W),
\end{equation}
where $\epsilon$ is the step-length. The gradient $\nabla J(\W)$ was defined in
\exref{ex:grad-matrix}. It was there pointed out that the gradient defines a
first order approximation to the perturbed objective function $J(\W+\epsilon
\H)$. With \eqref{eq:grad-matrix},
\begin{align}
  J(\W - \epsilon \nabla J(\W)) & = J(\W) - \epsilon \tr(\nabla J(\W)^\top \nabla J(\W)) + O(\epsilon^2)
\end{align}
For any (nonzero) matrix $\M$, it holds that
\begin{align}
  \tr(\M^\top \M) &= \sum_{i} (\M^\top\M)_{ii}\\
                  &= \sum_i \sum_j (\M^\top)_{ij} (\M)_{ji}\\
                  &= \sum_i \sum_j M_{ji} M_{ji}\\
                  &= \sum_{ij} (M_{ji})^2\\
                  &> 0, \label{eq:trace-nonnegativity}
\end{align}
which means that $\tr(\nabla J(\W)^\top \nabla J(\W)) > 0$ if the gradient is nonzero, and hence
\begin{equation}
  J(\W - \epsilon \nabla J(\W)) < J(\W)
\end{equation}
for small enough $\epsilon$. Consequently, $\nabla J(\W)$ is a descent
direction. Show that $\A^\top \A \nabla J(\W)\B\B^\top$ for non-zero matrices
$\A$ and $\B$ is also a descent direction or leaves the leaves the objective invariant.

\begin{solution}
  As in the introduction to the question, we appeal to \eqref{eq:grad-matrix} to
  obtain
  \begin{align}
    J(\W - \epsilon \nabla J(\W)\A^\top\A \B\B^\top) & = J(\W) - \epsilon \tr(\nabla J(\W)^\top \A^\top \A \nabla J(\W)\B\B^\top) + O(\epsilon^2)\\
                                                     & =  J(\W) - \epsilon \tr(\B^\top \nabla J(\W)^\top \A^\top \A \nabla J(\W)\B) + O(\epsilon^2),
  \end{align}
  where $\tr(\B^\top \nabla J(\W)^\top\A^\top \A \nabla J(\W)\B)$ takes the form
  $\tr(\M^\top \M)$ with $\M= \A \nabla J(\W)\B$. With
  \eqref{eq:trace-nonnegativity}, we thus have $\tr(\B^\top \nabla J(\W)^\top \A^\top
  \A \nabla J(\W)\B) > 0$ if $\A \nabla J(\W)\B$ is non-zero, and hence
  \begin{equation}
    J(\W - \epsilon \A^\top \A \nabla J(\W) \B\B^\top) < J(\W)
  \end{equation}
  for small enough $\epsilon$. We have equality if $\A \nabla J(\W)\B = 0$,
  e.g.\ if the columns of $\B$ are all in the null space of $\nabla J$.
\end{solution}


\chapter{Directed Graphical Models} 
\minitoc

\ex{Directed graph concepts}
\label{ex:directed-graph-concepts}
Consider the following directed graph:

\begin{center}
    \scalebox{1}{ 
      \begin{tikzpicture}[dgraph]
        \node[cont] (a) at (0,2) {$a$};
        \node[cont] (z) at (2,2) {$z$};
        \node[cont] (q) at (1,1) {$q$};
        \node[cont] (e) at (1,-0.4) {$e$};
        \node[cont] (h) at (3,1) {$h$};
        \draw(a) -- (q);
        \draw(z) -- (h);
        \draw(z) -- (q);
        \draw(q) -- (e);
    \end{tikzpicture}}
  \end{center}

\begin{exenumerate}
\item List all trails in the graph (of maximal length)

  \begin{solution}
    We have
  $$(a,q,e) \quad \quad (a,q,z,h) \quad \quad  (h,z,q,e)$$
    and the corresponding ones with swapped start and end nodes.
  \end{solution}
    
\item List all directed paths in the graph (of maximal length)
  \begin{solution}
$(a,q,e) \quad \quad (z,q,e) \quad \quad (z,h) $
  \end{solution}
  
\item What are the descendants of $z$?
  
  \begin{solution}
$\desc(z) =\{q,e,h\}$
  \end{solution}

\item What are the non-descendants of $q$?
    
  \begin{solution}
$\nondesc(q) = \{a,z,h,e\} \setminus \{e\} = \{a,z,h\}$
  \end{solution}

\item Which of the following orderings are topological to the graph?
  \begin{itemize}
    \item (a,z,h,q,e)
    \item (a,z,e,h,q)
    \item (z,a,q,h,e)
    \item (z,q,e,a,h)
  \end{itemize}
  
  \begin{solution}
    \begin{itemize}
    \item (a,z,h,q,e): yes
    \item (a,z,e,h,q): no ($q$ is a parent of $e$ and thus has to come before $e$ in the ordering)
    \item (z,a,q,h,e): yes
    \item (z,q,e,a,h): no ($a$ is a parent of $q$ and thus has to come before $q$ in the ordering)
    \end{itemize}
  \end{solution}

%
  
\end{exenumerate}

\begin{figure}[h!]
  \begin{center}
    \subfloat[Serial connection]{
      \scalebox{0.9}{ 
        \begin{tikzpicture}[dgraph]
          \node[cont] (x) at (0,0) {$x$};
          \node[cont] (z) at (2,0) {$z$};
          \node[cont] (y) at (4,0) {$y$};
          \draw(x) -- (z);
          \draw(z) -- (y);
      \end{tikzpicture}}
    }\hspace{2ex}
    \subfloat[Diverging connection]{
      \scalebox{0.9}{ 
        \begin{tikzpicture}[dgraph]
          \node[cont] (x) at (0,0) {$x$};
          \node[cont] (z) at (2,0) {$z$};
          \node[cont] (y) at (4,0) {$y$};
          \draw(z) -- (x);
          \draw(z) -- (y);
      \end{tikzpicture}}
    }\hspace{2ex}
    \subfloat[Converging connection]{
      \scalebox{0.9}{ 
        \begin{tikzpicture}[dgraph]
          \node[cont] (x) at (0,0) {$x$};
          \node[cont] (z) at (2,0) {$z$};
          \node[cont] (y) at (4,0) {$y$};
          \draw(x) -- (z);
          \draw(y) -- (z);
        \end{tikzpicture}}
    }
  \end{center}
  \caption{\label{fig:canonical-connections}The three canonical connections in DAGs.}
\end{figure}
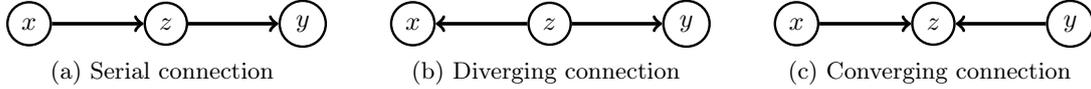

\ex{Canonical connections}
\label{ex:canonical-connections}
We here derive the independencies that hold in the three canonical
connections that exist in DAGs, shown in Figure \ref{fig:canonical-connections}.

\begin{exenumerate}
  \item For the serial connection, use the ordered Markov property to show that 
    $x \independent y \,|\, z$.
    \begin{solution}
     The only topological ordering is $x, z, y$. The predecessors of
     $y$ are $\pre_y = \{x, z\}$ and its parents $\pa_y = \{z\}$. The
     ordered Markov property
     \begin{equation}
       y \independent ( \pre_y \setminus \pa_y ) \mid \pa_y
     \end{equation}
     thus becomes $y \independent (\{x, z\} \setminus z) \mid z$. Hence we have
     \begin{equation}
      y \independent x \mid z,
     \end{equation}
     which is the same as $x \independent y \mid z$ since the
     independency relationship is symmetric.

     This means that if the state or value of $z$ is known (i.e.\ if
     the random variable $z$ is ``instantiated''), evidence about $x$
     will not change our belief about $y$, and vice versa. We say that
     the $z$ node is ``closed'' and that the trail between $x$ and $y$
     is ``blocked'' by the instantiated $z$. In other words, knowing
     the value of $z$ blocks the flow of evidence \emph{between} $x$
     and $y$.
     
    \end{solution}
   \item For the serial connection, show that the marginal $p(x,y)$ does generally not factorise into $p(x)p(y)$, i.e.\ that $x \independent y$ does not hold.
     \begin{solution}
       There are several ways to show the result. One is to
       present an example where the independency does not
       hold. Consider for instance the following model
       \begin{align}
         x & \sim \normal(x; 0, 1)\label{eq:serial-x}\\
         z &= x + n_z\\
         y &= z + n_y \label{eq:serial-y}
       \end{align}
       where $n_z \sim \normal(n_z; 0, 1)$ and $n_y \sim \normal(n_y;
       0, 1)$, both being statistically independent from $x$. Here
       $\normal(\cdot; 0, 1)$ denotes the Gaussian pdf with mean 0 and
       variance 1, and $x \sim \normal(x; 0, 1)$ means that we sample
       $x$ from the distribution $\normal(x; 0, 1)$. Hence $p(z|x) =
       \normal(z; x, 1)$, $p(y|z) = \normal(y; z, 1)$ and $p(x, y,z) =
       p(x) p(z|x) p(y|z) = \normal(x; 0, 1) \normal(z; x,
       1)\normal(y; z, 1)$.

       Whilst we could manipulate the pdfs to show the result, it's
       here easier to work with the generative model in Equations
       \eqref{eq:serial-x} to \eqref{eq:serial-y}. Eliminating $z$
       from the equations, by plugging the definition of $z$ into
       \eqref{eq:serial-y} we have
       \begin{equation}
         y = x + n_z + n_y,
       \end{equation}
       which describes the marginal distribution of $(x, y)$. We see
       that $\E[xy]$ is
       \begin{align}
         \E[xy] & = \E[x^2 + xn_z + x n_y]\\
         & =  \E[x^2] + \E[x]\E[n_z] + \E[x]\E[n_y]\\
         & = 1 + 0 + 0
       \end{align}
       where we have use the linearity of expectation, that $x$ is
       independent from $n_z$ and $n_y$, and that $x$ has zero
       mean. If $x$ and $y$ were independent (or only uncorrelated),
       we had $\E[xy] = \E[x]\E[y] = 0$. However, since $\E[xy] \neq \E[x]
       \E[y]$, $x$ and $y$ are not independent.

       In plain English, this means that if the state of $z$ is
       unknown, then evidence or information about $x$ will influence
       our belief about $y$, and the other way around. Evidence can
       flow through $z$ between $x$ and $y$. We say that the $z$ node
       is ``open'' and the trail between $x$ and $y$ is ``active''.
       
     \end{solution}
 \item For the diverging connection, use the ordered Markov property
   to show that $x \independent y \,|\, z$.
    \begin{solution}
     A topological ordering is $z, x, y$. The predecessors of
     $y$ are $\pre_y = \{x, z\}$ and its parents $\pa_y = \{z\}$. The
     ordered Markov property
     \begin{equation}
       y \independent ( \pre_y \setminus \pa_y ) \mid \pa_y
     \end{equation}
     thus becomes again
     \begin{equation}
       y \independent x \mid z,
     \end{equation}
     which is, since the independence relationship is symmetric, the
     same as $x \independent z \mid z$.

     As in the serial connection, if the state or value $z$ is known,
     evidence about $x$ will not change our belief about $y$, and vice
     versa. Knowing $z$ closes the $z$ node, which blocks the trail
     between $x$ and $y$.
    \end{solution}

 \item For the diverging connection, show that the marginal $p(x,y)$
   does generally not factorise into $p(x)p(y)$, i.e.\ that $x \independent y$ does not
   hold.

   \begin{solution}
     As for the serial connection, it suffices to give an example
     where $x \independent y$ does not hold. We consider the following
     generative model
     \begin{align}
         z & \sim \normal(z; 0, 1)\\
         x &= z + n_x\\
         y &= z + n_y
       \end{align}
     where $n_x \sim \normal(n_x; 0, 1)$ and $n_y \sim \normal(n_y; 0,
     1)$, and they are independent of each other and the other
     variables. We have $\E[x] = \E[z + n_x] = \E[z]+\E[n_x]=0$. On
     the other hand
     \begin{align}
       \E[x y] & = \E[(z + n_x)( z + n_y)]\\
       & = \E[z^2 + z(n_x+n_y) + n_xn_y]\\
       & = \E[z^2] +  \E[z(n_x+n_y)] +  \E[n_xn_y]\\
       & = 1 + 0 + 0
     \end{align}
     Hence, $\E[xy] \neq \E[x]\E[y]$ and we do not have that $x
     \independent y$ holds.

     In a diverging connection, as in the serial connection, if
     the state of $z$ is unknown, then evidence or information about
     $x$ will influence our belief about $y$, and the other way
     around. Evidence can flow through $z$ between $x$ and $y$. We say
     that the $z$ node is open and the trail between $x$ and $y$ is
     active.
   \end{solution}

 \item For the converging connection, show that $x \independent y$.

   \begin{solution}
     We can here again use the ordered Markov property with the
     ordering $y, x, z$. Since $\pre_x = \{y\}$ and $\pa_x =
     \varnothing$, we have
     \begin{equation}
       x \independent ( \pre_x \setminus \pa_x ) \mid \pa_x = x \independent y.
     \end{equation}
     Alternatively, we can use the basic definition of directed
     graphical models, i.e.\
     \begin{align}
       p(x,y,z) & = k(x) k(y) k(z \mid x, y)
     \end{align}
     together with the result that the kernels (factors) are valid
     (conditional) pdfs/pmfs and equal to the conditionals/marginals
     with respect to the joint distribution $p(x,y,z)$, i.e.\
     \begin{align}
       k(x) & = p(x)\\
       k(y) &= p(y) \\
       k(z | x, y) & = p(z|x,y) \quad \text{\small (not needed in the proof below)}
     \end{align}
     Integrating out $z$ gives
     \begin{align}
       p(x,y) & = \int p(x,y,z) \ud z\\
       & =  \int  k(x) k(y) k(z \mid x, y) \ud z\\
       & = k(x) k(y) \underbrace{\int k(z \mid x, y) \ud z}_{1}\\
       & = p(x) p(y)
     \end{align}
     Hence $p(x,y)$ factorises into its marginals, which means that $x
     \independent y$.

     Hence, when we do not have evidence about $z$, evidence about $x$
     will not change our belief about $y$, and vice versa. For the
     converging connection, if no evidence about $z$ is available, the
     $z$ node is closed, which blocks the trail between $x$ and $y$.

   \end{solution}

 \item For the converging connection, show that $x \independent y \mid
   z$ does generally not hold.

   \begin{solution}
     We give a simple example where $x \independent y \mid
     z$ does not hold.
     
     Consider
     \begin{align}
       x & \sim \normal(x;0, 1)\\ y & \sim \normal(y;0, 1)\\ z&= xy +
       n_z
     \end{align}
     where $n_z \sim \normal(n_z; 0, 1)$, independent from the other
     variables. From the last equation, we have
     \begin{equation}
       x y = z - n_z
     \end{equation}
     We thus have
     \begin{align}
       \E[xy \mid z] & = \E[z - n_z \mid z] \\ &= z - 0
     \end{align}
     On the other hand, $\E[x y] = \E[x]\E[y] = 0$. Since $\E[xy \mid z] \neq \E[x y]$, $x \independent y \mid z$ cannot hold.

     The intuition here is that if you know the value of the product
     $xy$, even if subject to noise, knowing the value of $x$ allows
     you to guess the value of $y$ and vice versa.
     
     More generally, for converging connections, if evidence or
     information about $z$ is available, evidence about $x$ will
     influence the belief about $y$, and vice versa. We say that
     information about $z$ opens the $z$-node, and evidence can flow
     between $x$ and $y$.
     
     Note: information about $z$ means that \emph{$z$ or one of its
       descendents} is observed, see exercise \ref{ex:independencies2}.
 
\end{solution}
  
\end{exenumerate}


\ex{Ordered and local Markov properties, d-separation}
\label{ex:ordered-local-d-sep-independencies01}
We continue with the investigation of the graph from \exref{ex:directed-graph-concepts} shown below for reference.

\begin{center}
    \scalebox{1}{ 
      \begin{tikzpicture}[dgraph]
        \node[cont] (a) at (0,2) {$a$};
        \node[cont] (z) at (2,2) {$z$};
        \node[cont] (q) at (1,1) {$q$};
        \node[cont] (e) at (1,-0.4) {$e$};
        \node[cont] (h) at (3,1) {$h$};
        \draw(a) -- (q);
        \draw(z) -- (h);
        \draw(z) -- (q);
        \draw(q) -- (e);
    \end{tikzpicture}}
  \end{center}

  \begin{exenumerate}
  \item The ordering $(z,h,a,q,e)$ is topological to the graph. What are the independencies that follow from the ordered Markov property?

    \begin{solution}
    A distribution that factorises over the graph satisfies the independencies
    $$ x_i \independent \left(\pre_i \setminus \pa_i\right) \mid \pa_i \text{ for all } i$$
    for all orderings of the variables that are topological to the graph. The ordering comes into play via the predecessors $\pre_i = \{x_1, \ldots, x_{i-1}\}$ of the variables $x_i$; the graph via the parent sets $\pa_i$.

  For the graph and the specified topological ordering, the predecessor sets are
      $$\pre_z = \varnothing, \pre_h = \{z\}, \pre_a = \{z,h\}, \pre_q = \{z,h,a\}, \pre_e =\{z,h,a,q\}$$
      The parent sets only depend on the graph and not the topological ordering. They are:
      $$\pa_z = \varnothing, \pa_h = \{z\}, \pa_a = \varnothing, \pa_q =\{a,z\}, \pa_e = \{q\}, $$
      The ordered Markov property reads  $x_i \independent \left(\pre_i \setminus \pa_i \right)\mid \pa_i$ where the $x_i$ refer to the ordered variables, e.g.\ $x_1 = z, x_2 = h, x_3 = a,$ etc.

      With
      $$ \pre_h \setminus \pa_h = \varnothing \quad \pre_a \setminus \pa_a = \{z,h\} \quad \pre_q \setminus \pa_q = \{h\} \quad \pre_e \setminus \pa_e = \{z,h,a\}$$
      we thus obtain
      $$ h \independent \varnothing \mid z \quad \quad
      a \independent \{z,h\} \quad \quad q \independent
      h \mid \{a,z\} \quad \quad e \independent \{z,h,a\} \mid q$$ The
      relation $ h \independent \varnothing \mid z$ should be understood as
      ``there is no variable from which $h$ is independent given $z$'' and
      should thus be dropped from the list. Note that we can possibly
      obtain more independence relations for variables that occur later in
      the topological ordering. This is because the set
      $\pre \setminus \pa$ can only increase when the predecessor set
      $\pre$ becomes larger.
      \end{solution}
    
  \item What are the independencies that follow from the local Markov property?

    \begin{solution}
      The non-descendants are
      $$\nondesc(a) = \{z,h\} \quad \nondesc(z) = \{a\} \quad \nondesc(h) = \{a,z,q,e\}$$
      $$\quad \nondesc(q) = \{a,z,h\} \quad \nondesc(e) = \{a,q,z,h\}$$
With the parent sets as before, the independencies that follow from the local Markov property are $x_i \independent \left( \nondesc(x_i)\setminus \pa_i \right) \mid \pa_i$, i.e.\:
      $$a \independent \{z,h\} \quad \quad z \independent a \quad \quad h \independent \{a,q,e\} \mid z \quad \quad q \independent  h \mid \{a,z\} \quad \quad e \independent \{a,z,h\} \mid q$$

    \end{solution}

  \item The independency relations obtained via the ordered and local
    Markov property include $q \independent h \mid \{a,z\}$. Verify
    the independency using d-separation.

    \begin{solution}

     The only trail from $q$ to $h$ goes through $z$ which is in a tail-tail configuration. Since $z$ is part of the conditioning set, the trail is blocked and the result follows.
    \end{solution}

  \item Use d-separation to check whether $a \independent h \mid e$ holds.
    \begin{solution}
      The trail from $a$ to $h$ is shown below in red together with the default states of the nodes along the trail.
\begin{center}
    \scalebox{1}{ 
      \begin{tikzpicture}[dgraph]
        \node[cont] (a) at (0,2) {$a$};
        \node[cont] (z) at (2,2) {$z$};
        \node[cont] (q) at (1,1) {$q$};
        \node[cont] (e) at (1,-0.4) {$e$};
        \node[cont] (h) at (3,1) {$h$};
        \draw (a) -- (q);
        \draw[color=red,-,style=dashed] (a) -- (q);
        \draw(z) -- (h);
        \draw[color=red,-,style=dashed] (z) -- (h);
        \draw(z) -- (q);
        \draw[color=red,-,style=dashed](z) -- (q);
        \draw(q) -- (e);
        \node[left=0.01of q] {closed};
        \node[above=0.01of z] {open};
    \end{tikzpicture}}
\end{center}
Conditioning on $e$ opens the $q$ node since $q$ in a collider configuration on the path. 
\begin{center}
    \scalebox{1}{ 
      \begin{tikzpicture}[dgraph]
        \node[cont] (a) at (0,2) {$a$};
        \node[cont] (z) at (2,2) {$z$};
        \node[cont] (q) at (1,1) {$q$};
        \node[contobs] (e) at (1,-0.4) {$e$};
        \node[cont] (h) at (3,1) {$h$};
        \draw (a) -- (q);
        \draw[color=red,-,style=dashed] (a) -- (q);
        \draw(z) -- (h);
        \draw[color=red,-,style=dashed] (z) -- (h);
        \draw(z) -- (q);
        \draw[color=red,-,style=dashed](z) -- (q);
        \draw(q) -- (e);
        \node[left=0.01of q] {open};
        \node[above=0.01of z] {open};
    \end{tikzpicture}}
\end{center}
    The trail from $a$ to $h$ is thus active, which means that the relationship does not hold because  $a \notind h \mid e$ for some distributions that factorise over the graph.
    \end{solution}

  \item Assume all variables in the graph are binary. How many numbers do you need to specify, or learn from data, in order to fully specify the probability distribution?
    \begin{solution}
      The graph defines a set of probability mass functions (pmf) that factorise as
      $$ p(a,z,q,h,e) = p(a) p(z) p(q|a,z) p(h|z) p(e|q)$$
      To specify a member of the set, we need to specify the (conditional) pmfs on the right-hand side. The (conditional) pmfs can be seen as tables, and the number of elements that we need to specified in the tables are:\\
      - 1 for $p(a)$ \\
      - 1 for $p(z)$ \\
      - 4 for $p(q | a,z)$ \\
      - 2 for $p(h | z)$\\
      - 2 for $p(e|q)$ \\
      In total, there are 10 numbers to specify. This is in contrast to $2^5-1 = 31$ for a distribution without independencies.
      Note that the number of parameters to specify could be further reduced by making parametric assumptions. 
      
      \end{solution}
    
  \end{exenumerate}

\ex{More on ordered and local Markov properties, d-separation}
We continue with the investigation of the graph below

\begin{center}
    \scalebox{1}{ 
      \begin{tikzpicture}[dgraph]
        \node[cont] (a) at (0,2) {$a$};
        \node[cont] (z) at (2,2) {$z$};
        \node[cont] (q) at (1,1) {$q$};
        \node[cont] (e) at (1,-0.4) {$e$};
        \node[cont] (h) at (3,1) {$h$};
        \draw(a) -- (q);
        \draw(z) -- (h);
        \draw(z) -- (q);
        \draw(q) -- (e);
    \end{tikzpicture}}
  \end{center}

\begin{exenumerate}
  \item Why can the ordered or local Markov property not be used to check whether $a \independent h \mid e$ may hold? 
    \begin{solution}
      The independencies that follow from the ordered or local Markov property require conditioning on parent sets. However, $e$ is not a parent of any node so that the above independence assertion cannot be checked via the ordered or local Markov property. 
      \end{solution}

  \item The independency relations obtained via the ordered and local Markov property include $a \independent \{z,h\}$. Verify the independency using d-separation.

    \begin{solution}
      All paths from $a$ to $z$ or $h$ pass through the node $q$ that forms a head-head connection along that trail. Since neither $q$ nor its descendant $e$ is part of the conditioning set, the trail is blocked and the independence relation follows.
  \end{solution}

  \item Determine the Markov blanket of $z$.
    \begin{solution}
      The Markov blanket is given by the parents, children, and co-parents. Hence: $\MB(z) = \{a,q,h\}$.
    \end{solution}

  \item Verify that  $ q \independent  h \mid \{a,z\}$  holds by manipulating the probability distribution induced by the graph.
    \begin{solution}

      A basic definition of conditional statistical independence $x_1
      \independent x_2 \mid x_3$ is that the (conditional) joint
      $p(x_1, x_2 \mid x_3)$ equals the product of the (conditional)
      marginals $p(x_1 \mid x_3)$ and $p(x_2 \mid x_3)$. In other
      words, for discrete random variables,
      \begin{align}
        x_1 &\independent x_2 \mid x_3 &  &\Longleftrightarrow &  p(x_1, x_2 \mid x_3) &= \left( \sum_{x_2}  p(x_1, x_2 \mid x_3) \right) \left( \sum_{x_1}  p(x_1, x_2 \mid x_3) \right)
      \end{align}
      We thus answer the question by showing that (use integrals in
      case of continuous random variables)
      \begin{align}
      p(q,h | a,z) &= \left(\sum_h  p(q,h | a,z)\right) \left( \sum_q  p(q,h | a,z)\right)
      \end{align}

      First, note that the graph defines a set of probability density
      or mass functions that factorise as
      $$ p(a,z,q,h,e) = p(a) p(z) p(q|a,z) p(h|z) p(e|q)$$ We then use the sum-rule to
      compute the joint distribution of $(a,z,q,h)$, i.e.\ the
      distribution of all the variables that occur in $p(q,h|a,z)$ 
        \begin{align}
          p(a,z,q,h) &= \sum_{e} p(a,z,q,h,e)\\
          &= \sum_{e}  p(a) p(z) p(q|a,z) p(h|z) p(e|q)\\
          & =  p(a) p(z) p(q|a,z) p(h|z) \underbrace{\sum_{e}  p(e|q)}_{1}\\
          & =  p(a) p(z) p(q|a,z) p(h|z),
        \end{align}
        where $\sum_{e}  p(e|q) = 1$ because (conditional) pdfs/pmfs are normalised so that the integrate/sum to one. We further have
        \begin{align}
          p(a,z) & = \sum_{q,h} p(a,z,q,h)\\
          & = \sum_{q,h} p(a) p(z) p(q|a,z) p(h|z)\\
          & = p(a) p(z) \sum_{q}  p(q|a,z) \sum_h p(h|z)\\
          & = p(a) p(z)
        \end{align}
        so that
        \begin{align}
          p(q,h | a,z) & = \frac{ p(a,z,q,h)}{ p(a,z)} \\
          & = \frac{ p(a) p(z) p(q|a,z) p(h|z)}{ p(a) p(z)}\\
          & =  p(q|a,z) p(h|z).
        \end{align}        
        We further see that $p(q| a,z)$ and $p(h|z)$ are the marginals of $p(q,h|a,z)$, i.e.
        \begin{align}
          p(q| a,z) & = \sum_h  p(q,h | a,z)\\
          p(h|z) & = \sum_q  p(q,h | a,z).
        \end{align}
        This means that
        \begin{equation}
           p(q,h | a,z) =\left( \sum_h  p(q,h | a,z) \right) \left(\sum_q  p(q,h | a,z) \right),
        \end{equation}
        which shows that $q \independent h | a,z$.

        We see that using the graph to determine the independency is
        easier than manipulating the pmf/pdf.
        
    \end{solution}
    
\end{exenumerate}

\begin{figure}[h!]
  \centering
  \includegraphics[width=0.9\textwidth]{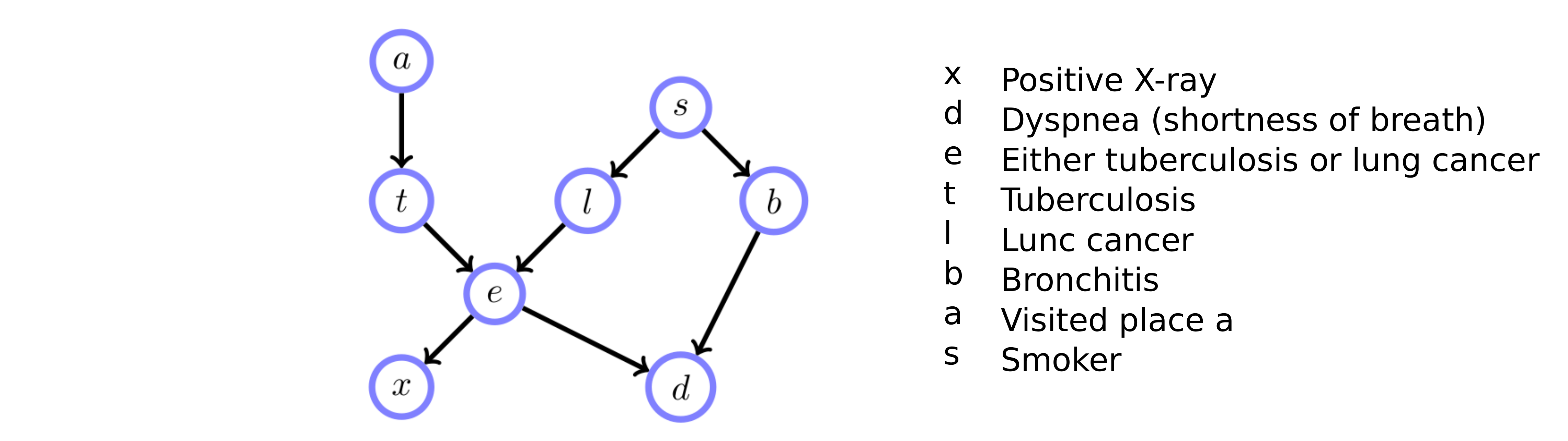}
  \caption{\label{fig:chest-clinic}Graphical model for \exref{ex:chest-clinic} (Barber Figure 3.15).}
\end{figure}

\ex{Chest clinic {\small \citep[based on][Exercise 3.3]{Barber2012}}} 
\label{ex:chest-clinic}
The directed graphical model in Figure \ref{fig:chest-clinic} is about
the diagnosis of lung disease (t=tuberculosis or l=lung cancer). In
this model, a visit to some place ``$a$'' is thought to increase the
probability of tuberculosis.

\begin{exenumerate}
\item Explain which of the following independence relationships hold for all distributions that factorise over the graph.
  \begin{enumerate}
  \item $t \independent s \mid d$

    \begin{solution}
      \begin{itemize}
        \item There are two trails from $t$ to $s$: $(t,e,l,s)$ and $(t,e,d,b,s)$. 
        \item The trail $(t,e,l,s)$ features a collider node $e$ that is opened by the conditioning variable $d$. The trail is thus active and we do not need to check the second trail because for independence all trails needed to be blocked.
        \item The independence relationship does thus generally not hold.
      \end{itemize}
      
      \end{solution}
    
  \item $l \independent b \mid s$

\begin{solution}
  \begin{itemize}
  \item There are two trails from $l$ to $b$: $(l,s,b)$ and $(l,e,d,b)$
  \item The trail $(l,s,b)$ is blocked by $s$ ($s$ is in a tail-tail configuration and part of the conditioning set)
  \item The trail $(l,e,d,b)$ is blocked by the collider configuration for node $d$.
  \item  All trails are blocked so that the independence relation holds.
  \end{itemize}
\end{solution}

  \end{enumerate}
  
\item Can we simplify $p(l|b,s)$ to $p(l|s)$?

  \begin{solution}
    Since $l \independent b \mid s$, we have $p(l|b,s) = p(l|s)$.
  \end{solution}

\end{exenumerate}

\ex{More on the chest clinic {\small \citep[based on][Exercise 3.3]{Barber2012}}} 
\label{ex:chest-clinic2}
Consider the directed graphical model in Figure \ref{fig:chest-clinic}.

\begin{exenumerate}
\item Explain which of the following independence relationships hold for all distributions that factorise over the graph.
  \begin{enumerate}
\item $a \independent s \mid l$
  \begin{solution}
    \begin{itemize}
      \item There are two trails from $a$ to $s$: $(a,t,e,l,s)$ and $(a,t,e,d,b,s)$
      \item The trail $(a,t,e,l,s)$ features a collider node $e$ that blocks the trail (the trail is also blocked by $l$).
      \item The trail $(a,t,e,d,b,s)$ is blocked by the collider node $d$. 
      \item All trails are blocked so that the independence relation holds.
    \end{itemize}
  \end{solution}
  
\item $a \independent s \mid l, d$

  \begin{solution}
    \begin{itemize}
      \item There are two trails from $a$ to $s$: $(a,t,e,l,s)$ and $(a,t,e,d,b,s)$
      \item The trail $(a,t,e,l,s)$ features a collider node $e$ that is opened by the conditioning variable $d$ but the $l$ node is closed by the conditioning variable $l$: the trail is blocked
      \item The trail $(a,t,e,d,b,s)$ features a collider node $d$ that is opened by conditioning on $d$. On this trail, $e$ is not in a head-head (collider) configuration) so that all nodes are open and the trail active.
      \item Hence, the independence relation does generally not hold.
    \end{itemize}
  \end{solution}          
  
  \end{enumerate}

\item Let $g$ be a (deterministic) function of $x$ and $t$. Is the expected value $\E[g(x,t) \mid l,b]$ equal to $\E[g(x,t) \mid l]$?

  \begin{solution}
    The question boils down to checking whether $x,t \independent b \mid l$. For the independence relation to hold, all trails from both $x$ and $t$ to $b$ need to be blocked by $l$.

    \begin{itemize}
      \item For $x$, we have the trails $(x,e,l,s,b)$ and $(x,e,d,b)$
      \item Trail $(x,e,l,s,b)$ is blocked by $l$
      \item Trail $(x,e,d,b)$ is blocked by the collider configuration of node $d$.
      \item For $t$, we have the trails $(t,e,l,s,b)$ and $(t,e,d,b)$
      \item Trail $(t,e,l,s,b)$ is blocked by $l$.
      \item Trail $(t,e,d,b)$ is blocked by the collider configuration of node $d$.
    \end{itemize}
    As all trails are blocked we have $x,t \independent b \mid l$ and  $\E[ g(x,t) \mid l,b] = \E[ g(x,t) \mid l]$.
    
  \end{solution}
\end{exenumerate}

\ex{Hidden Markov models}
\label{ex:HMM}
This exercise is about directed graphical models that are specified by the following DAG:
\begin{center}
  \scalebox{0.8}{
    \begin{tikzpicture}[dgraph]
      \node[cont] (y1) at (0,0) {$y_1$};
      \node[cont] (y2) at (2,0) {$y_2$};
      \node[cont] (y3) at (4,0) {$y_3$};
      \node[cont] (y4) at (6,0) {$y_4$};
      \node[cont] (x1) at (0,2) {$x_1$};
      \node[cont] (x2) at (2,2) {$x_2$};
      \node[cont] (x3) at (4,2) {$x_3$};
      \node[cont] (x4) at (6,2) {$x_4$};
      \draw(x1)--(y1);\draw(x2)--(y2);\draw(x3)--(y3);\draw(x4)--(y4);
      \draw(x1)--(x2);\draw(x2)--(x3);\draw(x3)--(x4);
  \end{tikzpicture}}
\end{center}
These models are called ``hidden'' Markov models because we typically assume to only observe the $y_i$ and not the $x_i$ that follow a Markov model.

\begin{exenumerate}
  \item Show that all probabilistic models specified by the DAG factorise as
    $$p(x_1,y_1, x_2, y_2, \ldots,x_4,y_4) = p(x_1)p(y_1|x_1)p(x_2 | x_1)p(y_2|x_2)p(x_3|x_2)p(y_3|x_3)p(x_4|x_3)p(y_4|x_4)$$
    \begin{solution}
      From the definition of directed graphical models it follows that
      $$p(x_1,y_1, x_2, y_2, \ldots,x_4,y_4) = \prod_{i=1}^4 p(x_i |
      \pa(x_i)) \prod_{i=1}^4 p(y_i | \pa(y_i)).$$ The result is then
      obtained by noting that the parent of $y_i$ is given by $x_i$ for all $i$,
      and that the parent of $x_i$ is $x_{i-1}$ for $i=2,3,4$ and that $x_1$ does not have a parent ($\pa(x_1) = \varnothing$).
    \end{solution}
  \item Derive the independencies implied by the ordered Markov property with the topological ordering $(x_1, y_1, x_2, y_2, x_3, y_3, x_4, y_4)$

    \begin{solution}
      $$y_i \independent x_1, y_1, \ldots, x_{i-1}, y_{i-1} \mid x_i \quad \quad  x_i \independent x_1, y_1, \ldots, x_{i-2},y_{i-2},y_{i-1} \mid x_{i-1}$$

      \end{solution}

  \item Derive the independencies implied by the ordered Markov property with the topological ordering $(x_1, x_2, \ldots, x_4, y_1, \ldots, y_4)$.
    \begin{solution} For the $x_i$, we use that for $i \ge 2$: $\pre(x_i) = \{x_1, \ldots, x_{i-1}\}$ and $\pa(x_i) = x_{i-1}$. For the $y_i$, we use that
      $\pre(y_1) = \{x_1,\ldots, x_4\}$, that $\pre(y_i) = \{x_1,\ldots, x_4, y_1, \ldots, y_{i-1}\}$ for $i>1$, and that $\pa(y_i) = x_i$. The ordered Markov property then gives:
  \begin{align*}
    x_3 \independent x_1 \mid x_2 && x_4 \independent \{x_1,x_2\} \mid x_3\\
    y_1 \independent \{x_2, x_3, x_4\} \mid x_1 &&  y_2 \independent \{x_1, x_3, x_4, y_1\} \mid x_2 \\
    y_3 \independent \{x_1, x_2, x_4, y_1, y_2\} \mid x_3 &&  y_4 \independent \{x_1, x_2, x_3, y_1, y_2, y_3\} \mid x_4
  \end{align*}

    \end{solution}

  \item Does $y_4 \independent y_1 \mid y_3$ hold?

    \begin{solution}
    The trail $y_1-x_1-x_2-x_3-x_4-y_4$ is active: none of the nodes
    is in a collider configuration, so that their default state is
    open and conditioning on $y_3$ does not block any of the nodes on
    the trail.

    While $x_1-x_2-x_3-x_4$ forms a Markov chain, where e.g.\ $x_4
    \independent x_1 \mid x_3$ holds, this not so for the distribution
    of the $y$'s.

    \end{solution}
\end{exenumerate}

\ex{Alternative characterisation of independencies}
\label{ex:independencies1}
We have seen that $x \independent y | z$ is characterised by
  $p(x,y | z) =p(x | z) p(y| z)$
  or, equivalently, by
  $p(x| y, z) = p(x | z)$.
  Show that further equivalent characterisations are
\begin{align}
  p(x,y,z) &= p(x|z) p(y|z) p(z) \quad \text{and} \label{eq:characterisation1} \\
  p(x,y,z) &= a(x,z) b(y,z) \quad \text{for some non-neg. functions \;} a(x,z) \text{\; and \;} b(x,z). \label{eq:characterisation2} 
  \end{align}
  The characterisation in Equation \eqref{eq:characterisation2} is particularly important for undirected graphical models.
\begin{solution}
  We first show the equivalence of  $p(x,y | z) =p(x | z) p(y| z)$ and
  $p(x,y,z) = p(x|z) p(y|z) p(z)$: By the product rule, we have
  $$p(x,y,z) = p(x,y|z) p(z).$$ If $p(x,y | z) =p(x | z) p(y| z)$, it
  follows that $p(x,y,z) = p(x|z) p(y|z) p(z)$. To show the opposite
  direction assume that $p(x,y,z) = p(x|z) p(y|z) p(z)$ holds. By
  comparison with the decomposition in the product rule, it follows
  that we must have $p(x,y | z) = p(x | z) p(y| z)$ whenever $p(z)>0$
  (it suffices to consider this case because for $z$ where $p(z)=0$, $p(x,y | z)$ may not be
  uniquely defined in the first place).

  Equation \eqref{eq:characterisation1} implies
  \eqref{eq:characterisation2} with $a(x,z) = p(x | z)$ and $b(y,z) =
  p(y| z)p(z)$. We now show the inverse. Let us assume that $p(x,y,z)
  = a(x,z) b(y,z)$. By the product rule, we have
  \begin{align}
    p(x,y | z)p(z) & = a(x,z) b(y,z).\\
  \end{align}
  Summing over $y$ gives
  \begin{align}
    \sum_y  p(x,y | z)p(z) & = p(z) \sum_y p(x,y | z)\\
    & = p(z) p(x|z)
  \end{align}
  Moreover
  \begin{align}
     \sum_y  p(x,y | z)p(z) & = \sum_y  a(x,z) b(y,z)\\
    & = a(x,z) \sum_y b(y,z)
  \end{align}
  so that 
  \begin{equation}
    a(x,z) = \frac{p(z) p(x | z)}{\sum_y b(y,z)}
    \label{eq:a}
  \end{equation}
  Since the sum of $p(x | z)$ over $x$ equals one we have
  \begin{equation}
    \sum_x a(x,z) = \frac{p(z)}{\sum_y b(y,z)}.
    \label{eq:asum}
  \end{equation}
  
  Now, summing $p(x,y | z) p(z)$ over $x$ yields
   \begin{align}
     \sum_x  p(x,y | z) p(z) & =  p(z) \sum_x p(x,y|z). \\
     & = p(y | z) p(z)
   \end{align}
   We also have
    \begin{align}
      \sum_x  p(x,y | z) p(z) &= \sum_x a(x,z) b(y,z) \\
      &= b(y,z) \sum_x a(x,z) \\
    &\stackrel{\eqref{eq:asum}}{=} b(y,z) \frac{p(z)}{\sum_y b(y,z)}
    \end{align}
    so that
    \begin{equation}
      p(y | z) p(z) =  p(z) \frac{b(y,z)}{\sum_y b(y,z)}
      \label{eq:brel}
    \end{equation}
   We thus have
   \begin{align}
     p(x,y,z) & = a(x,z) b(y,z) \\
     &  \stackrel{\eqref{eq:a}}{=} \frac{p(z) p(x | z)}{\sum_y b(y,z)} b(y,z) \\
     & = p(x | z)p(z) \frac{ b(y,z)}{\sum_y b(y,z)} \\
     & \stackrel{\eqref{eq:brel}}{=} p(x|z) p(y|z) p(z)
   \end{align}
   which is Equation \eqref{eq:characterisation1}.
  \end{solution}

\ex{More on independencies}
\label{ex:independencies2}
This exercise is on further properties and characterisations of statistical independence.

\begin{exenumerate}
\item Without using d-separation, show that $x \independent \{y,w\} \mid z$ implies that $x \independent y \mid z$ and $x \independent w \mid z$.\\
\emph{Hint: use the definition of statistical independence in terms of the factorisation of pmfs/pdfs.}
  
  \begin{loesung}
    We consider the joint distribution $p(x,y,w | z)$. By assumption
    \begin{equation}
      p(x,y,w | z) = p(x |z) p(y,w | z)
    \end{equation}
    We have to show that $x \independent y| z$ and $x \independent w | z$. For simplicity, we assume that the variables are discrete valued. If not, replace the sum below with an integral.

    To show that $x \independent y| z$, we marginalise $p(x,y,w | z)$
    over $w$ to obtain
    \begin{align}
      p(x,y | z ) & = \sum_w p(x,y,w | z) \\
      & = \sum_w  p(x |z) p(y,w | z) \\
      & = p(x | z) \sum_w p(y,w | z)
    \end{align}
    Since $\sum_w p(y,w | z)$ is the marginal $p(y |z)$, we have
    \begin{equation}
       p(x,y | z ) =  p(x | z) p(y | z),
    \end{equation}
    which means that $x \independent y | z$.

    To show that $x \independent w| z$, we similarly marginalise
    $p(x,y,w | z)$ over $y$ to obtain $p(x,w | z) =
    p(x|z) p(w |z)$, which means that $x \independent w | z$.
    
  \end{loesung}

\item For the directed graphical model below, show that the following two statements hold without using d-separation:

  \begin{align}
  &x \independent y \quad \text{and} \label{eq:statement1} \\
  &x \notind y \mid w \label{eq:statement2}
  \end{align}

  \begin{center}
  \scalebox{0.9}{ 
    \begin{tikzpicture}[dgraph]
      \node[cont] (x) at (0,0) {$x$};
      \node[cont] (z) at (2,0) {$z$};
      \node[cont] (w) at (2,-1.5) {$w$};
      \node[cont] (y) at (4,0) {$y$};
      \draw(x) -- (z);
      \draw(y) -- (z);
      \draw(z) -- (w);
  \end{tikzpicture}}
  \end{center}
  The exercise shows that not only conditioning on a collider node but also on one of its descendents activates the trail between $x$ and $y$. You can use the result that $x \independent y |w \Leftrightarrow p(x,y,w) = a(x,w)b(y,w)$ for some non-negative functions $a(x,w)$ and $b(y,w)$.
  
  \begin{solution}
    The graphical model corresponds to the factorisation $$p(x,y,z,w) = p(x) p(y) p(z|x,y) p(w|z).$$
    For the marginal $p(x,y)$ we have to sum (integrate) over all $(z,w)$
    \begin{align}
      p(x,y) &= \sum_{z,w} p(x,y,z,w) \\
      & = \sum_{z,w} p(x) p(y) p(z|x,y) p(w|z)\\
      & = p(x) p(y) \sum_{z,w} p(z| x,y) p(w|z)\\
      & = p(x) p(y) \underbrace{\sum_{z} p(z| x,y)}_{1} \underbrace{\sum_{w} p(w|z)}_{1}\\
      & = p(x) p(y)
    \end{align}
    Since $p(x,y) = p(x) p(y)$ we have $x \independent y$.

    For $x \notind y | w$, compute $p(x,y,w)$ and use the result $x \independent y |w \Leftrightarrow p(x,y,w) = a(x,w)b(y,w)$.
    \begin{align}
      p(x,y,w) & = \sum_z p(x,y,z,w)\\
      & = \sum_z p(x) p(y) p(z|x,y) p(w|z) \\
      & = p(x) \underbrace{p(y) \sum_z p(z|x,y) p(w|z)}_{k(x,y,w)}
    \end{align}
    Since $p(x,y,w)$ cannot be factorised as $a(x,w) b(y,w)$, the relation $x \independent y | w$ cannot generally hold.
  \end{solution}

\end{exenumerate}

\ex{Independencies in directed graphical models}

Consider the following directed acyclic graph.
       \begin{center}
         \scalebox{0.9}{ 
           \begin{tikzpicture}[dgraph]
             \node[cont] (x1) at (0,0) {$x_1$};
             \node[cont] (x2) at (0,-2) {$x_2$};
             \node[cont] (x3) at (0,-4) {$x_3$};
             \node[cont] (x4) at (2,-1.5) {$x_4$};
             \node[cont] (x5) at (2,-3) {$x_5$};
             \node[cont] (x6) at (2,-4.5) {$x_6$};
             \node[cont] (x7) at (4,0) {$x_7$};
             \node[cont] (x8) at (4,-2) {$x_8$};
             \node[cont] (x9) at (6,-1) {$x_9$};

             \draw (x1) -- (x2);
             \draw (x2) -- (x3);
             \draw (x1) -- (x4);
             \draw (x4) -- (x5);
             \draw (x5) -- (x6);
             \draw (x7) -- (x4);
             \draw (x7) -- (x8);
             \draw (x3) -- (x6);
             \draw (x8) -- (x6);
             \draw (x7) -- (x9);
         \end{tikzpicture}}
       \end{center}
   For each of the statements below, determine whether it holds for
   all probabilistic models that factorise over the graph. Provide a
   justification for your answer.
   
   \begin{exenumerate}
   \item  $p(x_7 | x_2) = p(x_7)$

     \begin{solution}
 Yes, it holds. $x_2$ is a non-descendant of $x_7$,
 $\pa(x_7)=\varnothing$, and hence, by the local Markov property, $x_7
 \independent x_2$, so that $p(x_7 | x_2) = p(x_7)$.
     \end{solution}
     
   \item  $x_1 \notind x_3$  

     \begin{solution}
 No, does not hold. $x_1$ and $x_3$ are d-connected, which only implies independence for \emph{some} and not all distributions that factorise over the graph. The graph generally only allows us to read out independencies and not dependencies.

     \end{solution}

   \item  $p(x_1,x_2,x_4) \propto \phi_1(x_1,x_2) \phi_2(x_1,x_4)$ for some non-negative functions $\phi_1$ and $\phi_2$. 

     \begin{solution}
       Yes, it holds. The statement is equivalent to $x_2 \independent x_4 \mid  x_1$. There are three trails from $x_2$ to $x_4$, which are all blocked:
       \begin{enumerate}
       \item $x_2-x_1-x_4$: this trail is blocked because $x_1$ is in a tail-tail connection and it is observed, which closes the node.
       \item $x_2-x_3-x_6-x_5-x_4$: this trail is blocked because
         $x_3, x_6, x_5$ is in a collider configuration, and $x_6$ is
         not observed (and it does not have any descendants).
       \item $x_2-x_3-x_6- x_8- x_7- x_4$: this trail is blocked because
         $x_3, x_6, x_8$ is in a collider configuration, and $x_6$ is
         not observed (and it does not have any descendants).
       \end{enumerate}
       Hence, by the global Markov property (d-separation), the independency holds.
     \end{solution}

   \item  $x_2 \independent x_9 \mid \{x_6, x_8\}$ 

     \begin{solution}
       No, does not hold. Conditioning on $x_6$ opens the collider node $x_4$ on the trail $x_2 - x_1 - x_4 - x_7 - x_9$, so that the trail is active.
     \end{solution}

   \item  $x_8 \independent \{x_2,x_9\} \mid \{x_3,x_5, x_6, x_7\}$ 

     \begin{solution}
       Yes, it holds. $\{x_3,x_5,x_6,x_7\}$ is the Markov blanket of $x_8$, so that $x_8$ is independent of remaining nodes given the Markov blanket.
     \end{solution}

   \item  $\E[ x_2 \cdot x_3 \cdot x_4 \cdot x_5 \cdot x_8 \mid x_7] = 0$ if $\E[x_8 \mid x_7] = 0$

     \begin{solution}
       Yes, it holds. $\{x_2,x_3,x_4,x_5\}$ are non-descendants of $x_8$, and $x_7$ is the parent of $x_8$, so that $x_8 \independent \{x_2,x_3,x_4,x_5\} \mid x_7$. This means that $\E[ x_2 \cdot x_3 \cdot x_4 \cdot x_5 \cdot x_8 \mid x_7] = \E[ x_2 \cdot x_3 \cdot x_4 \cdot x_5 \mid x_7] \E[x_8 \mid x_7] =0$.
     \end{solution}

   \end{exenumerate}

\ex{Independencies in directed graphical models}
  Consider the following directed acyclic graph:
   \begin{center}
    \begin{tikzpicture}[dgraph]

      \node[cont] (m1) at (0,0) {$m_1$};
      \node[cont] (s1) at (2,0) {$s_1$};
      \node[cont] (u1) at (0,-1.5) {$u_1$};
      \node[cont] (v1) at (2,-1.5) {$v_1$};
      \node[cont] (x1) at (0,-3) {$x_1$};
      \node[cont] (y1) at (2,-3) {$y_1$};
      \node[cont] (theta1) at (1,-4) {$\theta_1$};
      
      \draw (m1) -- (u1);
      \draw (m1) -- (v1);
      \draw (s1) -- (u1);
      \draw (s1) -- (v1);
      \draw (u1) -- (x1);
      \draw (v1) -- (y1);
      \draw (theta1) -- (x1);
      \draw (theta1) -- (y1);

      \node[cont] (m2) at (4,0) {$m_2$};
      \node[cont] (s2) at (6,0) {$s_2$};
      \node[cont] (u2) at (4,-1.5) {$u_2$};
      \node[cont] (v2) at (6,-1.5) {$v_2$};
      \node[cont] (x2) at (4,-3) {$x_2$};
      \node[cont] (y2) at (6,-3) {$y_2$};
      \node[cont] (theta2) at (5,-4) {$\theta_2$};
      
      \draw (m2) -- (u2);
      \draw (m2) -- (v2);
      \draw (s2) -- (u2);
      \draw (s2) -- (v2);
      \draw (u2) -- (x2);
      \draw (v2) -- (y2);
      \draw (theta2) -- (x2);
      \draw (theta2) -- (y2);
      
      \draw (theta1) -- (theta2);
      
    \end{tikzpicture}
   \end{center}
   
    For each of the statements below, determine whether it holds for
   all probabilistic models that factorise over the graph. Provide a
   justification for your answer.
   
   \begin{exenumerate}
   \item $x_1 \independent x_2$

     \begin{solution}
       Does not hold. The trail $x_1 - \theta_1 - \theta_2 - x_2$ is
       active (unblocked) because none of the nodes is in a collider
       configuration or in the conditioning set.
     \end{solution}
     
   \item $p(x_1, y_1, \theta_1, u_1) \propto \phi_A(x_1, \theta_1, u_1)\phi_B(y_1, \theta_1, u_1)$ for some non-negative functions $\phi_A$ and $\phi_B$ 

     \begin{solution}
       Holds. The statement is equivalent to $x_1 \independent y_1
       \mid \{\theta_1, u_1\}$. The conditioning set $\{\theta_1,
       u_1\}$ blocks all trails from $x_1$ to $y_1$ because they are
       both only in serial configurations in all trails from $x_1$ to
       $y_1$, hence the independency holds by the global Markov
       property.  Alternative justification: the conditioning set is
       the Markov blanket of $x_1$, and $x_1$ and $y_1$ are not
       neighbours which implies the independency.
     \end{solution}
     
   \item $v_2 \independent \{u_1, v_1, u_2, x_2\} \mid \{m_2, s_2, y_2, \theta_2\}$

     \begin{solution}
        Holds. The conditioning set is the Markov blanket of $v_2$
        (the set of parents, children, and co-parents): the set of
        parents is $\pa(v_2)=\{m_2, s_2\}$, $y_2$ is the only child of
        $v_2$, and $\theta_2$ is the only other parent of
        $y_2$. And $v_2$ is independent of all other variables
        given its Markov blanket.
     \end{solution}

   \item $\E [ m_2 \mid  m_1 ] = \E[m_2]$

     \begin{solution}
       Holds. There are four trails from $m_1$ to $m_2$, namely via
       $x_1$, via $y_1$, via $x_2$, via $y_2$. In all trails the four
       variables are in a collider configuration, so that each of the
       trails is blocked. By the global Markov property
       (d-separation), this means that $m_1 \independent m_2$ which
       implies that $\E[m_2 \mid m_1] = \E[m_2]$.

       Alternative justification 1: $m_2$ is a non-descendent of $m_1$
       and $\pa(m_2) = \varnothing$. By the directed local Markov
       property, a variable is independent from its non-descendents
       given the parents, hence $m_2 \independent m_1$.

       Alternative justification 2: We can choose a topological
       ordering where $m_1$ and $m_2$ are the first two
       variables. Moreover, their parent sets are both empty. By the
       directed ordered Markov, we thus have $m_1 \independent m_2$.
       
     \end{solution}

   \end{exenumerate}

\chapter{Undirected Graphical Models}
\minitoc

\ex{Visualising and analysing Gibbs distributions via undirected graphs}
\label{ex:visualising-gibbs-distributions}
We here consider the Gibbs distribution
$$p(x_1, \ldots, x_5) \propto \phi_{12}(x_1,x_2)\phi_{13}(x_1,x_3)\phi_{14}(x_1,x_4)\phi_{23}(x_2,x_3)\phi_{25}(x_2,x_5)\phi_{45}(x_4,x_5)$$

\begin{exenumerate}
\item Visualise it as an undirected graph.

  \begin{solution}

   We draw a node for each random variable $x_i$. There is an edge between two nodes if the corresponding variables co-occur in a factor. 
\begin{center}
\scalebox{0.9}{
\begin{tikzpicture}[ugraph]

\node[cont] (x1) at (0,0) {$x_1$};
\node[cont] (x2) at (2,1) {$x_2$};
\node[cont] (x3) at (1,2) {$x_3$};
\node[cont] (x4) at (4,0) {$x_4$};
\node[cont] (x5) at (3,2) {$x_5$};

\draw(x1) -- (x2);
\draw(x1) -- (x3);
\draw(x1) -- (x4);
\draw(x2) -- (x3);
\draw(x2) -- (x5);
\draw(x4) -- (x5);

\end{tikzpicture}}
\end{center}

  \end{solution}

\item What are the neighbours of $x_3$ in the graph?

  \begin{solution}
    The neighbours are all the nodes for which there is a single connecting edge. Thus:   
    $\ne(x_3) = \{x_1, x_2\}$. (Note that sometimes, we may denote $\ne(x_3)$ by $\ne_3$.)
  \end{solution}

\item Do we have $x_3 \independent x_4 \mid x_1, x_2$?

  \begin{solution}
    Yes. The conditioning set $\{x_1,x_2\}$ equals $\ne_3$, which is also the Markov blanket of  $x_3$. This means that $x_3$ is conditionally independent of all the other variables given $\{x_1,x_2\}$, i.e.\ $x_3 \independent x_4,x_5 \mid x_1, x_2$, which implies that $x_3 \independent x_4 \mid x_1, x_2$. (One can also use graph separation to answer the question.)
  \end{solution}
  
\item What is the Markov blanket of $x_4$? 

  \begin{solution}
  
    The Markov blanket of a node in a undirected graphical model equals the set of its neighbours: $\MB(x_4) = \ne(x_4) = \ne_4 = \{x_1,x_5\}$.  This implies, for example, that
    $x_4 \independent x_2,x_3 \mid x_1,x_5$.
  \end{solution}
  
\item On which minimal set of variables $A$ do we need to condition to have $x_1 \independent x_5 \mid A$?

  \begin{solution}
We first identify all trails from $x_1$ to $x_5$. There are three such trails: $(x_1,x_2,x_5)$, $(x_1, x_3, x_2, x_5)$, and $(x_1,x_4, x_5)$. Conditioning on $x_2$ blocks the first two trails, conditioning on $x_4$ blocks the last. We thus have: $x_1 \independent x_5 \mid x_2,x_4$, so that $A=\{x_2,x_4\}$.
    
\end{solution}

\end{exenumerate}


\ex{Factorisation and independencies for undirected graphical models}
\label{ex:factorisation-independencies-undirected-graphical-model}
Consider the undirected graphical model defined by the graph in Figure \ref{fig:undirected-graphical-model}.
  \begin{figure}[h]
  \begin{center}
   \scalebox{0.9}{ 
     \begin{tikzpicture}[ugraph]
       \node[cont] (x1) at (0,0) {$x_1$};
       \node[cont] (x2) at (0,-2) {$x_2$};
       \node[cont] (x3) at (2,0) {$x_3$};
       \node[cont] (x4) at (2,-2) {$x_4$};
       \node[cont] (x5) at (3,-1) {$x_5$};
       \node[cont] (x6) at (5,-2) {$x_6$};
       \draw(x1) -- (x3);
       \draw(x1) -- (x2);
       \draw(x2) -- (x4);
       \draw(x1) -- (x4);
       \draw(x3) -- (x4);
       \draw(x3) -- (x5);
       \draw(x4) -- (x5);
       \draw(x5) -- (x6);
       \draw(x4) -- (x6);
   \end{tikzpicture}}
  \end{center}
  \caption{\label{fig:undirected-graphical-model} Graph for \exref{ex:factorisation-independencies-undirected-graphical-model}}
\end{figure}  

\begin{exenumerate}

\item What is the set of Gibbs distributions that is induced by the graph?

  \begin{solution}
  The graph in Figure \ref{fig:undirected-graphical-model} has four maximal cliques:
  $$(x_1,x_2,x_4) \quad (x_1,x_3,x_4) \quad (x_3,x_4,x_5) \quad (x_4, x_5, x_6)$$
  The Gibbs distributions are thus
  $$p(x_1,\ldots,x_6) \propto \phi_1(x_1,x_2,x_4) \phi_2(x_1,x_3,x_4) \phi_3(x_3,x_4,x_5) \phi_4(x_4, x_5, x_6)$$
\end{solution}

\item Let $p$ be a pdf that factorises according to the graph. Does $p(x_3 | x_2, x_4) = p(x_3 | x_4)$ hold?

  \begin{solution}
    $p(x_3 | x_2, x_4) = p(x_3 | x_4)$ means that $x_3 \independent
    x_2 \mid x_4$. We can use the graph to check whether this
    generally holds for pdfs that factorise according to the
    graph. There are multiple trails from $x_3$ to $x_2$, including
    the trail $(x_3,x_1,x_2)$, which is not blocked by $x_4$. From the
    graph, we thus cannot conclude that $x_3 \independent x_2 \mid x_4$,
    and $p(x_3 | x_2, x_4) = p(x_3 | x_4)$ will generally not hold
    (the relation may hold for some carefully defined factors $\phi_i$).
    
  \end{solution}

\item Explain why $x_2 \independent x_5 \mid x_1,x_3,x_4,x_6$ holds for all distributions that factorise over the graph.

  \begin{solution}
    Distributions that factorise over the graph satisfy the pairwise Markov property. Since $x_2$ and $x_5$ are not neighbours, and $x_1,x_3,x_4,x_6$
    are the remaining nodes in the graph, the independence relation follows from the pairwise Markov property.    
  \end{solution}
  
\item Assume you would like to approximate $\E (x_1 x_2 x_5 \mid
  x_3,x_4)$, i.e.\ the expected value of the product of $x_1$, $x_2$,
  and $x_5$ given $x_3$ and $x_4$, with a sample average. Do you
  need to have joint observations for all five variables $x_1, \ldots, x_5$?
  
  \begin{solution}

    In the graph, all trails from $\{x_1,x_2\}$ to $x_5$ are blocked by $\{x_3,x_4\}$, so that $x_1, x_2 \independent x_5 \mid x_3, x_4$. We thus have
    $$\E (x_1 x_2 x_5 \mid x_3,x_4) =  \E (x_1 x_2 \mid x_3,x_4)  \E (x_5 \mid x_3,x_4).$$
    Hence, we only need joint observations of $(x_1, x_2, x_3, x_4)$ and $(x_3, x_4,x_5)$. Variables $(x_1,x_2)$ and $x_5$ do not need to be jointly measured.

  \end{solution}
  
\end{exenumerate}

\ex{Factorisation and independencies for undirected graphical models}
\label{ex:factorisation-independencies-diamond}

Consider the undirected graphical model defined by the following graph, sometimes called a diamond configuration.

\begin{center}
  \scalebox{1}{ 
    \begin{tikzpicture}[ugraph]
      \node[cont] (w) at (0,1) {$w$};
      \node[cont] (x) at (1,2) {$x$};
      \node[cont] (y) at (2,1) {$y$};
      \node[cont] (z) at (1,0) {$z$};
      \draw(w) -- (x);
      \draw(w) -- (z);
      \draw(x) -- (y);
      \draw(y) -- (z);
    \end{tikzpicture}
  }
\end{center}

\begin{exenumerate}

\item How do the pdfs/pmfs of the undirected graphical model factorise?

  \begin{solution}
    The maximal cliques are $(x, w)$, $(w, z)$, $(z, y)$ and $(x,
    y)$. The undirected graphical model thus consists of pdfs/pmfs
    that factorise as follows
    \begin{equation}
      p(x, w, z, y) \propto \phi_1(x,w) \phi_2(w,z) \phi_3(z,y) \phi_4(x,y)
    \end{equation}
    
  \end{solution}

\item List all independencies that hold for the undirected graphical model.

  \begin{solution}
    We can generate the independencies by conditioning on
    progressively larger sets. Since there is a trail between any two
    nodes, there are no unconditional independencies. If we condition
    on a single variable, there is still a trail that connects the
    remaining ones. Let us thus consider the case where we condition
    on two nodes. By graph separation, we have
    \begin{equation}
      w \independent y \mid x,z \quad \quad x \independent z \mid w,y
    \end{equation}
    These are all the independencies that hold for the model, since
    conditioning on three nodes does not lead to any independencies in
    a model with four variables.
  \end{solution}

\end{exenumerate}
  

\ex{Factorisation from the Markov blankets I}
\label{ex:factorisation-from-the-Markov-blankets-I}

Assume you know the following Markov blankets for all variables $x_1,
\ldots, x_4, y_1, \ldots y_4$ of a pdf or pmf $p(x_1, \ldots, x_4,
y_1, \ldots,y_4)$. 
\begin{align}
  \MB(x_1) &= \{x_2, y_1\} & \MB(x_2)&=\{x_1, x_3, y_2\} & \MB(x_3)&=\{x_2, x_4, y_3\} & \MB(x_4)&=\{x_3, y_4\}\\
  \MB(y_1) &= \{x_1\}      & \MB(y_2)&= \{x_2\}         &  \MB(y_3)&= \{x_3\}         &  \MB(y_4) &= \{x_4\}
\end{align}
Assuming that $p$ is positive for all possible
values of its variables, how does $p$ factorise?

\begin{solution}
In undirected graphical models, the Markov blanket for a variable is
the same as the set of its neighbours. Hence, when we are given all
Markov blankets we know what local Markov property $p$ must
satisfy. For positive distributions we have an equivalence between $p$
satisfying the local Markov property and $p$ factorising over the
graph. Hence, to specify the factorisation of $p$ it suffices to
construct the undirected graph $H$ based on the Markov blankets and
then read out the factorisation.

We need to build a graph where the neighbours of each variable equals
the indicated Markov blanket. This can be easily done by starting with
an empty graph and connecting each variable to the variables in its
Markov blanket.

We see that each $y_i$ is only connected to $x_i$. Including those
Markov blankets we get the following graph:

\begin{center}
  \scalebox{1}{
    \begin{tikzpicture}[ugraph]
      \node[cont] (y1) at (0,0) {$y_1$};
      \node[cont] (y2) at (2,0) {$y_2$};
      \node[cont] (y3) at (4,0) {$y_3$};
      \node[cont] (y4) at (6,0) {$y_4$};
      \node[cont] (x1) at (0,2) {$x_1$};
      \node[cont] (x2) at (2,2) {$x_2$};
      \node[cont] (x3) at (4,2) {$x_3$};
      \node[cont] (x4) at (6,2) {$x_4$};
      \draw(x1)--(y1);\draw(x2)--(y2);\draw(x3)--(y3);\draw(x4)--(y4);
  \end{tikzpicture}}
\end{center}

Connecting the $x_i$ to their neighbours according to the Markov blanket thus gives:

\begin{center}
  \scalebox{1}{
    \begin{tikzpicture}[ugraph]
      \node[cont] (y1) at (0,0) {$y_1$};
      \node[cont] (y2) at (2,0) {$y_2$};
      \node[cont] (y3) at (4,0) {$y_3$};
      \node[cont] (y4) at (6,0) {$y_4$};
      \node[cont] (x1) at (0,2) {$x_1$};
      \node[cont] (x2) at (2,2) {$x_2$};
      \node[cont] (x3) at (4,2) {$x_3$};
      \node[cont] (x4) at (6,2) {$x_4$};
      \draw(x1)--(y1);\draw(x2)--(y2);\draw(x3)--(y3);\draw(x4)--(y4);
      \draw(x1)--(x2);\draw(x2)--(x3);\draw(x3)--(x4);
  \end{tikzpicture}}
\end{center}

The graph has maximal cliques of size two, namely the $x_i-y_i$ for
$i=1,\ldots, 4$, and the $x_i-x_{i+1}$ for $i=1, \ldots, 3$.  Given
the equivalence between the local Markov property and factorisation
for positive distributions, we know that $p$ must factorise as
\begin{align}
  p(x_1, \ldots, x_4, y_1, \ldots, y_4) = \frac{1}{Z} \prod_{i=1}^3 m_i(x_i,x_{i+1}) \prod_{i=1}^4 g_i(x_i,y_i),
\end{align}
where $m_i(x_i, x_{i+1})>0$, $g(x_i,y_i)>0$ are positive factors (potential functions).

The graphical model corresponds to an undirected version of a hidden
Markov model where the $x_i$ are the unobserved (latent, hidden)
variables and the $y_i$ are the observed ones. Note that the $x_i$
form a Markov chain.

\end{solution}

\ex{Factorisation from the Markov blankets II}
\label{ex:factorisation-from-the-Markov-blankets-II}

We consider the same setup as in Exercise \ref{ex:factorisation-from-the-Markov-blankets-I} but we now assume that we do not know all Markov blankets but only
\begin{align}
  \MB(x_1) &= \{x_2, y_1\} & \MB(x_2)&=\{x_1, x_3, y_2\} & \MB(x_3)&=\{x_2, x_4, y_3\} & \MB(x_4)&=\{x_3, y_4\}
\end{align}
Without inserting more independencies than those specified by the
Markov blankets, draw the graph over which $p$ factorises and state
the factorisation. (Again assume that $p$ is positive for all possible
values of its variables).

\begin{solution}
We take the same approach as in Exercise
\ref{ex:factorisation-from-the-Markov-blankets-I}. In particular, the
Markov blankets of a variable are its neighbours in the graph. But
since we are not given all Markov blankets and are not allowed to
insert additional independencies, we must assume that each $y_i$ is
connected to all the other $y$'s. For example, if we didn't connect
$y_1$ and $y_4$ we would assert the additional independency $y_1
\independent y_4 \mid x_1,x_2,x_3,x_4,y_2,y_3$.

We thus have a graph as follows:

\begin{center}
  \scalebox{1}{
    \begin{tikzpicture}[ugraph]
      \node[cont] (y1) at (0,0) {$y_1$};
      \node[cont] (y2) at (2,0) {$y_2$};
      \node[cont] (y3) at (4,0) {$y_3$};
      \node[cont] (y4) at (6,0) {$y_4$};
      \node[cont] (x1) at (0,2) {$x_1$};
      \node[cont] (x2) at (2,2) {$x_2$};
      \node[cont] (x3) at (4,2) {$x_3$};
      \node[cont] (x4) at (6,2) {$x_4$};
      \draw(x1)--(y1);\draw(x2)--(y2);\draw(x3)--(y3);\draw(x4)--(y4);
      \draw(x1)--(x2);\draw(x2)--(x3);\draw(x3)--(x4);
      \draw(y1)--(y2);\draw(y1) to [bend right= 45] (y3);\draw(y1)  to [bend right= 45] (y4);
      \draw(y2)--(y3);\draw(y2) to [bend right=45]  (y4);\draw(y3)--(y4);   
  \end{tikzpicture}}
\end{center}

The factorisation thus is
\begin{align}
  p(x_1, \ldots, x_4, y_1, \ldots, y_4) = \frac{1}{Z} g(y_1, \ldots, y_4) \prod_{i=1}^3 m_i(x_i,x_{i+1})\prod_{i=1}^4 g_i(x_i,y_i),
\end{align}
where the $m_i(x_i, x_{i+1})$, $g_i(x_i, y_i)$ and $g(y_1, \ldots, y_4)$ are positive
factors. Compared to the factorisation in Exercise
\ref{ex:factorisation-from-the-Markov-blankets-I}, we still have the
Markov structure for the $x_i$, but only a single factor for $(y_1,
y_2, y_3, y_4)$ to avoid inserting independencies beyond those
specified by the given Markov blankets.

\end{solution}


\ex{Undirected graphical model with pairwise potentials}
We here consider Gibbs distributions where the factors only depend on two variables at
a time. The probability density or mass functions over $d$ random
variables $x_1, \ldots, x_d$ then take the form
$$p(x_1, \ldots, x_d) \propto \prod_{i \le j} \phi_{ij}(x_i,x_j)$$
Such models are sometimes called pairwise Markov networks.

\begin{exenumerate} 
  
\item Let $p(x_1, \ldots, x_d) \propto \exp\left( -\frac{1}{2} \x^\top
  \A \x - \b^\top \x \right)$ where $\A$ is symmetric and $\x = (x_1,
  \ldots, x_d)^\top$. What are the corresponding factors $\phi_{ij}$
  for $i\le j$?
  \begin{solution}
    Denote the $(i,j)$-th element of $\A$ by $a_{ij}$. We have
    \begin{align}
      \x^\top \A \x & = \sum_{ij} a_{ij} x_i x_j \\
      &= \sum_{i<j} 2 a_{ij} x_i x_j + \sum_i a_{ii} x_i^2
    \end{align}
    where the second line follows from $\A^\top =\A$. Hence,
    \begin{align}
      -\frac{1}{2} \x^\top \A \x - \b^\top \x & = -\frac{1}{2} \sum_{i<j} 2a_{ij} x_i x_j - \frac{1}{2} \sum_i a_{ii} x_i^2 - \sum_i b_i x_i
    \end{align}
    so that
\begin{equation}
      \phi_{ij}(x_i,x_j) = \begin{cases}
        \exp\left(- a_{ij} x_i x_j\right) & \text{if } i < j\\
        \exp\left( - \frac{1}{2} a_{ii} x_i^2 - b_i x_i\right) & \text{if } i=j
      \end{cases}
      \label{eq:pairwisepot}
\end{equation}
For $\x \in \mathbb{R}^d$, the distribution is a Gaussian with $\A$
equal to the inverse covariance matrix. For binary $\x$, the model is
known as Ising model or Boltzmann machine. For $x_i \in \{-1,1\}$,
$x_i^2=1$ for all $i$, so that the $a_{ii}$ are constants
that can be absorbed into the normalisation constant. This means that
for $x_i \in \{-1,1\}$, we can work with matrices $\A$ that have zeros
on the diagonal.

  \end{solution}

\item For $p(x_1, \ldots, x_d) \propto \exp\left( -\frac{1}{2} \x^\top
  \A \x - \b^\top \x \right)$, show that $x_i \independent x_j \mid
  \{x_1, \ldots, x_d\} \setminus \{x_i,x_j\}$ if the $(i,j)$-th
  element of $\A$ is zero.

  \begin{solution}

    The previous question showed that we can write $p(x_1, \ldots,
    x_d) \propto \prod_{i \leq j} \phi_{ij}(x_i,x_j)$ with potentials as in
    Equation \eqref{eq:pairwisepot}.  Consider two variables $x_i$ and
    $x_j$ for fixed $(i,j)$. They only appear in the factorisation via
    the potential $\phi_{ij}$. If $a_{ij} = 0$, the factor $\phi_{ij}$
    becomes a constant, and no other factor contains $x_i$ and $x_j$,
    which means that there is no edge between $x_i$ and $x_j$ if
    $a_{ij}=0$. By the pairwise Markov property it then follows that
    $x_i \independent x_j \mid \{x_1, \ldots, x_d\} \setminus \{x_i,x_j\}$.

  \end{solution}
  
\end{exenumerate}


\ex{Restricted Boltzmann machine {\small \citep[based on][Exercise 4.4]{Barber2012}}}%
The restricted Boltzmann machine is an undirected
graphical model for binary variables $\v =(v_1, \ldots, v_n)^\top$ and
$\h=(h_1, \ldots, h_m)^\top$ with a probability mass function equal to
\begin{equation}
  p(\v,\h) \propto \exp\left( \v^\top \W \h + \a^\top\v + \b^\top\h \right),
\end{equation}
where $\W$ is a $n \times m$ matrix. Both the $v_i$ and $h_i$ take values in $\{0,1\}$. The
$v_i$ are called the ``visibles'' variables since they are assumed to
be observed while the $h_i$ are the hidden variables since it is
assumed that we cannot measure them.

\begin{exenumerate}
\item Use graph separation to show that the joint conditional $p(\h | \v)$ factorises as
  $$p(\h | \v) = \prod_{i=1}^m p(h_i | \v).$$

  \begin{solution}
    Figure \ref{fig:restricted-boltzmann-machine} on the left shows
    the undirected graph for $p(\v,\h)$ with $n=3, m=2$. We note that
    the graph is bi-partite: there are only direct connections
    between the $h_i$ and the $v_i$. Conditioning on $\v$ thus blocks
    all trails between the $h_i$ (graph on the right). This means that
    the $h_i$ are independent from each other given $\v$ so that
    $$p(\h | \v) = \prod_{i=1}^m p(h_i | \v).$$

    \begin{figure}[ht]
      \centering
      \scalebox{1}{
        \begin{tikzpicture}[ugraph]
          
          \node[cont] (h1) at (-1,0) {$h_1$};
          \node[cont] (h2) at (1,0) {$h_2$};
          \node[cont] (v1) at (-2,-2) {$v_1$};
          \node[cont] (v2) at (0,-2) {$v_2$};
          \node[cont] (v3) at (2,-2) {$v_3$};
          
          \draw(h1) -- (v1);
          \draw(h1) -- (v2);
          \draw(h1) -- (v3);
          \draw(h2) -- (v1);
          \draw(h2) -- (v2);
          \draw(h2) -- (v3);
          
          
          \node[cont] (h1) at (7,0) {$h_1$};
          \node[cont] (h2) at (9,0) {$h_2$};
          \node[contobs] (v1) at (6,-2) {$v_1$};
          \node[contobs] (v2) at (8,-2) {$v_2$};
          \node[contobs] (v3) at (10,-2) {$v_3$};
          
          \draw(h1) -- (v1);
          \draw(h1) -- (v2);
          \draw(h1) -- (v3);
          \draw(h2) -- (v1);
          \draw(h2) -- (v2);
          \draw(h2) -- (v3);
          
      \end{tikzpicture}}
      \caption{\label{fig:restricted-boltzmann-machine} Left: Graph for $p(\v,\h)$. Right: Graph for $p(\h | \v)$}
    \end{figure}
  \end{solution}

\item Show that
  \begin{equation}
    p(h_i = 1 | \v) = \frac{1}{1+\exp\left(-b_i-\sum_j W_{ji} v_j\right)}
    \label{eq:hcondv}
  \end{equation}
where $W_{ji}$ is the $(ji)$-th element of $\W$, so that $\sum_j W_{ji} v_j$ is the inner product (scalar product) between the $i$-th column of $\W$ and $\v$.
    \begin{solution}

     For the conditional pmf $p(h_i | \v)$ any quantity that does not
     depend on $h_i$ can be considered to be part of the normalisation constant. A general strategy is to first work out $p(h_i | \v)$ up to the normalisation constant and then to normalise it afterwards.

     We begin with $p(\h | \v)$:
      \begin{align}
        p(\h | \v) &= \frac{p(\h,\v)}{p(\v)} \\
        & \propto p(\h,\v) \\
        & \propto \exp\left( \v^\top \W \h + \a^\top\v + \b^\top\h \right)\\
        & \propto \exp\left( \v^\top \W \h + \b^\top\h \right)\\
        & \propto \exp\left( \sum_i \sum_j v_j W_{ji} h_i + \sum_i b_i h_i \right) 
     \end{align}
     As we are interested in $p(h_i | \v)$ for a fixed $i$, we can drop all the terms not depending on that $h_i$, so that
      \begin{align}
        p(h_i | \v) & \propto \exp\left( \sum_j v_j W_{ji} h_i + b_i h_i \right)
      \end{align}
      Since $h_i$ only takes two values, 0 and 1, normalisation is here straightforward. Call the unnormalised pmf $\tilde{p}(h_i | \v)$,
      \begin{equation}
        \tilde{p}(h_i | \v) = \exp\left( \sum_j v_j W_{ji} h_i + b_i h_i \right).
      \end{equation}
      We then have
      \begin{align}
        p(h_i | \v) &= \frac{\tilde{p}(h_i | \v)} { \tilde{p}(h_i=0 | \v) + \tilde{p}(h_i =1 | \v)}\\
        & = \frac{\tilde{p}(h_i | \v)} { 1+  \exp\left( \sum_j v_j W_{ji} + b_i \right)}\\
        & = \frac{\exp\left( \sum_j v_j W_{ji} h_i + b_i h_i \right)}{ 1+  \exp\left( \sum_j v_j W_{ji} + b_i \right)},
       \end{align}
      so that
      \begin{align}
        p(h_i=1 | \v) & =  \frac{\exp\left( \sum_j v_j W_{ji} + b_i \right)}{ 1+  \exp\left( \sum_j v_j W_{ji} + b_i \right)}\\
         & =  \frac{1}{ 1+  \exp\left(- \sum_j v_j W_{ji} - b_i \right)}.
      \end{align}
      The probability $p(h=0 | \v)$ equals $1 -  p(h_i=1 | \v)$, which is
      \begin{align}
        p(h_i =0 | \v) & =  \frac{1+ \exp\left( \sum_j v_j W_{ji} + b_i \right)}{ 1+  \exp\left( \sum_j v_j W_{ji} + b_i \right)} - \frac{\exp\left( \sum_j v_j W_{ji} + b_i \right)}{ 1+  \exp\left( \sum_j v_j W_{ji} + b_i \right)}\\
        & = \frac{1}{ 1+  \exp\left( \sum_j W_{ji} v_j + b_i \right)}
        \end{align}

      The function $x \mapsto 1/(1+\exp(-x))$ is called the logistic function. It is a sigmoid function and is thus sometimes denoted by $\sigma(x)$.
      For other versions of the sigmoid function, see \url{https://en.wikipedia.org/wiki/Sigmoid_function}.
      \begin{center}
        \vspace{5ex}
        \begin{tikzpicture}
          \begin{axis}[ xlabel={$x$}, ylabel={$\sigma(x)$},
              axis lines=middle, smooth , grid , thick, domain=-6:6]
            \addplot[no marks]{1/(1+exp(-\x))};
            \end{axis}
        \end{tikzpicture}
        \vspace{2ex}
      \end{center}
      With that notation, we have
      \begin{align*}
        p(h_i=1 | \v) &= \sigma\left(\sum_j W_{ji} v_j  + b_i\right).
      \end{align*}
      
    \end{solution}

  \item Use a symmetry argument to show that
    $$p(\v | \h) = \prod_i p(v_i | \h) \quad \text{ and }  \quad p(v_i = 1 | \h) = \frac{1}{1+\exp\left(-a_i-\sum_j W_{ij} h_j\right)}$$

    \begin{solution}
      Since $\v^\top \W \h$ is a scalar we have  $(\v^\top \W \h)^\top = \h^\top \W^\top \v =  \v^\top \W \h$, so that  
      \begin{align}
      p(\v,\h) & \propto \exp\left( \v^\top \W \h + \a^\top\v + \b^\top\h \right) \\
      & \propto \exp\left( \h^\top \W^\top \v + \b^\top\h + \a^\top\v  \right).
      \end{align}
      To derive the result, we note that $\v$ and $a$ now take the place of $\h$ and $\b$ from before, and that we now have $\W^\top$ rather than $\W$. In Equation \eqref{eq:hcondv}, we thus replace $h_i$ with $v_i$, $b_i$ with $a_i$, and $W_{ji}$ with $W_{ij}$ to obtain $p(v_i = 1 | \h)$. In terms of the sigmoid function, we have
       $$ p(v_i=1 | \h) = \sigma\left(\sum_j W_{ij} h_j + a_i\right).$$

      Note that while $p(\v | \h)$ factorises, the marginal $p(\v)$
      does generally not. The marginal $p(\v)$ can here be obtained in
      closed form up to its normalisation constant.
      \begin{align}
        p(\v) & = \sum_{\h \in \{0,1\}^m} p(\v,\h) \\
        & = \frac{1}{Z} \sum_{\h \in \{0,1\}^m} \exp\left( \v^\top \W \h + \a^\top\v + \b^\top\h \right)\\
        & =  \frac{1}{Z}  \sum_{\h \in \{0,1\}^m} \exp\left( \sum_{ij} v_i h_j W_{ij} + \sum_i a_i v_i + \sum_j b_j h_j \right)\\
        & =  \frac{1}{Z}  \sum_{\h \in \{0,1\}^m} \exp\left( \sum_{j=1}^m h_j \left[\sum_i v_i W_{ij} + b_j\right] + \sum_i a_i v_i \right)\\
        & =  \frac{1}{Z}  \sum_{\h \in \{0,1\}^m} \prod_{j=1}^m \exp\left( h_j \left[\sum_i v_i W_{ij} + b_j\right] \right) \exp\left( \sum_i a_i v_i \right)\\
        & =  \frac{1}{Z} \exp\left( \sum_i a_i v_i \right) \sum_{\h \in \{0,1\}^m} \prod_{j=1}^m \exp\left( h_j \left[\sum_i v_i W_{ij} + b_j\right] \right) \\
        & =  \frac{1}{Z} \exp\left( \sum_i a_i v_i \right) \sum_{h_1, \ldots, h_m} \prod_{j=1}^m  \exp\left( h_j \left[\sum_i v_i W_{ij} + b_j\right] \right)
      \end{align}
      Importantly, each term in the product only depends on a single $h_j$, so that by sequentially applying the distributive law, we have
      \begin{align}
        \sum_{h_1, \ldots, h_m} \prod_{j=1}^m  \exp\left( h_j \left[\sum_i v_i W_{ij} + b_j\right] \right)  =& \left[ \sum_{h_1, \ldots, h_{m-1}} \prod_{j=1}^{m-1}  \exp\left( h_j \left[\sum_i v_i W_{ij} + b_j\right] \right)\right] \cdot \nonumber \\
        &\sum_{h_m} \exp\left( h_m \left[\sum_i v_i W_{im} + b_m\right] \right)\\
         =& \ldots \nonumber \\
        =& \prod_{j=1}^m \left[\sum_{h_j} \exp\left( h_j \left[\sum_i v_i W_{ij} + b_j\right] \right)\right]
      \end{align}
      Since $h_j \in \{0,1\}$, we obtain
      \begin{align}
        \sum_{h_j} \exp\left( h_j \left[\sum_i v_i W_{ij} + b_j\right] \right) & = 1+ \exp\left(\sum_i v_i W_{ij} + b_j \right)
      \end{align}
      and thus
       \begin{align}
         p(\v) & = \frac{1}{Z}  \exp\left( \sum_i a_i v_i \right) \prod_{j=1}^m \left[1+ \exp\left(\sum_i v_i W_{ij} + b_j \right)\right].
      \end{align}
       Note that in the derivation of $p(\v)$ we have not used the
       assumption that the visibles $v_i$ are binary. The same
       expression would thus obtained if the visibles were defined in
       another space, e.g.\ the real numbers.
       
       While $p(\v)$ is written as a product, $p(\v)$ does not
       factorise into terms that depend on subsets of the
       $v_i$. On the contrary, all $v_i$ are present in all
       factors. Since $p(\v)$ does not factorise, computing the
       normalising $Z$ is expensive. For binary visibles
       $v_i \in \{0,1\}$, $Z$ equals
       \begin{equation}
         Z = \sum_{\v \in \{0,1\}^n} \exp\left( \sum_i a_i v_i \right) \prod_{j=1}^m \left[1+ \exp\left(\sum_i v_i W_{ij} + b_j \right)\right]
       \end{equation}
       where we have to sum over all $2^n$ configurations of the
       visibles $\v$. This is computationally expensive, or even
       prohibitive if $n$ is large ($2^{20} = 1048576,\, 2^{30}>
       10^9$). Note that different values of $a_i,b_i,W_{ij}$ yield
       different values of $Z$. (This is a reason why $Z$ is called
       the partition \emph{function} when the $a_i, b_i, W_{ij}$ are free parameters.)

       It is instructive to write $p(\v)$ in the log-domain,
       \begin{align}
         \log p(\v) & = \log Z + \sum_{i=1}^n a_i v_i + \sum_{j=1}^m \log\left[1+ \exp\left(\sum_i v_i W_{ij} + b_j \right)\right],
       \end{align}
       and to introduce the nonlinearity $f(u)$,
       \begin{align}
         f(u) & = \log\left[ 1+\exp(u) \right],
       \end{align}
       which is called the softplus function and plotted below. The softplus function is a smooth approximation of $\max(0,u)$, see e.g.\ \url{https://en.wikipedia.org/wiki/Rectifier_(neural_networks)}
       \begin{center}
         \vspace{3ex}
         \begin{tikzpicture}
           \begin{axis}[ xlabel={$u$}, ylabel={$f(u)$},
               axis lines=middle, smooth , grid , thick, domain=-6:6]
             \addplot[no marks]{ln(1+exp(\x))};
           \end{axis}
         \end{tikzpicture}
      \end{center}
       With the softplus function $f(u)$, we can write $\log p(\v)$ as
       \begin{align}
         \log p(\v) & = \log Z + \sum_{i=1}^n a_i v_i + \sum_{j=1}^m f\left(\sum_i v_i W_{ij} + b_j \right).
       \end{align}
       The parameter $b_j$ plays the role of a threshold as shown in
       the figure below. The terms $f\left(\sum_i v_i W_{ij} + b_j
       \right)$ can be interpreted in terms of feature detection. The
       sum $\sum_i v_i W_{ij}$ is the inner product between $\v$ and
       the $j$-th column of $\W$, and the inner product is largest if
       $\v$ equals the $j$-th column. We can thus consider the columns
       of $\W$ to be feature-templates, and the $f\left(\sum_i v_i
       W_{ij} + b_j \right)$ a way to measure how much of each feature
       is present in $\v$.

       Further, $\sum_i v_i W_{ij} +b_j$ is also the input to the
       sigmoid function when computing $p(h_j=1 | \v)$. Thus, the
       conditional probability for $h_j$ to be one, i.e.\ ``active'',
       can be considered to be an indicator of the presence of the
       $j$-th feature ($j$-th column of $\W$) in the input $\v$.

       If $v$ is such that $\sum_i v_i W_{ij} +b_j$ is large for many
       $j$, i.e.\ if many features are detected, then $f\left(\sum_i
       v_i W_{ij} + b_j \right)$ will be non-zero for many $j$, and
       $\log p(\v)$ will be large.

       \begin{center}
         \vspace{3ex}
         \begin{tikzpicture}
           \begin{axis}[ xlabel={$u$}, ylabel={$f(u)$},
               axis lines=middle, smooth , grid , thick, domain=-6:6]
             \addplot[no marks]{ln(1+exp(\x))} node[left]{\scriptsize$f(u)$};
             \addplot[no marks, dashed]{ln(1+exp(\x+2))} node[below left]{\scriptsize $f(u+2)$};
             \addplot[no marks, dashed]{ln(1+exp(\x-2))} node[below left]{\scriptsize $f(u-2)$};
           \end{axis}
         \end{tikzpicture}
      \end{center}

    \end{solution}

\end{exenumerate}

\ex{Hidden Markov models and change of measure}

Consider the following undirected graph for a hidden Markov model where the $y_i$ correspond to observed (visible) variables and the $x_i$ to unobserved (hidden/latent) variables.

\begin{center}
  \scalebox{1}{
    \begin{tikzpicture}[ugraph]
      \node[cont] (x1) at (0,2) {$x_1$};
      \node[cont] (x2) at (2,2) {$x_2$};
      \node[cont] (x3) at (4,2) {$x_3$};
      \node       (a) at (6,2) {};
      \node       (b) at (7,2) {$\ldots$};
      \node       (c) at (7,0) {$\ldots$};
      \node       (d) at (8,2) {};
      \node[cont] (x4) at (10,2) {$x_t$};
      \node[cont] (y1) at (0,0) {$y_1$};
      \node[cont] (y2) at (2,0) {$y_2$};
      \node[cont] (y3) at (4,0) {$y_3$};
      \node[cont] (y4) at (10,0) {$y_t$};
      \draw(x1)--(y1);\draw(x2)--(y2);\draw(x3)--(y3);\draw(x4)--(y4);
      \draw(x1)--(x2);\draw(x2)--(x3);\draw(x3)--(a);\draw(d)--(x4);
  \end{tikzpicture}}
\end{center}

The graph implies the following factorisation
\begin{equation}
  p( x_1, \ldots, x_t,y_1, \ldots, y_t) \propto \phi_1^y(x_1, y_1) \prod_{i=2}^t \phi_i^x(x_{i-1}, x_i) \phi_i^y(x_i, y_i),
\end{equation}
where  the $\phi_i^x$ and $\phi_i^y$ are non-negative factors.

Let us consider the situation where $\prod_{i=2}^t \phi^x_i(x_{i-1}, x_i)$ equals
\begin{equation}
  f(\x) = \prod_{i=2}^t \phi^x_i(x_{i-1}, x_i) = f_1(x_1) \prod_{i=2}^t f_i(x_i|x_{i-1}),
\end{equation}
with $\x=(x_1, \ldots, x_t)$ and where the $f_i$ are (conditional) pdfs. We thus have
\begin{align}
  p(x_1, \ldots, x_t, y_1, \ldots, y_t) \propto f_1(x_1) \prod_{i=2}^t f_i(x_i|x_{i-1}) \prod_{i=1}^t \phi_i^y(x_i, y_i).
  \label{eq:Markov-model-joint-def} 
\end{align}

\begin{exenumerate}
  
\item Provide a factorised expression for $ p(x_1, \ldots, x_t | y_1, \ldots, y_t)$

  \begin{solution}
    For fixed (observed) values of the $y_i$, $p(x_1, \ldots, x_t |
    y_1, \ldots, y_t)$ factorises as
    \begin{align}
      p(x_1, \ldots, x_t | y_1, \ldots, y_t) \propto f_1(x_1) g_1(x_1)
      \prod_{i=1}^t f_i(x_i|x_{i-1}) g_i(x_i).
    \end{align}
    where $g_i(x_i)$ is $\phi^y_i(x_i,y_i)$ for a fixed value of $y_i$. 
  \end{solution}

\item Draw the undirected graph for $p(x_1, \ldots, x_t | y_1, \ldots, y_t)$
  \begin{solution}
    Conditioning corresponds to removing nodes from an undirected
    graph. We thus have the following Markov chain for $p(x_1, \ldots,
    x_t | y_1, \ldots, y_t)$.

\begin{center}
  \scalebox{1}{
\begin{tikzpicture}[ugraph]
      \node[cont] (x1) at (0,0) {$x_1$};
      \node[cont] (x2) at (2,0) {$x_2$};
      \node[cont] (x3) at (4,0) {$x_3$};
      \node       (a) at (6,0) {};
      \node       (b) at (7,0) {$\ldots$};
      \node       (d) at (8,0) {};
      \node[cont] (x4) at (10,0) {$x_t$};
      \draw(x1)--(x2);\draw(x2)--(x3);\draw(x3)--(a);\draw(d)--(x4);
  \end{tikzpicture}
  }
\end{center}
   
 \end{solution}

\item Show that if $\phi_i^y(x_i, y_i)$ equals the conditional pdf of
  $y_i$ given $x_i$, i.e.\ $p(y_i|x_i)$, the marginal $p(x_1, \ldots, x_t)$,
  obtained by integrating out $y_1, \ldots, y_t$ from
  \eqref{eq:Markov-model-joint-def}, equals $f(\x)$.

  \begin{solution}
    In this setting all factors in \eqref{eq:Markov-model-joint-def}
    are conditional pdfs and we are dealing with a directed graphical
    model that factorises as
    \begin{equation}
      p(x_1, \ldots, x_t, y_1, \ldots, y_t) = f_1(x_1) \prod_{i=2}^t f_i(x_i|x_{i-1}) \prod_{i=1}^t p(y_i|x_i).
    \end{equation}
    By integrating over the $y_i$, we have
    \begin{align}
      p(x_1, \ldots, x_t) &= \int p(x_1, \ldots, x_t, y_1, \ldots, y_t) \ud y_1 \ldots \ud y_t\\
      & = f_1(x_1) \prod_{i=2}^t f_i(x_i|x_{i-1}) \int \prod_{i=1}^t p(y_i|x_i) \ud y_1 \ldots \ud y_t\\
      & = f_1(x_1) \prod_{i=2}^t f_i(x_i|x_{i-1}) \prod_{i=1}^t \underbrace{\int  p(y_i|x_i) \ud y_i}_{1}\\
      & = f_1(x_1) \prod_{i=2}^t f_i(x_i|x_{i-1})\\
      & = f(\x)
    \end{align}
    
  \end{solution}

\item Compute the normalising constant for $p(x_1, \ldots, x_t | y_1, \ldots, y_t)$ and express it as an expectation over $f(\x)$.
  \begin{solution}
    With
    \begin{align}
      p(x_1, \ldots, x_t, y_1, \ldots, y_t) \propto f_1(x_1) \prod_{2=1}^t f_i(x_i|x_{i-1}) \prod_{i=1}^t \phi_i^y(x_i, y_i).
    \end{align}
    The normalising constant is given by
    \begin{align}
      Z & = \int f_1(x_1) \prod_{2=1}^t f_i(x_i|x_{i-1})\prod_{i=1}^t g_i(x_i) \ud x_1 \ldots \ud x_t\\
      & = \E_f\left [\prod_{i=1}^t g_i(x_i) \right]
    \end{align}
    Since we can use ancestral sampling to sample from $f$, the
    above expectation can be easily computed via sampling.    
    
  \end{solution}

\item Express the expectation of a test function $h(\x)$ with respect
  to $p(x_1, \ldots, x_t | y_1, \ldots, y_t)$ as a reweighted
  expectation with respect to $f(\x)$.

  \begin{solution}

    By definition, the expectation over a test function $h(\x)$ is
    \begin{align}
      \E_{p(x_1, \ldots, x_t | y_1,
        \ldots, y_t)}[ h(\x) ] & = \frac{1}{Z} \int h(\x) f_1(x_1) \prod_{2=1}^t f(x_i|x_{i-1})\prod_{i=1}^t g_i(x_i) \ud x_1 \ldots \ud x_t\\
      & = \frac{\E_f \left[ h(\x) \prod_i g_i(x_i) \right]}{\E_f\left [\prod_i g_i(x_i) \right]}
    \end{align}
    Both the numerator and denominator can be approximated using
    samples from $f$.

    Since the $g_i(x_i) = \phi_i^y(x_i,y_i)$ involve the observed
    variables $y_i$, this has a nice interpretation: We can think we
    have two models for $\x$: $f(\x)$ that does not involve the
    observations and $p(x_1, \ldots, x_t | y_1,\ldots, y_t)$ that
    does. Note, however, that unless $\phi_i^y(x_i, y_i)$ is the
    conditional pdf $p(y_i|x_i)$, $f(\x)$ is \emph{not} the marginal
    $p(x_1, \ldots, x_t)$ that you would obtain by integrating out the
    $y$'s from the joint model . We can thus generally think it is a
    base distribution that got ``enhanced'' by a change of measure in our expression for
    $p(x_1, \ldots, x_t| y_1,\ldots, y_t)$. If $\phi_i^y(x_i, y_i)$ is
    the conditional pdf $p(y_i|x_i)$, the change of measure
    corresponds to going from the prior to the posterior by
    multiplication with the likelihood (the terms $g_i$).

    From the expression for the expectation, we can see that the
    ``enhancing'' leads to a corresponding introduction of weights in
    the expectation that depend via $g_i$ on the observations. This
    can be particularly well seen when we approximate the expectation as
    a sample average over $n$ samples $\x^{(k)} \sim f(\x)$:
    \begin{align}
      \E_{p(x_1, \ldots, x_t | y_1, \ldots, y_t)}[ h(\x) ] & \approx \sum_{k=1}^n W^{(k)} h(\x^{(k)})\\
      W^{(k)} & =  \frac{w^{(k)}}{\sum_{k=1}^n  w^{(k)}}\\
      w^{(k)} & =  \prod_i g_i(x^{(k)}_i)
    \end{align}
    where $x^{(k)}_i$ is the $i$-th dimension of the vector $\x^{(k)}$.
  \end{solution}
  
\end{exenumerate}

\chapter{Expressive Power of Graphical Models} 
\minitoc

\ex{I-equivalence}
\label{ex:I-equiv}

\begin{exenumerate}
\item Which of three graphs represent the same set of independencies? Explain.\\

  \begin{tikzpicture}[dgraph]
    \node[cont] (v) at (0,0) {$v$};
    \node[cont, below right= of v] (w) {$w$};
    \node[cont, right= of v] (x) {$x$};
    \node[cont, below right= of x] (y) {$y$};
    \node[cont, right= of x] (z) {$z$};

    \node[below= 0.2 of w] {Graph 1};

    \draw (v) -- (w);
    \draw (w) -- (x);
    \draw (x) -- (y);
    \draw (z) -- (y);   
  \end{tikzpicture}
  \hspace{5ex}
  \begin{tikzpicture}[dgraph]
    \node[cont] (v) at (0,0) {$v$};
    \node[cont, below right= of v] (w) {$w$};
    \node[cont, right= of v] (x) {$x$};
    \node[cont, below right= of x] (y) {$y$};
    \node[cont, right= of x] (z) {$z$};

    \node[below= 0.2 of w] {Graph 2};
    
    \draw (w) -- (v);
    \draw (x) -- (w);
    \draw (x) -- (y);
    \draw (z) -- (y);   
  \end{tikzpicture}
 \hspace{5ex}
  \begin{tikzpicture}[dgraph]
    \node[cont] (v) at (0,0) {$v$};
    \node[cont, below right= of v] (w) {$w$};
    \node[cont, right= of v] (x) {$x$};
    \node[cont, below right= of x] (y) {$y$};
    \node[cont, right= of x] (z) {$z$};

    \node[below= 0.2 of w] {Graph 3};
    
    \draw (w) -- (v);
    \draw (x) -- (w);
    \draw (y) -- (x);
    \draw (z) -- (y);

  \end{tikzpicture}

  \begin{solution}
    To check whether the graphs are I-equivalent, we have to check the skeletons and the immoralities. All have the same skeleton, but graph 1 and graph 2 also have the same immorality. The answer is thus: graph 1 and 2 encode the same independencies.\\
    \begin{center}
      \begin{tikzpicture}[ugraph]
        \node[cont] (v) at (0,0) {$v$};
        \node[cont, below right= of v] (w) {$w$};
        \node[cont, right= of v] (x) {$x$};
        \node[cont, below right= of x] (y) {$y$};
        \node[cont, right= of x] (z) {$z$};
        
        \node[below= 0.2 of w] {skeleton};
        
        \draw (w) -- (v);
        \draw (x) -- (w);
        \draw (y) -- (x);
        \draw (z) -- (y);
        
      \end{tikzpicture}
      \hspace{4ex}
      \begin{tikzpicture}[ugraph]
        \node[cont] (v) at (0,0) {$v$};
        \node[cont, below right= of v] (w) {$w$};
        \node[cont, right= of v] (x) {$x$};
        \node[cont, below right= of x] (y) {$y$};
        \node[cont, right= of x] (z) {$z$};
        
        \node[below= 0.2 of w] {immorality};
        
        \draw (w) -- (v);
        \draw (x) -- (w);
        \draw[->, red] (x) -- (y);
        \draw[->,red] (z) -- (y);
        
      \end{tikzpicture}
    \end{center}
    \end{solution}

\item Which of three graphs represent the same set of independencies? Explain.\\

  \begin{tikzpicture}[dgraph]
    \node[cont] (v) at (0,0) {$v$};
    \node[cont, right= of v] (x) {$x$};
    \node[cont, below = of x] (w) {$w$};
    \node[cont, right= of w] (y) {$y$};
    \node[cont, below= of y] (z) {$z$};

    \node[below= 2 of w] {Graph 1};

    \draw (v) -- (x);
    \draw (v) -- (w);
    \draw (x) -- (w);
    \draw (w) -- (z);
    \draw (y) -- (z);   
  \end{tikzpicture}
  \hspace{5ex}
  \begin{tikzpicture}[dgraph]
    \node[cont] (v) at (0,0) {$v$};
    \node[cont, right= of v] (x) {$x$};
    \node[cont, below = of x] (w) {$w$};
    \node[cont, right= of w] (y) {$y$};
    \node[cont, below= of y] (z) {$z$};

    \node[below= 2 of w] {Graph 2};

    \draw (x) -- (v);
    \draw (w) -- (v);
    \draw (x) -- (w);
    \draw (w) -- (z);
    \draw (y) -- (z);   
  \end{tikzpicture}
  \hspace{5ex}
  \begin{tikzpicture}[dgraph]
    \node[cont] (v) at (0,0) {$v$};
    \node[cont, right= of v] (x) {$x$};
    \node[cont, below = of x] (w) {$w$};
    \node[cont, right= of w] (y) {$y$};
    \node[cont, below= of y] (z) {$z$};
    
    \node[below= 2 of w] {Graph 3};
    
    \draw (v) -- (w);
    \draw (x) -- (w);
    \draw (w) -- (z);
    \draw (y) -- (z);   
  \end{tikzpicture}
  \hspace{5ex}
  
  \begin{solution}
    The skeleton of graph 3 is different from the skeleton of graphs 1
    and 2, so that graph 3 cannot be I-equivalent to graph 1 or 2, and
    we do not need to further check the immoralities for graph 3. Graph 1 and 2
    have the same skeleton, and they also have the same
    immorality. Hence, graph 1 and 2 are I-equivalent. Note that node
    $w$ in graph 1 is in a collider configuration along trail $v-w-x$
    but it is not an immorality because its parents are connected
    (covering edge); equivalently for node $v$ in graph 2.
    
    \begin{center}
      \begin{tikzpicture}[ugraph]
        \node[cont] (v) at (0,0) {$v$};
        \node[cont, right= of v] (x) {$x$};
        \node[cont, below = of x] (w) {$w$};
        \node[cont, right= of w] (y) {$y$};
        \node[cont, below= of y] (z) {$z$};
        
        \node[below= 2 of w] {skeleton};
        
        \draw (v) -- (x);
        \draw (v) -- (w);
        \draw (x) -- (w);
        \draw (w) -- (z);
        \draw (y) -- (z);   
      \end{tikzpicture}
      \hspace{4ex}
      \begin{tikzpicture}[ugraph]
        \node[cont] (v) at (0,0) {$v$};
        \node[cont, right= of v] (x) {$x$};
        \node[cont, below = of x] (w) {$w$};
        \node[cont, right= of w] (y) {$y$};
        \node[cont, below= of y] (z) {$z$};
        
        \node[below= 2 of w] {immorality};
        
        \draw (v) -- (x);
        \draw (v) -- (w);
        \draw (x) -- (w);
        \draw[->, red] (w) -- (z);
        \draw[->, red] (y) -- (z);   
      \end{tikzpicture}
    \end{center}
    \end{solution}

\item Assume the graph below is a perfect map for a set of independencies $\mathcal{U}$.

\begin{center}
  \begin{tikzpicture}[dgraph]
    \node[cont] (x1) at (0,0) {$x_1$};
    \node[cont, right= of x1] (x2) {$x_2$};
    \node[cont, below = of x1] (x3) {$x_3$};
    \node[cont, right= of x3] (x4) {$x_4$};
    \node[cont, right= of x4] (x5) {$x_5$};
    \node[cont, below= of x3] (x6) {$x_6$};
    \node[cont, below= of x4] (x7) {$x_7$};
    
    \node[below= 0.2 of x6] {Graph 0};

    \draw (x1) -- (x3);
    \draw (x2) -- (x4);
    \draw (x2) -- (x5);
    \draw (x3) -- (x6);
    \draw (x4) -- (x7);
    \draw (x5) -- (x7);
    \draw (x7) -- (x6);
    \draw (x3) -- (x7);
  \end{tikzpicture}
\end{center}

For each of the three graphs below, explain whether the graph is a perfect map, an I-map, or not an I-map for $\mathcal{U}$.

\begin{center}
  \begin{tikzpicture}[dgraph]
    \node[cont] (x1) at (0,0) {$x_1$};
    \node[cont, right= of x1] (x2) {$x_2$};
    \node[cont, below = of x1] (x3) {$x_3$};
    \node[cont, right= of x3] (x4) {$x_4$};
    \node[cont, right= of x4] (x5) {$x_5$};
    \node[cont, below= of x3] (x6) {$x_6$};
    \node[cont, below= of x4] (x7) {$x_7$};
    
    \node[below= 0.2 of x6] {Graph 1};

    \draw (x1) -- (x3);
    \draw (x2) -- (x4);
    \draw (x2) -- (x5);
    \draw (x3) -- (x6);
    \draw (x4) -- (x7);
    \draw (x7) -- (x5);
    \draw (x7) -- (x6);
    \draw (x3) -- (x7);
  \end{tikzpicture}
  \begin{tikzpicture}[dgraph]
    \node[cont] (x1) at (0,0) {$x_1$};
    \node[cont, right= of x1] (x2) {$x_2$};
    \node[cont, below = of x1] (x3) {$x_3$};
    \node[cont, right= of x3] (x4) {$x_4$};
    \node[cont, right= of x4] (x5) {$x_5$};
    \node[cont, below= of x3] (x6) {$x_6$};
    \node[cont, below= of x4] (x7) {$x_7$};
    
    \node[below= 0.2 of x6] {Graph 2};
    
    \draw (x1) -- (x3);
    \draw (x2) -- (x4);
    \draw (x2) -- (x5);
    \draw (x3) -- (x6);
    \draw (x4) -- (x7);
    \draw (x5) -- (x7);
    \draw (x7) -- (x6);
    \draw (x7) -- (x3);
  \end{tikzpicture}
  \begin{tikzpicture}[dgraph]
    \node[cont] (x1) at (0,0) {$x_1$};
    \node[cont, right= of x1] (x2) {$x_2$};
    \node[cont, below = of x1] (x3) {$x_3$};
    \node[cont, right= of x3] (x4) {$x_4$};
    \node[cont, right= of x4] (x5) {$x_5$};
    \node[cont, below= of x3] (x6) {$x_6$};
    \node[cont, below= of x4] (x7) {$x_7$};
    
    \node[below= 0.2 of x6] {Graph 3};

    \draw (x1) -- (x3);
    \draw (x2) -- (x4);
    \draw (x5) -- (x2);
    \draw (x3) -- (x6);
    \draw (x4) -- (x7);
    \draw (x5) -- (x7);
    \draw (x7) -- (x6);
    \draw (x3) -- (x7);
  \end{tikzpicture}
\end{center}

\begin{solution}
  \begin{itemize}
    \item Graph 1 has an immorality $x_2 \rightarrow x_5
      \leftarrow x_7$ which graph 0 does not have. The graph is thus not I-equivalent to graph 0
      and can thus not be a perfect map. Moreover, graph 1 asserts
      that $x_2 \independent x_7 | x_4$ which is not case for graph
      0. Since graph 0 is a perfect map for $\mathcal{U}$, graph 1
      asserts an independency that does not hold for $\mathcal{U}$ and
      can thus not be an I-map for $\mathcal{U}.$
    \item Graph 2 has an immorality $x_1 \rightarrow x_3 \leftarrow x_7$ which graph 0 does not have. Graph 2 thus asserts that $x_1 \independent x_7$, which is not the case for graph 0. Hence, for the same reason as for graph 1, graph 2 is not an I-map for $\mathcal{U}$.
    \item Graph 3 has the same skeleton and set of immoralities as graph 0. It is thus I-equivalent to graph 0, and hence also a perfect map.
  \end{itemize}
\end{solution}

\end{exenumerate}

\ex{Minimal I-maps}
\label{ex:minimal-I-maps-1}
\begin{exenumerate}
\item Assume that the graph $G$ in Figure \ref{fig:I-map} is a perfect I-map for $p(a,z,q,e,h)$. Determine the minimal 
  directed I-map using the ordering $(e,h,q,z,a)$. Is the obtained graph I-equivalent to $G$?
  \begin{figure}[ht]
    \begin{center}
      \scalebox{0.9}{ 
        \begin{tikzpicture}[dgraph]
          \node[cont] (a) at (0,2) {$a$};
          \node[cont] (z) at (2,2) {$z$};
          \node[cont] (q) at (1,1) {$q$};
          \node[cont] (e) at (1,-0.4) {$e$};
          \node[cont] (h) at (3,1) {$h$};
          \draw(a) -- (q);
          \draw(z) -- (h);
          \draw(z) -- (q);
          \draw(q) -- (e);
      \end{tikzpicture}}
    \end{center}
    \caption{\label{fig:I-map} Perfect I-map $G$ for \exref{ex:minimal-I-maps-1}, question \ref{q:directed-I-maps}.}
    \end{figure}
\label{q:directed-I-maps}
  \begin{solution} Since the graph $G$ is a perfect I-map for $p$, we can use
    $G$ to check whether $p$ satisfies a certain
    independency. This gives the following
    recipe to construct the minimal directed I-map:
    \begin{enumerate}
    \item Assume an ordering of the
    variables. Denote the ordered random variables by $x_1, \ldots, x_d$.
    \item For each $i$, find a minimal subset of variables
    $\pap_i \subseteq \pre_i$ such that $$
    x_i \independent \{\pre_i \setminus \pap_i \} \mid \pap_i$$ is in
    $\Ind(G)$ (only works if $G$ is a perfect I-map for
    $\Ind(p)$)
    \item Construct a graph with parents $\pa_i=\pap_i$.
    \end{enumerate}
   
    Note: For I-maps $G$ that are not perfect, if the graph does not
    indicate that a certain independency holds, we have to check
    that the independency indeed does not hold for $p$. If we
    don't, we won't obtain a minimal I-map but just an I-map for
    $\Ind(p)$. This is because $p$ may have independencies that are
    not encoded in the graph $G$. 

    Given the ordering $(e,h,q,z,a)$, we build a graph where $e$ is the
    root. From Figure \ref{fig:I-map} (and the perfect map
    assumption), we see that $h \independent e$ does not hold. We
    thus set $e$ as parent of $h$, see first graph in Figure
    \ref{fig:I-map2}. Then:
    \begin{itemize}
      \item We consider $q$: $\pre_q = \{e,h\}$. There is no subset
        $\pap_q$ of $\pre_q$ on which we could condition to make $q$
        independent of $\pre_q \setminus \pap_q$, so that we set the
        parents of $q$ in the graph to $\pa_q = \{e,h\}$. (Second
        graph in Figure \ref{fig:I-map2}.)
      \item We consider $z$: $\pre_z = \{e,h,q\}$. From the graph in
        Figure \ref{fig:I-map}, we see that for $\pap_z = \{q,h\}$ we
        have $z \independent \pre_z \setminus \pap_z | \pap_z$. Note
        that $\pap_z = \{q\}$ does not work because $z \independent
        e,h | q$ does not hold. We thus set $\pa_z =\{q,h\}$. (Third
        graph in Figure \ref{fig:I-map2}.)
      \item We consider $a$: $\pre_a = \{e,h,q,z\}$. This is the last
        node in the ordering. To find the minimal set $\pap_a$ for
        which $a \independent \pre_a \setminus \pap_a | \pap_a$, we
        can determine its Markov blanket $\MB(a)$. The Markov blanket
        is the set of parents (none), children ($q$), and co-parents
        of $a$ ($z$) in Figure \ref{fig:I-map}, so that $\MB(a) = \{q,z\}$. We
        thus set $\pa_a = \{q,z\}$.(Fourth graph in Figure
        \ref{fig:I-map2}.)
\end{itemize}
    \begin{figure}[h]
      \centering
      \scalebox{0.9}{ 
        \begin{tikzpicture}[dgraph]
          \node[cont] (a) at (0,2) {$a$};
          \node[cont] (z) at (2,2) {$z$};
          \node[cont] (q) at (1,1) {$q$};
          \node[cont] (e) at (1,-0.4) {$e$};
          \node[cont] (h) at (3,1) {$h$};
          \draw(e) -- (h);

          \draw[-,gray,dotted] (3.7,-0.4) -- (3.7,2);
      \end{tikzpicture}}
      \hspace{1ex}
      \scalebox{0.9}{ 
        \begin{tikzpicture}[dgraph]
          \node[cont] (a) at (0,2) {$a$};
          \node[cont] (z) at (2,2) {$z$};
          \node[cont] (q) at (1,1) {$q$};
          \node[cont] (e) at (1,-0.4) {$e$};
          \node[cont] (h) at (3,1) {$h$};
          \draw(e) -- (h);
          \draw(e) -- (q);
          \draw(h) -- (q);
          
          \draw[-,gray,dotted] (3.7,-0.4) -- (3.7,2);
      \end{tikzpicture}}
      \hspace{1ex}
      \scalebox{0.9}{ 
        \begin{tikzpicture}[dgraph]
          \node[cont] (a) at (0,2) {$a$};
          \node[cont] (z) at (2,2) {$z$};
          \node[cont] (q) at (1,1) {$q$};
          \node[cont] (e) at (1,-0.4) {$e$};
          \node[cont] (h) at (3,1) {$h$};
          \draw(e) -- (h);
          \draw(e) -- (q);
          \draw(h) -- (q);
          \draw(q) -- (z);
          \draw(h) -- (z);

          \draw[-,gray,dotted] (3.7,-0.4) -- (3.7,2);
      \end{tikzpicture}}
      \hspace{1ex}
      \scalebox{0.9}{ 
        \begin{tikzpicture}[dgraph]
          \node[cont] (a) at (0,2) {$a$};
          \node[cont] (z) at (2,2) {$z$};
          \node[cont] (q) at (1,1) {$q$};
          \node[cont] (e) at (1,-0.4) {$e$};
          \node[cont] (h) at (3,1) {$h$};
          \draw(e) -- (h);
          \draw(e) -- (q);
          \draw(h) -- (q);
          \draw(q) -- (z);
          \draw(h) -- (z);
          \draw(q) -- (a);
          \draw(z) -- (a);    
      \end{tikzpicture}}
      \caption{\label{fig:I-map2} \exref{ex:minimal-I-maps-1}, Question \ref{q:directed-I-maps}:Construction of a minimal directed I-map for the ordering $(e,h,q,z,a)$. }
    \end{figure}

Since the skeleton in the obtained minimal I-map is different from the
skeleton of $G$, we do not have I-equivalence. Note that the ordering
$(e,h,q,z,a)$ yields a denser graph (Figure \ref{fig:I-map2}) than the
graph in Figure \ref{fig:I-map}. Whilst a minimal I-map, the graph does
e.g.\ not show that $a \independent z$. Furthermore, the causal
interpretation of the two graphs is different.

  \end{solution}

\item For the collection of random variables $(a,z,h,q,e)$ you are given the following Markov blankets for each variable:
  \begin{itemize}
  \item \MB(a) = \{q,z\}
  \item \MB(z) = \{a,q,h\}
  \item \MB(h) = \{z\}
  \item \MB(q) = \{a,z,e\}
  \item \MB(e) = \{q\}
  \end{itemize}
  \begin{exenumerate}
    \item Draw the undirected minimal I-map representing the independencies.
    \item Indicate a Gibbs distribution that satisfies the independence relations specified by the Markov blankets.
  \end{exenumerate}

  \begin{solution} Connecting each variable to all variables in its
    Markov blanket yields the desired undirected minimal I-map. Note
    that the Markov blankets are not mutually
    disjoint.  \begin{figure}[h] \begin{center} \scalebox{0.9}{ %
    x,y \begin{tikzpicture}[ugraph] \node[cont] (a) at (0,2)
    {$a$}; \node[cont] (z) at (2,2) {$z$}; \node[cont] (q) at (1,1)
    {$q$}; \node[cont] (e) at (1,-0.4) {$e$}; \node[cont] (h) at (3,1)
    {$h$}; \draw(a) -- (q); \draw(a) -- (z); \draw[-,gray,dotted]
    (3.7,-0.4) -- (3.7,2);

          \node[below = of e] {After $\MB(a)$};
        \end{tikzpicture}
        \vspace{3ex}
        \begin{tikzpicture}[ugraph]
          \node[cont] (a) at (0,2) {$a$};
          \node[cont] (z) at (2,2) {$z$};
          \node[cont] (q) at (1,1) {$q$};
          \node[cont] (e) at (1,-0.4) {$e$};
          \node[cont] (h) at (3,1) {$h$};
          \draw(a) -- (q);
          \draw(a) -- (z);
          \draw(z) -- (h);
          \draw(z) -- (q);
          \draw[-,gray,dotted] (3.7,-0.4) -- (3.7,2);
          \node[below = of e] {After $\MB(z)$};
        \end{tikzpicture}
        \vspace{3ex}
      \begin{tikzpicture}[ugraph]
        \node[cont] (a) at (0,2) {$a$};
        \node[cont] (z) at (2,2) {$z$};
        \node[cont] (q) at (1,1) {$q$};
        \node[cont] (e) at (1,-0.4) {$e$};
        \node[cont] (h) at (3,1) {$h$};
        \draw(a) -- (q);
        \draw(a) -- (z);
        \draw(z) -- (h);
        \draw(z) -- (q);
        \draw(q) -- (e);
        \node[below = of e] {After $\MB(q)$};
      \end{tikzpicture}}
    \end{center}
    \end{figure}

    For positive distributions, the set of distributions that satisfy
    the local Markov property relative to a graph (as given by the
    Markov blankets) is the same as the set of Gibbs distributions
    that factorise according to the graph. Given the I-map, we can now
    easily find the Gibbs distribution
    $$p(a,z,h,q,e) = \frac{1}{Z} \phi_1(a,z,q) \phi_2(q,e) \phi_3(z,h),$$
    where the $\phi_i$ must take positive values on their domain. Note that we
    used the maximal clique $(a,z,q)$.
      
  \end{solution}
\end{exenumerate}

\ex{I-equivalence between directed and undirected graphs}

\begin{exenumerate}
\item Verify that the following two graphs are I-equivalent by listing and
  comparing the independencies that each graph implies.

\begin{center}
  \scalebox{1}{ 
    \begin{tikzpicture}[ugraph]
      \node[cont] (z) at (0,0) {$z$};
      \node[cont] (y) at (1,1) {$y$};
      \node[cont] (x) at (0,2) {$x$};
      \node[cont] (u) at (-1,1) {$u$};
      \draw(x) -- (y);
      \draw(y) -- (z);
      \draw(z) -- (u);
      \draw(u) -- (x);
      \draw(u) -- (y);
    \end{tikzpicture}
    \hspace{6ex}
    \begin{tikzpicture}[dgraph]
      \node[cont] (z) at (0,0) {$z$};
      \node[cont] (y) at (1,1) {$y$};
      \node[cont] (x) at (0,2) {$x$};
      \node[cont] (u) at (-1,1) {$u$};
      \draw(x) -- (y);
      \draw(y) -- (z);
      \draw(u) -- (z);
      \draw(x) -- (u);
      \draw(u) -- (y);
    \end{tikzpicture}
  }
\end{center}

\begin{solution}
  First, note that both graphs share the same skeleton and the only
  reason that they are not fully connected is the missing edge between
  $x$ and $z$.

  For the DAG, there is also only one ordering that is topological to
  the graph: $x, u, y, z$. The missing edge between $x$ and $y$
  corresponds to the only independency encoded by the graph: $z
  \independent \pre_z \setminus \pa_z | \pa_z$, i.e.
  $$z \independent x | u, y.$$ This is the same independency that we
  get from the directed local Markov property.

  For the undirected graph, 
  $$z \independent x | u, y$$ holds because $u, y$ block all paths
  between $z$ and $x$. All variables but $z$ and $x$ are connected to
  each other, so that no further independency can hold.

  Hence both graphs only encode $z \independent x | u, y$ and they are
  thus I-equivalent.
  
\end{solution}

\item Are the following two graphs, which are directed and undirected hidden Markov models, I-equivalent?
  \begin{center}
  \scalebox{0.8}{
    \begin{tikzpicture}[dgraph]
      \node[cont] (y1) at (0,0) {$y_1$};
      \node[cont] (y2) at (2,0) {$y_2$};
      \node[cont] (y3) at (4,0) {$y_3$};
      \node[cont] (y4) at (6,0) {$y_4$};
      \node[cont] (x1) at (0,2) {$x_1$};
      \node[cont] (x2) at (2,2) {$x_2$};
      \node[cont] (x3) at (4,2) {$x_3$};
      \node[cont] (x4) at (6,2) {$x_4$};
      \draw(x1)--(y1);\draw(x2)--(y2);\draw(x3)--(y3);\draw(x4)--(y4);
      \draw(x1)--(x2);\draw(x2)--(x3);\draw(x3)--(x4);
  \end{tikzpicture}}
  \hspace{6ex}
  \scalebox{0.8}{
    \begin{tikzpicture}[ugraph]
      \node[cont] (y1) at (0,0) {$y_1$};
      \node[cont] (y2) at (2,0) {$y_2$};
      \node[cont] (y3) at (4,0) {$y_3$};
      \node[cont] (y4) at (6,0) {$y_4$};
      \node[cont] (x1) at (0,2) {$x_1$};
      \node[cont] (x2) at (2,2) {$x_2$};
      \node[cont] (x3) at (4,2) {$x_3$};
      \node[cont] (x4) at (6,2) {$x_4$};
      \draw(x1)--(y1);\draw(x2)--(y2);\draw(x3)--(y3);\draw(x4)--(y4);
      \draw(x1)--(x2);\draw(x2)--(x3);\draw(x3)--(x4);
  \end{tikzpicture}}
  \end{center}

  \begin{solution}
    The skeleton of the two graphs is the same and there are no
    immoralities. Hence, the two graphs are I-equivalent. 
  \end{solution}

\item Are the following two graphs I-equivalent?
  \begin{center}
  \scalebox{0.8}{
    \begin{tikzpicture}[dgraph]
      \node[cont] (y1) at (0,0) {$y_1$};
      \node[cont] (y2) at (2,0) {$y_2$};
      \node[cont] (y3) at (4,0) {$y_3$};
      \node[cont] (y4) at (6,0) {$y_4$};
      \node[cont] (x1) at (0,2) {$x_1$};
      \node[cont] (x2) at (2,2) {$x_2$};
      \node[cont] (x3) at (4,2) {$x_3$};
      \node[cont] (x4) at (6,2) {$x_4$};
      \draw(x1)--(y1);\draw(x2)--(y2);\draw(x3)--(y3);\draw(x4)--(y4);
      \draw(x1)--(x2);\draw(x3)--(x2);\draw(x3)--(x4);
  \end{tikzpicture}}
  \hspace{6ex}
  \scalebox{0.8}{
    \begin{tikzpicture}[ugraph]
      \node[cont] (y1) at (0,0) {$y_1$};
      \node[cont] (y2) at (2,0) {$y_2$};
      \node[cont] (y3) at (4,0) {$y_3$};
      \node[cont] (y4) at (6,0) {$y_4$};
      \node[cont] (x1) at (0,2) {$x_1$};
      \node[cont] (x2) at (2,2) {$x_2$};
      \node[cont] (x3) at (4,2) {$x_3$};
      \node[cont] (x4) at (6,2) {$x_4$};
      \draw(x1)--(y1);\draw(x2)--(y2);\draw(x3)--(y3);\draw(x4)--(y4);
      \draw(x1)--(x2);\draw(x2)--(x3);\draw(x3)--(x4);
  \end{tikzpicture}}
  \end{center}

  \begin{solution}
    The two graphs are not I-equivalent because $x_1-x_2-x_3$ forms an
    immorality. Hence, the undirected graph encodes $x_1 \independent
    x_3 | x_2$ which is not represented in the directed graph. On the
    other hand, the directed graph asserts $x_1 \independent x_3$
    which is not represented in the undirected graph.
  \end{solution}

\end{exenumerate}

\ex{Moralisation: Converting DAGs to undirected minimal I-maps}
\label{ex:DAG-to-undirected}
The following recipe constructs undirected minimal I-maps for
$\Ind(p)$:
\begin{itemize}
\item Determine the Markov blanket for each variable $x_i$
\item Construct a graph where the neighbours of $x_i$ are given by its
  Markov blanket.
\end{itemize}
We can adapt the recipe to construct an undirected minimal I-map for
the independencies $\Ind(G)$ encoded by a DAG $G$. What we need to do
is to use $G$ to read out the Markov blankets for the variables $x_i$
rather than determining the Markov blankets from the distribution $p$.

Show that this procedure leads to the following recipe to
convert DAGs to undirected minimal I-maps:
\begin{enumerate}
\item For all immoralities in the graph: add edges between \emph{all} parents of the collider node.
\item Make all edges in the graph undirected.
\end{enumerate}
The first step is sometimes called ``moralisation'' because we
``marry'' all the parents in the graph that are not already directly
connected by an edge. The resulting undirected graph is called the
moral graph of $G$, sometimes denoted by $\mathcal{M}(G)$.

\begin{solution}
  The Markov blanket of a variable $x$ is the set of its parents,
  children, and co-parents, as shown in the graph below in sub-figure (a). The parents and children are connected
  to $x$ in the directed graph, but the co-parents are not directly
  connected to $x$. Hence, according to ``Construct a graph where the
  neighbours of $x_i$ are its Markov blanket.'', we need to introduce
  edges between $x$ and all its co-parents. This gives the
  intermediate graph in sub-figure (b).

  Now, considering the top-left parent of $x$, we see that for that node,
  the Markov blanket includes the other parents of $x$. This means
  that we need to connect all parents of $x$, which gives the graph in
  sub-figure (c). This is sometimes called ``marrying''
  the parents of $x$. Continuing in this way, we see that we need to
  ``marry'' all parents in the graph that are not already married.

  Finally, we need to make all edges in the graph undirected, which
  gives sub-figure (d).

  A simpler approach is to note that the DAG specifies the
  factorisation $p(\x) = \prod_i p(x_i | \pa_i)$. We can consider each
  conditional $p(x_i | \pa_i)$ to be a factor $\phi_i(x_i, \pa_i)$ so
  that we obtain the Gibbs distribution $p(\x) = \prod_i \phi_i(x_i |
  \pa_i)$. Visualising the distribution by connecting all variables in
  the same factor $\phi_i(x_i | \pa_i)$ leads to the ``marriage'' of
  all parents of $x_i$. This corresponds to the first step in the
  recipe because $x_i$ is in a collider configuration with respect to
  the parent nodes. Not all parents form an immorality but this does
  here not matter because those that do not form an immorality are
  already connected by a covering edge in the first place.
  
  \begin{figure}[htb]
    \centering
    \begin{tabular}[b]{c}
      \scalebox{0.7}{ 
        \begin{tikzpicture}[dgraph]
          
          \node[cont] (lowest1) at ( 2,0) {};
          \node[cont] (lowest2) at ( 3.5,0) {};
          \node[cont] (lowest3) at ( 4.5,0) {};
          
          \node[cont] (belowleft) at ( 2,1) {};
          \node[cont] (belowright) at ( 4,1) {};
          
          \node[cont] (copleft) at ( 1,2) {};
          \node[cont] (x) at ( 3,2) {$x$};
          \node[cont] (copright) at ( 5,2) {};
          
          \node (abovemostleft) at ( 1,3) {};
          \node[cont] (aboveleft) at ( 2,3) {};
          \node[cont] (above) at ( 3,3) {};
          \node[cont] (aboveright) at ( 4,3) {};
          \node (abovemostright) at ( 5,3) {};
          
          \node (top1) at ( 1,4) {};
          \node[cont] (top2) at ( 2,4) {};
          \node (top3) at ( 3,4) {};
          \node[cont] (top4) at ( 4,4) {};
          \node (top5) at ( 5,4) {};
          
          \draw (top1) -- (aboveleft);
          \draw (top2) -- (aboveleft);
          \draw (top3) -- (above);
          \draw (top4) -- (aboveright);
          \draw (top5) -- (aboveright);
          
          \draw (aboveright) -- (x);
          \draw (aboveleft) -- (x);
          \draw (above) -- (x);
          
          \draw (x) -- (belowright);
          \draw (x) -- (belowleft);
          
          \draw (copleft) -- (belowleft);
          \draw (copright) -- (belowright);
          
          \draw (abovemostleft) -- (copleft);
          \draw (abovemostright) -- (copright);
          
          \draw (belowright) -- (lowest2);
          \draw (belowright) -- (lowest3);
          \draw (belowleft) -- (lowest1);

          \node[contobs] at ( 2,3) {};
          \node[contobs] at ( 3,3) {};
          \node[contobs] at ( 4,3) {};
          
          \node[contobs] at ( 2,1) {};
          \node[contobs] at ( 4,1) {};
          
          \node[contobs] at ( 1,2) {};
          \node[contobs] at ( 5,2) {};
      \end{tikzpicture}}\\
      {\small (a) DAG}
    \end{tabular}
    \begin{tabular}[b]{c}
      \scalebox{0.7}{ 
        \begin{tikzpicture}[dgraph]
          
          \node[cont] (lowest1) at ( 2,0) {};
          \node[cont] (lowest2) at ( 3.5,0) {};
          \node[cont] (lowest3) at ( 4.5,0) {};
          
          \node[cont] (belowleft) at ( 2,1) {};
          \node[cont] (belowright) at ( 4,1) {};
          
          \node[cont] (copleft) at ( 1,2) {};
          \node[cont] (x) at ( 3,2) {$x$};
          \node[cont] (copright) at ( 5,2) {};
          
          \node (abovemostleft) at ( 1,3) {};
          \node[cont] (aboveleft) at ( 2,3) {};
          \node[cont] (above) at ( 3,3) {};
          \node[cont] (aboveright) at ( 4,3) {};
          \node (abovemostright) at ( 5,3) {};
          
          \node (top1) at ( 1,4) {};
          \node[cont] (top2) at ( 2,4) {};
          \node (top3) at ( 3,4) {};
          \node[cont] (top4) at ( 4,4) {};
          \node (top5) at ( 5,4) {};
          
          \draw (top1) -- (aboveleft);
          \draw (top2) -- (aboveleft);
          \draw (top3) -- (above);
          \draw (top4) -- (aboveright);
          \draw (top5) -- (aboveright);
          
          \draw (aboveright) -- (x);
          \draw (aboveleft) -- (x);
          \draw (above) -- (x);
          
          \draw (x) -- (belowright);
          \draw (x) -- (belowleft);
          
          \draw (copleft) -- (belowleft);
          \draw (copright) -- (belowright);
          
          \draw (abovemostleft) -- (copleft);
          \draw (abovemostright) -- (copright);
          
          \draw (belowright) -- (lowest2);
          \draw (belowright) -- (lowest3);
          \draw (belowleft) -- (lowest1);

          \node[contobs] at ( 2,3) {};
          \node[contobs] at ( 3,3) {};
          \node[contobs] at ( 4,3) {};
          
          \node[contobs] at ( 2,1) {};
          \node[contobs] at ( 4,1) {};
          
          \node[contobs] at ( 1,2) {};
          \node[contobs] at ( 5,2) {};

          \draw[-] (copleft) -- (x);
          \draw[-] (copright) -- (x);
      \end{tikzpicture}}\\
    {\small  (b) Intermediate step 1}
    \end{tabular}
    \begin{tabular}[b]{c}
      \scalebox{0.7}{ 
        \begin{tikzpicture}[dgraph]
          
          \node[cont] (lowest1) at ( 2,0) {};
          \node[cont] (lowest2) at ( 3.5,0) {};
          \node[cont] (lowest3) at ( 4.5,0) {};
          
          \node[cont] (belowleft) at ( 2,1) {};
          \node[cont] (belowright) at ( 4,1) {};
          
          \node[cont] (copleft) at ( 1,2) {};
          \node[cont] (x) at ( 3,2) {$x$};
          \node[cont] (copright) at ( 5,2) {};
          
          \node (abovemostleft) at ( 1,3) {};
          \node[cont] (aboveleft) at ( 2,3) {};
          \node[cont] (above) at ( 3,3) {};
          \node[cont] (aboveright) at ( 4,3) {};
          \node (abovemostright) at ( 5,3) {};
          
          \node (top1) at ( 1,4) {};
          \node[cont] (top2) at ( 2,4) {};
          \node (top3) at ( 3,4) {};
          \node[cont] (top4) at ( 4,4) {};
          \node (top5) at ( 5,4) {};
          
          \draw (top1) -- (aboveleft);
          \draw (top2) -- (aboveleft);
          \draw (top3) -- (above);
          \draw (top4) -- (aboveright);
          \draw (top5) -- (aboveright);
          
          \draw (aboveright) -- (x);
          \draw (aboveleft) -- (x);
          \draw (above) -- (x);
          
          \draw (x) -- (belowright);
          \draw (x) -- (belowleft);
          
          \draw (copleft) -- (belowleft);
          \draw (copright) -- (belowright);
          
          \draw (abovemostleft) -- (copleft);
          \draw (abovemostright) -- (copright);
          
          \draw (belowright) -- (lowest2);
          \draw (belowright) -- (lowest3);
          \draw (belowleft) -- (lowest1);

          \node[contobs] at ( 2,3) {};
          \node[contobs] at ( 3,3) {};
          \node[contobs] at ( 4,3) {};
          
          \node[contobs] at ( 2,1) {};
          \node[contobs] at ( 4,1) {};
          
          \node[contobs] at ( 1,2) {};
          \node[contobs] at ( 5,2) {};

          \draw[-] (copleft) -- (x);
          \draw[-] (copright) -- (x);
          \draw[-] (aboveleft) -- (above);
          \draw[-] (aboveleft) to [bend right=-45] (aboveright);
          \draw[-] (above) -- (aboveright);
      \end{tikzpicture}}\\
     {\small (c) Intermediate step 2}
    \end{tabular}
    \begin{tabular}[b]{c}
      \scalebox{0.7}{ 
        \begin{tikzpicture}[ugraph]
           
          \node[cont] (lowest1) at ( 2,0) {};
          \node[cont] (lowest2) at ( 3.5,0) {};
          \node[cont] (lowest3) at ( 4.5,0) {};
          
          \node[cont] (belowleft) at ( 2,1) {};
          \node[cont] (belowright) at ( 4,1) {};
          
          \node[cont] (copleft) at ( 1,2) {};
          \node[cont] (x) at ( 3,2) {$x$};
          \node[cont] (copright) at ( 5,2) {};
          
          \node (abovemostleft) at ( 1,3) {};
          \node[cont] (aboveleft) at ( 2,3) {};
          \node[cont] (above) at ( 3,3) {};
          \node[cont] (aboveright) at ( 4,3) {};
          \node (abovemostright) at ( 5,3) {};
          
          \node (top1) at ( 1,4) {};
          \node[cont] (top2) at ( 2,4) {};
          \node (top3) at ( 3,4) {};
          \node[cont] (top4) at ( 4,4) {};
          \node (top5) at ( 5,4) {};
          
          \draw (top1) -- (aboveleft);
          \draw (top2) -- (aboveleft);
          \draw (top3) -- (above);
          \draw (top4) -- (aboveright);
          \draw (top5) -- (aboveright);
          
          \draw (aboveright) -- (x);
          \draw (aboveleft) -- (x);
          \draw (above) -- (x);
          
          \draw (x) -- (belowright);
          \draw (x) -- (belowleft);
          
          \draw (copleft) -- (belowleft);
          \draw (copright) -- (belowright);
          
          \draw (abovemostleft) -- (copleft);
          \draw (abovemostright) -- (copright);
          
          \draw (belowright) -- (lowest2);
          \draw (belowright) -- (lowest3);
          \draw (belowleft) -- (lowest1);

          \node[contobs] at ( 2,3) {};
          \node[contobs] at ( 3,3) {};
          \node[contobs] at ( 4,3) {};
          
          \node[contobs] at ( 2,1) {};
          \node[contobs] at ( 4,1) {};
          
          \node[contobs] at ( 1,2) {};
          \node[contobs] at ( 5,2) {};

          \draw[-] (copleft) -- (x);
          \draw[-] (copright) -- (x);
          \draw[-] (aboveleft) -- (above);
          \draw[-] (aboveleft) to [bend right=-45] (aboveright);
          \draw[-] (above) -- (aboveright);
          \draw[-] (top1) to [bend right=-45] (top2);
          \draw[-] (top4) to [bend right=-45] (top5); 
      \end{tikzpicture}}\\
     {\small (d) Undirected graph}
    \end{tabular}
    \caption{Answer to \exref{ex:DAG-to-undirected}: Illustrating the moralisation process}
  \end{figure}
\end{solution}

\ex{Moralisation exercise}
\label{ex:moralisation}

For the DAG $G$ below find the minimal undirected I-map for $\Ind(G)$.
  
 \begin{center}
  \scalebox{0.8}{
    \begin{tikzpicture}[dgraph]

      \node[cont] (x2) at (2,0) {$x_2$};
      \node[cont, left = of x2] (x1) {$x_1$};
      \node[cont, right = of x2] (x3) {$x_3$};
      \node[cont, below = of x2] (x4) {$x_4$};
      \node[cont, right = of x4] (x5) {$x_5$};
      \node[cont, below = of x4] (x6) {$x_6$};
      \node[cont, below = of x5] (x7) {$x_7$};

      \draw (x1) -- (x4);
      \draw (x2) -- (x4);
      \draw (x3) -- (x4);
      \draw (x4) -- (x6);
      \draw (x4) -- (x7);
      \draw (x5) -- (x7);
      
    \end{tikzpicture}
    }
\end{center}
  
  \begin{solution}

    To derive an undirected minimal I-map from a directed one, we have
    to construct the moralised graph where the ``unmarried'' parents
    are connected by a covering edge. This is because each conditional
    $p(x_i | \pa_i)$ corresponds to a factor $\phi_i(x_i,\pa_i)$ and
    we need to connect all variables that are arguments of the same factor with edges.

    Statistically, the reason for marrying the parents is as follows:
    An independency $x \independent y | \{\text{child, other nodes}\}$
    does not hold in the directed graph in case of collider connections
    but would hold in the undirected graph if we didn't marry the
    parents. Hence links between the parents must be added.

    It is important to add edges between \emph{all} parents of a
    node. Here, $p(x_4 | x_1, x_2, x_3)$ corresponds to a factor
    $\phi(x_4, x_1, x_2, x_3)$ so that all four variables need to be
    connected. Just adding edges $x_1 - x_2$ and $x_2 - x_3$ would not be enough.

    The moral graph, which is the requested minimal undirected I-map,
    is shown below.

     \begin{center}
       \scalebox{0.8}{
         \begin{tikzpicture}[ugraph]
           
           \node[cont] (x2) at (2,0) {$x_2$};
           \node[cont, left = of x2] (x1) {$x_1$};
           \node[cont, right = of x2] (x3) {$x_3$};
           \node[cont, below = of x2] (x4) {$x_4$};
           \node[cont, right = of x4] (x5) {$x_5$};
           \node[cont, below = of x4] (x6) {$x_6$};
           \node[cont, below = of x5] (x7) {$x_7$};

           \draw (x1) -- (x4);
           \draw (x2) -- (x4);
           \draw (x3) -- (x4);
           \draw (x4) -- (x6);
           \draw (x4) -- (x7);
           \draw (x5) -- (x7);

           \draw (x1) -- (x2);
           \draw (x2) -- (x3);
           \draw (x1) to [bend left=45] (x3);
           \draw (x4) -- (x5);
           
         \end{tikzpicture}
       }
     \end{center}
     
  \end{solution}

  \ex{Moralisation exercise}

  Consider the DAG $G$:
  \begin{center}
    \scalebox{0.9}{ 
      \begin{tikzpicture}[dgraph]
        \node[cont] (y) at (0,0) {$y$};
        \node[cont] (z1) at (-1.75,1.5) {$z_1$};
        \node[cont] (z2) at (1.75,1.5) {$z_2$};
        \node[cont] (x1) at (-3,3) {$x_1$};
        \node[cont] (x2) at (-1.75,3) {$x_2$};
        \node[cont] (x3) at (-0.5,3) {$x_3$};
        \node[cont] (x4) at (0.5,3) {$x_4$};
        \node[cont] (x5) at (1.75,3) {$x_5$};
        \node[cont] (x6) at (3,3) {$x_6$};
        
        \draw (z1) -- (y);
        \draw (z2) -- (y);
        \draw (x1) -- (z1);
        \draw (x2) -- (z1);
        \draw (x3) -- (z1);
        \draw (x4) -- (z2);
        \draw (x5) -- (z2);
        \draw (x6) -- (z2);
    \end{tikzpicture}}
  \end{center}
  A friend claims that the undirected graph below is the moral
  graph $\mathcal{M}(G)$ of $G$. Is your friend correct? If not,
  state which edges needed to be removed or added, and explain, in
  terms of represented independencies, why the changes are
  necessary for the graph to become the moral graph of
  $G$. 

  \begin{center}
    \scalebox{0.9}{ 
      \begin{tikzpicture}[ugraph]
        \node[cont] (y) at (0,0) {$y$};
        \node[cont] (z1) at (-1.75,1.5) {$z_1$};
        \node[cont] (z2) at (1.75,1.5) {$z_2$};
        \node[cont] (x1) at (-3,3) {$x_1$};
        \node[cont] (x2) at (-1.75,3) {$x_2$};
        \node[cont] (x3) at (-0.5,3) {$x_3$};
        \node[cont] (x4) at (0.5,3) {$x_4$};
        \node[cont] (x5) at (1.75,3) {$x_5$};
        \node[cont] (x6) at (3,3) {$x_6$};
        
        \draw (z1) -- (y);
        \draw (z2) -- (y);
        \draw (x1) -- (z1);
        \draw (x2) -- (z1);
        \draw (x3) -- (z1);
        \draw (x4) -- (z2);
        \draw (x5) -- (z2);
        \draw (x6) -- (z2);

        \draw (x1) -- (x2);
        \draw (x2) -- (x3);
        \draw (x4) -- (x5);
        \draw (x5) -- (x6);

        \draw (x1) to [out=40,in=140] (x6);

        \draw (z1) -- (z2);
    \end{tikzpicture}}
  \end{center}

  \begin{solution}
    The moral graph $\mathcal{M}(G)$ is an undirected minimal I-map of
    the independencies represented by $G$. Following the procedure of
    connecting ``unmarried'' parents of colliders, we obtain the following moral graph of $G$:
 
    \begin{center}
      \begin{tikzpicture}[ugraph]
        \node[cont] (y) at (0,0) {$y$};
        \node[cont] (z1) at (-1.75,1.5) {$z_1$};
        \node[cont] (z2) at (1.75,1.5) {$z_2$};
        \node[cont] (x1) at (-3,3) {$x_1$};
        \node[cont] (x2) at (-1.75,3) {$x_2$};
        \node[cont] (x3) at (-0.5,3) {$x_3$};
        \node[cont] (x4) at (0.5,3) {$x_4$};
        \node[cont] (x5) at (1.75,3) {$x_5$};
        \node[cont] (x6) at (3,3) {$x_6$};
        
        \draw (z1) -- (y);
        \draw (z2) -- (y);
        \draw (x1) -- (z1);
        \draw (x2) -- (z1);
        \draw (x3) -- (z1);
        \draw (x4) -- (z2);
        \draw (x5) -- (z2);
        \draw (x6) -- (z2);

        \draw (x1) -- (x2);
        \draw (x2) -- (x3);
        \draw (x1) to [out=45,in=135] (x3);

        \draw (x4) -- (x5);
        \draw (x5) -- (x6);
        \draw (x4) to [out=45,in=135] (x6);

        \draw (z1) -- (z2);
      \end{tikzpicture}
    \end{center}
    We can thus see that the friend's undirected graph is not the
    moral graph of $G$.

    The edge between $x_1$ and $x_6$ can be removed. This is because
    for $G$, we have e.g.\ the independencies $x_1 \independent x_6 |
    z_1$, $x_1 \independent x_6 | z_2$, $x_1 \independent x_6 | z_1,
    z_2$ which is not represented by the drawn undirected
    graph.
    
    We need to add edges between $x_1$ and $x_3$, and between $x_4$
    and $x_6$. Otherwise, the undirected graph makes the wrong
    independency assertion that $x_1 \independent x_3 | x_2, z_1$ (and
    equivalent for $x_4$ and $x_6$).
   
  \end{solution}

\ex{Triangulation: Converting undirected graphs to directed minimal I-maps}
\label{ex:undirected-to-DAG}

In \exref{ex:DAG-to-undirected} we adapted a recipe for constructing
undirected minimal I-maps for $\Ind(p)$ to the case of $\Ind(G)$,
where $G$ is a DAG. The key difference was that we used the graph $G$
to determine independencies rather than the distribution $p$.

We can similarly adapt the recipe for constructing a directed minimal
I-map for $\Ind(p)$ to build a directed minimal I-map for $\Ind(H)$,
where $H$ is an undirected graph:
\begin{enumerate}
  \item Choose an ordering of the random variables.
  \item For all variables $x_i$, use $H$ to determine a \emph{minimal}
    subset $\pap_i$ of the predecessors $\pre_i$ such that
    $$ x_i \independent \left(\pre_i \setminus \pap_i\right) \mid \pap_i $$
    holds.
  \item Construct a DAG with the $\pap_i$ as parents $\pa_i$ of $x_i$.
\end{enumerate}
Remarks: (1) Directed minimal I-maps obtained with different orderings
are generally not I-equivalent. (2) The directed minimal I-maps
obtained with the above method are always chordal graphs. Chordal
graphs are graphs where the longest trail without shortcuts is a
triangle (\url{https://en.wikipedia.org/wiki/Chordal_graph}). They are
thus also called triangulated graphs. We obtain chordal graphs because
if we had trails without shortcuts that involved more than 3 nodes, we would
necessarily have an immorality in the graph. But immoralities encode
independencies that an undirected graph cannot represent, which would
make the DAG not an I-map for $\Ind(H)$ any more.

\begin{exenumerate}
\item Let $H$ be the undirected graph below. Determine the directed
  minimal I-map for $\Ind(H)$ with the variable ordering $x_1, x_2,
  x_3, x_4, x_5$.
  \label{q:undirected-to-directed-I-map}
  \begin{center}
    \scalebox{0.9}{ 
      \begin{tikzpicture}[ugraph]
        \node[cont] (x1) at (0,0) {$x_1$};
        \node[cont] (x2) at (2,0) {$x_2$};
        \node[cont] (x3) at (0,-1.5) {$x_3$};
        \node[cont] (x4) at (2,-1.5) {$x_4$};
        \node[cont] (x5) at (1,-3) {$x_5$};

        \draw(x1) -- (x2);
        \draw(x1) -- (x3);
        \draw(x3) -- (x5);
        \draw(x2) -- (x4);
        \draw(x4) -- (x5);
    \end{tikzpicture}}
  \end{center}

  \begin{solution}
    We use the ordering $x_1, x_2, x_3, x_4, x_5$ and follow the
    conversion procedure:
    
    \begin{itemize}
      \item $x_2$ is not independent from $x_1$ so that we set $\pa_2 =\{x_1\}$. See first graph in Figure \ref{fig:undirected-to-directed-I-map}.
      \item Since $x_3$ is connected to both $x_1$ and $x_2$, we
        don't have $x_3 \independent x_2, x_1$. We cannot
        make $x_3$ independent from $x_2$ by conditioning on $x_1$
        because there are two paths from $x_3$ to $x_2$ and $x_1$ only
        blocks the upper one. Moreover, $x_1$ is a neighbour of $x_3$
        so that conditioning on $x_2$ does make them
        independent. Hence we must set $\pa_3 = \{x_1, x_2\}$.  See
        second graph in Figure \ref{fig:undirected-to-directed-I-map}.
      \item For $x_4$, we see from the undirected graph, that $x_4
        \independent x_1 \mid x_3, x_2$. The graph further shows that
        removing either $x_3$ or $x_2$ from the conditioning set is
        not possible and conditioning on $x_1$ won't make $x_4$
        independent from $x_2$ or $x_3$. We thus have $\pa_4 = \{x_2,
        x_3\}$.  See fourth graph in Figure \ref{fig:undirected-to-directed-I-map}.
      \item The same reasoning shows that $\pa_5 = \{x_3,x_4\}$. See last graph in Figure \ref{fig:undirected-to-directed-I-map}.
    \end{itemize}
    This results in the triangulated directed graph in Figure
    \ref{fig:undirected-to-directed-I-map} on the right.
 
    \begin{figure}[h!]
    \centering
    \scalebox{0.9}{ 
       \begin{tikzpicture}[dgraph]
        \node[cont] (x1) at (0,0) {$x_1$};
        \node[cont] (x2) at (2,0) {$x_2$};
        \node[cont] (x3) at (0,-1.5) {$x_3$};
        \node[cont] (x4) at (2,-1.5) {$x_4$};
        \node[cont] (x5) at (1,-3) {$x_5$};

        \draw(x1) -- (x2);

        \draw[-,gray,dotted] (2.5,-3) -- (2.5,0);
       \end{tikzpicture}
       \hspace{2ex}
      \begin{tikzpicture}[dgraph]
        \node[cont] (x1) at (0,0) {$x_1$};
        \node[cont] (x2) at (2,0) {$x_2$};
        \node[cont] (x3) at (0,-1.5) {$x_3$};
        \node[cont] (x4) at (2,-1.5) {$x_4$};
        \node[cont] (x5) at (1,-3) {$x_5$};

        \draw(x1) -- (x2);
        \draw(x1) -- (x3);
        \draw(x2) -- (x3);

        \draw[-,gray,dotted] (2.5,-3) -- (2.5,0);
      \end{tikzpicture}
      \hspace{2ex}
      \begin{tikzpicture}[dgraph]
        \node[cont] (x1) at (0,0) {$x_1$};
        \node[cont] (x2) at (2,0) {$x_2$};
        \node[cont] (x3) at (0,-1.5) {$x_3$};
        \node[cont] (x4) at (2,-1.5) {$x_4$};
        \node[cont] (x5) at (1,-3) {$x_5$};

        \draw(x1) -- (x2);
        \draw(x1) -- (x3);
        \draw(x2) -- (x3);
        \draw(x2) -- (x4);
        \draw(x3) -- (x4);

        \draw[-,gray,dotted] (2.5,-3) -- (2.5,0);
      \end{tikzpicture}
      \hspace{2ex}
      \begin{tikzpicture}[dgraph]
        \node[cont] (x1) at (0,0) {$x_1$};
        \node[cont] (x2) at (2,0) {$x_2$};
        \node[cont] (x3) at (0,-1.5) {$x_3$};
        \node[cont] (x4) at (2,-1.5) {$x_4$};
        \node[cont] (x5) at (1,-3) {$x_5$};

        \draw(x1) -- (x2);
        \draw(x1) -- (x3);
        \draw(x2) -- (x3);
        \draw(x2) -- (x4);
        \draw(x3) -- (x4);
        \draw(x3) -- (x5);
        \draw(x4) -- (x5);
    \end{tikzpicture}
    }
    \caption{\label{fig:undirected-to-directed-I-map}. Answer to \exref{ex:undirected-to-DAG}, Question \ref{q:undirected-to-directed-I-map}. }
    
    \end{figure}

    To see why triangulation is necessary consider the case where we
    didn't have the edge between $x_2$ and $x_3$ as in Figure
    \ref{fig:triangulation}. The directed graph would then imply that
    $x_3 \independent x_2 \mid x_1$ (check!). But this independency assertion
    does not hold in the undirected graph so that the graph in Figure
    \ref{fig:triangulation} is not an I-map.
  
     \begin{figure}[h!]
       \centering
       \begin{tikzpicture}[dgraph]
         \node[cont] (x1) at (0,0) {$x_1$};
         \node[cont] (x2) at (2,0) {$x_2$};
         \node[cont] (x3) at (0,-1.5) {$x_3$};
         \node[cont] (x4) at (2,-1.5) {$x_4$};
         \node[cont] (x5) at (1,-3) {$x_5$};
         
         \draw(x1) -- (x2);
         \draw(x1) -- (x3);
         \draw(x2) -- (x4);
         \draw(x3) -- (x4);
         \draw(x3) -- (x5);
         \draw(x4) -- (x5);
       \end{tikzpicture}
       \caption{\label{fig:triangulation} Not a directed I-map for the undirected graphical model defined by the graph in \exref{ex:undirected-to-DAG}, Question \ref{q:undirected-to-directed-I-map}.}
     \end{figure}
  \end{solution}

\item For the undirected graph from question
  \ref{q:undirected-to-directed-I-map} above, which variable ordering
  yields the directed minimal I-map below?
  \label{q:undirected-to-directed-I-map-chordal-graph-variable-ordering}
  \begin{center}
  \begin{tikzpicture}[dgraph]
    \node[cont] (x1) at (0,0) {$x_1$};
    \node[cont] (x2) at (2,0) {$x_2$};
    \node[cont] (x3) at (0,-1.5) {$x_3$};
    \node[cont] (x4) at (2,-1.5) {$x_4$};
    \node[cont] (x5) at (1,-3) {$x_5$};
    
    \draw(x1) -- (x2);
    \draw(x1) -- (x3);
    \draw(x1) -- (x4);
    \draw(x2) -- (x4);
    \draw(x4) -- (x3);
    \draw(x3) -- (x5);
    \draw(x4) -- (x5);
  \end{tikzpicture}
  \end{center}

  \begin{solution}
    $x_1$ is the root of the DAG, so it comes first. Next in the ordering
    are the children of $x_1$: $x_2, x_3, x_4$. Since $x_3$ is a child
    of $x_4$, and $x_4$ a child of $x_2$, we must have $x_1, x_2, x_4, x_3$. Furthermore, $x_3$ must come before $x_5$ in the ordering
    since $x_5$ is a child of $x_3$, hence the ordering used must have
    been: $x_1, x_2, x_4, x_3, x_5$.
    
  \end{solution}
  
\end{exenumerate}

\ex{I-maps, minimal I-maps, and I-equivalency}
Consider the following probability density function for random variables $x_1, \ldots, x_6$.
$$ p_a(x_1, \ldots, x_6) = p(x_1) p(x_2) p(x_3 | x_1,x_2) p(x_4 | x_2)
p(x_5| x_1) p(x_6 | x_3, x_4, x_5)$$ For each of the two graphs below,
explain whether it is a minimal I-map, not a minimal I-map but still
an I-map, or not an I-map for the independencies that hold for $p_a$.

  \begin{center}
    \scalebox{0.95}{ 
      \begin{tikzpicture}[dgraph]
        \node[cont] (x1) at (0,0) {$x_1$};
        \node[cont] (x2) at (2,0) {$x_2$};
        \node[cont] (x3) at (1,-1.5) {$x_3$};
        \node[cont] (x4) at (3,-1.5) {$x_4$};
        \node[cont] (x5) at (-1,-1.5) {$x_5$};
        \node[cont] (x6) at (1,-3) {$x_6$};
        
        \draw (x1) -- (x3);
        \draw (x2) -- (x3);
        \draw (x2) -- (x4);
        \draw (x1) -- (x5);
        \draw (x3) -- (x6);
        \draw (x5) -- (x6);
        \draw (x4) -- (x6);

        \draw (x3) -- (x4);
        \draw (x3) -- (x5);

        \node[] (text) at (1,-4) {graph 1};
      \end{tikzpicture}\hspace{12ex}
      \begin{tikzpicture}[ugraph]
        \node[cont] (x1) at (0,0) {$x_1$};
        \node[cont] (x2) at (2,0) {$x_2$};
        \node[cont] (x3) at (1,-1.5) {$x_3$};
        \node[cont] (x4) at (3,-1.5) {$x_4$};
        \node[cont] (x5) at (-1,-1.5) {$x_5$};
        \node[cont] (x6) at (1,-3) {$x_6$};
        
        \draw (x1) -- (x3);
        \draw (x2) -- (x3);
        \draw (x2) -- (x4);
        \draw (x1) -- (x5);
        \draw (x3) -- (x6);
        \draw (x5) -- (x6);
        \draw (x4) -- (x6);
        
        \draw (x3) -- (x4);
        \draw (x3) -- (x5);
        \draw (x1) -- (x2);
        \node[] (text) at (1,-4) {graph 2};
    \end{tikzpicture}}
  \end{center}

  \begin{solution}
    The pdf can be visualised as the following directed graph, which is a minimal I-map for it.\\
    \begin{center}
      \scalebox{1}{ 
        \begin{tikzpicture}[dgraph]
          \node[cont] (x1) at (0,0) {$x_1$};
          \node[cont] (x2) at (2,0) {$x_2$};
          \node[cont] (x3) at (1,-1.5) {$x_3$};
          \node[cont] (x4) at (3,-1.5) {$x_4$};
          \node[cont] (x5) at (-1,-1.5) {$x_5$};
          \node[cont] (x6) at (1,-3) {$x_6$};
          
          \draw (x1) -- (x3);
          \draw (x2) -- (x3);
          \draw (x2) -- (x4);
          \draw (x1) -- (x5);
          \draw (x3) -- (x6);
          \draw (x5) -- (x6);
          \draw (x4) -- (x6);
          
      \end{tikzpicture}}
    \end{center}
    Graph 1 defines distributions that factorise as
    \begin{equation}
      p_b(\x) = p(x_1) p(x_2) p(x_3|x_1, x_2) p(x_4|x_2, x_3) p(x_5|x_1, x_3) p(x_6|x_3,x_4,x_5).
    \end{equation}
    Comparing with $p_a(x_1, \ldots, x_6)$, we see that only the
    conditionals $p(x_4|x_2, x_3)$ and $p(x_5|x_1, x_3)$ are
    different. Specifically, their conditioning set includes $x_3$,
    which means that Graph 1 encodes fewer independencies than what
    $p_a(x_1, \ldots, x_6)$ satisfies. In particular $x_4 \independent
    x_3 | x_2$ and $x_5 \independent x_3 | x_1$ are not represented in
    the graph. This means that we could remove $x_3$ from the
    conditioning sets, or equivalently remove the edges $x_3
    \rightarrow x_4$ and $x_3 \rightarrow x_5$ from the graph without
    introducing independence assertions that do not hold for
    $p_a$. This means graph 1 is an I-map but not a minimal I-map.
    
    Graph 2 is not an I-map. To be an undirected minimal I-map, we had
    to connect variables $x_5$ and $x_4$ that are parents of
    $x_6$. Graph 2 wrongly claims that $x_5 \independent x_4 \mid
    x_1,x_3,x_6$.
    
  \end{solution}

\ex{Limits of directed and undirected graphical models}

We here consider the probabilistic model $p(y_1,y_2,x_1,x_2) = p(y_1,
y_2 | x_1, x_2)p(x_1) p(x_2)$ where $p(y_1, y_2 | x_1, x_2)$
factorises as
\begin{equation}
  p(y_1, y_2 | x_1, x_2) = p(y_1 | x_1) p(y_2 | x_2) \phi(y_1,y_2) n(x_1,x_2)
\end{equation}
with  $n(x_1,x_2)$ equal to
\begin{equation}
  n(x_1,x_2) = \left(\int p(y_1 | x_1) p(y_2 | x_2) \phi(y_1,y_2) \ud y_1 \ud y_2\right)^{-1}.
  \label{eq:ndef}
\end{equation}
In the model, $x_1$ and $x_2$ are two independent inputs that each control
the interacting variables $y_1$ and $y_2$ (see graph below). However, the
nature of the interaction between $y_1$ and $y_2$ is not modelled. In
particular, we do not assume a directionality, i.e. $y_1 \rightarrow y_2$, or $y_2 \rightarrow y_1$.

\begin{center}
  \scalebox{1}{ 
    \begin{tikzpicture}[dgraph]
          
      \node[cloud,opacity=0.5, fill=gray!20, cloud puffs=12, cloud puff arc= 100,
        minimum width=4cm, minimum height=2cm, aspect=1] at (1,-2) {};
          \node[] at (1,-2.5) {\tiny some interaction};
          \node[cont] (x1) at (0,-0.5) {$x_1$};
          \node[cont] (x2) at (2,-0.5) {$x_2$};
          \node[cont] (y1) at (0,-2) {$y_1$};
          \node[cont] (y2) at (2,-2) {$y_2$};
          \draw(x1) -- (y1);
          \draw(x2) -- (y2);
          \draw[-,dotted](y1) -- (y2);
  \end{tikzpicture}}
\end{center}

\begin{exenumerate}

\item Use the basic characterisations of statistical independence
  \begin{align}
  u \independent v | z &\Longleftrightarrow p(u,v | z) = p(u|z) p(v | z) \label{eq:ind1}\\
  u \independent v | z &\Longleftrightarrow p(u,v | z) = a(u,z) b(v,z) \quad \quad \quad (a(u,z) \ge0, b(v,z) \ge 0) \label{eq:ind2}
  \end{align}
  to show that $p(y_1,y_2,x_1,x_2)$ satisfies the following independencies
\begin{align*}
  x_1 \independent x_2 &&  x_1 \independent y_2 \mid y_1, x_2 && x_2 \independent y_1 \mid y_2, x_1
\end{align*}

  \begin{solution}
    The pdf/pmf is
    $$p(y_1,y_2,x_1,x_2) = p(y_1 | x_1) p(y_2 | x_2) \phi(y_1,y_2) n(x_1,x_2) p(x_1) p(x_2)$$
    
    For $\mathbf{x_1 \independent x_2}$\\
    We compute $p(x_1,x_2)$ as
    \begin{align}
      p(x_1,x_2) & = \int p(y_1,y_2,x_1,x_2) \ud y_1 \ud y_2\\
      & =  \int p(y_1 | x_1) p(y_2 | x_2) \phi(y_1,y_2) n(x_1,x_2) p(x_1) p(x_2) \ud y_1 \ud y_2 \\
      & =  n(x_1,x_2) p(x_1) p(x_2) \int p(y_1 | x_1) p(y_2 | x_2) \phi(y_1,y_2) \ud y_1 \ud y_2\\
      & \stackrel{\eqref{eq:ndef}}=  n(x_1,x_2) p(x_1) p(x_2) \frac{1}{n(x_1,x_2)}\\
      & = p(x_1) p(x_2).
    \end{align}
    Since $p(x_1)$ and $p(x_2)$ are the univariate marginals of $x_1$ and $x_2$, respectively, it follows from \eqref{eq:ind1} that $x_1 \independent x_2$.

    \vspace{2ex}
    For $\mathbf{x_1 \independent y_2 \mid y_1, x_2}$\\
    We rewrite $p(y_1,y_2,x_1,x_2)$ as
    \begin{align}
      p(y_1,y_2,x_1,x_2) & =  p(y_1 | x_1) p(y_2 | x_2) \phi(y_1,y_2) n(x_1,x_2) p(x_1) p(x_2)\\
      & =  \left[p(y_1 | x_1) p(x_1)  n(x_1,x_2)\right]\left[ p(y_2 | x_2)\phi(y_1,y_2)p(x_2)\right]\\
      & = \phi_A(x_1,y_1,x_2) \phi_B(y_2,y_1,x_2)
    \end{align}
    With \eqref{eq:ind2}, we have that $x_1 \independent y_2 \mid y_1,x_2$. Note that $p(x_2)$ can be associated either with $\phi_A$ or with $\phi_B$.

    \vspace{2ex}
    For $\mathbf{x_2 \independent y_1 \mid y_2, x_1}$\\ We use here the
    same approach as for $x_1 \independent y_2 \mid y_1, x_2$. (By
    symmetry considerations, we could immediately see that the
    relation holds but let us write it out for clarity).  We rewrite $p(y_1,y_2,x_1,x_2)$ as
    \begin{align}
      p(y_1,y_2,x_1,x_2) & =  p(y_1 | x_1) p(y_2 | x_2) \phi(y_1,y_2) n(x_1,x_2) p(x_1) p(x_2)\\
      & =  \left[  p(y_2 | x_2)  n(x_1,x_2) p(x_2)p(x_1) )\right]\left[ p(y_1 | x_1) \phi(y_1,y_2) ]\right)\\
      & = \tilde{\phi}_A(x_2,x_1,y_2) \tilde{\phi}_B(y_1,y_2,x_1)
    \end{align}
    With \eqref{eq:ind2}, we have that $x_2 \independent y_1 \mid y_2,x_1$. 

  \end{solution}

\item Is there an undirected perfect map for the independencies
  satisfied by $p(y_1,y_2,x_1,x_2)$?

  \begin{solution}
    We write 
    $$p(y_1,y_2,x_1,x_2) = p(y_1 | x_1) p(y_2 | x_2) \phi(y_1,y_2) n(x_1,x_2) p(x_1) p(x_2)$$
    as a Gibbs distribution
    \begin{align}
      p(y_1,y_2,x_1,x_2) & = \phi_1(y_1,x_1) \phi_2(y_2,x_2) \phi_3(y_1,y_2) \phi_4(x_1,x_2) \quad \quad \text{with} \\
      \phi_1(y_1,x_1) & =p(y_1 | x_1)p(x_1)   \\
      \phi_2(y_2,x_2)& =  p(y_2 | x_2)p(x_2)\\
      \phi_3(y_1,y_2)& =  \phi(y_1,y_2) \\
      \phi_4(x_1,x_2)& = n(x_1,x_2). 
    \end{align}
    Visualising it as an undirected graph gives an I-map:
     \begin{center}
      \scalebox{1}{ 
        \begin{tikzpicture}[ugraph]
          \node[cont] (x1) at (0,-0.5) {$x_1$};
          \node[cont] (x2) at (2,-0.5) {$x_2$};
          \node[cont] (y1) at (0,-2) {$y_1$};
          \node[cont] (y2) at (2,-2) {$y_2$};
          \draw(x1) -- (y1);
          \draw(x2) -- (y2);
          \draw(y1) -- (y2);
          \draw(x1) -- (x2);
      \end{tikzpicture}}
     \end{center}
   While the graph implies $x_1 \independent y_2 \mid y_1, x_2$ and
   $x_2 \independent y_1 \mid y_2, x_1$, the independency $x_1
   \independent x_2$ is not represented. Hence the graph is not a
   perfect map. Note further that removing any edge would result in a
   graph that is not an I-map for $\Ind(p)$ anymore. Hence the graph
   is a minimal I-map for $\Ind(p)$ but that we cannot obtain a
   perfect I-map.

  \end{solution}
\item Is there a directed perfect map for the independencies
  satisfied by $p(y_1,y_2,x_1,x_2)$?
   
   \begin{solution} We construct directed minimal I-maps for
   $p(y_1,y_2,x_1,x_2) = p(y_1, y_2 | x_1, x_2)p(x_1) p(x_2)$ for different
   orderings. We will see that they do not represent all independencies in
   $\Ind(p)$ and hence that they are not perfect I-maps.

   To guarantee unconditional independence of $x_1$ and $x_2$, the two
   variables must come first in the orderings (either $x_1$ and then
   $x_2$ or the other way around).
   
   If we use the ordering $x_1,x_2,y_1,y_2$, and that 
   \begin{itemize}
   \item $x_1 \independent x_2$
   \item $y_2 \independent x_1 | y_1, x_2$, which is $y_2 \independent \pre(y_2) \setminus \pi | \pi$ for $\pi = (y_1, x_2)$
   \end{itemize}
   are in $\Ind(p)$, we obtain the following directed minimal I-map:
   \begin{center}
     \scalebox{0.9}{ 
       \begin{tikzpicture}[dgraph]
         \node[cont] (x1) at (0,0) {$x_1$};
         \node[cont] (x2) at (2,0) {$x_2$};
         \node[cont] (y1) at (0,-2) {$y_1$};
         \node[cont] (y2) at (2,-2) {$y_2$};
           \draw(x1) -- (y1);
           \draw(x2) -- (y2);
           \draw(x2) -- (y1);
           \draw(y1) -- (y2);
       \end{tikzpicture}
     }
   \end{center}
   The graphs misses $x_2 \independent y_1 \mid y_2, x_1$.
   
   If we use the ordering $x_1,x_2,y_2,y_1$, and that
   \begin{itemize}
   \item $x_1 \independent x_2$ 
   \item $y_1 \independent x_2 | x_1, y_2$, which is $y_1 \independent \pre(y_1) \setminus \pi | \pi$ for $\pi = (x_1, y_2)$
   \end{itemize}
   are in $\Ind(p)$, we obtain the following directed minimal I-map:
    \begin{center}
      \scalebox{0.9}{ 
        \begin{tikzpicture}[dgraph]
          \node[cont] (x1) at (0,0) {$x_1$};
          \node[cont] (x2) at (2,0) {$x_2$};
          \node[cont] (y1) at (0,-2) {$y_1$};
          \node[cont] (y2) at (2,-2) {$y_2$};
          \draw(x1) -- (y1);
          \draw(x2) -- (y2);
          \draw(y2) -- (y1);
          \draw(x1) -- (y2);
        \end{tikzpicture}
      }
    \end{center}
    The graph misses $x_1\independent y_2 \mid y_1, x_2$.
    
    Moreover, the graphs imply a directionality between $y_1$ and
    $y_2$, or a direct influence of $x_1$ on $y_2$, or of $x_2$ on
    $y_1$, in contrast to the original modelling goals.
    
   \end{solution}

 \item \emph{(advanced)} The following factor graph represents $p(y_1,y_2,x_1,x_2)$:
\begin{center}
      \scalebox{0.75}{ 
        \begin{tikzpicture}[dgraph]
          \node[] (middle) at (1.5,0) {};
          \node[fact, label=left: $p(x_1)$] (f1) at (0,0) {};
          \node[cont, below=of f1] (x1) {$x_1$};
          \node[fact, label=right: $p(x_2)$] (f2) at (3,0) {};
          \node[cont, below=of f2] (x2) {$x_2$};

          \node[fact, below =of x1, label=left: $p(y_1 | x_1)$] (fy1) {};
          \node[fact, below =of x2, label=right: $p(y_2 | x_2)$] (fy2) {};

          \node[cont, below =of fy1] (y1) {$y_1$};
          \node[cont, below =of fy2] (y2) {$y_2$};

          \node[fact, below= 2 of middle, label=above: $n(x_1 \,x_2)$] (n) {};
          \node[fact, below= 1 of n, label={[label distance=0.25cm]below: $\phi(y_1\,y_2)$}] (phi) {};

          \draw[-] (f1) -- (x1);
          \draw[-] (f2) -- (x2);
          \draw  (x1) -- (fy1);
          \draw  (x2) -- (fy2);
          \draw  (x1) -- (n);
          \draw  (x2) -- (n);
          \draw  (fy1) -- (y1);
          \draw  (fy2) -- (y2);
          \draw  (phi) -- (y1);
          \draw  (phi) -- (y2);
          \draw[dashed] (n) -- (y1);
          \draw[dashed] (n) -- (y2);
      \end{tikzpicture}}
    \end{center}

Use the separation rules for factor graphs to verify that we can find all independence relations. The separation rules are \citep[see][Section 4.4.1]{Barber2012}, or the original paper by \citet{Frey2003}:\\[1ex]
``If all paths are blocked, the variables are conditionally independent. A path is blocked if one or more of the following conditions is satisfied:
\begin{enumerate}
\item One of the variables in the path is in the conditioning set.
\item One of the variables or factors in the path has two incoming edges that are part of the path (variable or factor collider), and neither the variable or factor nor any of its descendants are in the conditioning set.''
\end{enumerate}

Remarks:
\begin{itemize}
  \item ``one or more of the following'' should best be read as ``one of the following''. 
  \item ``incoming edges'' means directed incoming edges
  \item the descendants of a variable or factor node are all the variables that you can reach by following a path (containing directed or directed edges, but for directed edges, all directions have to be consistent)
  \item In the graph we have dashed directed edges: they do count when you determine the descendants but they do not contribute to paths. For example, $y_1$ is a descendant of the $n(x_1,x_2)$ factor node but $x_1 - n - y_2$ is not a path.
\end{itemize}

\begin{solution}
  $\mathbf{x_1 \independent x_2}$\\
  There are two paths from $x_1$ to $x_2$ marked with red and blue below:
\begin{center}
      \scalebox{0.75}{ 
        \begin{tikzpicture}[dgraph]
          \node[] (middle) at (1.5,0) {};
          \node[fact, label=left: $p(x_1)$] (f1) at (0,0) {};
          \node[cont, below=of f1] (x1) {$x_1$};
          \node[fact, label=right: $p(x_2)$] (f2) at (3,0) {};
          \node[cont, below=of f2] (x2) {$x_2$};

          \node[fact, below =of x1, label=left: $p(y_1 | x_1)$] (fy1) {};
          \node[fact, below =of x2, label=right: $p(y_2 | x_2)$] (fy2) {};

          \node[cont, below =of fy1] (y1) {$y_1$};
          \node[cont, below =of fy2] (y2) {$y_2$};

          \node[fact, below= 2 of middle, label=above: $n(x_1 \,x_2)$] (n) {};
          \node[fact, below= 1 of n, label={[label distance=0.25cm]below: $\phi(y_1\,y_2)$}] (phi) {};

          \draw[-] (f1) -- (x1);
          \draw[-] (f2) -- (x2);
          \draw[blue]  (x1) -- (fy1);
          \draw[blue]  (x2) -- (fy2);
          \draw[red]  (x1) -- (n);
          \draw[red]  (x2) -- (n);
          \draw[blue]  (fy1) -- (y1);
          \draw[blue]  (fy2) -- (y2);
          \draw[blue]  (phi) -- (y1);
          \draw[blue]  (phi) -- (y2);
          \draw[dashed] (n) -- (y1);
          \draw[dashed] (n) -- (y2);
      \end{tikzpicture}}
    \end{center}
  Both the blue and red path are blocked by condition 2. 

$\mathbf{x_1 \independent y_2 \mid y_1, x_2}$\\
There are two paths from $x_1$ to $y_2$ marked with red and blue below:
\begin{center}
      \scalebox{0.75}{ 
        \begin{tikzpicture}[dgraph]
          \node[] (middle) at (1.5,0) {};
          \node[fact, label=left: $p(x_1)$] (f1) at (0,0) {};
          \node[cont, below=of f1] (x1) {$x_1$};
          \node[fact, label=right: $p(x_2)$] (f2) at (3,0) {};
          \node[contobs, below=of f2] (x2) {$x_2$};

          \node[fact, below =of x1, label=left: $p(y_1 | x_1)$] (fy1) {};
          \node[fact, below =of x2, label=right: $p(y_2 | x_2)$] (fy2) {};

          \node[contobs, below =of fy1] (y1) {$y_1$};
          \node[cont, below =of fy2] (y2) {$y_2$};

          \node[fact, below= 2 of middle, label=above: $n(x_1 \,x_2)$] (n) {};
          \node[fact, below= 1 of n, label={[label distance=0.25cm]below: $\phi(y_1\,y_2)$}] (phi) {};

          \draw[-] (f1) -- (x1);
          \draw[-] (f2) -- (x2);
          \draw[blue]  (x1) -- (fy1);
          \draw[red]  (x2) -- (fy2);
          \draw[red]  (x1) -- (n);
          \draw[red]  (x2) -- (n);
          \draw[blue]  (fy1) -- (y1);
          \draw[red]  (fy2) -- (y2);
          \draw[blue]  (phi) -- (y1);
          \draw[blue]  (phi) -- (y2);
          \draw[dashed] (n) -- (y1);
          \draw[dashed] (n) -- (y2);
      \end{tikzpicture}}
\end{center}
The observed variables are marked in blue. For the red path, the
observed $x_2$ blocks the path (condition 1). Note that the
$n(x_1,x_2)$ node would be open by condition 2. The blue path is
blocked by condition 1 too. In directed graphical models, the $y_1$
node would be open, but here while condition 2 does not apply,
condition 1 still applies (note the \emph{one or more of ...} in the separation rules), so that
the path is blocked.

$\mathbf{x_2 \independent y_1 \mid y_2, x_1}$\\
There are two paths from $x_2$ to $y_1$ marked with red and blue below:
\begin{center}
      \scalebox{0.75}{ 
        \begin{tikzpicture}[dgraph]
          \node[] (middle) at (1.5,0) {};
          \node[fact, label=left: $p(x_1)$] (f1) at (0,0) {};
          \node[contobs, below=of f1] (x1) {$x_1$};
          \node[fact, label=right: $p(x_2)$] (f2) at (3,0) {};
          \node[cont, below=of f2] (x2) {$x_2$};

          \node[fact, below =of x1, label=left: $p(y_1 | x_1)$] (fy1) {};
          \node[fact, below =of x2, label=right: $p(y_2 | x_2)$] (fy2) {};

          \node[cont, below =of fy1] (y1) {$y_1$};
          \node[contobs, below =of fy2] (y2) {$y_2$};

          \node[fact, below= 2 of middle, label=above: $n(x_1 \,x_2)$] (n) {};
          \node[fact, below= 1 of n, label={[label distance=0.25cm]below: $\phi(x_1\,x_2)$}] (phi) {};

          \draw[-] (f1) -- (x1);
          \draw[-] (f2) -- (x2);
          \draw[blue]  (x1) -- (fy1);
          \draw[red]  (x2) -- (fy2);
          \draw[blue]  (x1) -- (n);
          \draw[blue]  (x2) -- (n);
          \draw[blue]  (fy1) -- (y1);
          \draw[red]  (fy2) -- (y2);
          \draw[red]  (phi) -- (y1);
          \draw[red]  (phi) -- (y2);
          \draw[dashed] (n) -- (y1);
          \draw[dashed] (n) -- (y2);
      \end{tikzpicture}}
\end{center}
The same reasoning as before yields the result.

Finally note that $x_1$ and $x_2$ are not independent given $y_1$ or
$y_2$ because the upper path through $n(x_1,x_2)$ is not blocked
whenever $y_1$ or $y_2$ are observed (condition 2).\\[1ex]
{\small Credit: this example is discussed in the original paper by B. Frey (Figure 6).}

\end{solution}   
\end{exenumerate}

\chapter{Factor Graphs and Message Passing} 
\minitoc

\ex{Conversion to factor graphs}
\label{ex:DGM-vs-UGM}
\begin{exenumerate}

   \item Draw an undirected graph and an undirected factor graph for
     $p(x_1,x_2,x_3) = p(x_1) p(x_2) p(x_3 | x_1,x_2)$

     \begin{solution}
       \begin{center}
         \scalebox{0.9}{ 
           \begin{tikzpicture}[ugraph]
             \node[cont] (x1) at (0,2) {$x_1$};
             \node[cont] (x2) at (2,2) {$x_2$};
             \node[cont] (x3) at (1,-0.2) {$x_3$};
    
             \draw(x1) -- (x2);
             \draw(x2) -- (x3);
             \draw(x3) -- (x1);
           \end{tikzpicture}
           \begin{tikzpicture}[ugraph]
             \node[cont] (x1) at (0,2) {$x_1$};
             \node[cont] (x2) at (2,2) {$x_2$};
             \node[fact, label=right: $p(x_3 | x_1\, x_2)$] (f3) at (1,1) {};
             \node[cont] (x3) at (1,-0.2) {$x_3$};
             \node[fact, label=left: $p(x_1)$] (f1) at (0,3) {};
             \node[fact, label=right: $p(x_2)$] (f2) at (2,3) {};

                \draw(x1) -- (f3);
                \draw(x2) -- (f3);
                \draw(f3) -- (x3);
                \draw[-](f1) -- (x1);
                \draw[-](f2) -- (x2);
           \end{tikzpicture}
         }
\end{center}
     \end{solution}
     
     \item Draw an undirected factor graph for the directed graphical model defined by the graph below.

  \begin{center}
  \scalebox{0.9}{
    \begin{tikzpicture}[dgraph]
      \node[cont] (y1) at (0,0) {$y_1$};
      \node[cont] (y2) at (2,0) {$y_2$};
      \node[cont] (y3) at (4,0) {$y_3$};
      \node[cont] (y4) at (6,0) {$y_4$};
      \node[cont] (x1) at (0,2) {$x_1$};
      \node[cont] (x2) at (2,2) {$x_2$};
      \node[cont] (x3) at (4,2) {$x_3$};
      \node[cont] (x4) at (6,2) {$x_4$};
      \draw(x1)--(y1);\draw(x2)--(y2);\draw(x3)--(y3);\draw(x4)--(y4);
      \draw(x1)--(x2);\draw(x2)--(x3);\draw(x3)--(x4);
  \end{tikzpicture}}
\end{center}

  \begin{solution}
    The graph specifies probabilistic models that factorise as
    $$p(x_1, \ldots, x_4, y_1, \ldots, y_4) = p(x_1) p(y_1 | x_1)
    \prod_{i=2}^4 p(y_i |x_i) p(x_i | x_{i-1})$$
   It is the graph for a hidden Markov model. The corresponding factor graph is shown below.
\begin{center}
  \scalebox{0.9}{
    \begin{tikzpicture}[ugraph]
      \node[cont] (y1) at (0,0) {$y_1$};
      \node[fact, label={[xshift=0.15cm,  yshift=0cm]left: $p(y_1 | x_1)$}] (fy1) at (0,1) {};
      \node[cont] (y2) at (2,0) {$y_2$};
      \node[fact, label={[xshift=0.15cm,  yshift=0cm]left: $p(y_2 | x_2)$}] (fy2) at (2,1) {};
      \node[cont] (y3) at (4,0) {$y_3$};
      \node[fact, label={[xshift=0.15cm,  yshift=0cm]left: $p(y_3 | x_3)$}] (fy3) at (4,1) {};
      \node[cont] (y4) at (6,0) {$y_4$};
      \node[fact, label={[xshift=0.15cm,  yshift=0cm]left: $p(y_4 | x_4)$}] (fy4) at (6,1) {};
      
      \node[fact, label=left: $p(x_1)$] (f1) at (-1,2) {};
      \node[cont] (x1) at (0,2) {$x_1$};
      \node[fact, label=above: $p(x_2 | x_1)$] (f2) at (1,2) {};
      \node[cont] (x2) at (2,2) {$x_2$};
      \node[fact, label=above: $p(x_3 | x_2)$] (f3) at (3,2) {};
      \node[cont] (x3) at (4,2) {$x_3$};
      \node[fact, label=above: $p(x_4 | x_3)$] (f4) at (5,2) {};
      \node[cont] (x4) at (6,2) {$x_4$};
      \draw(x1)--(y1);\draw(x2)--(y2);\draw(x3)--(y3);\draw(x4)--(y4);
      \draw(x1)--(x2);\draw(x2)--(x3);\draw(x3)--(x4);

      \draw(f1) -- (x1);
      
  \end{tikzpicture}}
\end{center}

  \end{solution}

\item Draw the moralised graph and an undirected factor graph for directed graphical models defined by the graph below (this kind of graph is called a polytree: there are no loops but a node may have more than one parent).

  \begin{center}
    \scalebox{0.9}{ 
      \begin{tikzpicture}[dgraph]
        \node[cont] (x1) at (0,0) {$x_1$};
        \node[cont] (x2) at (2,0) {$x_2$};
        \node[cont] (x3) at (-1,-1.5) {$x_3$};
        \node[cont] (x4) at (1,-1.5) {$x_4$};
        \node[cont] (x5) at (0,-3) {$x_5$};
        \node[cont] (x6) at (2,-3) {$x_6$};
        
        \draw(x1) -- (x3);
        \draw(x1) -- (x4);
        \draw(x2) -- (x4);
        \draw(x4) -- (x5);
        \draw(x4) -- (x6);
      \end{tikzpicture}
      }
  \end{center}

  \begin{solution}
    The moral graph is obtained by connecting the parents of the collider node $x_4$. See the graph on the left in the figure below.

    For the factor graph, we note that the directed graph defines the
    following class of probabilistic models
    $$p(x_1, \ldots x_6) = p(x_1) p(x_2) p(x_3 | x_1) p(x_4|x_1,x_2) p(x_5|x_4) p(x_6|x_4)$$ 
    This gives the factor graph on right in the figure below.
    
    \begin{center}
      \scalebox{0.9}{ 
        \begin{tikzpicture}[ugraph]
          \node[cont] (x1) at (0,0) {$x_1$};
          \node[cont] (x2) at (2,0) {$x_2$};
          \node[cont] (x3) at (-1,-1.5) {$x_3$};
          \node[cont] (x4) at (1,-1.5) {$x_4$};
          \node[cont] (x5) at (0,-3) {$x_5$};
          \node[cont] (x6) at (2,-3) {$x_6$};
          
          \draw(x1) -- (x2);
          \draw(x1) -- (x3);
          \draw(x1) -- (x4);
          \draw(x2) -- (x4);
          \draw(x4) -- (x5);
          \draw(x4) -- (x6);
        \end{tikzpicture}
        \hspace{10ex}
        \begin{tikzpicture}[ugraph]
          \node[fact, label=left: $p(x_1)$] (f1) at (-1,0) {};     
          \node[cont] (x1) at (0,0) {$x_1$};

          \node[fact, label=right: $p(x_2)$] (f2) at (3,0) {};     
          \node[cont] (x2) at (2,0) {$x_2$};

          \node[fact, label=left: $p(x_3|x_1)$] (f3) at (-1,-0.7) {};     
          \node[cont] (x3) at (-1,-1.5) {$x_3$};

          \node[fact, label=right: $p(x_4 | x_1\,x_2)$] (f4) at (1,-0.7) {};
          \node[cont] (x4) at (1,-1.5) {$x_4$};

          \node[fact, label=left: $p(x_5 | x_4)$] (f5) at (0,-2.2) {};
          \node[cont] (x5) at (0,-3) {$x_5$};

          \node[fact, label=right: $p(x_6 | x_4)$] (f6) at (2,-2.2) {};
          \node[cont] (x6) at (2,-3) {$x_6$};

          \draw(f1) -- (x1);
          \draw(f2) -- (x2);
          \draw(x1) -- (f4);
          \draw(x1) -- (f3);
          
          \draw(x2) -- (f4);
          \draw(f4) -- (x4);

          \draw(f3) -- (x3);
          
          \draw(x4) -- (f5);
          \draw(f5) -- (x5);

          \draw(x4) -- (f6);
          \draw(f6) -- (x6);
        \end{tikzpicture}
      }
    \end{center}

    Note:
    \begin{itemize}
    \item The moral graph contains a loop while the factor graph does not. The factor graph is still a polytree. This can be exploited for inference.
    \item One may choose to group some factors together in order to obtain a factor graph with a particular structure (see factor graph below)
    \end{itemize}
    \vspace{1ex}

    \begin{center}
        \begin{tikzpicture}[ugraph]
          \node[cont] (x1) at (0,0) {$x_1$};
          
          \node[cont] (x2) at (2,0) {$x_2$};

          \node[fact, label=left: $p(x_3|x_1)$] (f3) at (-1,-0.7) {};     
          \node[cont] (x3) at (-1,-1.5) {$x_3$};

          \node[fact, label=right: $p(x_4 | x_1\,x_2)p(x_1)p(x_2)$] (f4) at (1,-0.7) {};
          \node[cont] (x4) at (1,-1.5) {$x_4$};

          \node[fact, label=right: $p(x_5 | x_4)p(x_6|x_4)$] (f5) at (1,-2.2) {};
          \node[cont] (x5) at (0,-3) {$x_5$};
          \node[cont] (x6) at (2,-3) {$x_6$};

          \draw(x1) -- (f4);
          \draw(x1) -- (f3);
          
          \draw(x2) -- (f4);
          \draw(f4) -- (x4);

          \draw(f3) -- (x3);
          
          \draw(x4) -- (f5);
          \draw(f5) -- (x5);
          \draw(f5) -- (x6);
        \end{tikzpicture}
  \end{center}
    
  \end{solution}

\end{exenumerate}

\ex{Sum-product message passing}
\label{ex:sum-product-message-passing}

We here consider the following factor tree:

  \begin{center}
    \begin{tikzpicture}[ugraph]
      \node[fact, label=above: $\phi_A$] (fa) at (0,0) {} ; %
      \node[cont] (x1) at (1.5,0)  {$x_1$} ; %
      \node[fact, label=below: $\phi_C$] (fc) at (3,0) {} ; %
      \node[cont] (x2) at (3,1)  {$x_2$} ; %
      \node[fact, label=left: $\phi_B$] (fb) at (3,2) {} ; %
      \node[cont] (x3) at (4.5,0)  {$x_3$} ; %
      \node[fact, label=above: $\phi_D$] (fd) at (5.5,1) {} ; %
      \node[cont] (x4) at (7,1)  {$x_4$} ; %
      \node[fact, label=above: $\phi_E$] (fe) at (5.5,-1) {} ; %
      \node[cont] (x5) at (7,-1)  {$x_5$} ; %
      \node[fact, label=above: $\phi_F$] (ff) at (8.5,-1) {} ; %
      \draw (fa) -- (x1);
      \draw (x1) -- (fc);
      \draw (fc) -- (x2);
      \draw (x2) -- (fb);
      \draw (fc) -- (x3);
      \draw (x3) -- (fd);
      \draw (x3) -- (fe);
      \draw (fd) -- (x4);
      \draw (fe) -- (x5);
      \draw (x5) -- (ff);
    \end{tikzpicture}
  \end{center}
  Let all variables be binary, $x_i \in \{0,1\}$, and the factors be defined as follows:
  \begin{center}
    \begin{tabular}{ll}
      \toprule
      $x_1$ & $\phi_A$\\
      \midrule
      0 & 2\\
      1 & 4\\
      \bottomrule
    \end{tabular}
    \hfill
    \begin{tabular}{ll}
      \toprule
      $x_2$ & $\phi_B$\\
      \midrule
      0 & 4\\
      1 & 4\\
      \bottomrule
    \end{tabular}
    \hfill
    \begin{tabular}{llll}
      \toprule
      $x_1$ & $x_2$ & $x_3$ & $\phi_C$\\
      \midrule
    0 & 0 & 0 & 4 \\
    1 & 0 & 0 & 2 \\
    0 & 1 & 0 & 2 \\
    1 & 1 & 0 & 6 \\
    0 & 0 & 1 & 2 \\
    1 & 0 & 1 & 6 \\
    0 & 1 & 1 & 6 \\
    1 & 1 & 1 & 4 \\
      \bottomrule
    \end{tabular}
    \hfill
    \begin{tabular}{lll}
      \toprule
      $x_3$ & $x_4$ & $\phi_D$\\
      \midrule
    0 & 0 &  8 \\
    1 & 0 &  2 \\
    0 & 1 &  2 \\
    1 & 1 &  6 \\
      \bottomrule
    \end{tabular}
    \hfill
    \begin{tabular}{lll}
      \toprule
      $x_3$ & $x_5$ & $\phi_E$\\
      \midrule
    0 & 0 &  3 \\
    1 & 0 &  6 \\
    0 & 1 &  6 \\
    1 & 1 &  3 \\
      \bottomrule
    \end{tabular}
   \hfill
    \begin{tabular}{ll}
      \toprule
      $x_5$ & $\phi_F$\\
      \midrule
      0 & 1\\
      1 & 8\\
      \bottomrule
    \end{tabular}

  \end{center}
  
\begin{exenumerate}

\item Mark the graph with arrows indicating all messages that need to
  be computed for the computation of $p(x_1)$.

  \begin{solution}
    
    \begin{center}
      \begin{tikzpicture}[ugraph]
        \node[fact, label=above: $\phi_A$] (fa) at (0,0) {} ; %
        \node[cont] (x1) at (1.5,0)  {$x_1$} ; %
        \node[fact, label=below: $\phi_C$] (fc) at (3,0) {} ; %
        \node[cont] (x2) at (3,1)  {$x_2$} ; %
        \node[fact, label=left: $\phi_B$] (fb) at (3,2) {} ; %
        \node[cont] (x3) at (4.5,0)  {$x_3$} ; %
        \node[fact, label=above: $\phi_D$] (fd) at (5.5,1) {} ; %
        \node[cont] (x4) at (7,1)  {$x_4$} ; %
        \node[fact, label=above: $\phi_E$] (fe) at (5.5,-1) {} ; %
        \node[cont] (x5) at (7,-1)  {$x_5$} ; %
        \node[fact, label=above: $\phi_F$] (ff) at (8.5,-1) {} ; %
        \draw (fa) -- (x1) node[midway,above,sloped] {$\rightarrow$};
        \draw (x1) -- (fc) node[midway,below,sloped] {$\leftarrow$};
        \draw (fc) -- (x2) node[midway,left] {$\downarrow$} ;
        \draw (x2) -- (fb) node[midway,left] {$\downarrow$};
        \draw (fc) -- (x3) node[midway,below] {$\leftarrow$};
        \draw (x3) -- (fd) node[midway,above,sloped] {$\leftarrow$};
        \draw (fd) -- (x4) node[midway,above] {$\leftarrow$};
        \draw (x3) -- (fe) node[midway,below,sloped] {$\leftarrow$};
        \draw (fe) -- (x5) node[midway,below] {$\leftarrow$};
        \draw (ff) -- (x5) node[midway,below] {$\leftarrow$};
      \end{tikzpicture}
    \end{center}
    
  \end{solution}
  
\item \label{q:messages} Compute the messages that you have
  identified.

  Assuming that the computation of the messages is scheduled according
  to a common clock, group the messages together so that all messages
  in the same group can be computed in parallel during a clock
  cycle.
     
  \begin{solution}
    Since the variables are binary, each message can be represented as a two-dimensional vector. We use the convention that the first element of the vector corresponds to the message for $x_i =0$ and the second element to the message for $x_i=1$. For example,
    \begin{equation}
      \fxmessb{\phi_A}{x_1}= \begin{pmatrix}
        2 \\
        4 \\
      \end{pmatrix}
    \end{equation}
means that the message $\fxmess{\phi_A}{x_1}{1}(x_1)$ equals 2 for $x_1=0$, i.e.\ $\fxmess{\phi_A}{x_1}{1}(0)=2$.

The following figure shows a grouping (scheduling) of the computation of the messages. 
    \begin{figure}[h]
    \begin{center}
      \begin{tikzpicture}[ugraph]
        \node[fact, label=above: $\phi_A$] (fa) at (0,0) {} ; %
        \node[cont] (x1) at (1.5,0)  {$x_1$} ; %
        \node[fact, label=below: $\phi_C$] (fc) at (3,0) {} ; %
        \node[cont] (x2) at (3,1)  {$x_2$} ; %
        \node[fact, label=left: $\phi_B$] (fb) at (3,2) {} ; %
        \node[cont] (x3) at (4.5,0)  {$x_3$} ; %
        \node[fact, label=above: $\phi_D$] (fd) at (5.5,1) {} ; %
        \node[cont] (x4) at (7,1)  {$x_4$} ; %
        \node[fact, label=above: $\phi_E$] (fe) at (5.5,-1) {} ; %
        \node[cont] (x5) at (7,-1)  {$x_5$} ; %
        \node[fact, label=above: $\phi_F$] (ff) at (8.5,-1) {} ; %
        \draw (fa) -- (x1) node[midway,above,sloped] {$\substack{[1] \\ \rightarrow}$};
        \draw (x1) -- (fc) node[midway,below,sloped] {$\substack{\leftarrow\\ [5]}$};
        
        \draw (fc) -- (x2) node[midway,left] {{\scriptsize $[2]$}$\downarrow$} ;
        \draw (x2) -- (fb) node[midway,left] {{\scriptsize $[1]$}$\downarrow$};
        \draw (fc) -- (x3) node[midway,below] {$\substack{\leftarrow \\ [4]}$};
        \draw (x3) -- (fd) node[midway,above,sloped] {$\substack{[2] \\\leftarrow}$};
        \draw (fd) -- (x4) node[midway,above] {$\substack{[1] \\ \leftarrow}$};
        \draw (x3) -- (fe) node[midway,below,sloped] {$\substack{\leftarrow \\ [3]}$};
        \draw (fe) -- (x5) node[midway,below] {$\substack{\leftarrow \\ [2]}$};
        \draw (ff) -- (x5) node[midway,below] {$\substack{\leftarrow \\ [1]}$};
      \end{tikzpicture}
    \end{center}
    \end{figure}

    { \bf Clock cycle 1:}\\
    \begin{align}
      \fxmessb{\phi_A}{x_1} & = \begin{pmatrix}
        2 \\
        4 \\
      \end{pmatrix}
      & 
      \fxmessb{\phi_B}{x_2} & = \begin{pmatrix}
        4 \\
        4 \\
      \end{pmatrix}
      &
      \xfmessb{x_4}{\phi_D} & = \begin{pmatrix}
        1 \\
        1 \\
      \end{pmatrix}
      &
      \fxmessb{\phi_F}{x_5} & = \begin{pmatrix}
        1 \\
        8 \\
      \end{pmatrix}
    \end{align}
    
    { \bf Clock cycle 2:}\\
    \begin{align}
      \xfmessb{x_2}{\phi_C} = \fxmessb{\phi_B}{x_2}  & = \begin{pmatrix}
        4 \\
        4 \\
      \end{pmatrix}
      &
      \xfmessb{x_5}{\phi_E} =  \fxmessb{\phi_F}{x_5} & = \begin{pmatrix}
        1 \\
        8 \\
      \end{pmatrix}
    \end{align}
    Message $\fxmess{\phi_D}{x_3}{1}$ is defined as
    \begin{align}
      \fxmess{\phi_D}{x_3}{1}(x_3) & = \sum_{x_4} \phi_D(x_3,x_4) \xfmess{x_4}{\phi_D}{1}(x_4)\end{align}
    so that
    \begin{align}
      \fxmess{\phi_D}{x_3}{1}(0) & = \sum_{x_4=0}^1 \phi_D(0,x_4) \xfmess{x_4}{\phi_D}{1}(x_4) \\
      & = \phi_D(0,0) \xfmess{x_4}{\phi_D}{1}(0) + \phi_D(0,1) \xfmess{x_4}{\phi_D}{1}(1) \\
      & = 8 \cdot 1  + 2 \cdot 1\\
      & = 10\\
      \fxmess{\phi_D}{x_3}{1}(1) & = \sum_{x_4=0}^1 \phi_D(1,x_4) \xfmess{x_4}{\phi_D}{1}(x_4) \\
      & = \phi_D(1,0) \xfmess{x_4}{\phi_D}{1}(0) + \phi_D(1,1) \xfmess{x_4}{\phi_D}{1}(1) \\
      & = 2 \cdot 1 + 6 \cdot 1\\
      & = 8
    \end{align}
    and thus
    \begin{align}
      \fxmessb{\phi_D}{x_3} &=  \begin{pmatrix}
        10 \\
        8
      \end{pmatrix}.
    \end{align}
    The above computations can be written more compactly in matrix notation. Let $\pmb{\phi_D}$ be the matrix that contains the outputs of $\phi_D(x_3, x_4)$ 
    \begin{align}
      \pmb{\phi_D} & = \begin{pmatrix}
        \phi_D(x_3=0,x_4=0) &  \phi_D(x_3=0,x_4=1) \\
        \phi_D(x_3=1,x_4=0) & \phi_D(x_3=1,x_4=1)  
      \end{pmatrix}
       = \begin{pmatrix}
        8 & 2 \\
        2 & 6
      \end{pmatrix}.
    \end{align}
    We can then write $\fxmessb{\phi_D}{x_3}$ in terms of a matrix vector product,
    \begin{align}
      \fxmessb{\phi_D}{x_3} & = \pmb{\phi_D} \xfmessb{x_4}{\phi_D}.
    \end{align}
    
    { \bf Clock cycle 3:}\\
    Representing the factor $\phi_E$ as matrix $\pmb{\phi_E}$,
    \begin{align}
      \pmb{\phi_E} & = \begin{pmatrix}
        \phi_E(x_3=0,x_5=0) &  \phi_E(x_3=0,x_5=1) \\
        \phi_E(x_3=1,x_5=0) & \phi_E(x_3=1,x_5=1)  
      \end{pmatrix}
      = \begin{pmatrix}
        3 & 6 \\
        6 & 3
      \end{pmatrix},
    \end{align}
      we can write
      \begin{align}
        \fxmess{\phi_E}{x_3}{1}(x_3) & = \sum_{x_5} \phi_E(x_3,x_5) \xfmess{x_5}{\phi_E}{1}(x_5)
      \end{align}
      as a matrix vector product,
      \begin{align}
        \fxmessb{\phi_E}{x_3} & =  \pmb{\phi_E} \xfmessb{x_5}{\phi_E}\\
        & = \begin{pmatrix}
          3 & 6 \\
          6 & 3
        \end{pmatrix}
        \begin{pmatrix}
          1 \\
          8 \\
        \end{pmatrix}\\
        & =  \begin{pmatrix}
          51 \\
          30 \\
        \end{pmatrix}.
      \end{align}
      
      { \bf Clock cycle 4:}\\
      Variable node $x_3$ has received all incoming messages, and can thus output $\xfmess{x_3}{\phi_C}{1}$,
      \begin{align}
        \xfmess{x_3}{\phi_C}{1}(x_3) & = \fxmess{\phi_D}{x_3}{1}(x_3)\fxmess{\phi_E}{x_3}{1}(x_3).
      \end{align}
      Using $\odot$ to denote element-wise multiplication of two vectors, we have 
      \begin{align}
        \xfmessb{x_3}{\phi_C} & = \fxmessb{\phi_D}{x_3} \odot \fxmessb{\phi_E}{x_3}\\
        & = \begin{pmatrix}
        10 \\
        8
        \end{pmatrix}
        \odot
        \begin{pmatrix}
          51 \\
          30 \\
        \end{pmatrix}\\
        & = \begin{pmatrix}
          510 \\
          240
        \end{pmatrix}.
      \end{align}

      { \bf Clock cycle 5:}\\
      Factor node $\phi_C$ has received all incoming messages, and can thus output $\fxmess{\phi_C}{x_1}{1}$,
      \begin{align}
        \fxmess{\phi_C}{x_1}{1}(x_1) & = \sum_{x_2,x_3} \phi_C(x_1,x_2,x_3) \xfmess{x_2}{\phi_C}{1}(x_2)\xfmess{x_3}{\phi_C}{1}(x_3).
      \end{align}      
      Writing out the sum for $x_1=0$ and $x_1=1$ gives
      \begin{align}
        \fxmess{\phi_C}{x_1}{1}(0) = &  \sum_{x_2,x_3} \phi_C(0,x_2,x_3) \xfmess{x_2}{\phi_C}{1}(x_2) \xfmess{x_3}{\phi_C}{1}(x_3)\\
                                   = &  \phi_C(0,x_2,x_3) \xfmess{x_2}{\phi_C}{1}(x_2) \xfmess{x_3}{\phi_C}{1}(x_3) \mid_{(x_2,x_3)=(0,0)} +\\
                                   & \phi_C(0,x_2,x_3) \xfmess{x_2}{\phi_C}{1}(x_2) \xfmess{x_3}{\phi_C}{1}(x_3) \mid_{(x_2,x_3)=(1,0)} +\\
                                   & \phi_C(0,x_2,x_3) \xfmess{x_2}{\phi_C}{1}(x_2) \xfmess{x_3}{\phi_C}{1}(x_3) \mid_{(x_2,x_3)=(0,1)} +\\
                                   & \phi_C(0,x_2,x_3) \xfmess{x_2}{\phi_C}{1}(x_2) \xfmess{x_3}{\phi_C}{1}(x_3) \mid_{(x_2,x_3)=(1,1)}\\
                                   = & 4 \cdot 4 \cdot 510 +\\
                                   &  2 \cdot 4 \cdot 510 +  \\
                                   &  2 \cdot 4 \cdot 240  + \\
                                   &  6 \cdot 4 \cdot 240 \\
                                   = & 19920\\
        \fxmess{\phi_C}{x_1}{1}(1) = &  \sum_{x_2,x_3} \phi_C(1,x_2,x_3) \xfmess{x_2}{\phi_C}{1}(x_2) \xfmess{x_3}{\phi_C}{1}(x_3)\\
                                   = &  \phi_C(1,x_2,x_3) \xfmess{x_2}{\phi_C}{1}(x_2) \xfmess{x_3}{\phi_C}{1}(x_3) \mid_{(x_2,x_3)=(0,0)} +\\
                                   & \phi_C(1,x_2,x_3) \xfmess{x_2}{\phi_C}{1}(x_2) \xfmess{x_3}{\phi_C}{1}(x_3) \mid_{(x_2,x_3)=(1,0)} +\\
                                   & \phi_C(1,x_2,x_3) \xfmess{x_2}{\phi_C}{1}(x_2) \xfmess{x_3}{\phi_C}{1}(x_3) \mid_{(x_2,x_3)=(0,1)} +\\
                                   & \phi_C(1,x_2,x_3) \xfmess{x_2}{\phi_C}{1}(x_2) \xfmess{x_3}{\phi_C}{1}(x_3) \mid_{(x_2,x_3)=(1,1)}\\
                                   = & 2 \cdot 4 \cdot 510 +\\
                                   &  6 \cdot 4 \cdot 510 +  \\
                                   &  6 \cdot 4 \cdot 240  + \\
                                   &  4 \cdot 4 \cdot 240 \\
                                   = & 25920                 
      \end{align}
      and hence
      \begin{align}
        \fxmessb{\phi_C}{x_1} &= \begin{pmatrix}
          19920\\
          25920
        \end{pmatrix}
      \end{align}
     
      After step 5, variable node $x_1$ has received all incoming messages and the marginal can be computed.

In addition to the messages needed for computation of $p(x_1)$ one can
compute \emph{all} messages in the graph in five clock cycles, see
Figure \ref{fig:all_messages}. This means that \emph{all} marginals,
as well as the joints of those variables sharing a factor node, are
available after five clock cycles.

 \begin{figure}[h]
   \begin{center}
     \scalebox{1.6}{
      \begin{tikzpicture}[ugraph]
        \node[fact, label=above: {\tiny $\phi_A$}] (fa) at (0,0) {} ; %
        \node[cont] (x1) at (1.5,0)  {$x_1$} ; %
        \node[fact, label=below:  {\tiny$\phi_C$}] (fc) at (3,0) {} ; %
        \node[cont] (x2) at (3,1)  {$x_2$} ; %
        \node[fact, label=left: {\tiny $\phi_B$}] (fb) at (3,2) {} ; %
        \node[cont] (x3) at (4.5,0)  {$x_3$} ; %
        \node[fact, label=above: {\tiny  $\phi_D$}] (fd) at (5.5,1) {} ; %
        \node[cont] (x4) at (7,1)  {$x_4$} ; %
        \node[fact, label=above: {\tiny  $\phi_E$}] (fe) at (5.5,-1) {} ; %
        \node[cont] (x5) at (7,-1)  {$x_5$} ; %
        \node[fact, label=above:  {\tiny $\phi_F$}] (ff) at (8.5,-1) {} ; %
        \draw (fa) -- (x1) node[midway,above,sloped] {$\substack{[1] \\ \rightarrow}$};
        \draw (x1) -- (fc) node[midway,below,sloped] {$\substack{\leftarrow\\ [5]}$} node[near start,above,sloped] {$\substack{[2] \\ \rightarrow}$};
        
        \draw (fc) -- (x2) node[midway,left] {{\tiny $[2]$}$\downarrow$}  node[midway,right] {$\uparrow${\tiny $[5]$}} ;
        \draw (x2) -- (fb) node[midway,left] {{\tiny $[1]$}$\downarrow$};
        \draw (fc) -- (x3) node[midway,below] {$\substack{\leftarrow \\ [4]}$} node[near end,above] {$\substack{[3] \\ \rightarrow }$};
        \draw (x3) -- (fd) node[midway,above,sloped] {$\substack{[2] \\\leftarrow}$} node[midway,below,sloped] {$\substack{\rightarrow \\ [4]}$};
        \draw (fd) -- (x4) node[midway,above] {$\substack{[1] \\ \leftarrow}$} node[midway,below] {$\substack{\rightarrow \\ [5]}$}; 
        \draw (x3) -- (fe) node[midway,below,sloped] {$\substack{\leftarrow \\ [3]}$}  node[midway,above,sloped] {$\substack{[4] \\ \rightarrow}$};
        \draw (fe) -- (x5) node[midway,below] {$\substack{\leftarrow \\ [2]}$} node[midway,above] {$\substack{ [5] \\ \rightarrow }$}; 
        \draw (ff) -- (x5) node[midway,below] {$\substack{\leftarrow \\ [1]}$};
      \end{tikzpicture}}
   \end{center}
   \caption{\label{fig:all_messages} Answer to Exercise \ref{ex:sum-product-message-passing} Question \ref{q:messages}: Computing all messages in five clock cycles. If we also computed the messages toward the leaf factor nodes, we needed six cycles, but they are not necessary for computation of the marginals so they are omitted.}
    \end{figure}
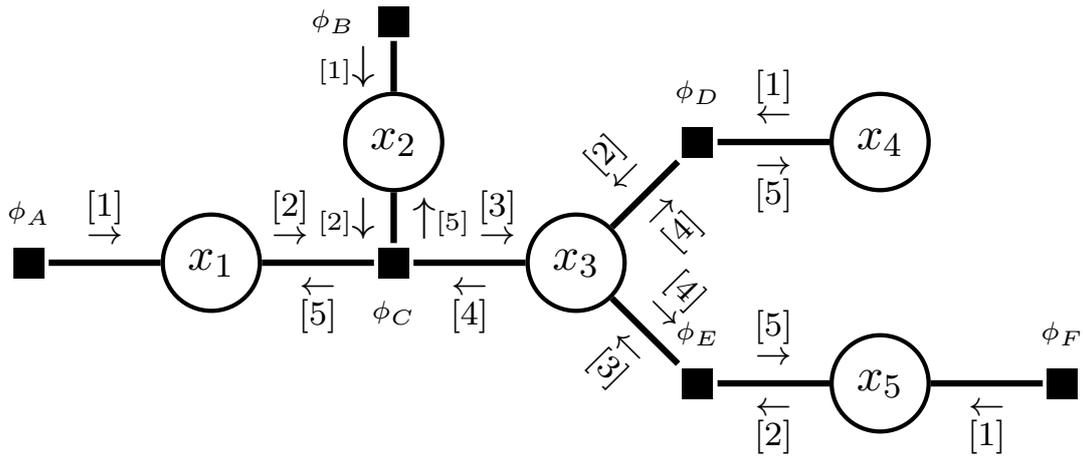

    \end{solution}
  
\item What is $p(x_1=1)$?
  
    \begin{solution}
      We compute the marginal $p(x_1)$ as
      \begin{align}
        p(x_1) \propto \fxmess{\phi_A}{x_1}{1}(x_1) \fxmess{\phi_C}{x_1}{1}(x_1)
      \end{align}
      which is in vector notation
      \begin{align}
        \begin{pmatrix}
          p(x_1=0)\\
          p(x_1=1)
        \end{pmatrix}
        & \propto 
        \fxmessb{\phi_A}{x_1} \odot \fxmessb{\phi_C}{x_1}\\
          & \propto  
          \begin{pmatrix}
            2 \\
            4 \\
          \end{pmatrix}
          \odot
          \begin{pmatrix}
            19920\\
            25920
          \end{pmatrix}\\
          & \propto
          \begin{pmatrix}
            39840\\
            103680
          \end{pmatrix}.
      \end{align}
      Normalisation gives
      \begin{align}
        \begin{pmatrix}
          p(x_1=0)\\
          p(x_1=1)
        \end{pmatrix}
     & =  \frac{1}{39840+103680}\begin{pmatrix}
            39840\\
            103680
      \end{pmatrix}\\
     & = \begin{pmatrix}
        0.2776\\
        0.7224
      \end{pmatrix}
      \end{align}
      so that $p(x_1=1) = 0.7224$.

      Note the relatively large numbers in the messages that we
      computed. In other cases, one may obtain very small ones
      depending on the scale of the factors. This can cause
      numerical issues that can be addressed by working in the
      logarithmic domain.
  \end{solution}
  
\item Draw the factor graph corresponding to $p(x_1, x_3, x_4, x_5 | x_2 =1)$ and provide the numerical values for all factors.
  
  \begin{solution}
    The pmf represented by the original factor graph is
    $$p(x_1, \ldots, x_5) \propto \phi_A(x_1) \phi_B(x_2) \phi_C(x_1,x_2,x_3) \phi_D(x_3,x_4) \phi_E(x_3,x_5) \phi_F(x_5)$$
    The conditional $p(x_1, x_3, x_4, x_5 | x_2 =1)$ is proportional to $p(x_1, \ldots, x_5)$ with $x_2$ fixed to $x_2=1$, i.e.\
    \begin{align}
      p(x_1, x_3, x_4, x_5 | x_2 =1) &\propto p(x_1,x_2=1, x_3, x_4, x_5) \\
      & \propto \phi_A(x_1) \phi_B(x_2=1) \phi_C(x_1,x_2=1,x_3) \phi_D(x_3,x_4) \phi_E(x_3,x_5) \phi_F(x_5)\\
      & \propto \phi_A(x_1) \phi^{x_2}_C(x_1,x_3) \phi_D(x_3,x_4) \phi_E(x_3,x_5) \phi_F(x_5)
    \end{align}
    where $\phi^{x_2}_C(x_1,x_3) = \phi_C(x_1,x_2=1,x_3)$. The numerical values of  $\phi^{x_2}_C(x_1,x_3)$ can be read from the table defining $\phi_C(x_1,x_2,x_3)$, extracting those rows where $x_2=1$,
    \begin{center}
      \begin{tabular}{lllll}
        \toprule
        & $x_1$ & $x_2$ & $x_3$ & $\phi_C$\\
      \midrule
      &0 & 0 & 0 & 4 \\
      &1 & 0 & 0 & 2 \\
      $\rightarrow$&0 & 1 & 0 & 2 \\
      $\rightarrow$& 1 & 1 & 0 & 6 \\
      &0 & 0 & 1 & 2 \\
      &1 & 0 & 1 & 6 \\
       $\rightarrow$&0 & 1 & 1 & 6 \\
       $\rightarrow$&1 & 1 & 1 & 4 \\
      \bottomrule
      \end{tabular}
      \hspace{3ex} \text{so that} \hspace{3ex}
      \begin{tabular}{lll}
        \toprule
        $x_1$ & $x_3$ & $\phi^{x_2}_C$\\
      \midrule
      0 & 0 & 2 \\
      1 & 0 & 6 \\
      0 & 1 & 6 \\
      1 & 1 & 4 \\
      \bottomrule
      \end{tabular}
    \end{center}
  
    The factor graph for $p(x_1, x_3, x_4, x_5 | x_2 =1)$ is shown below. Factor $\phi_B$ has disappeared since it only depended on $x_2$ and thus became a constant. Factor $\phi_C$ is replaced by $\phi_C^{x_2}$ defined above. The remaining factors are the same as in the original factor graph.
        
    \begin{center}
      \begin{tikzpicture}[ugraph]
        \node[fact, label=above: $\phi_A$] (fa) at (0,0) {} ; %
        \node[cont] (x1) at (1.5,0)  {$x_1$} ; %
        \node[fact, label=below: $\phi^{x_2}_C$] (fc) at (3,0) {} ; %
        \node[cont] (x3) at (4.5,0)  {$x_3$} ; %
        \node[fact, label=above: $\phi_D$] (fd) at (5.5,1) {} ; %
        \node[cont] (x4) at (7,1)  {$x_4$} ; %
        \node[fact, label=above: $\phi_E$] (fe) at (5.5,-1) {} ; %
        \node[cont] (x5) at (7,-1)  {$x_5$} ; %
        \node[fact, label=above: $\phi_F$] (ff) at (8.5,-1) {} ; %
        \draw (fa) -- (x1);
        \draw (x1) -- (fc);
        \draw (fc) -- (x3);
        \draw (x3) -- (fd);
        \draw (x3) -- (fe);
        \draw (fd) -- (x4);
        \draw (fe) -- (x5);
        \draw (x5) -- (ff);
      \end{tikzpicture}
    \end{center}
    
  \end{solution}

\item Compute $p(x_1 =1 | x_2 = 1)$, re-using messages that you have already computed for the evaluation of $p(x_1 =1)$.

  \begin{solution}
    The message $\fxmess{\phi_A}{x_1}{1}$ is the same as in the
    original factor graph and $\xfmess{x_3}{\phi_C^{x_2}}{1} =
    \xfmess{x_3}{\phi_C}{1}$. This is because the outgoing message
    from $x_3$ corresponds to the effective factor obtained by summing
    out all variables in the sub-trees attached to $x_3$ (without the
    $\phi_C^{x_2}$ branch), and these sub-trees do not depend on $x_2$.

    The message $\fxmess{\phi_C^{x_2}}{x_1}{1}$ needs to be newly
    computed. We have
    \begin{align}
      \fxmess{\phi_C^{x_2}}{x_1}{1}(x_1) & = \sum_{x_3} \phi_C^{x_2}(x_1,x_3) \xfmess{x_3}{\phi_C^{x_2}}{1}
    \end{align}
    or in vector notation
    \begin{align}
      \fxmessb{\phi_C^{x_2}}{x_1} & = \pmb{\phi_C^{x_2}} \xfmessb{x_3}{\phi_C^{x_2}}\\
      &= \begin{pmatrix}
          \phi_C^{x_2}(x_1=0,x_3=0) &  \phi_C^{x_2}(x_1=0,x_3=1) \\
          \phi_C^{x_2}(x_1=1,x_3=0) & \phi_C^{x_2}(x_1=1,x_3=1)  
        \end{pmatrix}
      \xfmessb{x_3}{\phi_C^{x_2}}\\
      & = \begin{pmatrix}
        2 & 6 \\
        6 & 4
      \end{pmatrix}
      \begin{pmatrix}
        510\\
        240
      \end{pmatrix}\\
      &=
      \begin{pmatrix}
        2460\\
        4020
      \end{pmatrix}
    \end{align}
    We thus obtain for the marginal posterior of $x_1$ given $x_2=1$:
      \begin{align}
        \begin{pmatrix}
          p(x_1=0 | x_2=1)\\
          p(x_1=1 | x_2=1)
        \end{pmatrix}
        & \propto 
        \fxmessb{\phi_A}{x_1} \odot \fxmessb{\phi^{x_2}_C}{x_1}\\
          & \propto  
          \begin{pmatrix}
            2 \\
            4 \\
          \end{pmatrix}
          \odot
          \begin{pmatrix}
            2460\\
            4020
          \end{pmatrix}\\
          & \propto
          \begin{pmatrix}
            4920\\
            16080
          \end{pmatrix}.
      \end{align}
      Normalisation gives
      \begin{align}
        \begin{pmatrix}
          p(x_1=0 | x_2=1)\\
          p(x_1=1 | x_2=1)
        \end{pmatrix}
        & =  \begin{pmatrix}
        0.2343\\
        0.7657
        \end{pmatrix}
      \end{align}
      and thus $p(x_1=1 | x_2 =1) =  0.7657$. The posterior probability is slightly larger than the prior probability, $p(x_1=1) = 0.7224$.
  \end{solution}
  
\end{exenumerate}

\ex{Sum-product message passing}
The following factor graph represents a Gibbs distribution
  over four binary variables $x_i \in \{0,1\}$.
  \begin{figure}[h]
    \begin{center}
      \scalebox{1}{ 
        \begin{tikzpicture}[ugraph]
          \node[fact, label=above: $\phi_a$] (fa) at (-1.5,0) {};
          \node[cont] (x1) at (0,0) {$x_1$};
     
          \node[fact, label=above: $\phi_b$] (fb) at (0,1.5) {};
          \node[cont] (x2) at (-1.5,1.5) {$x_2$};

          \node[fact, label=above: $\phi_c$] (fc) at (1.5,0) {};
          \node[cont] (x3) at (3,1.5) {$x_3$};
          \node[cont] (x4) at (3,-1.5) {$x_4$};

          \node[fact, label=above: $\phi_d$] (fd) at (4.5,1.5) {};
          \node[fact, label=above: $\phi_e$] (fe) at (4.5,-1.5) {};

          \draw (fa) -- (x1);
          \draw (fb) -- (x1);
          \draw (fb) -- (x2);
          \draw (x1) -- (fc);
          \draw (fc) -- (x3);
          \draw (fc) -- (x4);
          \draw (x3) -- (fd);
          \draw (x4) -- (fe);
      \end{tikzpicture}}
    \end{center}
  \end{figure}\\
  The factors $\phi_a, \phi_b, \phi_d$ are defined as follows:\\
  \begin{center}
  \begin{tabular}{l l}
    \toprule
    $x_1$ & $\phi_a$\\
    \midrule
    0 & 2 \\
    1 & 1\\
    \bottomrule
  \end{tabular}
  \hspace{10ex}
  \begin{tabular}{l l l}
    \toprule
    $x_1$ & $x_2$ & $\phi_b$\\
    \midrule
    0 & 0 & 5 \\
    1 & 0 & 2 \\
    0 & 1 & 2 \\
    1 & 1 & 6 \\
    \bottomrule
  \end{tabular}
  \hspace{10ex}
  \begin{tabular}{l l}
    \toprule
    $x_3$ & $\phi_d$\\
    \midrule
    0 & 1 \\
    1 & 2\\
    \bottomrule
  \end{tabular}\\
  \end{center}
  and $\phi_c(x_1,x_3,x_4) = 1$ if $x_1=x_3=x_4$, and is zero otherwise.\\[2ex]
  For all questions below, justify your answer:
  \begin{exenumerate}
  \item Compute the values of $\xfmess{x_2}{\phi_b}{1}(x_2)$ for $x_2=0$ and $x_2=1$.

    \begin{solution}
      Messages from leaf-variable nodes to factor nodes are equal to one, so that $\xfmess{x_2}{\phi_b}{1}(x_2)=1$ for all $x_2$.
    \end{solution}

  \item Assume the message $\xfmess{x_4}{\phi_c}{1}(x_4)$ equals
    $$\xfmess{x_4}{\phi_c}{1}(x_4) = \begin{cases}
    1 & \text{if } x_4=0\\
    3 & \text{if } x_4=1\\
  \end{cases}$$
    Compute the values of $\phi_e(x_4)$ for $x_4=0$ and $x_4=1$. 

    \begin{solution}

      Messages from leaf-factors to their variable nodes are
      equal to the leaf-factors, and variable nodes with single
      incoming messages copy the message. We thus have
      \begin{align}
        \fxmess{\phi_e}{x_4}{1}(x_4) & = \phi_e(x_4)\\
        \xfmess{x_4}{\phi_c}{1}(x_4) & =  \fxmess{\phi_e}{x_4}{1}(x_4)
      \end{align}
      and hence
      \begin{align}
        \phi_e(x_4) &= \begin{cases}
          1 & \text{if } x_4=0\\
          3 & \text{if } x_4=1\\
        \end{cases}
      \end{align}
  
    \end{solution}

  \item Compute the values of $\fxmess{\phi_c}{x_1}{1}(x_1)$ for $x_1=0$ and $x_1=1$.

    \begin{solution}

      We first compute $\xfmess{x_3}{\phi_c}{1}(x_3)$:
      \begin{align}
        \xfmess{x_3}{\phi_c}{1}(x_3) & = \fxmess{\phi_d}{x_3}{1}(x_3)\\
        & = \begin{cases}
          1 & \text{if } x_3=0\\
           2 & \text{if } x_3=1
        \end{cases}
      \end{align}
      The desired message $\fxmess{\phi_c}{x_1}{1}(x_1)$ is by definition
      \begin{align}
        \fxmess{\phi_c}{x_1}{1}(x_1) & = \sum_{x_3,x_4} \phi_c(x_1,x_3,x_4) \xfmess{x_3}{\phi_c}{1}(x_3) \xfmess{x_4}{\phi_c}{1}(x_4)
      \end{align}
      Since $\phi_c(x_1,x_3,x_4)$ is only non-zero if $x_1=x_3=x_4$, where it equals one, the computations simplify:
      \begin{align}
        \fxmess{\phi_c}{x_1}{1}(x_1=0) & = \phi_c(0,0,0) \xfmess{x_3}{\phi_c}{1}(0) \xfmess{x_4}{\phi_c}{1}(0)\\
        & = 1 \cdot 1 \cdot 1\\
        & = 1
      \end{align}
      \begin{align}
        \fxmess{\phi_c}{x_1}{1}(x_1=1) & = \phi_c(1,1,1) \xfmess{x_3}{\phi_c}{1}(1) \xfmess{x_4}{\phi_c}{1}(1)\\
        & = 1 \cdot 2 \cdot 3\\
        & = 6
      \end{align}
    \end{solution}

  \item The message $\fxmess{\phi_b}{x_1}{1}(x_1)$ equals
    $$ \fxmess{\phi_b}{x_1}{1}(x_1) = \begin{cases}
    7 & \text{if } x_1=0\\
    8 & \text{if } x_1=1\\
  \end{cases} $$
    What is the probability that $x_1=1$, i.e.\ $p(x_1=1)$?

    \begin{solution}
      The unnormalised marginal $p(x_1)$ is given by the product of the three incoming messages
      \begin{align}
        p(x_1) & \propto \fxmess{\phi_a}{x_1}{1}(x_1)\fxmess{\phi_b}{x_1}{1}(x_1) \fxmess{\phi_c}{x_1}{1}(x_1)
      \end{align}
      With
      \begin{align}
        \fxmess{\phi_b}{x_1}{1}(x_1) & = \sum_{x_2} \phi_b(x_1,x_2)
      \end{align}
      it follows that
      \begin{align}
        \fxmess{\phi_b}{x_1}{1}(x_1=0) & = \sum_{x_2} \phi_b(0,x_2)\\
        & = 5+2 \\
        & = 7\\
        \fxmess{\phi_b}{x_1}{1}(x_1=1) & = \sum_{x_2} \phi_b(1,x_2)\\
        & = 2+6\\
        & = 8
      \end{align}
      Hence, we obtain
      \begin{align}
        p(x_1=0) &\propto 2 \cdot 7 \cdot 1 = 14\\
        p(x_1=1) &\propto 1 \cdot 8 \cdot 6 = 48
      \end{align}
      and normalisation yields the desired result
      \begin{align}
        p(x_1=1) & = \frac{48}{14+48} = \frac{48}{62} = \frac{24}{31} = 0.774
      \end{align}
     
    \end{solution}

  \end{exenumerate}

\ex{Max-sum message passing}
\label{ex:max-sum-message-passing}
We here compute most probable states for the factor graph and factors below.

  \begin{center}
    \begin{tikzpicture}[ugraph]
      \node[fact, label=above: $\phi_A$] (fa) at (0,0) {} ; %
      \node[cont] (x1) at (1.5,0)  {$x_1$} ; %
      \node[fact, label=below: $\phi_C$] (fc) at (3,0) {} ; %
      \node[cont] (x2) at (3,1)  {$x_2$} ; %
      \node[fact, label=left: $\phi_B$] (fb) at (3,2) {} ; %
      \node[cont] (x3) at (4.5,0)  {$x_3$} ; %
      \node[fact, label=above: $\phi_D$] (fd) at (5.5,1) {} ; %
      \node[cont] (x4) at (7,1)  {$x_4$} ; %
      \node[fact, label=above: $\phi_E$] (fe) at (5.5,-1) {} ; %
      \node[cont] (x5) at (7,-1)  {$x_5$} ; %
      \node[fact, label=above: $\phi_F$] (ff) at (8.5,-1) {} ; %
      \draw (fa) -- (x1);
      \draw (x1) -- (fc);
      \draw (fc) -- (x2);
      \draw (x2) -- (fb);
      \draw (fc) -- (x3);
      \draw (x3) -- (fd);
      \draw (x3) -- (fe);
      \draw (fd) -- (x4);
      \draw (fe) -- (x5);
      \draw (x5) -- (ff);
    \end{tikzpicture}
  \end{center}
 
 Let all variables be binary, $x_i \in \{0,1\}$, and the factors be defined as follows:
  \begin{center}
    \begin{tabular}{ll}
      \toprule
      $x_1$ & $\phi_A$\\
      \midrule
      0 & 2\\
      1 & 4\\
      \bottomrule
    \end{tabular}
    \hfill
    \begin{tabular}{ll}
      \toprule
      $x_2$ & $\phi_B$\\
      \midrule
      0 & 4\\
      1 & 4\\
      \bottomrule
    \end{tabular}
    \hfill
    \begin{tabular}{llll}
      \toprule
      $x_1$ & $x_2$ & $x_3$ & $\phi_C$\\
      \midrule
    0 & 0 & 0 & 4 \\
    1 & 0 & 0 & 2 \\
    0 & 1 & 0 & 2 \\
    1 & 1 & 0 & 6 \\
    0 & 0 & 1 & 2 \\
    1 & 0 & 1 & 6 \\
    0 & 1 & 1 & 6 \\
    1 & 1 & 1 & 4 \\
      \bottomrule
    \end{tabular}
    \hfill
    \begin{tabular}{lll}
      \toprule
      $x_3$ & $x_4$ & $\phi_D$\\
      \midrule
    0 & 0 &  8 \\
    1 & 0 &  2 \\
    0 & 1 &  2 \\
    1 & 1 &  6 \\
      \bottomrule
    \end{tabular}
    \hfill
    \begin{tabular}{lll}
      \toprule
      $x_3$ & $x_5$ & $\phi_E$\\
      \midrule
    0 & 0 &  3 \\
    1 & 0 &  6 \\
    0 & 1 &  6 \\
    1 & 1 &  3 \\
      \bottomrule
    \end{tabular}
   \hfill
    \begin{tabular}{ll}
      \toprule
      $x_5$ & $\phi_F$\\
      \midrule
      0 & 1\\
      1 & 8\\
      \bottomrule
    \end{tabular}

  \end{center}
 
  \begin{exenumerate}
  \item Will we need to compute the normalising constant $Z$ to determine $\argmax_{\x} p(x_1, \ldots, x_5)$?

    \begin{solution}
      This is not necessary since $\argmax_{\x} p(x_1, \ldots, x_5) =
      \argmax_{\x} c p(x_1, \ldots, x_5)$ for any constant
      $c$. Algorithmically, the backtracking algorithm is also
      invariant to any scaling of the factors.
    \end{solution}
  \item Compute $\argmax_{x_1, x_2, x_3} p(x_1, x_2, x_3 | x_4=0, x_5=0)$ via max-sum message passing.
    \begin{solution}
      We first derive the factor graph and corresponding factors for
      $p(x_1, x_2, x_3 | x_4=0, x_5=0)$.

      For fixed values of $x_4, x_5$, the two variables are removed
      from the graph, and the factors $\phi_D(x_3,x_4)$ and
      $\phi_E(x_3, x_5)$ are reduced to univariate factors
      $\phi_D^{x_4}(x_3)$ and $\phi_D^{x_5}(x_3)$ by retaining those
      rows in the table where $x_4=0$ and $x_5=0$, respectively:
      \begin{center}
        \begin{tabular}{ll}
          \toprule
          $x_3$ & $\phi_D^{x_4}$\\
          \midrule
          0 & 8\\
          1 & 2\\
          \bottomrule
        \end{tabular}
        \hspace{4ex}
        \begin{tabular}{ll}
          \toprule
          $x_3$ & $\phi_E^{x_5}$\\
          \midrule
          0 & 3\\
          1 & 6\\
          \bottomrule
        \end{tabular}   
      \end{center}
      Since both factors only depend on $x_3$, they can be combined into a new factor $\tilde{\phi}(x_3)$ by element-wise multiplication.
      \begin{center}
        \begin{tabular}{ll}
          \toprule
          $x_3$ & $\tilde{\phi}$\\
          \midrule
          0 & 24\\
          1 & 12\\
          \bottomrule
        \end{tabular}
      \end{center}
      Moreover, since we work with an unnormalised model, we can rescale
      the factor so that the maximum value is one, so that
      \begin{center}
        \begin{tabular}{ll}
          \toprule
          $x_3$ & $\tilde{\phi}$\\
          \midrule
          0 & 2\\
          1 & 1\\
          \bottomrule
        \end{tabular}
      \end{center}
      Factor $\phi_F(x_5)$ is a constant for fixed value of $x_5$ and can be ignored. The factor graph for $p(x_1, x_2, x_3 | x_4=0, x_5=0)$ thus is
      \begin{center}
        \begin{tikzpicture}[ugraph]
          \node[fact, label=above: $\phi_A$] (fa) at (0,0) {} ; %
          \node[cont] (x1) at (1.5,0)  {$x_1$} ; %
          \node[fact, label=below: $\phi_C$] (fc) at (3,0) {} ; %
          \node[cont] (x2) at (3,1)  {$x_2$} ; %
          \node[fact, label=left: $\phi_B$] (fb) at (3,2) {} ; %
          \node[cont] (x3) at (4.5,0)  {$x_3$} ; %
          \node[fact, label=above: $\tilde{\phi}$] (fnew) at (6,0) {} ; %
          \draw (fa) -- (x1);
          \draw (x1) -- (fc);
          \draw (fc) -- (x2);
          \draw (x2) -- (fb);
          \draw (fc) -- (x3);
          \draw (x3) -- (fnew);
        \end{tikzpicture}
      \end{center}

      Let us fix $x_1$ as root towards which we compute the messages. The messages that we need to compute are shown in the following graph
      \begin{center}
        \begin{tikzpicture}[ugraph]
          \node[fact, label=above: $\phi_A$] (fa) at (0,0) {} ; %
          \node[cont] (x1) at (1.5,0)  {$x_1$} ; %
          \node[fact, label=below: $\phi_C$] (fc) at (3,0) {} ; %
          \node[cont] (x2) at (3,1)  {$x_2$} ; %
          \node[fact, label=left: $\phi_B$] (fb) at (3,2) {} ; %
          \node[cont] (x3) at (4.5,0)  {$x_3$} ; %
          \node[fact, label=above: $\tilde{\phi}$] (fnew) at (6,0) {} ; %
          \draw (fa) -- (x1) node[midway,above] {$\rightarrow$}; 
          \draw (x1) -- (fc) node[midway,above] {$\leftarrow$}; 
          \draw (fc) -- (x2) node[midway,left] {$\downarrow$}; 
          \draw (x2) -- (fb) node[midway,left] {$\downarrow$}; 
          \draw (fc) -- (x3) node[midway,above] {$\leftarrow$}; 
          \draw (x3) -- (fnew) node[midway,above] {$\leftarrow$}; 
        \end{tikzpicture}
      \end{center}
      
      Next, we compute the leaf (log) messages. We only have factor nodes as leaf nodes so that
      \begin{align}
        \lambdab_{\phi_A \to x_1} & =  \begin{pmatrix}
            \log \phi_A(x_1=0)\\
            \log \phi_A(x_1=1)
          \end{pmatrix}
          = \begin{pmatrix}
            \log 2\\
            \log 4
          \end{pmatrix}
      \end{align}
      and similarly
      \begin{align}
        \lambdab_{\phi_B \to x_2} & =   \begin{pmatrix}
            \log \phi_B(x_2=0)\\
            \log \phi_B(x_2=1)
        \end{pmatrix} = \begin{pmatrix}
          \log 4\\
          \log 4
        \end{pmatrix}
        &
        \lambdab_{\tilde{\phi} \to x_3} & =   \begin{pmatrix}
            \log \tilde{\phi}(x_3=0)\\
            \log \tilde{\phi}(x_3=1)
          \end{pmatrix}= \begin{pmatrix}
            \log 2\\
            \log 1
          \end{pmatrix}
      \end{align}
      Since the variable nodes $x_2$ and $x_3$ only have one incoming
      edge each, we obtain
      \begin{align}
        \lambdab_{x_2 \to \phi_C} = \lambdab_{\phi_B \to x_2} & =  \begin{pmatrix}
          \log 4\\
          \log 4
        \end{pmatrix}
        &
        \lambdab_{x_3 \to \phi_C} = \lambdab_{\tilde{\phi} \to x_3} & =  \begin{pmatrix}
          \log 2\\
          \log 1
        \end{pmatrix}
      \end{align}
      The message $\lambda_{\phi_C \to x_1}(x_1)$ equals
      \begin{align}
        \lambda_{\phi_C \to x_1}(x_1) & = \max_{x_2, x_3} \log \phi_C(x_1, x_2, x_3) +  \lambda_{x_2 \to \phi_C}(x_2) + \lambda_{x_3 \to \phi_C}(x_3)
      \end{align}
      where we wrote the messages in non-vector notation to highlight
      their dependency on the variables $x_2$ and $x_3$. We now have to consider all combinations of $x_2$ and $x_3$
      \begin{center}
        \begin{tabular}{lll}
          \toprule
          $x_2$ & $x_3$ & $\log \phi_C(x_1=0, x_2, x_3)$\\
          \midrule
          0 & 0 & $\log 4$ \\
          1 & 0 &  $\log 2$\\
          0 & 1 & $\log 2$ \\
          1 & 1 & $\log 6$ \\
          \bottomrule
        \end{tabular}
        \hspace{4ex}
        \begin{tabular}{lll}
          \toprule
          $x_2$ & $x_3$ & $\log \phi_C(x_1=1, x_2, x_3)$\\
          \midrule
          0 & 0 & $\log 2$  \\
          1 & 0 & $\log 6 $ \\
          0 & 1 & $\log 6$  \\
          1 & 1 & $\log 4$ \\
          \bottomrule
        \end{tabular}
       \end{center}
      Furthermore
      \begin{center}
        \begin{tabular}{lll}
          \toprule
          $x_2$ & $x_3$ & $\lambda_{x_2 \to \phi_C}(x_2) + \lambda_{x_3 \to \phi_C}(x_3)$\\
          \midrule
          0 & 0 & $\log 4 + \log 2 = \log 8$  \\
          1 & 0 & $\log 4 + \log 2 = \log 8$\\
          0 & 1 & $\log 4$\\
          1 & 1 & $\log 4$\\
          \bottomrule
        \end{tabular}
      \end{center}
      Hence for $x_1=0$, we have
       \begin{center}
         \begin{tabular}{lll}
           \toprule
           $x_2$ & $x_3$ & $\log \phi_C(x_1=0, x_2, x_3) + \lambda_{x_2 \to \phi_C}(x_2) + \lambda_{x_3 \to \phi_C}(x_3)$\\
           \midrule
            0 & 0 & $\log 4 + \log 8 = \log 32$  \\
            1 & 0 & $\log 2 + \log 8 = \log 16$ \\
            0 & 1 & $\log 2 + \log 4 = \log 8$  \\
            1 & 1 & $\log 6 + \log 4 = \log 24$ \\
           \bottomrule
         \end{tabular}
       \end{center}
       The maximal value is $\log 32$ and for backtracking, we also
       need to keep track of the $\argmax$ which is here
       $\hat{x}_2=\hat{x}_3=0$.

       For $x_1=1$, we have
        \begin{center}
         \begin{tabular}{lll}
           \toprule
           $x_2$ & $x_3$ & $\log \phi_C(x_1=1, x_2, x_3) + \lambda_{x_2 \to \phi_C}(x_2) + \lambda_{x_3 \to \phi_C}(x_3)$\\
           \midrule
           0 & 0 & $\log 2 + \log 8 = \log 16 $  \\
           1 & 0 & $\log 6 + \log 8 = \log 48$\\
           0 & 1 & $\log 6 + \log 4 = \log 24$ \\
           1 & 1 & $\log 4 + \log 4 = \log 16$ \\
           \bottomrule
         \end{tabular}
       \end{center}
        The maximal value is $\log 48$ and the $\argmax$ is $(\hat{x}_2=1, \hat{x}_3=0)$.
        
        So overall, we have
        \begin{equation}
          \lambdab_{\phi_C \to x_1} = \begin{pmatrix}
            \lambda_{\phi_C \to x_1}(x_1=0)\\
            \lambda_{\phi_C \to x_1}(x_1=1)
          \end{pmatrix}
          = \begin{pmatrix}
            \log 32\\
            \log 48
      \end{pmatrix}
    \end{equation}
    and the $\argmax$ back-tracking function is
    \begin{equation}
      \lambda^*_{\phi_C \to x_1}(x_1) = \begin{cases}
        (\hat{x}_2 = 0, \hat{x}_3 = 0) & \text{if } x_1=0\\
        (\hat{x}_2=1, \hat{x}_3=0) & \text{if } x_1=1
      \end{cases}
    \end{equation}
    We now have all incoming messages to the assigned root node $x_1$. \emph{Ignoring the normalising constant}, we obtain
    \begin{align}
      \boldsymbol{\gamma}&=\begin{pmatrix}
      \gamma^*(x_1=0) \\
      \gamma^*(x_1=1)
      \end{pmatrix}
      =  \lambdab_{\phi_A \to x_1} +  \lambdab_{\phi_C \to x_1}\\
      &=  \begin{pmatrix}
        \log 2\\
        \log 4
      \end{pmatrix} +
      \begin{pmatrix}
        \log 32\\
        \log 48
      \end{pmatrix}
      =
      \begin{pmatrix}
        \log 64\\
        \log 192
      \end{pmatrix}
    \end{align}
    The value $x_1$ for which $\gamma^*(x_1)$ is largest is thus
    $\hat{x}_1 = 1$. Plugging $\hat{x}_1 = 1$ into the backtracking
    function $\lambda^*_{\phi_C \to x_1}(x_1)$ gives 
    \begin{equation}
      (\hat{x}_1, \hat{x}_2, \hat{x}_3) = \argmax_{x_1, x_2, x_3} p(x_1, x_2, x_3 | x_4=0, x_5=0) = (1,1,0).
    \end{equation}
    
    In this low-dimensional example, we can verify the solution by
    computing the unnormalised pmf for all combinations of
    $x_1,x_2,x_3$. This is done in the following table where we start
    with the table for $\phi_C$ and then multiply-in the further
    factors $\phi_A$, $\tilde{\phi}$ and $\phi_B$. 
    \begin{center}
      \begin{tabular}{lllllll}
        \toprule
        $x_1$ & $x_2$ & $x_3$ & $\phi_C$ & $\phi_C \phi_A$ & $\phi_C \phi_A\tilde{\phi}$& $\phi_C \phi_A\tilde{\phi}\phi_B$ \\
        \midrule
        0 & 0 & 0 & 4 & 8 & 16 & $16\cdot 4$\\
        1 & 0 & 0 & 2 & 8 & 16 & $16 \cdot 4$\\
        0 & 1 & 0 & 2 & 8 & 16 & $16\cdot 4$\\
        1 & 1 & 0 & 6 & 24& 48 & $48 \cdot 4$\\
        0 & 0 & 1 & 2 & 8 & 8  & $8 \cdot 4$\\
        1 & 0 & 1 & 6 & 24 & 24 & $24 \cdot 4$\\
        0 & 1 & 1 & 6 & 12 & 12 & $12 \cdot 4$\\
        1 & 1 & 1 & 4 & 16 & 16 & $16\cdot 4$\\
        \bottomrule
      \end{tabular}
    \end{center}
    For example, for the column $\phi_c \phi_A$, we multiply each
    value of $\phi_C(x_1, x_2, x_3)$ by $\phi_A(x_1)$, so that the
    rows with $x_1=0$ get multiplied by 2, and the rows with $x_1=1$
    by 4.

    The maximal value in the final column is achieved for $x_1=1,
    x_2=1, x_3=0$, in line with the result above (and $48\cdot 4 =
    192$). Since $\phi_B(x_2)$ is a constant, being equal to 4 for all
    values of $x_2$, we could have ignored it in the computation. The
    formal reason for this is that since the model is unnormalised, we
    are allowed to rescale each factor by an arbitrary
    (factor-dependent) \emph{constant}. This operation does not change
    the model. So we could divide $\phi_B$ by 4 which would give a
    value of 1, so that the factor can indeed be ignored.
   
    \end{solution}
  \item Compute $\argmax_{x_1, \ldots, x_5} p(x_1, \ldots, x_5)$ via max-sum message passing with $x_1$ as root.
    \label{q:x_1-root}
    \begin{solution}
      As discussed in the solution to the answer above, we can drop
      factor $\phi_B(x_2)$ since it takes the same value for all
      $x_2$. Moreover, we can rescale the individual factors by a
      constant so they are more amenable to calculations by hand. We
      normalise them such that the largest value is one, which gives
      the following factors. Note that this is entirely optional.
      \begin{center}
        \begin{tabular}{ll}
          \toprule
          $x_1$ & $\phi_A$\\
          \midrule
          0 & 1\\
          1 & 2\\
          \bottomrule
        \end{tabular}
        \hfill
        \begin{tabular}{llll}
          \toprule
          $x_1$ & $x_2$ & $x_3$ & $\phi_C$\\
          \midrule
          0 & 0 & 0 & 2 \\
          1 & 0 & 0 & 1 \\
          0 & 1 & 0 & 1 \\
          1 & 1 & 0 & 3 \\
          0 & 0 & 1 & 1 \\
          1 & 0 & 1 & 3 \\
          0 & 1 & 1 & 3 \\
          1 & 1 & 1 & 2 \\
          \bottomrule
        \end{tabular}
        \hfill
        \begin{tabular}{lll}
          \toprule
          $x_3$ & $x_4$ & $\phi_D$\\
          \midrule
          0 & 0 &  4 \\
          1 & 0 &  1 \\
          0 & 1 &  1 \\
          1 & 1 &  3 \\
          \bottomrule
        \end{tabular}
        \hfill
        \begin{tabular}{lll}
          \toprule
          $x_3$ & $x_5$ & $\phi_E$\\
          \midrule
          0 & 0 &  1 \\
          1 & 0 &  2 \\
          0 & 1 &  2 \\
          1 & 1 &  1 \\
          \bottomrule
        \end{tabular}
        \hfill
        \begin{tabular}{ll}
          \toprule
          $x_5$ & $\phi_F$\\
          \midrule
          0 & 1\\
          1 & 8\\
          \bottomrule
        \end{tabular}   
      \end{center}
      The factor graph without $\phi_B$ together with the messages that we need to compute is:
      \begin{center}
        \begin{tikzpicture}[ugraph]
          \node[fact, label=above: $\phi_A$] (fa) at (0,0) {} ; %
          \node[cont] (x1) at (1.5,0)  {$x_1$} ; %
          \node[fact, label=below: $\phi_C$] (fc) at (3,0) {} ; %
          \node[cont] (x2) at (3,1)  {$x_2$} ; %
          \node[cont] (x3) at (4.5,0)  {$x_3$} ; %
          \node[fact, label=above: $\phi_D$] (fd) at (5.5,1) {} ; %
          \node[cont] (x4) at (7,1)  {$x_4$} ; %
          \node[fact, label=above: $\phi_E$] (fe) at (5.5,-1) {} ; %
          \node[cont] (x5) at (7,-1)  {$x_5$} ; %
          \node[fact, label=above: $\phi_F$] (ff) at (8.5,-1) {} ; %
          \draw (fa) -- (x1) node[midway,above] {$\rightarrow$}; 
          \draw (x1) -- (fc) node[midway,above] {$\leftarrow$}; 
          \draw (fc) -- (x2) node[midway,left] {$\downarrow$}; 
          \draw (fc) -- (x3) node[midway,above] {$\leftarrow$}; 
          \draw (x3) -- (fd) node[midway,above, sloped] {$\leftarrow$}; 
          \draw (x3) -- (fe) node[midway,below, sloped] {$\leftarrow$}; 
          \draw (fd) -- (x4) node[midway,above, sloped] {$\leftarrow$}; 
          \draw (fe) -- (x5) node[midway,below, sloped] {$\leftarrow$}; 
          \draw (x5) -- (ff) node[midway,below, sloped] {$\leftarrow$}; 
        \end{tikzpicture}
      \end{center}
      The leaf (log) messages are (using vector notation where the top element corresponds to $x_i=0$ and the bottom one to $x_i=1$):
      \begin{align}
        \lambdab_{\phi_A \to x_1} & = \begin{pmatrix}
          0\\
          \log 2
        \end{pmatrix}
        &
        \lambdab_{x_2 \to \phi_C} & = \begin{pmatrix}
          0\\
          0
        \end{pmatrix}
        &
        \lambdab_{x_4 \to \phi_D} &  = \begin{pmatrix}
          0\\
          0
        \end{pmatrix}
        &
        \lambdab_{\phi_F \to x_5} &  = \begin{pmatrix}
          0\\
          \log 8
        \end{pmatrix}
      \end{align}
      The variable node $x_5$ only has one incoming edge so that
      $\lambdab_{x_5 \to \phi_E} = \lambdab_{\phi_F \to x_5}$. The
      message $\lambda_{\phi_E \to x_3}(x_3)$ equals
      \begin{align}
        \lambda_{\phi_E \to x_3}(x_3) & = \max_{x_5} \log \phi_E(x_3, x_5) +  \lambda_{x_5 \to \phi_E}(x_5)
      \end{align}
      Writing out $\log \phi_E(x_3, x_5) +  \lambda_{x_5 \to \phi_E}(x_5)$ for all $x_5$ as a function of $x_3$ we have
      \begin{center}
        \begin{tabular}{ll}
          \toprule
          $x_5$ &  $\log \phi_E(x_3=0, x_5) +  \lambda_{x_5 \to \phi_E}(x_5)$\\
          \midrule
          0  &  $\log 1 + 0 =0$\\
          1  &  $\log 2 + \log 8 = \log 16$\\
          \bottomrule
        \end{tabular}
        \hspace{3ex}
        \begin{tabular}{ll}
          \toprule
          $x_5$ &  $\log \phi_E(x_3=1, x_5) +  \lambda_{x_5 \to \phi_E}(x_5)$\\
          \midrule
          0  &  $\log 2 + 0 =\log 2$\\
          1  &  $\log 1 + \log 8 = \log 8$\\
          \bottomrule
        \end{tabular}
      \end{center}
      Taking the maximum over $x_5$ as a function of $x_3$, we obtain
      \begin{equation}
        \lambdab_{\phi_E \to x_3} = \begin{pmatrix}
          \log 16\\
          \log 8
        \end{pmatrix}
      \end{equation}
      and the backtracking function that indicates the maximiser
      $\hat{x}_5 = \argmax_{x_5} \log \phi_E(x_3, x_5) + \lambda_{x_5
        \to \phi_E}(x_5)$ as a function of $x_3$ equals
      \begin{equation}
        \lambda^*_{\phi_E \to x_3}(x_3) = \begin{cases}
          \hat{x}_5=1 & \text{if } x_3= 0\\
          \hat{x}_5=1 & \text{if } x_3= 1
          \end{cases}
      \end{equation}
      We perform the same kind of operation for $\lambda_{\phi_D \to x_3}(x_3)$ 
      \begin{align}
        \lambda_{\phi_D \to x_3}(x_3) & = \max_{x_4} \log \phi_D(x_3, x_4) +  \lambda_{x_4 \to \phi_D}(x_4)
      \end{align}
      Since $\lambda_{x_4 \to \phi_D}(x_4)=0$ for all $x_4$, the table with all values of $\log \phi_D(x_3, x_4) +  \lambda_{x_4 \to \phi_D}(x_4)$ is
      \begin{center}
        \begin{tabular}{lll}
          \toprule
          $x_3$ & $x_4$ & $\log \phi_D(x_3, x_4) +  \lambda_{x_4 \to \phi_D}(x_4)$ \\
          \midrule
          0 & 0 &  $\log 4 + 0 = \log 4$ \\
          1 & 0 &  $\log 1 +0 = 0$\\
          0 & 1 &  $\log 1 +0 =0$\\
          1 & 1 &  $\log 3 +0 = \log3$ \\
          \bottomrule
        \end{tabular}
      \end{center}
      Taking the maximum over $x_4$ as a function of $x_3$ we thus obtain
      \begin{equation}
        \lambdab_{\phi_D \to x_3} = \begin{pmatrix}
          \log 4\\
          \log 3
        \end{pmatrix}
      \end{equation}
      and the backtracking function that indicates the maximiser
      $\hat{x}_4 = \argmax_{x_4} \log \phi_D(x_3, x_4) +  \lambda_{x_4 \to \phi_D}(x_4)$ as a function of $x_3$ equals
      \begin{equation}
        \lambda^*_{\phi_D \to x_3}(x_3) = \begin{cases}
          \hat{x}_4=0 & \text{if } x_3= 0\\
          \hat{x}_4=1 & \text{if } x_3= 1
          \end{cases}
      \end{equation}
      For the message $\lambda_{x_3 \to \phi_C}(x_3)$ we add together
      the messages $\lambda_{\phi_E \to x_3}(x_3)$ and
      $\lambda_{\phi_D \to x_3}(x_3)$ which gives
      \begin{equation}
        \lambdab_{ x_3 \to \phi_C} = \begin{pmatrix}
          \log 16 + \log 4\\
          \log 8 + \log 3\\
        \end{pmatrix}
        =\begin{pmatrix}
        \log 64\\
        \log 24\\
        \end{pmatrix}
      \end{equation}
      
    Next we compute the message $\lambda_{\phi_C \to x_1}(x_1)$ by maximising over $x_2$ and $x_3$,
    \begin{equation}
      \lambda_{\phi_C \to x_1}(x_1)  = \max_{x_2, x_3} \log \phi_C(x_1, x_2, x_3) +  \lambda_{x_2 \to \phi_C}(x_2) + \lambda_{x_3 \to \phi_C}(x_3)
    \end{equation}
    Since $\lambda_{x_2 \to \phi_C}(x_2) = 0$, the problem becomes
    \begin{equation}
      \lambda_{\phi_C \to x_1}(x_1)  = \max_{x_2, x_3} \log \phi_C(x_1, x_2, x_3) + \lambda_{x_3 \to \phi_C}(x_3)
    \end{equation}
    Building on the table for $\phi_C$, we form a table with all
    values of $\log \phi_C(x_1, x_2, x_3) + \lambda_{x_3 \to
      \phi_C}(x_3)$
    \begin{center}
      \begin{tabular}{llll}
        \toprule
        $x_1$ & $x_2$ & $x_3$ & $\log \phi_C(x_1, x_2, x_3) + \lambda_{x_3 \to \phi_C}(x_3)$\\
        \midrule
        0 & 0 & 0 & $\log 2 + \log 64 = \boldsymbol{\log 128}$ \\
        1 & 0 & 0 & $0 + \log 64 = \log 64$ \\
        0 & 1 & 0 & $0 + \log 64 = \log 64$ \\
        1 & 1 & 0 & $\log 3 + \log 64 = \boldsymbol{\log 192}$ \\
        0 & 0 & 1 & $\log 24$ \\
        1 & 0 & 1 & $\log 3 + \log 24 = \log 72$ \\
        0 & 1 & 1 & $\log 3 + \log 24 = \log 72 $ \\
        1 & 1 & 1 & $\log 2 + \log 24 = \log 48$ \\
        \bottomrule
      \end{tabular}
    \end{center}
    The maximal value as a function of $x_1$ are highlighted in the table, which gives the message
    \begin{equation}
      \lambdab_{\phi_C \to x_1} = \begin{pmatrix}
        \log 128\\
        \log 192
      \end{pmatrix}
    \end{equation}
    and the backtracking function
    \begin{equation}
      \lambda^*_{\phi_C \to x_1}(x_1) = \begin{cases}
        (\hat{x}_2=0, \hat{x}_3=0) & \text{if } x_1=0\\
        (\hat{x}_2=1, \hat{x}_3=0) & \text{if } x_1=1
      \end{cases}
    \end{equation}
    We now have all incoming messages to the assigned root node $x_1$. \emph{Ignoring the normalising constant}, we obtain
    \begin{align}
      \boldsymbol{\gamma}&=\begin{pmatrix}
      \gamma^*(x_1=0) \\
      \gamma^*(x_1=1)
      \end{pmatrix}
      = \begin{pmatrix}
        0 + \log 128 \\
        \log 2 + \log 192 
      \end{pmatrix}
    \end{align}
    We can now start the backtracking to compute the desired
    $\argmax_{x_1, \ldots, x_5} p(x_1, \ldots, x_5)$. Starting at the
    root we have $\hat{x}_1 = \argmax_{x_1} \gamma^*(x_1) =
    1$. Plugging this value into the look-up table $\lambda^*_{\phi_C
      \to x_1}(x_1)$, we obtain $(\hat{x}_2=1, \hat{x}_3=0)$. With the
    look-up table $\lambda^*_{\phi_E \to x_3}(x_3)$ we find
    $\hat{x}_5=1$ and $\lambda^*_{\phi_D \to x_3}(x_3)$ gives
    $\hat{x}_4=0$ so that overall
    \begin{equation}
      \argmax_{x_1, \ldots, x_5} p(x_1, \ldots, x_5) = (1, 1, 0, 0, 1).
    \end{equation}

    \end{solution}
    
  \item Compute $\argmax_{x_1, \ldots, x_5} p(x_1, \ldots, x_5)$ via max-sum message passing with $x_3$ as root.
    \begin{solution}
      With $x_3$ as root, we need the following messages:
        \begin{center}
        \begin{tikzpicture}[ugraph]
          \node[fact, label=above: $\phi_A$] (fa) at (0,0) {} ; %
          \node[cont] (x1) at (1.5,0)  {$x_1$} ; %
          \node[fact, label=below: $\phi_C$] (fc) at (3,0) {} ; %
          \node[cont] (x2) at (3,1)  {$x_2$} ; %
          \node[cont] (x3) at (4.5,0)  {$x_3$} ; %
          \node[fact, label=above: $\phi_D$] (fd) at (5.5,1) {} ; %
          \node[cont] (x4) at (7,1)  {$x_4$} ; %
          \node[fact, label=above: $\phi_E$] (fe) at (5.5,-1) {} ; %
          \node[cont] (x5) at (7,-1)  {$x_5$} ; %
          \node[fact, label=above: $\phi_F$] (ff) at (8.5,-1) {} ; %
          \draw (fa) -- (x1) node[midway,above] {$\rightarrow$}; 
          \draw (x1) -- (fc) node[midway,above] {$\rightarrow$}; 
          \draw (fc) -- (x2) node[midway,left] {$\downarrow$}; 
          \draw (fc) -- (x3) node[midway,above] {$\rightarrow$}; 
          \draw (x3) -- (fd) node[midway,above, sloped] {$\leftarrow$}; 
          \draw (x3) -- (fe) node[midway,below, sloped] {$\leftarrow$}; 
          \draw (fd) -- (x4) node[midway,above, sloped] {$\leftarrow$}; 
          \draw (fe) -- (x5) node[midway,below, sloped] {$\leftarrow$}; 
          \draw (x5) -- (ff) node[midway,below, sloped] {$\leftarrow$}; 
        \end{tikzpicture}
      \end{center}
        The following messages are the same as when $x_1$ was the root:
        \begin{align}
          \lambdab_{\phi_D \to x_3} &= \begin{pmatrix}
          \log 4\\
          \log 3
        \end{pmatrix}
          &
          \lambdab_{\phi_E \to x_3} &= \begin{pmatrix}
            \log 16\\
            \log 8
          \end{pmatrix}
          &
          \lambdab_{\phi_A \to x_1} & = \begin{pmatrix}
            0\\
            \log 2
          \end{pmatrix}
          &
           \lambdab_{x_2 \to \phi_C} & = \begin{pmatrix}
          0\\
          0
        \end{pmatrix}
        \end{align}
        Since $x_1$ has only one incoming message, we further have
        \begin{equation}
          \lambdab_{x_1 \to \phi_C} = \lambdab_{\phi_A \to x_1} =  \begin{pmatrix}
            0\\
            \log 2
          \end{pmatrix}.
        \end{equation}
       We next compute $\lambda_{\phi_C \to x_3}(x_3)$,
       \begin{equation}
         \lambda_{\phi_C \to x_3}(x_3) = \max_{x_1, x_2} \log \phi_C(x_1,x_2,x_3) + \lambda_{x_1 \to \phi_C}(x_1) + \lambda_{x_2 \to \phi_C}(x_2).
       \end{equation}
       We first form a table for $ \log \phi_C(x_1,x_2,x_3) + \lambda_{x_1 \to \phi_C}(x_1) + \lambda_{x_2 \to \phi_C}(x_2)$ noting that $\lambda_{x_2 \to \phi_C}(x_2)=0$
       \begin{center}
         \begin{tabular}{llll}
          \toprule
          $x_1$ & $x_2$ & $x_3$ & $\log \phi_C(x_1,x_2,x_3) + \lambda_{x_1 \to \phi_C}(x_1) + \lambda_{x_2 \to \phi_C}(x_2)$  \\
          \midrule
          0 & 0 & 0 & $\log 2 + 0 = \log 2$ \\
          1 & 0 & 0 & $0+\log 2 = \log 2$ \\
          0 & 1 & 0 & $0+0 = 0$ \\
          1 & 1 & 0 & $\log 3+ \log 2 = \boldsymbol{\log 6}$ \\
          0 & 0 & 1 & $0+0 = 0 $ \\
          1 & 0 & 1 & $\log 3+ \log 2 = \boldsymbol{\log 6}$ \\
          0 & 1 & 1 & $\log 3+0 = \log 3$ \\
          1 & 1 & 1 & $\log 2+\log 2 = \log 4$ \\
          \bottomrule
        \end{tabular}
       \end{center}
       The maximal value as a function of $x_3$ are highlighted in the table, which gives the message
       \begin{equation}
         \lambdab_{\phi_C \to x_3} = \begin{pmatrix}
           \log 6\\
           \log 6
         \end{pmatrix}
       \end{equation}
       and the backtracking function
       \begin{equation}
         \lambda^*_{\phi_C \to x_3}(x_3) = \begin{cases}
           (\hat{x}_1=1, \hat{x}_2=1) & \text{if } x_3=0\\
           (\hat{x}_1=1, \hat{x}_2=0) & \text{if } x_3=1
         \end{cases}
       \end{equation}
       We have now all incoming messages for $x_3$ and can compute
       $\gamma^*(x_3)$ up the normalising constant $-\log Z$ (which is
       not needed if we are interested in the $\argmax$ only:
       \begin{align}
         \boldsymbol{\gamma}&=\begin{pmatrix}
         \gamma^*(x_3=0) \\
         \gamma^*(x_3=1)
         \end{pmatrix}
         =  \lambdab_{\phi_C \to x_3} +  \lambdab_{\phi_D \to x_3} +  \lambdab_{\phi_E \to x_3}\\
         & = \begin{pmatrix}
           \log 6 + \log 4 + \log 16 = \log 384 \\
           \log 6 + \log 3 + \log 8 = \log 144
         \end{pmatrix}
       \end{align}
       We can now start the backtracking which gives: $\hat{x}_3 = 0$,
       so that $\lambda^*_{\phi_C \to x_3}(0) = (\hat{x}_1=1,
       \hat{x}_2=1)$. The backtracking functions $\lambda^*_{\phi_E
         \to x_3}(x_3)$ and $\lambda^*_{\phi_D \to x_3}(x_3)$ are the
       same for question \ref{q:x_1-root}, which gives
       $\lambda^*_{\phi_E \to x_3}(0) = \hat{x}_5=1$ and
       $\lambda^*_{\phi_D \to x_3}(0) = \hat{x}_4=0$. Hence, overall, we
       find
       \begin{equation}
         \argmax_{x_1, \ldots, x_5} p(x_1, \ldots, x_5) = (1, 1, 0, 0, 1).
       \end{equation}
       Note that this matches the result from question
       \ref{q:x_1-root} where $x_1$ was the root. This is because the
       output of the max-sum algorithm is invariant to the choice of
       the root.

    \end{solution}
    
  \end{exenumerate}
  
\ex{Choice of elimination order in factor graphs}
Consider the following factor graph, which contains a loop:

\begin{center}
  \begin{tikzpicture}[ugraph]
    \node[cont] (x1) at (0,0)  {$x_1$} ; %
    \node[fact, label=above: $\phi_A$] (fa) at (1.5,0) {} ; %
    \node[cont] (x2) at (2.5,1)  {$x_2$} ; %
    \node[cont] (x3) at (2.5,-1)  {$x_3$} ; %
    \node[fact, label=above: $\phi_B$] (fb) at (3.5,0) {} ; %
    \node[cont] (x4) at (5,0)  {$x_4$} ; %
    \node[fact, label=above: $\phi_C$] (fc) at (6,1) {} ; %
    \node[cont] (x5) at (7.5,1)  {$x_5$} ; %
    \node[cont] (x6) at (7.5,-1)  {$x_6$} ;
    \node[fact, label=above: $\phi_D$] (fd) at (6,-1) {} ; %

    \draw (fd) -- (x6);
    \draw (x1) -- (fa);
    \draw (fa) -- (x2);
    \draw (fa) -- (x3);
    \draw (x2) -- (fb);
    \draw (x3) -- (fb);
    \draw (fb) -- (x4);
    \draw (x4) -- (fc);
    \draw (x4) -- (fd);
    \draw (fc) -- (x5);
  \end{tikzpicture}
\end{center}

Let all variables be binary, $x_i \in \{0,1\}$, and the factors be defined as follows:
\begin{center}
  \hfill
  %
  \begin{tabular}{llll}
    \toprule
    $x_1$ & $x_2$ & $x_3$ & $\phi_A$\\
    \midrule
    0 & 0 & 0 & 4 \\
    1 & 0 & 0 & 2 \\
    0 & 1 & 0 & 2 \\
    1 & 1 & 0 & 6 \\
    0 & 0 & 1 & 2 \\
    1 & 0 & 1 & 6 \\
    0 & 1 & 1 & 6 \\
    1 & 1 & 1 & 4 \\
    \bottomrule
  \end{tabular}
  \hfill
  %
  \begin{tabular}{llll}
    \toprule
    $x_2$ & $x_3$ & $x_4$ & $\phi_B$\\
    \midrule
    0 & 0 & 0 & 2 \\
    1 & 0 & 0 & 2 \\
    0 & 1 & 0 & 4 \\
    1 & 1 & 0 & 2 \\
    0 & 0 & 1 & 6 \\
    1 & 0 & 1 & 8 \\
    0 & 1 & 1 & 4 \\
    1 & 1 & 1 & 2 \\
    \bottomrule
  \end{tabular}
  \hfill
  %
  \begin{tabular}{lll}
    \toprule
    $x_4$ & $x_5$ & $\phi_C$\\
    \midrule
    0 & 0 & 8 \\
    1 & 0 & 2 \\
    0 & 1 & 2 \\
    1 & 1 & 6 \\
    \bottomrule
  \end{tabular}
  \hfill
  %
  \begin{tabular}{lll}
    \toprule
    $x_4$ & $x_6$ & $\phi_D$\\
    \midrule
    0 & 0 & 3 \\
    1 & 0 & 6 \\
    0 & 1 & 6 \\
    1 & 1 & 3 \\
    \bottomrule
  \end{tabular}
  \hfill
\end{center}

\begin{exenumerate}

  %
  \item Draw the factor graph corresponding to $p(x_2, x_3, x_4, x_5 \mid x_1=0, x_6=1)$ and give the tables defining the new factors $\phi_A^{x_1=0}(x_2, x_3)$ and $\phi_D^{x_6=1}(x_4)$ that you obtain.

  %
  \begin{solution}
    First condition on $x_1 = 0$:

    Factor node $\phi_A(x_1, x_2, x_3)$ depends on $x_1$, thus we create a new factor $\phi_A^{x_1=0}(x_2, x_3)$ from the table for $\phi_A$ using the rows where $x_1 = 0$.

    \begin{center}
      \begin{tikzpicture}[ugraph]
        \node[fact, label=above: $\phi_A^{x_1=0}$] (fa) at (1.5,0) {} ; %
        \node[cont] (x2) at (2.5,1)  {$x_2$} ; %
        \node[cont] (x3) at (2.5,-1)  {$x_3$} ; %
        \node[fact, label=above: $\phi_B$] (fb) at (3.5,0) {} ; %
        \node[cont] (x4) at (5,0)  {$x_4$} ; %
        \node[fact, label=above: $\phi_C$] (fc) at (6,1) {} ; %
        \node[cont] (x5) at (7.5,1)  {$x_5$} ; %
        \node[cont] (x6) at (7.5,-1)  {$x_6$} ;
        \node[fact, label=above: $\phi_D$] (fd) at (6,-1) {} ; %

        \draw (fd) -- (x6);
        \draw (fa) -- (x2);
        \draw (fa) -- (x3);
        \draw (x2) -- (fb);
        \draw (x3) -- (fb);
        \draw (fb) -- (x4);
        \draw (x4) -- (fc);
        \draw (x4) -- (fd);
        \draw (fc) -- (x5);
      \end{tikzpicture}
    \end{center}

    \begin{center}
      %
      \begin{tabular}{lllll}
        \toprule
        & $x_1$ & $x_2$ & $x_3$ & $\phi_A$\\
        \midrule
        $\rightarrow$ & 0 & 0 & 0 & 4 \\
                      & 1 & 0 & 0 & 2 \\
        $\rightarrow$ & 0 & 1 & 0 & 2 \\
                      & 1 & 1 & 0 & 6 \\
        $\rightarrow$ & 0 & 0 & 1 & 2 \\
                      & 1 & 0 & 1 & 6 \\
        $\rightarrow$ & 0 & 1 & 1 & 6 \\
                      & 1 & 1 & 1 & 4 \\
        \bottomrule
      \end{tabular}
      \hspace{3ex} \text{so that} \hspace{3ex}
      %
      \begin{tabular}{lll}
        \toprule
        $x_2$ & $x_3$ & $\phi_A^{x_1=0}$\\
        \midrule
        0 & 0 & 4 \\
        1 & 0 & 2 \\
        0 & 1 & 2 \\
        1 & 1 & 6 \\
        \bottomrule
      \end{tabular}
    \end{center}

    Next condition on $x_6 = 1$:

    Factor node $\phi_D(x_4, x_6)$ depends on $x_6$, thus we create a new factor $\phi_D^{x_6=1}(x_4)$ from the table for $\phi_D$ using the rows where $x_6 = 1$.

    \begin{center}
      \begin{tikzpicture}[ugraph]
        \node[fact, label=above: $\phi_A^{x_1=0}$] (fa) at (1.5,0) {} ; %
        \node[cont] (x2) at (2.5,1)  {$x_2$} ; %
        \node[cont] (x3) at (2.5,-1)  {$x_3$} ; %
        \node[fact, label=above: $\phi_B$] (fb) at (3.5,0) {} ; %
        \node[cont] (x4) at (5,0)  {$x_4$} ; %
        \node[fact, label=above: $\phi_C$] (fc) at (6,1) {} ; %
        \node[cont] (x5) at (7.5,1)  {$x_5$} ; %
        \node[fact, label=right: $\phi_D^{x_6=1}$] (fd) at (6,-1) {} ; %

        \draw (fa) -- (x2);
        \draw (fa) -- (x3);
        \draw (x2) -- (fb);
        \draw (x3) -- (fb);
        \draw (fb) -- (x4);
        \draw (x4) -- (fc);
        \draw (x4) -- (fd);
        \draw (fc) -- (x5);
      \end{tikzpicture}
    \end{center}

    \begin{center}
      %
      \begin{tabular}{llll}
        \toprule
        & $x_4$ & $x_6$ & $\phi_D$\\
        \midrule
                      & 0 & 0 & 3 \\
                      & 1 & 0 & 6 \\
        $\rightarrow$ & 0 & 1 & 6 \\
        $\rightarrow$ & 1 & 1 & 3 \\
        \bottomrule
      \end{tabular}
      \hspace{3ex} \text{so that} \hspace{3ex}
      %
      \begin{tabular}{ll}
        \toprule
        $x_4$ & $\phi_D^{x_6=1}$\\
        \midrule
        0 & 6 \\
        1 & 3 \\
        \bottomrule
      \end{tabular}
    \end{center}

  \end{solution}

  %
  \item Find $p(x_2 \mid x_1=0, x_6=1)$ using the elimination ordering $(x_4, x_5, x_3)$:

  \begin{exenumerate}
    \item Draw the graph for $p(x_2, x_3, x_5 \mid x_1=0, x_6=1)$ by marginalising $x_4$ \\
    Compute the table for the new factor $\tilde{\phi}_4(x_2, x_3, x_5)$ \\

    \item Draw the graph for $p(x_2, x_3 \mid x_1=0, x_6=1)$ by marginalising $x_5$ \\
    Compute the table for the new factor $\tilde{\phi}_{45}(x_2, x_3)$ \\

    \item Draw the graph for $p(x_2 \mid x_1=0, x_6=1)$ by marginalising $x_3$ \\
    Compute the table for the new factor $\tilde{\phi}_{453}(x_2)$ 
  \end{exenumerate}

  %
  \begin{solution}

    Starting with the factor graph for $p(x_2, x_3, x_4, x_5 \mid x_1=0, x_6=1)$

    \begin{center}
      \begin{tikzpicture}[ugraph]
        \node[fact, label=above: $\phi_A^{x_1=0}$] (fa) at (1.5,0) {} ; %
        \node[cont] (x2) at (2.5,1)  {$x_2$} ; %
        \node[cont] (x3) at (2.5,-1)  {$x_3$} ; %
        \node[fact, label=above: $\phi_B$] (fb) at (3.5,0) {} ; %
        \node[cont] (x4) at (5,0)  {$x_4$} ; %
        \node[fact, label=above: $\phi_C$] (fc) at (6,1) {} ; %
        \node[cont] (x5) at (7.5,1)  {$x_5$} ; %
        \node[fact, label=right: $\phi_D^{x_6=1}$] (fd) at (6,-1) {} ; %

        \draw (fa) -- (x2);
        \draw (fa) -- (x3);
        \draw (x2) -- (fb);
        \draw (x3) -- (fb);
        \draw (fb) -- (x4);
        \draw (x4) -- (fc);
        \draw (x4) -- (fd);
        \draw (fc) -- (x5);
      \end{tikzpicture}
    \end{center}

    Marginalising $x_4$ combines the three factors $\phi_B$, $\phi_C$ and $\phi_D^{x_6=1}$

    \begin{center}
      \begin{tikzpicture}[ugraph]
        \node[fact, label=above: $\phi_A^{x_1=0}$] (fa) at (1.5,0) {} ; %
        \node[cont] (x2) at (2.5,1)  {$x_2$} ; %
        \node[cont] (x3) at (2.5,-1)  {$x_3$} ; %
        \node[fact, label=above: $\tilde{\phi}_4$] (f4) at (3.5,0) {} ; %
        \node[cont] (x5) at (5,0)  {$x_5$} ; %

        \draw (fa) -- (x2);
        \draw (fa) -- (x3);
        \draw (x2) -- (f4);
        \draw (x3) -- (f4);
        \draw (f4) -- (x5);
      \end{tikzpicture}
    \end{center}

    Marginalising $x_5$ modifies the factor $\tilde{\phi}_4$

    \begin{center}
      \begin{tikzpicture}[ugraph]
        \node[fact, label=above: $\phi_A^{x_1=0}$] (fa) at (1.5,0) {} ; %
        \node[cont] (x2) at (2.5,1)  {$x_2$} ; %
        \node[cont] (x3) at (2.5,-1)  {$x_3$} ; %
        \node[fact, label=right: $\tilde{\phi}_{45}$] (f45) at (3.5,0) {} ; %

        \draw (fa) -- (x2);
        \draw (fa) -- (x3);
        \draw (x2) -- (f45);
        \draw (x3) -- (f45);
      \end{tikzpicture}
    \end{center}

    Marginalising $x_3$ combines the factors $\phi_A^{x_1=0}$ and $\tilde{\phi}_{45}$

    \begin{center}
      \begin{tikzpicture}[ugraph]
        \node[cont] (x2) at (0,0)  {$x_2$} ; %
        \node[fact, label=right: $\tilde{\phi}_{453}$] (f453) at (2,0) {} ; %

        \draw (x2) -- (f453);
      \end{tikzpicture}
    \end{center}

    We now compute the tables for the new factors $\tilde{\phi}_4$, $\tilde{\phi}_{45}$, $\tilde{\phi}_{453}$.

    First find $\tilde{\phi}_4(x_2, x_3, x_5)$
    \begin{center}
      %
      \begin{tabular}{llll}
        \toprule
        $x_2$ & $x_3$ & $x_4$ & $\phi_B$\\
        \midrule
        0 & 0 & 0 & 2 \\
        1 & 0 & 0 & 2 \\
        0 & 1 & 0 & 4 \\
        1 & 1 & 0 & 2 \\
        0 & 0 & 1 & 6 \\
        1 & 0 & 1 & 8 \\
        0 & 1 & 1 & 4 \\
        1 & 1 & 1 & 2 \\
        \bottomrule
      \end{tabular}\hspace{2ex}
      %
      \begin{tabular}{lll}
        \toprule
        $x_4$ & $x_5$ & $\phi_C$\\
        \midrule
        0 & 0 & 8 \\
        1 & 0 & 2 \\
        0 & 1 & 2 \\
        1 & 1 & 6 \\
        \bottomrule
      \end{tabular}\hspace{2ex}
      %
      \begin{tabular}{ll}
        \toprule
        $x_4$ & $\phi_D^{x_6=1}$\\
        \midrule
        0 & 6 \\
        1 & 3 \\
        \bottomrule
      \end{tabular}
    \end{center}
    so that  $\phi_*(x_2,x_3,x_4,x_5) = \phi_B(x_2, x_3, x_4) \phi_C(x_4, x_5) \phi_D^{x_6=1}(x_4)$ equals
   %
    \begin{center}
      \begin{tabular}{lllll}
        \toprule
        $x_2$ & $x_3$ & $x_4$& $x_5$ & $\phi_*(x_2,x_3,x_4,x_5)$\\
        \midrule
        0 & 0 & 0 & 0 & 2 * 8 * 6 \\
        1 & 0 & 0 & 0 & 2 * 8 * 6 \\
        0 & 1 & 0 & 0 & 4 * 8 * 6 \\
        1 & 1 & 0 & 0 & 2 * 8 * 6 \\
        0 & 0 & 1 & 0 & 6 * 2 * 3\\
        1 & 0 & 1 & 0 & 8 * 2 * 3\\
        0 & 1 & 1 & 0 & 4 * 2 * 3\\
        1 & 1 & 1 & 0 & 2 * 2 * 3\\
        0 & 0 & 0 & 1 & 2 * 2 * 6\\
        1 & 0 & 0 & 1 & 2 * 2 * 6\\
        0 & 1 & 0 & 1 & 4 * 2 * 6\\
        1 & 1 & 0 & 1 & 2 * 2 * 6\\
        0 & 0 & 1 & 1 & 6 * 6 * 3\\
        1 & 0 & 1 & 1 & 8 * 6 * 3\\
        0 & 1 & 1 & 1 & 4 * 6 * 3\\
        1 & 1 & 1 & 1 & 2 * 6 * 3\\
        \bottomrule
      \end{tabular}
    \end{center}
and
    %
    \begin{center}
      \begin{tabular}{llllll}
        \toprule
        $x_2$ & $x_3$ & $x_5$ & $\sum_{x_4} \phi_B(x_2, x_3, x_4) \phi_C(x_4, x_5) \phi_D^{x_6=1}(x_4)$ &  & $\tilde{\phi}_4$\\
        \midrule
        0 & 0 & 0 & (2 * 8 * 6) + (6 * 2 * 3) & = & 132 \\
        1 & 0 & 0 & (2 * 8 * 6) + (8 * 2 * 3) & = & 144 \\
        0 & 1 & 0 & (4 * 8 * 6) + (4 * 2 * 3) & = & 216 \\
        1 & 1 & 0 & (2 * 8 * 6) + (2 * 2 * 3) & = & 108 \\
        0 & 0 & 1 & (2 * 2 * 6) + (6 * 6 * 3) & = & 132 \\
        1 & 0 & 1 & (2 * 2 * 6) + (8 * 6 * 3) & = & 168 \\
        0 & 1 & 1 & (4 * 2 * 6) + (4 * 6 * 3) & = & 120 \\
        1 & 1 & 1 & (2 * 2 * 6) + (2 * 6 * 3) & = & 60 \\
        \bottomrule
      \end{tabular}
    \end{center}

    Next find $\tilde{\phi}_{45}(x_2, x_3)$
    \begin{center}
      %
      \begin{tabular}{llll}
        \toprule
        $x_2$ & $x_3$ & $x_5$ & $\tilde{\phi}_4$\\
        \midrule
        0 & 0 & 0 & 132 \\
        1 & 0 & 0 & 144 \\
        0 & 1 & 0 & 216 \\
        1 & 1 & 0 & 108 \\
        0 & 0 & 1 & 132 \\
        1 & 0 & 1 & 168 \\
        0 & 1 & 1 & 120 \\
        1 & 1 & 1 & 60 \\
        \bottomrule
      \end{tabular}
      \hspace{3ex} \text{so that} \hspace{3ex}
      %
      \begin{tabular}{lllll}
        \toprule
        $x_2$ & $x_3$ & $\sum_{x_5} \tilde{\phi}_4(x_2, x_3, x_5)$ &  & $\tilde{\phi}_{45}$\\
        \midrule
        0 & 0 & 132 + 132 & = & 264 \\
        1 & 0 & 144 + 168 & = & 312 \\
        0 & 1 & 216 + 120 & = & 336 \\
        1 & 1 & 108 + 60  & = & 168 \\
        \bottomrule
      \end{tabular}
    \end{center}

    Finally find $\tilde{\phi}_{453}(x_2)$
    \begin{center}
      %
      \begin{tabular}{lll}
        \toprule
        $x_2$ & $x_3$ & $\phi_A^{x_1=0}$\\
        \midrule
        0 & 0 & 4 \\
        1 & 0 & 2 \\
        0 & 1 & 2 \\
        1 & 1 & 6 \\
        \bottomrule
      \end{tabular}\hspace{2ex}
      %
      \begin{tabular}{lll}
        \toprule
        $x_2$ & $x_3$ & $\tilde{\phi}_{45}$\\
        \midrule
        0 & 0 & 264 \\
        1 & 0 & 312 \\
        0 & 1 & 336 \\
        1 & 1 & 168 \\
        \bottomrule
      \end{tabular}
      \end{center}
    so that
    \begin{center}
      \begin{tabular}{llll}
        \toprule
        $x_2$ & $\sum_{x_3} \tilde{\phi}_{45}(x_2, x_3) \phi_A^{x_1=0}(x_2,x_3)$  &  & $\tilde{\phi}_{453}$\\
        \midrule
        0 & (4 * 264) + (2 * 336) & = & 1728 \\
        1 & (2 * 312) + (6 * 168) & = & 1632 \\
        \bottomrule
      \end{tabular}
    \end{center}\vspace{1ex}

   The normalising constant is $Z = 1728 + 1632$. Our conditional marginal is thus:
    \begin{equation}
      p(x_2 \mid x_1=0, x_6=1) = 
      \begin{pmatrix}
        1728 / Z \\
        1632 / Z \\
      \end{pmatrix} = 
      \begin{pmatrix}
        0.514 \\
        0.486 \\
      \end{pmatrix}
    \end{equation}

  \end{solution}

  %
  \item Now determine $p(x_2 \mid x_1=0, x_6=1)$ with the elimination ordering $(x_5, x_4, x_3)$:

  \begin{exenumerate}
    \item Draw the graph for $p(x_2, x_3, x_4, \mid x_1=0, x_6=1)$ by marginalising $x_5$ \\
    Compute the table for the new factor $\tilde{\phi}_{5}(x_4)$ \\

    \item Draw the graph for $p(x_2, x_3 \mid x_1=0, x_6=1)$ by marginalising $x_4$ \\
    Compute the table for the new factor $\tilde{\phi}_{54}(x_2, x_3)$ \\

    \item Draw the graph for $p(x_2 \mid x_1=0, x_6=1)$ by marginalising $x_3$ \\
      Compute the table for the new factor $\tilde{\phi}_{543}(x_2)$ 
  \end{exenumerate}

  %
  \begin{solution}

    Starting with the factor graph for $p(x_2, x_3, x_4, x_5 \mid x_1=0, x_6=1)$

    \begin{center}
      \begin{tikzpicture}[ugraph]
        \node[fact, label=above: $\phi_A^{x_1=0}$] (fa) at (1.5,0) {} ; %
        \node[cont] (x2) at (2.5,1)  {$x_2$} ; %
        \node[cont] (x3) at (2.5,-1)  {$x_3$} ; %
        \node[fact, label=above: $\phi_B$] (fb) at (3.5,0) {} ; %
        \node[cont] (x4) at (5,0)  {$x_4$} ; %
        \node[fact, label=above: $\phi_C$] (fc) at (6,1) {} ; %
        \node[cont] (x5) at (7.5,1)  {$x_5$} ; %
        \node[fact, label=right: $\phi_D^{x_6=1}$] (fd) at (6,-1) {} ; %

        \draw (fa) -- (x2);
        \draw (fa) -- (x3);
        \draw (x2) -- (fb);
        \draw (x3) -- (fb);
        \draw (fb) -- (x4);
        \draw (x4) -- (fc);
        \draw (x4) -- (fd);
        \draw (fc) -- (x5);
      \end{tikzpicture}
    \end{center}

    Marginalising $x_5$ modifies the factor $\phi_C$

    \begin{center}
      \begin{tikzpicture}[ugraph]
        \node[fact, label=above: $\phi_A^{x_1=0}$] (fa) at (1.5,0) {} ; %
        \node[cont] (x2) at (2.5,1)  {$x_2$} ; %
        \node[cont] (x3) at (2.5,-1)  {$x_3$} ; %
        \node[fact, label=above: $\phi_B$] (fb) at (3.5,0) {} ; %
        \node[cont] (x4) at (5,0)  {$x_4$} ; %
        \node[fact, label=right: $\tilde{\phi}_5$] (f5) at (6,1) {} ; %
        \node[fact, label=right: $\phi_D^{x_6=1}$] (fd) at (6,-1) {} ; %

        \draw (fa) -- (x2);
        \draw (fa) -- (x3);
        \draw (x2) -- (fb);
        \draw (x3) -- (fb);
        \draw (fb) -- (x4);
        \draw (x4) -- (f5);
        \draw (x4) -- (fd);
      \end{tikzpicture}
    \end{center}

    Marginalising $x_4$ combines the three factors $\phi_B$, $\tilde{\phi}_5$ and $\phi_D^{x_6=1}$

    \begin{center}
      \begin{tikzpicture}[ugraph]
        \node[fact, label=above: $\phi_A^{x_1=0}$] (fa) at (1.5,0) {} ; %
        \node[cont] (x2) at (2.5,1)  {$x_2$} ; %
        \node[cont] (x3) at (2.5,-1)  {$x_3$} ; %
        \node[fact, label=right: $\tilde{\phi}_{54}$] (f54) at (3.5,0) {} ; %

        \draw (fa) -- (x2);
        \draw (fa) -- (x3);
        \draw (x2) -- (f54);
        \draw (x3) -- (f54);
      \end{tikzpicture}
    \end{center}

    Marginalising $x_3$ combines the factors $\phi_A^{x_1=0}$ and $\tilde{\phi}_{54}$

    \begin{center}
      \begin{tikzpicture}[ugraph]
        \node[cont] (x2) at (0,0)  {$x_2$} ; %
        \node[fact, label=right: $\tilde{\phi}_{543}$] (f543) at (2,0) {} ; %

        \draw (x2) -- (f543);
      \end{tikzpicture}
    \end{center}

    We now compute the tables for the new factors $\tilde{\phi}_5$, $\tilde{\phi}_{54}$, and $\tilde{\phi}_{543}$.

    First find $\tilde{\phi}_5(x_4)$
    \begin{center}
      %
      \begin{tabular}{lll}
        \toprule
        $x_4$ & $x_5$ & $\phi_C$\\
        \midrule
        0 & 0 & 8 \\
        1 & 0 & 2 \\
        0 & 1 & 2 \\
        1 & 1 & 6 \\
        \bottomrule
      \end{tabular}
      \hspace{3ex} \text{so that} \hspace{3ex}
      %
      \begin{tabular}{llll}
        \toprule
        $x_4$ & $\sum_{x_5} \phi_C(x_4, x_5)$ &  & $\tilde{\phi}_5$\\
        \midrule
        0 & 8 + 2 & = & 10 \\
        1 & 2 + 6 & = & 8 \\
        \bottomrule
      \end{tabular}
    \end{center}

    Next find $\tilde{\phi}_{54}(x_2, x_3)$
    \begin{center}
      %
      \begin{tabular}{llll}
        \toprule
        $x_2$ & $x_3$ & $x_4$ & $\phi_B$\\
        \midrule
        0 & 0 & 0 & 2 \\
        1 & 0 & 0 & 2 \\
        0 & 1 & 0 & 4 \\
        1 & 1 & 0 & 2 \\
        0 & 0 & 1 & 6 \\
        1 & 0 & 1 & 8 \\
        0 & 1 & 1 & 4 \\
        1 & 1 & 1 & 2 \\
        \bottomrule
      \end{tabular}\hspace{2ex}
      %
      \begin{tabular}{ll}
        \toprule
        $x_4$ & $\tilde{\phi}_5$\\
        \midrule
        0 & 10 \\
        1 & 8 \\
        \bottomrule
      \end{tabular}\hspace{2ex}
      %
      \begin{tabular}{ll}
        \toprule
        $x_4$ & $\phi_D^{x_6=1}$\\
        \midrule
        0 & 6 \\
        1 & 3 \\
        \bottomrule
      \end{tabular}
    \end{center}
      so that $\phi_*(x_2,x_3,x_4) = \phi_B(x_2, x_3, x_4) \tilde{\phi}_5(x_4) \phi_D^{x_6=1}(x_4)$ equals
      \begin{center}
        \begin{tabular}{lllll}
          \toprule
          $x_2$ & $x_3$ & $x_4$ & $\phi_*(x_2,x_3,x_4)$\\
          \midrule
          0 & 0 & 0 & 2 * 10 * 6 \\ 
          1 & 0 & 0 & 2 * 10 * 6 \\
          0 & 1 & 0 & 4 * 10 * 6 \\
          1 & 1 & 0 & 2 * 10 * 6 \\
          0 & 0 & 1 & 6 * 8 * 3 \\
          1 & 0 & 1 & 8 * 8 * 3 \\
          0 & 1 & 1 & 4 * 8 * 3 \\
          1 & 1 & 1 & 2 * 8 * 3 \\
          \bottomrule
        \end{tabular}
      \end{center}
    and
    \begin{center}
      \begin{tabular}{lllll}
        \toprule
        $x_2$ & $x_3$ & $\sum_{x_4} \phi_B(x_2, x_3, x_4) \tilde{\phi}_5(x_4) \phi_D^{x_6=1}(x_4)$ &  & $\tilde{\phi}_{54}$\\
        \midrule
        0 & 0 & (2 * 10 * 6) + (6 * 8 * 3) & = & 264 \\
        1 & 0 & (2 * 10 * 6) + (8 * 8 * 3) & = & 312 \\
        0 & 1 & (4 * 10 * 6) + (4 * 8 * 3) & = & 336 \\
        1 & 1 & (2 * 10 * 6) + (2 * 8 * 3) & = & 168 \\
        \bottomrule
      \end{tabular}
    \end{center}

    Finally find $\tilde{\phi}_{543}(x_2)$
    \begin{center}
      %
      \begin{tabular}{lll}
        \toprule
        $x_2$ & $x_3$ & $\phi_A^{x_1=0}$\\
        \midrule
        0 & 0 & 4 \\
        1 & 0 & 2 \\
        0 & 1 & 2 \\
        1 & 1 & 6 \\
        \bottomrule
      \end{tabular}\hspace{2ex}
      %
      \begin{tabular}{lll}
        \toprule
        $x_2$ & $x_3$ & $\tilde{\phi}_{54}$\\
        \midrule
        0 & 0 & 264 \\
        1 & 0 & 312 \\
        0 & 1 & 336 \\
        1 & 1 & 168 \\
        \bottomrule
      \end{tabular}
      \end{center}
    so that
    \begin{center}
      %
      \begin{tabular}{llll}
        \toprule
        $x_2$ & $\sum_{x_3} \tilde{\phi}_{54}(x_2, x_3) \phi_A^{x_1=0}(x_2,x_3)$ &  & $\tilde{\phi}_{543}$\\
        \midrule
        0 & (4 * 264) + (2 * 336) & = & 1728 \\
        1 & (2 * 312) + (6 * 168) & = & 1632 \\
        \bottomrule
      \end{tabular}
    \end{center}

    As with the ordering in the previous part, we should come to the
    same result for our conditional marginal distribution.The
    normalising constant is $Z = 1728 + 1632$, so that the conditional
    marginal is
    \begin{equation}
      p(x_2 \mid x_1=0, x_6=1) = 
      \begin{pmatrix}
        1728 / Z \\
        1632 / Z \\
      \end{pmatrix} = 
      \begin{pmatrix}
        0.514 \\
        0.486 \\
      \end{pmatrix}
    \end{equation}

  \end{solution}

\item Which variable ordering, $(x_4, x_5, x_3)$ or $(x_5, x_4, x_3)$ do you prefer?

  \begin{solution}
    The ordering $(x_5, x_4, x_3)$ is cheaper and should be preferred
    over the ordering $(x_4, x_5, x_3)$ .

    The reason for the difference in the cost is that $x_4$ has three
    neighbours in the factor graph for $p(x_2, x_3, x_4, x_5 \mid
    x_1=0, x_6=1)$. However, after elimination of $x_5$, which has
    only one neighbour, $x_4$ has only two neighbours
    left. Eliminating variables with more neighbours leads to
    larger (temporary) factors and hence a larger cost. We can see
    this from the tables that were generated during the computation
    (or numbers that we needed to add together): for the ordering
    $(x_4, x_5, x_3)$, the largest table had $2^4$ entries while for
    $(x_5, x_4, x_3)$, it had $2^3$ entries.

    Choosing a reasonable variable ordering has a direct effect on the
    computational complexity of variable elimination. This effect
    becomes even more pronounced when the domain of our discrete
    variables has a size greater than 2 (binary variables), or if the
    variables are continuous.
    
    \begin{center}
      \begin{tikzpicture}[ugraph]
        \node[fact, label=above: $\phi_A^{x_1=0}$] (fa) at (1.5,0) {} ; %
        \node[cont] (x2) at (2.5,1)  {$x_2$} ; %
        \node[cont] (x3) at (2.5,-1)  {$x_3$} ; %
        \node[fact, label=above: $\phi_B$] (fb) at (3.5,0) {} ; %
        \node[cont] (x4) at (5,0)  {$x_4$} ; %
        \node[fact, label=above: $\phi_C$] (fc) at (6,1) {} ; %
        \node[cont] (x5) at (7.5,1)  {$x_5$} ; %
        \node[fact, label=right: $\phi_D^{x_6=1}$] (fd) at (6,-1) {} ; %

        \draw (fa) -- (x2);
        \draw (fa) -- (x3);
        \draw (x2) -- (fb);
        \draw (x3) -- (fb);
        \draw (fb) -- (x4);
        \draw (x4) -- (fc);
        \draw (x4) -- (fd);
        \draw (fc) -- (x5);
      \end{tikzpicture}
    \end{center}

  \end{solution}
  
\end{exenumerate}

\ex{Choice of elimination order in factor graphs}

We would like to compute the marginal $p(x_1)$ by variable
elimination for a joint pmf represented by the following factor
graph. All variables $x_i$ can take $K$ different values. \\
\begin{center}
  \scalebox{1}{ 
    \begin{tikzpicture}[ugraph]
      \node[cont] (x1) at (0,0) {$x_1$};
      \node[fact, label=above: $\phi_a$] (fa) at (1,-1) {};
      \node[cont] (x2) at (2,0) {$x_2$};
      \node[fact, label={[xshift=0.15cm,  yshift=0cm]left: $\phi_b$}] (fb) at (2,-1) {};
      \node[cont] (x3) at (4,0) {$x_3$};
      \node[fact, label=above: $\phi_c$] (fc) at (3,-1) {};

      \node[cont] (x4) at (2,-2) {$x_4$};

      \node[cont] (x5) at (0,-4) {$x_5$};
      \node[fact, label=above: $\phi_d$] (fd) at (1,-3) {};
      \node[cont] (x6) at (2,-4) {$x_6$};
      \node[fact, label={[xshift=0.15cm,  yshift=0cm]left: $\phi_e$}] (fe) at (2,-3) {};
      \node[cont] (x7) at (4,-4) {$x_7$};
      \node[fact, label=above: $\phi_f$] (ff) at (3,-3) {};
      
      \draw(x1) -- (fa);
      \draw(x2) -- (fb);
      \draw(x3) -- (fc);
      \draw(x4) -- (fa);
      \draw(x4) -- (fb);
      \draw(x4) -- (fc);

      \draw(x4) -- (fc);

      \draw(x4) -- (fd);
      \draw(x4) -- (fe);
      \draw(x4) -- (ff);

      \draw(fd) -- (x5);
      \draw(fe) -- (x6);
      \draw(ff) -- (x7);
  \end{tikzpicture}}
\end{center}
\begin{exenumerate}
\item A friend proposes the elimination order $x_4, x_5, x_6, x_7, x_3, x_2$, i.e.\ to do $x_4$ first and $x_2$ last. Explain why this is computationally inefficient.

  \begin{solution}

    According to the factor graph, $p(x_1, \ldots, x_7)$ factorises as
    \begin{align}
      p(x_1, \ldots, x_7) &\propto \phi_a(x_1,x_4)\phi_b(x_2,x_4)\phi_c(x_3,x_4)\phi_d(x_5,x_4)\phi_e(x_6,x_4)\phi_f(x_7,x_4)
    \end{align}
    If we choose to eliminate $x_4$ first, i.e. compute
    \begin{align}
      p(x_1,x_2,x_3,x_5, x_6, x_7) &= \sum_{x_4}   p(x_1, \ldots, x_7) \\
      & \propto \sum_{x_4} \phi_a(x_1,x_4)\phi_b(x_2,x_4)\phi_c(x_3,x_4)\phi_d(x_5,x_4)\phi_e(x_6,x_4)\phi_f(x_7,x_4)
    \end{align}
    we cannot pull any of the factors out of the sum since each of them depends on $x_4$. This means the
    cost to sum out $x_4$ for all combinations of the six variables
    $(x_1,x_2,x_3,x_5, x_6, x_7)$ is $K^7$. Moreover, the new factor
    \begin{equation}
      \tilde{\phi}(x_1,x_2,x_3,x_5, x_6, x_7) =  \sum_{x_4} \phi_a(x_1,x_4)\phi_b(x_2,x_4)\phi_c(x_3,x_4)\phi_d(x_5,x_4)\phi_e(x_6,x_4)\phi_f(x_7,x_4)
    \end{equation}
    does not factorise anymore so that subsequent variable eliminations will be expensive too.
  \end{solution}
  
\item Propose an elimination ordering that achieves $O(K^2)$ computational cost per variable elimination and explain why it does so.

  \begin{solution}

    Any ordering where $x_4$ is eliminated last will do. At any stage,
    elimination of one of the variables $x_2, x_3, x_5, x_6, x_7$ is
    then a $O(K^2)$ operation. This is because e.g.\
  \begin{align}
    p(x_1, \ldots, x_6) & = \sum_{x_7} p(x_1, \ldots, x_7)\\
    & \propto \phi_a(x_1,x_4)\phi_b(x_2,x_4)\phi_c(x_3,x_4)\phi_d(x_5,x_4)\phi_e(x_6,x_4) \underbrace{\sum_{x_7} \phi_f(x_7,x_4)}_{\tilde{\phi}_7(x_4)}\\
    & \propto \phi_a(x_1,x_4)\phi_b(x_2,x_4)\phi_c(x_3,x_4)\phi_d(x_5,x_4)\phi_e(x_6,x_4) \tilde{\phi}_7(x_4)
  \end{align}
  where computing $\tilde{\phi}_7(x_4)$ for all values of $x_4$ is $O(K^2)$. Further,
  \begin{align}
    p(x_1, \ldots, x_5) & = \sum_{x_6} p(x_1, \ldots, x_6)\\
    & \propto \phi_a(x_1,x_4)\phi_b(x_2,x_4)\phi_c(x_3,x_4)\phi_d(x_5,x_4) \tilde{\phi}_7(x_4) \sum_{x_6}\phi_e(x_6,x_4)\\
    & \propto \phi_a(x_1,x_4)\phi_b(x_2,x_4)\phi_c(x_3,x_4)\phi_d(x_5,x_4) \tilde{\phi}_7(x_4) \tilde{\phi}_6(x_4), 
  \end{align}
  where computation of $\tilde{\phi}_6(x_4)$ for all values of $x_4$ is again $O(K^2)$. Continuing in this manner, one obtains
  \begin{align}
    p(x_1, x_4) &\propto \phi_a(x_1,x_4) \tilde{\phi}_2(x_4)\tilde{\phi}_3(x_4)\tilde{\phi}_5(x_4)\tilde{\phi}_6(x_4)\tilde{\phi}_7(x_4).
  \end{align}
  where each derived factor $\tilde{\phi}$ has $O(K^2)$ cost. Summing out $x_4$ and normalising the pmf is again a $O(K^2)$ operation.
  \end{solution}
  
\end{exenumerate}

\chapter{Inference for Hidden Markov Models}
\minitoc

\ex{Predictive distributions for hidden Markov models}
\label{ex:predictive-distributions-for-hidden-markov-models}
For the hidden Markov model
$$ p(h_{1:d},v_{1:d}) = p(v_1|h_1)p(h_1)\prod_{i=2}^d
p(v_i|h_i)p(h_i|h_{i-1})$$ assume you have observations for $v_i$,
$i=1, \ldots, u < d$.

\begin{exenumerate}
\item \label{q:h-predictive-hmm} Use message passing to compute $p(h_t|v_{1:u})$ for $u<t\le
  d$. For the sake of concreteness, you may consider the case $d=6, u=2,
  t=4$.

  \begin{solution}
    The factor graph for $d=6, u=2$, with messages that are required
    for the computation of $p(h_t|v_{1:u})$ for $t=4$, is as follows.
    \begin{center}
      \scalebox{1}{
        \begin{tikzpicture}[ugraph,minimum size=1cm, inner sep=3pt]
          
          \node[fact, label=above:$\phi_1$] (f1) at (-1,2) {};;
          \node[cont] (h1) at (0,2) {$h_1$};
          \node[fact, label=above:$\phi_2$] (f2) at (1,2) {};
          \node[cont] (h2) at (2,2) {$h_2$};
          \node[fact, label=above: {\footnotesize$p(h_3 | h_2)$}] (f3) at (3,2) {};
          \node[cont] (h3) at (4,2) {$h_3$};
          \node[fact, label=above: {\footnotesize$p(h_4 | h_3)$}] (f4) at (5,2) {};
          \node[cont] (h4) at (6,2) {$h_4$};
          \node[fact, label=above:$p(h_5 | h_4)$] (f5) at (7,2) {};
          \node[cont] (h5) at (8,2) {$h_5$};
          \node[fact, label=above:$p(h_6 | h_5)$] (f6) at (9,2) {};
          \node[cont] (h6) at (10,2) {$h_6$};

          \node[cont] (v3) at (4,0) {$v_3$};
          \node[fact, label={[xshift=0.1cm,  yshift=0cm]left: {\footnotesize$p(v_3 | h_3)$}}] (fv3) at (4,1) {};
          \node[cont] (v4) at (6,0) {$v_4$};
          \node[fact, label={[xshift=0.1cm,  yshift=0cm]left: {\footnotesize$p(v_4 | h_4)$}}] (fv4) at (6,1) {};
          \node[cont] (v5) at (8,0) {$v_5$};
          \node[fact, label={[xshift=0.1cm,  yshift=0cm]left: {\footnotesize$p(v_5 | h_5)$}}] (fv5) at (8,1) {};
          \node[cont] (v6) at (10,0) {$v_6$};
          \node[fact, label={[xshift=0.1cm,  yshift=0cm]left: {\footnotesize$p(v_6 | h_6)$}}] (fv6) at (10,1) {};
          
          \draw(h5)--(v5);
          \draw(h6)--(v6);
          \draw(h1)--(h2);
          \draw(h2)--(h3);
          \draw(h3)--(h4);
          \draw(h4)--(h5);
          \draw(h5)--(h6);
          \draw(f1) -- (h1);    

          \draw(f1)--(h1) node[midway, above, yshift=-7pt] {$\rightarrow$};
          \draw(h1)--(f2) node[midway, above, yshift=-7pt] {$\rightarrow$};
          \draw(f2)--(h2) node[midway, above, yshift=-7pt] {$\rightarrow$};
          \draw(h2)--(f3) node[midway, above, yshift=-7pt] {$\rightarrow$};
          \draw(f3)--(h3) node[midway, above, yshift=-7pt] {$\rightarrow$};
          \draw(h3)--(f4) node[midway, above, yshift=-7pt] {$\rightarrow$};
          \draw(f4)--(h4) node[midway, above, yshift=-7pt] {\red{$\rightarrow$}};

          \draw(h4)--(f5) node[midway, above, yshift=-7pt] {\red{$\leftarrow$}};
          \draw(f5)--(h5) node[midway, above, yshift=-7pt] {$\leftarrow$};
          \draw(h5)--(f6) node[midway, above, yshift=-7pt] {$\leftarrow$};
          \draw(f6)--(h6) node[midway, above, yshift=-7pt] {$\leftarrow$};

          \draw(fv3)--(h3) node[midway, right, xshift=-7pt] {$\uparrow$};
          \draw(v3)--(fv3) node[midway, right, xshift=-7pt] {$\uparrow$};
          \draw(fv4)--(h4) node[midway, right, xshift=-7pt] {$\uparrow$};
          \draw(v4)--(fv4) node[midway, right, xshift=-7pt] {$\uparrow$};
          \draw(fv5)--(h5) node[midway, right, xshift=-7pt] {$\uparrow$};
          \draw(v5)--(fv5) node[midway, right, xshift=-7pt] {$\uparrow$};
          \draw(fv6)--(h6) node[midway, right, xshift=-7pt] {$\uparrow$};
          \draw(v6)--(fv6) node[midway, right, xshift=-7pt] {$\uparrow$};
          
        \end{tikzpicture}
      }
    \end{center}
    The messages from the unobserved visibles $v_i$ to their
    corresponding $h_i$, e.g.\ $v_3$ to $h_3$, are all one. Moreover,
    the message from the $p(h_5|h_4)$ node to $h_4$ equals one as well. This
    is because all involved factors, $p(v_i|h_i)$ and
    $p(h_i|h_{i-1})$, sum to one. Hence the factor graph reduces to a chain:
    \begin{center}
      \scalebox{1}{
        \begin{tikzpicture}[ugraph]         
          \node[fact, label=above: $\phi_1$] (f1) at (-1.5,2) {};
          \node[cont] (h1) at (0,2) {$h_1$};
          \node[fact, label=above: $\phi_2$] (f2) at (1.5,2) {};
          \node[cont] (h2) at (3,2) {$h_2$};
          \node[fact, label=above: $p(h_3|h_2)$] (f3) at (4.5,2) {};
          \node[cont] (h3) at (6,2) {$h_3$};
          \node[fact, label=above: $p(h_4|h_3)$] (f4) at (7.5,2) {};
          \node[cont] (h4) at (9,2) {$h_4$};
          \draw(f1) -- (h1)  node[midway,above,sloped] {$\rightarrow$};
          \draw(h1)--(f2);
          \draw(f2)--(h2) node[midway,above,sloped] {$\rightarrow$};
          \draw(h2)--(f3);
          \draw(f3)--(h3) node[midway,above,sloped] {$\rightarrow$};
          \draw(h3)--(f4);
          \draw(f4)--(h4) node[midway,above,sloped] {\red{$\rightarrow$}}; 
      \end{tikzpicture}}
    \end{center}
    Since the variable nodes copy the messages in case of a chain, we only show
    the factor-to-variable messages.

    The graph shows that we are essentially in the same situation as
    in filtering, with the difference that we use the factors
    $p(h_s|h_{s-1})$ for $s\ge u+1$. Hence, we can use filtering to compute
    the messages until time $s=u$ and then compute the further
    messages with the $p(h_s|h_{s-1})$ as factors. This gives the
    following algorithm:
    \begin{enumerate}
    \item Compute $\alpha(h_u)$ by filtering.
    \item For $s=u+1, \ldots, t$, compute
      \begin{equation}
        \alpha(h_s) = \sum_{h_{s-1}} p(h_s|h_{s-1}) \alpha(h_{s-1})
      \end{equation}
    \item The required predictive distribution is
      \begin{equation}
        p(h_t | v_{1:u}) = \frac{1}{Z} \alpha(h_t) \quad \quad Z = \sum_{h_t} \alpha(h_t)
      \end{equation}
    \end{enumerate}
    For $s \ge u+1$, we have that
    \begin{align}
      \sum_{h_s} \alpha(h_s) &= \sum_{h_s} \sum_{h_{s-1}} p(h_s|h_{s-1}) \alpha(h_{s-1})\\
      & =  \sum_{h_{s-1}} \alpha(h_{s-1})
    \end{align}
    since $p(h_s|h_{s-1})$ is normalised. This means that the normalising constant $Z$ above equals
    \begin{equation}
      Z = \sum_{h_u} \alpha(h_u) = p(v_{1:u})
    \end{equation}
    which is the likelihood.

    For filtering, we have seen that $\alpha(h_s) \propto
    p(h_s|v_{1:s})$, $s\le u$. The $\alpha(h_s)$ for all $s>u$ are
    proportional to $p(h_s| v_{1:u})$. This may be seen by noting that
    the above arguments hold for any $t>u$. 
  \end{solution}

\item Use message passing to compute $p(v_t|v_{1:u})$ for $u<t\le
  d$. For the sake of concreteness, you may consider the case $d=6,
  u=2, t=4$.

  \begin{solution}
    The factor graph for $d=6, u=2$, with messages that are required
    for the computation of $p(v_t|v_{1:u})$ for $t=4$, is as follows.
    \begin{center}
      \scalebox{1}{
        \begin{tikzpicture}[ugraph,minimum size=1cm, inner sep=3pt]
          
          \node[fact, label=above:$\phi_1$] (f1) at (-1,2) {};;
          \node[cont] (h1) at (0,2) {$h_1$};
          \node[fact, label=above:$\phi_2$] (f2) at (1,2) {};
          \node[cont] (h2) at (2,2) {$h_2$};
          \node[fact, label=above: {\footnotesize$p(h_3 | h_2)$}] (f3) at (3,2) {};
          \node[cont] (h3) at (4,2) {$h_3$};
          \node[fact, label=above: {\footnotesize$p(h_4 | h_3)$}] (f4) at (5,2) {};
          \node[cont] (h4) at (6,2) {$h_4$};
          \node[fact, label=above:$p(h_5 | h_4)$] (f5) at (7,2) {};
          \node[cont] (h5) at (8,2) {$h_5$};
          \node[fact, label=above:$p(h_6 | h_5)$] (f6) at (9,2) {};
          \node[cont] (h6) at (10,2) {$h_6$};

          \node[cont] (v3) at (4,0) {$v_3$};
          \node[fact, label={[xshift=0.1cm,  yshift=0cm]left: {\footnotesize$p(v_3 | h_3)$}}] (fv3) at (4,1) {};
          \node[cont] (v4) at (6,0) {$v_4$};
          \node[fact, label={[xshift=0.1cm,  yshift=0cm]left: {\footnotesize$p(v_4 | h_4)$}}] (fv4) at (6,1) {};
          \node[cont] (v5) at (8,0) {$v_5$};
          \node[fact, label={[xshift=0.1cm,  yshift=0cm]left: {\footnotesize$p(v_5 | h_5)$}}] (fv5) at (8,1) {};
          \node[cont] (v6) at (10,0) {$v_6$};
          \node[fact, label={[xshift=0.1cm,  yshift=0cm]left: {\footnotesize$p(v_6 | h_6)$}}] (fv6) at (10,1) {};
          
          \draw(h5)--(v5);
          \draw(h6)--(v6);
          \draw(h1)--(h2);
          \draw(h2)--(h3);
          \draw(h3)--(h4);
          \draw(h4)--(h5);
          \draw(h5)--(h6);
          \draw(f1) -- (h1);    

          \draw(f1)--(h1) node[midway, above, yshift=-7pt] {$\rightarrow$};
          \draw(h1)--(f2) node[midway, above, yshift=-7pt] {$\rightarrow$};
          \draw(f2)--(h2) node[midway, above, yshift=-7pt] {$\rightarrow$};
          \draw(h2)--(f3) node[midway, above, yshift=-7pt] {$\rightarrow$};
          \draw(f3)--(h3) node[midway, above, yshift=-7pt] {$\rightarrow$};
          \draw(h3)--(f4) node[midway, above, yshift=-7pt] {$\rightarrow$};
          \draw(f4)--(h4) node[midway, above, yshift=-7pt] {$\rightarrow$};

          \draw(h4)--(f5) node[midway, above, yshift=-7pt] {$\leftarrow$};
          \draw(f5)--(h5) node[midway, above, yshift=-7pt] {$\leftarrow$};
          \draw(h5)--(f6) node[midway, above, yshift=-7pt] {$\leftarrow$};
          \draw(f6)--(h6) node[midway, above, yshift=-7pt] {$\leftarrow$};

          \draw(fv3)--(h3) node[midway, right, xshift=-7pt] {$\uparrow$};
          \draw(v3)--(fv3) node[midway, right, xshift=-7pt] {$\uparrow$};
          \draw(fv4)--(h4) node[midway, right, xshift=-7pt] {$\downarrow$};
          \draw(v4)--(fv4) node[midway, right, xshift=-7pt] {\red{$\downarrow$}};
          \draw(fv5)--(h5) node[midway, right, xshift=-7pt] {$\uparrow$};
          \draw(v5)--(fv5) node[midway, right, xshift=-7pt] {$\uparrow$};
          \draw(fv6)--(h6) node[midway, right, xshift=-7pt] {$\uparrow$};
          \draw(v6)--(fv6) node[midway, right, xshift=-7pt] {$\uparrow$};
          
        \end{tikzpicture}
      }
    \end{center}
    Due to the normalised factors, as above, the messages to the right
    of $h_t$ are all one. Moreover the messages that go up from the
    $v_i$ to the $h_i, i\neq t$, are also all one. Hence the graph simplifies to a chain.
    \begin{center}
      \scalebox{1}{
        \begin{tikzpicture}[ugraph]         
          \node[fact, label=above: $\phi_1$] (f1) at (-1.5,2) {};
          \node[cont] (h1) at (0,2) {$h_1$};
          \node[fact, label=above: $\phi_2$] (f2) at (1.5,2) {};
          \node[cont] (h2) at (3,2) {$h_2$};
          \node[fact, label=above: $p(h_3|h_2)$] (f3) at (4.5,2) {};
          \node[cont] (h3) at (6,2) {$h_3$};
          \node[fact, label=above: $p(h_4|h_3)$] (f4) at (7.5,2) {};
          \node[cont] (h4) at (9,2) {$h_4$};
          \node[fact, label=above: $p(v_4|h_4)$] (fv4) at (10.5,2) {};
          \node[cont] (v4) at (12,2) {$v_4$};

          \draw(f1) -- (h1)  node[midway,above,sloped] {$\rightarrow$};
          \draw(h1)--(f2);
          \draw(f2)--(h2) node[midway,above,sloped] {$\rightarrow$};
          \draw(h2)--(f3);
          \draw(f3)--(h3) node[midway,above,sloped] {$\rightarrow$};
          \draw(h3)--(f4);
          \draw(f4)--(h4) node[midway,above,sloped] {\blue{$\rightarrow$}};
          \draw(h4)--(fv4);
          \draw(fv4)--(v4) node[midway,above,sloped] {\red{$\rightarrow$}};
      \end{tikzpicture}}
    \end{center}
    The message in blue is proportional to $p(h_t|v_{1:u})$ computed
    in question \ref{q:h-predictive-hmm}. Thus assume that we have
    computed $p(h_t|v_{1:u})$. The predictive distribution on the
    level of the visibles thus is
    \begin{equation}
      p(v_t | v_{1:u}) = \sum_{h_t} p(v_t|h_t) p(h_t|v_{1:u}).
    \end{equation}
    This follows from message passing since the last node ($h_4$ in
    the graph) just copies the (normalised) message and the next
    factor equals $p(v_t|h_t)$.

    An alternative derivation follows from basic definitions and
    operations, together with the independencies in HMMs:
    \begin{align}
    &\text{\scriptsize (sum rule)}&  p(v_t | v_{1:u}) &= \sum_{h_t} p(v_t, h_t|v_{1:u})\\
    &\text{\scriptsize (product rule)}&  & = \sum_{h_t} p(v_t|h_t, v_{1:u}) p(h_t|v_{1:u})\\
    &\text{\scriptsize ($v_t \independent v_{1:u} \mid h_t$)}&  & = \sum_{h_t} p(v_t|h_t) p(h_t|v_{1:u})
    \end{align}
  \end{solution}
  
\end{exenumerate}

\ex{Viterbi algorithm}

For the hidden Markov model
$$ p(h_{1:t},v_{1:t}) = p(v_1|h_1)p(h_1)\prod_{i=2}^t
p(v_i|h_i)p(h_i|h_{i-1})$$ assume you have observations for $v_i$,
$i=1, \ldots, t$. Use the max-sum algorithm to derive an iterative algorithm to compute
\begin{equation}
  \hat{\h} = \argmax_{h_1, \ldots, h_t}  p(h_{1:t}|v_{1:t})
\end{equation}
Assume that the latent variables $h_i$ can take $K$ different values,
e.g.\ $h_i \in \{0,\ldots, K-1\}$. The resulting algorithm is known as
Viterbi algorithm.

\begin{solution}
  
  We first form the factors
  \begin{align}
    \phi_1(h_1) &= p(v_1|h_1) p(h_1)& \phi_2(h_1, h_2) &= p(v_2|h_2) p(h_2|h_1)\\
    \ldots      &                  & \phi_t(h_{t-1},h_t) &= p(v_t|h_t) p(h_{t}|h_{t-1})
  \end{align}
 where the $v_i$ are known and fixed. The posterior $p(h_1, \ldots,
 h_t | v_1, \ldots, v_t)$ is then represented by the following factor
 graph (assuming $t=4$).
  
  \begin{center}
    \scalebox{1}{
      \begin{tikzpicture}[ugraph]         
        \node[fact, label=above: $\phi_1$] (f1) at (-1.5,2) {};
        \node[cont] (h1) at (0,2) {$h_1$};
        \node[fact, label=above: $\phi_2$] (f2) at (1.5,2) {};
        \node[cont] (h2) at (3,2) {$h_2$};
        \node[fact, label=above: $\phi_3$] (f3) at (4.5,2) {};
        \node[cont] (h3) at (6,2) {$h_3$};
        \node[fact, label=above: $\phi_4$] (f4) at (7.5,2) {};
        \node[cont] (h4) at (9,2) {$h_4$};
        
        \draw(f1)--(h1);
        \draw(h1)--(f2);
        \draw(f2)--(h2);
        \draw(h2)--(f3);
        \draw(f3)--(h3);
        \draw(h3)--(f4);
        \draw(f4)--(h4);
    \end{tikzpicture}}
  \end{center}

  For the max-sum algorithm, we here choose $h_t$ to be the root. We
  thus initialise the algorithm with $\mfxmess{\phi_1}{h_1}{1}(h_1) =
  \log \phi_1(h_1) = \log p(v_1|h_1) + \log p(h_1)$ and then compute the
  messages from left to right, moving from the leaf $\phi_1$ to the
  root $h_t$.

  Since we are dealing with a chain, the variable nodes, much like in
  the sum-product algorithm, just copy the incoming messages. It thus
  suffices to compute the factor to variable messages shown in the
  graph, and then backtrack to $h_1$.
  
 \begin{center}
      \scalebox{1}{
        \begin{tikzpicture}[ugraph]         
          \node[fact, label=above: $\phi_1$] (f1) at (-1.5,2) {};
          \node[cont] (h1) at (0,2) {$h_1$};
          \node[fact, label=above: $\phi_2$] (f2) at (1.5,2) {};
          \node[cont] (h2) at (3,2) {$h_2$};
          \node[fact, label=above: $\phi_3$] (f3) at (4.5,2) {};
          \node[cont] (h3) at (6,2) {$h_3$};
          \node[fact, label=above: $\phi_4$] (f4) at (7.5,2) {};
          \node[cont] (h4) at (9,2) {$h_4$};

          \draw(f1) -- (h1)  node[midway,above,sloped] {$\rightarrow$};
          \draw(h1)--(f2);
          \draw(f2)--(h2) node[midway,above,sloped] {$\rightarrow$};
          \draw(h2)--(f3);
          \draw(f3)--(h3) node[midway,above,sloped] {$\rightarrow$};
          \draw(h3)--(f4);
          \draw(f4)--(h4) node[midway,above,sloped] {$\rightarrow$};
      \end{tikzpicture}}
 \end{center}

 With $\mxfmess{h_{i-1}}{\phi_i}{1}(h_{i-1})=\mfxmess{\phi_{i-1}}{h_{i-1}}{1}(h_{i-1})$, the factor-to-variable update equation is
 \begin{align}
   \mfxmess{\phi_i}{h_i}{1}(h_i) &= \max_{h_{i-1}} \log \phi_i(h_{i-1}, h_i) + \mxfmess{h_{i-1}}{\phi_i}{1}(h_{i-1})\\
   &= \max_{h_{i-1}} \log \phi_i(h_{i-1}, h_i) + \mfxmess{\phi_{i-1}}{h_{i-1}}{1}(h_{i-1})
 \end{align}
 To simplify notation, denote $\mfxmess{\phi_i}{h_i}{1}(h_i)$ by $V_i(h_i)$. We thus have
 \begin{align}
   V_1(h_1) & = \log p(v_1|h_1) + \log p(h_1) \\
   V_i(h_i) & =  \max_{h_{i-1}} \log \phi_i(h_{i-1}, h_i) + V_{i-1}(h_{i-1}) \quad \quad i=2, \ldots, t
 \end{align}
 In general, $V_1(h_1)$ and $V_i(h_i)$ are functions that depend on
 $h_1$ and $h_i$, respectively.  Assuming that the $h_i$ can take on
 the values $0, \ldots, K-1$, the above equations can be written as
\begin{align}
  v_{1,k} &= \log p(v_1|k) + \log p(k) &  k&=0, \ldots, K-1\\
  v_{i,k} & =  \max_{m \in{0, \ldots, K-1}} \log \phi_i(m, k) + v_{i-1,m} &  k&=0, \ldots, K-1, \quad i=2, \ldots, t,
 \end{align}
At the end of the algorithm, we thus have a $t \times K$ matrix $\V$
with elements $v_{i,k}$.

The maximisation can be performed by computing the temporary matrix
$\A$ (via broadcasting) where the $(m,k)$-th element is $\log
\phi_i(m, k) + v_{i-1,m}$. Maximisation then corresponds to
determining the maximal value in each column.

To support the backtracking, when we compute $V_i(h_i)$ by maximising
over $h_{i-1}$, we compute at the same time the look-up table
\begin{equation}
  \gamma_i^*(h_i)   =  \argmax_{h_{i-1}} \log \phi_i(h_{i-1}, h_i) + V_{i-1}(h_{i-1})
\end{equation}
When $h_i$ takes on the values $0, \ldots, K-1$, this can be written as
\begin{equation}
  \gamma^*_{i,k} = \argmax_{m \in{0, \ldots, K-1}} \log \phi_i(m, k) + v_{i-1,m}
\end{equation}
This is the (row) index of the maximal element in each column of the temporary matrix $\A$.

After computing $v_{t,k}$ and $\gamma^*_{t,k}$, we then perform
backtracking via
\begin{align}
  \hat{h}_t & = \argmax_{k}   v_{t,k}\\
  \hat{h}_i & = \gamma^*_{i+1,\hat{h}_{i+1}} \quad \quad i=t-1, \ldots, 1
\end{align}
This gives recursively $\hat{\h} = (\hat{h}_1, \ldots, \hat{h}_t) = \argmax_{h_1, \ldots, h_t}  p(h_{1:t}|v_{1:t})$.

\end{solution}


\ex{Forward filtering backward sampling for hidden Markov models}
\label{ex:FFBS}

Consider the hidden Markov model specified by the following DAG.

\begin{center}
  \scalebox{0.8}{
    \begin{tikzpicture}[dgraph, minimum size=0.5cm, inner sep=3pt]
      \node[cont] (h1) at (0,2) {$h_1$};
      \node       (aa) at (1,2) {};
      \node       (bb) at (2,2) {$\ldots$};
      \node       (cc) at (2,0) {$\ldots$};
      \node       (dd) at (3,2) {};

      \node[cont, inner sep=0em] (h2) at (5,2) {$h_{t-1}$};
      \node[cont] (h3) at (7,2) {$h_{t}$};
      
      \node       (a) at (8,2) {};
      \node       (b) at (9,2) {$\ldots$};
      \node       (c) at (9,0) {$\ldots$};
      \node       (d) at (10,2) {};
      
      \node[cont] (h4) at (12,2) {$h_n$};

      \node[cont] (v1) at (0,0) {$v_1$};
      \node[cont, inner sep=0em] (v2) at (5,0) {$v_{t-1}$};
      \node[cont] (v3) at (7,0) {$v_t$};
      \node[cont] (v4) at (12,0) {$v_n$};
      \draw(h1)--(v1);\draw(h2)--(v2);\draw(h3)--(v3);\draw(h4)--(v4);
      \draw(h1)--(aa);\draw(dd)--(h2);\draw(h2)--(h3);\draw(h3)--(a);\draw(d)--(h4);
  \end{tikzpicture}}
\end{center}
We assume that have already run the alpha-recursion (filtering) and
can compute $p(h_t | v_{1:t})$ for all $t$. The goal is now to
generate samples $p(h_1, \ldots, h_n | v_{1:n})$, i.e.\ entire
trajectories $(h_1, \ldots, h_n)$ from the posterior. Note that this is
not the same as sampling from the $n$ filtering distributions $p(h_t |
v_{1:t})$. Moreover, compared to the Viterbi algorithm, the sampling approach
generates samples from the full posterior rather than just returning
the most probable state and its corresponding probability.

\begin{exenumerate}
\item Show that $p(h_1, \ldots, h_n|v_{1:n})$ forms a
  first-order Markov chain.

  \begin{solution}
    There are several ways to show this. The simplest is to notice
    that the undirected graph for the hidden Markov model is the same
    as the DAG but with the arrows removed as there are no colliders
    in the DAG. Moreover, conditioning corresponds to removing nodes
    from an undirected graph. This leaves us with a chain that
    connects the $h_i$.
    \begin{center}
      \scalebox{0.8}{
        \begin{tikzpicture}[ugraph]
          \node[cont] (h1) at (0,2) {$h_1$};
          \node[cont] (h2) at (2,2) {$h_2$};
          \node[cont] (h3) at (4,2) {$h_3$};
          \node       (a) at (6,2) {};
          \node       (b) at (7,2) {$\ldots$};
          \node       (d) at (8,2) {};
          \node[cont] (h4) at (10,2) {$h_n$};
          \draw(h1)--(h2);\draw(h2)--(h3);\draw(h3)--(a);\draw(d)--(h4);
      \end{tikzpicture}}
    \end{center}
    
  By graph separation, we see that $p(h_1,\ldots, h_n|v_{1:n})$ forms
  a first-order Markov chain so that e.g.\ $h_{1:t-1} \independent
  h_{t+1:n} | h_t$ (past independent from the future given the
  present).
    
  \end{solution}
  
  \item Since $p(h_1, \ldots, h_n|v_{1:n})$ is a first-order Markov
    chain, it suffices to determine $p(h_{t-1}|h_t, v_{1:n})$, the
    probability mass function for $h_{t-1}$ given $h_t$ and
    all the data $v_{1:n}$. Use message passing to show
    that
    \begin{equation}
      p(h_{t-1}, h_t| v_{1:n}) \propto \alpha(h_{t-1}) \beta(h_t) p(h_t|h_{t-1})p(v_t|h_t) 
    \end{equation}

    \begin{solution}
      Since all visibles are in the conditioning set, i.e.\ assumed
      observed, we can represent the conditional model $p(h_1, \ldots,
      h_n|v_{1:n})$ as a chain factor tree, e.g.\ as follows in case
      of $n=4$
      \begin{center}
        \scalebox{1}{
          \begin{tikzpicture}[ugraph]         
            \node[fact, label=above: $\phi_1$] (f1) at (-1.5,2) {};
            \node[cont] (h1) at (0,2) {$h_1$};
            \node[fact, label=above: $\phi_2$] (f2) at (1.5,2) {};
            \node[cont] (h2) at (3,2) {$h_2$};
            \node[fact, label=above: $\phi_3$] (f3) at (4.5,2) {};
            \node[cont] (h3) at (6,2) {$h_3$};
            \node[fact, label=above: $\phi_4$] (f4) at (7.5,2) {};
            \node[cont] (h4) at (9,2) {$h_4$};
            \draw(f1) -- (h1);
            \draw(h1)--(f2);
            \draw(f2)--(h2);
            \draw(h2)--(f3);
            \draw(f3)--(h3);
            \draw(h3)--(f4);
            \draw(f4)--(h4);
        \end{tikzpicture}}
      \end{center}
      Combining the emission distributions $p(v_s|h_s)$ (and marginal
      $p(h_1)$) with the transition distributions $p(h_s|h_{s-1})$ we
      obtain the factors
       \begin{align}
         \phi_1(h_1) & = p(h_1) p(v_1|h_1)\\
         \phi_s(h_{s-1}, h_s) &= p(h_s|h_{s-1})p(v_s|h_s) \quad \text{for } t=2, \ldots, n
       \end{align}
       We see from the factor tree that $h_{t-1}$ and $h_t$ are
       neighbours, being attached to the same factor node
       $\phi_t(h_{t-1}, h_t)$, e.g. $\phi_3$ in case of $p(h_{2}, h_3 | v_{1:4})$.

       By the rules of message passing, the joint $p(h_{t-1}, h_{t} |
       v_{1:n})$ is thus proportional to $\phi_t$ times the messages
       into $\phi_t$. The following graph shows the messages for the
       case of $p(h_{2}, h_3 | v_{1:4})$.
       \begin{center}
         \scalebox{1}{
           \begin{tikzpicture}[ugraph]         
             \node[fact, label=above: $\phi_1$] (f1) at (-1.5,2) {};
             \node[cont] (h1) at (0,2) {$h_1$};
             \node[fact, label=above: $\phi_2$] (f2) at (1.5,2) {};
             \node[cont] (h2) at (3,2) {$h_2$};
             \node[fact, label=above: $\phi_3$] (f3) at (4.5,2) {};
             \node[cont] (h3) at (6,2) {$h_3$};
             \node[fact, label=above: $\phi_4$] (f4) at (7.5,2) {};
             \node[cont] (h4) at (9,2) {$h_4$};
             \draw(f1) -- (h1);
             \draw(h1)--(f2);
             \draw(f2)--(h2);
             \draw(h2)--(f3) node[midway,above,sloped] {$\rightarrow$};
             \draw(f3)--(h3) node[midway,above,sloped] {$\leftarrow$}; 
             \draw(h3)--(f4);
             \draw(f4)--(h4);
         \end{tikzpicture}}
       \end{center}
       Since the variable nodes only receive single messages from any
       direction, they copy the messages so that the messages into
       $\phi_t$ are given by $\alpha(h_{t-1})$ and $\beta(h_t)$ shown
       below in red and blue, respectively.
       \begin{center}
         \scalebox{1}{
           \begin{tikzpicture}[ugraph]         
             \node[fact, label=above: $\phi_1$] (f1) at (-1.5,2) {};
             \node[cont] (h1) at (0,2) {$h_1$};
             \node[fact, label=above: $\phi_2$] (f2) at (1.5,2) {};
             \node[cont] (h2) at (3,2) {$h_2$};
             \node[fact, label=above: $\phi_3$] (f3) at (4.5,2) {};
             \node[cont] (h3) at (6,2) {$h_3$};
             \node[fact, label=above: $\phi_4$] (f4) at (7.5,2) {};
             \node[cont] (h4) at (9,2) {$h_4$};
             \draw(f1) -- (h1);
             \draw(h1)--(f2);
             \draw(f2)--(h2) node[midway,above,sloped] {\red{$\rightarrow$}};
             \draw(h2)--(f3);
             \draw(f3)--(h3);
             \draw(h3)--(f4) node[midway,above,sloped] {\blue{$\leftarrow$}}; 
             \draw(f4)--(h4);
         \end{tikzpicture}}
       \end{center}
       Hence, 
       \begin{align}
         p(h_{t-1}, h_t| v_{1:n}) &\propto \alpha(h_{t-1}) \beta(h_t) \phi_t(h_{t-1}, h_t)\\
         &\propto \alpha(h_{t-1}) \beta(h_t) p(h_t|h_{t-1}) p(v_t|h_t)
       \end{align}
       which is the result that we want to show.
    \end{solution}

  \item Show that $p(h_{t-1}|h_t, v_{1:n}) = \frac{\alpha(h_{t-1})}{\alpha(h_t)} p(h_t|h_{t-1}) p(v_t|h_t)$.
   
    \begin{solution}
      The conditional $p(h_{t-1}|h_t, v_{1:n})$ can be written as the ratio
      \begin{equation}
        p(h_{t-1}|h_t, v_{1:n}) = \frac{p(h_{t-1}, h_t| v_{1:n})}{p(h_t|v_{1:n})}.
      \end{equation}
      Above, we have shown that the numerator satisfies
      \begin{equation}
        p(h_{t-1}, h_t| v_{1:n}) \propto \alpha(h_{t-1}) \beta(h_t) p(h_t|h_{t-1}) p(v_t|h_t).
      \end{equation}
      The denominator $p(h_t|v_{1:n})$ is proportional to $\alpha(h_t)
      \beta(h_t)$ since it is the smoothing distribution than can be
      determined via the alpha-beta recursion.

      Normally, we needed to sum the messages over all values of
      $(h_{t-1},h_t)$ to find the normalising constant of the
      numerator. For the denominator, we had to sum over all values of
      $h_t$. Next, I will argue qualitatively that this summation is
      not needed; the normalising constants are both equal to
      $p(v_{1:t})$. A more mathematical argument is given below.

      We started with a factor graph and factors that represent the
      joint $p(h_{1:n}, v_{1:n})$. The conditional $p(h_{1:n},
      v_{1:n})$ equals
      \begin{equation}
        \label{eq:h-given-v-hmm}
        p(h_{1:n} | v_{1:n}) = \frac{p(h_{1:n}, v_{1:n})}{p(v_{1:n})}
      \end{equation}
      Message passing is variable elimination. Hence, when computing
      $p(h_t|v_{1:n})$ as $\alpha(h_t) \beta(h_t)$ from a factor graph
      for $p(h_{1:n}, v_{1:n})$, we only need to divide by
      $p(v_{1:n})$ for normalisation; explicitly summing out $h_t$ is
      not needed. In other words,
      \begin{equation}
        p(h_t|v_{1:n}) =  \frac{\alpha(h_t) \beta(h_t)}{p(v_{1:n})}.
      \end{equation}
      Similarly, $p(h_{t-1}, h_t|v_{1:n})$ is also obtained from
      \eqref{eq:h-given-v-hmm} by marginalisation/variable
      elimination. Again, when computing $p(h_{t-1}, h_t|v_{1:n})$ as
      $\alpha(h_{t-1}) \beta(h_t) p(h_t|h_{t-1}) p(v_t|h_t)$ from a
      factor graph for $p(h_{1:n}, v_{1:n})$, we do not need to
      explicitly sum over all values of $h_t$ and $h_{t-1}$ for
      normalisation. The definition of the factors in the factor graph
      together with \eqref{eq:h-given-v-hmm} shows that we can simply
      divide by $p(v_{1:n})$. This gives
      \begin{align}
        p(h_{t-1}, h_t| v_{1:n}) & = \frac{1}{p(v_{1:n})} \alpha(h_{t-1}) \beta(h_t) p(h_t|h_{t-1}) p(v_t|h_t).
      \end{align}

      The desired conditional thus is
      \begin{align}
        p(h_{t-1} | h_t, v_{1:n}) & = \frac{p(h_{t-1}, h_t| v_{1:n})}{p(h_t|v_{1:n})} \\
        & = \frac{\alpha(h_{t-1}) \beta(h_t) p(h_t|h_{t-1}) p(v_t|h_t)}{\alpha(h_t) \beta(h_t)}\\
        & = \frac{\alpha(h_{t-1}) p(h_t|h_{t-1}) p(v_t|h_t)}{\alpha(h_t)}
      \end{align}
      which is the result that we wanted to show. Note that
      $\beta(h_t)$ cancels out and that $p(h_{t-1} | h_t, v_{1:n})$
      only involves the $\alpha$'s, the (forward) transition
      distribution $p(h_t|h_{t-1})$ and the emission distribution at
      time $t$.

      \emph{Alternative solution:} An alternative, mathematically rigorous
      solution is as follows.  The conditional $p(h_{t-1}|h_t,
      v_{1:n})$ can be written as the ratio
      \begin{equation}
        p(h_{t-1}|h_t, v_{1:n}) = \frac{p(h_{t-1}, h_t| v_{1:n})}{p(h_t|v_{1:n})}.
      \end{equation}
      We first determine the denominator. From the properties of the
      alpha and beta recursion, we know that
      \begin{align}
        \alpha(h_{t}) &= p(h_{t}, v_{1:t}) & \beta(h_t) = p(v_{t+1:n}|h_t)
      \end{align}
      Using that $v_{t+1:n} \independent v_{1:t} | h_t$, we can thus
      express the denominator $p(h_t|v_{1:n})$ as
      \begin{align}
        p(h_t|v_{1:n}) &= \frac{p(h_t, v_{1:n})}{p(v_{1:n})}\\
        & = \frac{p(h_t, v_{1:t}) p(v_{t+1:n}|h_t)}{p(v_{1:n})}\\
        & = \frac{\alpha(h_t) \beta(h_t)}{p(v_{1:n})}
      \end{align}
      
      For the numerator, we have
      \begin{align}
        p(h_{t-1},h_t | v_{1:n}) & =  \frac{p(h_{t-1},h_t, v_{1:n})}{p(v_{1:n})} \\
        & = \frac{p(h_{t-1}, v_{1:t-1}, h_t, v_{t:n})}{p(v_{1:n})} \\
        & = \frac{p(h_{t-1}, v_{1:t-1}) p(h_t, v_{t:n} | h_{t-1}, v_{1:t-1})}{p(v_{1:n})} \\
        & = \frac{p(h_{t-1}, v_{1:t-1}) p(h_t, v_{t:n} | h_{t-1})}{p(v_{1:n})} \quad \quad \text{\small (using $h_t, v_{1:t} \independent v_{1:t-1} | h_{t-1}$)}\\
        & = \frac{\alpha(h_{t-1}) p(h_t, v_{t:n} | h_{t-1})}{p(v_{1:n})} \quad \quad \text{\small (using $\alpha(h_{t-1}) = p(h_{t-1}, v_{1:t-1})$)}
      \end{align}
      With the product rule, we have  $p(h_t, v_{t:n} | h_{t-1}) =  p(v_t|h_t, h_{t-1}, v_{t+1:n}) p(h_t, v_{t+1:n} | h_{t-1})$ so that
      \begin{align}   
  p(h_{t-1},h_t | v_{1:n})& = \frac{\alpha(h_{t-1}) p(v_t|h_t, h_{t-1}, v_{t+1:n}) p(h_t, v_{t+1:n} | h_{t-1})}{p(v_{1:n})}\\
       & = \frac{\alpha(h_{t-1}) p(v_t|h_t) p(h_t, v_{t+1:n} | h_{t-1})}{p(v_{1:n})} \quad \quad \text{\small (using $v_t \independent h_{t-1}, v_{t+1:n} |h_t$)}\\
        & = \frac{\alpha(h_{t-1}) p(v_t|h_t) p(h_t | h_{t-1}) p(v_{t+1:n} | h_{t-1}, h_t)}{p(v_{1:n})} 
        \end{align}
        Hence
        \begin{align}
       p(h_{t-1},h_t | v_{1:n}) & = \frac{\alpha(h_{t-1}) p(v_t|h_t) p(h_t | h_{t-1}) p(v_{t+1:n} | h_t)}{p(v_{1:n})} \quad \quad \text{\small (using $v_{t+1:n} \independent h_{t-1} | h_t$)}\\
& = \frac{\alpha(h_{t-1}) p(v_t|h_t) p(h_t | h_{t-1}) \beta(h_t)}{p(v_{1:n})} \quad \quad \text{\small (using $\beta(h_t) = p(v_{t+1:n}|h_t)$)}
      \end{align}

      The desired conditional thus is
      \begin{align}
        p(h_{t-1} | h_t, v_{1:n}) & = \frac{p(h_{t-1}, h_t| v_{1:n})}{p(h_t|v_{1:n})} \\
        & = \frac{\alpha(h_{t-1}) \beta(h_t) p(h_t|h_{t-1}) p(v_t|h_t)}{\alpha(h_t) \beta(h_t)}\\
        & = \frac{\alpha(h_{t-1}) p(h_t|h_{t-1}) p(v_t|h_t)}{\alpha(h_t)}
      \end{align}
      which is the result that we wanted to show.
    \end{solution}

    We thus obtain the following algorithm to generate samples from
    $p(h_1, \ldots, h_n|v_{1:n})$:
    \begin{enumerate}
    \item Run the alpha-recursion (filtering) to determine all
      $\alpha(h_t)$ forward in time for $t=1, \ldots, n$.
    \item Sample $h_n$ from $p(h_n | v_{1:n}) \propto \alpha(h_n)$
    \item Go backwards in time using
      \begin{equation}
        p(h_{t-1}|h_t, v_{1:n}) = \frac{\alpha(h_{t-1})}{\alpha(h_t)} p(h_t|h_{t-1}) p(v_t|h_t)
      \end{equation}
      to generate samples $h_{t-1}|h_t, v_{1:n}$ for $t=n, \ldots, 2$.
    \end{enumerate}
    This algorithm is known as forward filtering backward sampling (FFBS). 
    
\end{exenumerate}


 \ex{Prediction exercise}
 
 Consider a hidden Markov model with three visibles $v_1, v_2, v_3$
 and three hidden variables $h_1, h_2, h_3$ which can be represented
 with the following factor graph:
 
 \begin{center}
   \scalebox{0.9}{
     \begin{tikzpicture}[ugraph,minimum size=1cm, inner sep=3pt]
       \node[cont] (v1) at (0,0) {$v_1$};
       \node[fact, label={[xshift=0.1cm,  yshift=0cm]left: $p(v_1 | h_1)$}] (fv1) at (0,1) {};
       \node[cont] (v2) at (3,0) {$v_2$};
       \node[fact, label={[xshift=0.1cm,  yshift=0cm]left: $p(v_2 | h_2)$}] (fv2) at (3,1) {};
       \node[cont] (v3) at (6,0) {$v_3$};
       \node[fact, label={[xshift=0.1cm,  yshift=0cm]left: $p(v_3 | h_3)$}] (fv3) at (6,1) {};
       
       \node[fact, label=left: $p(h_1)$] (f1) at (-1.5,2) {};
       \node[cont] (h1) at (0,2) {$h_1$};
       \node[fact, label={[label distance=-7pt]above: $p(h_2 | h_1)$}] (f2) at (1.5,2) {};
       \node[cont] (h2) at (3,2) {$h_2$};
       \node[fact, label={[label distance=-7pt]above: $p(h_3 | h_2)$}] (f3) at (4.5,2) {};
       \node[cont] (h3) at (6,2) {$h_3$};

       \draw(h1)--(v1);
       \draw(h2)--(v2);
       \draw(h3)--(v3);
       \draw(h1)--(h2);
       \draw(h2)--(h3);
       \draw(f1) -- (h1);
       
   \end{tikzpicture}}
 \end{center}
 This question is about computing the predictive probability $p(v_3=1 | v_1 =1)$.

\begin{exenumerate}
  
\item The factor graph below represents $p(h_1, h_2, h_3, v_2, v_3
  \mid v_1 =1)$. Provide an equation that defines $\phi_A$ in terms of
  the factors in the factor graph above. 

  \begin{center}
    \scalebox{0.9}{
      \begin{tikzpicture}[ugraph,minimum size=1cm, inner sep=3pt]
        \node[cont] (v2) at (3,0) {$v_2$};
        \node[fact, label={[xshift=0.1cm,  yshift=0cm]left: $p(v_2 | h_2)$}] (fv2) at (3,1) {};
        \node[cont] (v3) at (6,0) {$v_3$};
        \node[fact, label={[xshift=0.1cm,  yshift=0cm]left: $p(v_3 | h_3)$}] (fv3) at (6,1) {};
        
        \node[cont] (h1) at (0,2) {$h_1$};
        \node[fact, label={[label distance=-7pt]above: $\phi_A$}] (f2) at (1.5,2) {};
        \node[cont] (h2) at (3,2) {$h_2$};
        \node[fact, label={[label distance=-7pt]above: $p(h_3 | h_2)$}] (f3) at (4.5,2) {};
        \node[cont] (h3) at (6,2) {$h_3$};
        
        \draw(h2)--(v2);
        \draw(h3)--(v3);
        \draw(h1)--(h2);
        \draw(h2)--(h3);
                
    \end{tikzpicture}}
  \end{center}

  \begin{solution}
    
    $\phi_A(h_1, h_2) \propto p(v_1 | h_1) p(h_1) p(h_2 | h_1)$ with $v_1=1$.
    
  \end{solution}
  
\item Assume further that all variables are
  binary, $h_i \in \{0,1\}$, $v_i \in \{0,1\}$; that $p(h_1=1)=0.5$, and that the transition and emission distributions are, for all $i$, given by:

  \begin{center}
    \begin{tabular}{lll}
      \toprule
      $p(h_{i+1}|h_i)$ & $h_{i+1}$ &$h_i$ \\
      \midrule
      0 & 0 & 0\\
      1 & 1 & 0\\
      1 & 0 & 1\\
      0 & 1 & 1\\
      \bottomrule
    \end{tabular}
    \hspace{5ex}
    \begin{tabular}{lll}
      \toprule
      $p(v_i|h_i)$ & $v_i$ &$h_i$ \\
      \midrule
      0.6 & 0 & 0\\
      0.4 & 1 & 0\\
      0.4 & 0 & 1\\
      0.6 & 1 & 1\\
      \bottomrule
    \end{tabular}
  \end{center}
  Compute the numerical values of the factor $\phi_A$.

  \begin{solution}

  \item Given the definition of the transition and emission
    probabilities, we have $\phi_A(h_1,h_2) = 0$ if $h_1=h_2$. For $h_1=0, h_2 =1$, we obtain
    \begin{align}
      \phi_A(h_1=0, h_2=1) &= p(v_1=1 | h_1=0) p(h_1=0) p(h_2=1 | h_1=0) \\
      & = 0.4 \cdot 0.5 \cdot 1\\
      & = \frac{4}{10} \cdot \frac{1}{2}\\
      & = \frac{2}{10} = 0.2
    \end{align}
    For $h_1=1, h_2 =0$, we obtain
    \begin{align}
      \phi_A(h_1=1, h_2=0) &= p(v_1=1 | h_1=1) p(h_1=1) p(h_2=0 | h_1=1) \\
      & = 0.6 \cdot 0.5 \cdot 1\\
      & = \frac{6}{10} \cdot \frac{1}{2}\\
      & = \frac{3}{10} =0.3
    \end{align}
    Hence
     \begin{center}
     \begin{tabular}{lll}
      \toprule
   $\phi_A(h_1, h_2)$ & $h_1$ &$h_2$ \\
      \midrule
      0 & 0 & 0\\
      0.3 & 1 & 0\\
      0.2 & 0 & 1\\
      0 & 1 & 1\\
      \bottomrule
    \end{tabular}
  \end{center}
    
  \end{solution}

\item  Denote the message from variable node $h_2$ to factor node $p(h_3 |
  h_2)$ by $\alpha(h_2)$. Use message passing to compute $\alpha(h_2)$
  for $h_2=0$ and $h_2=1$. Report the values of any intermediate
  messages that need to be computed for the computation of
  $\alpha(h_2)$.

  \begin{solution}
    
    The message from $h_1$ to $\phi_A$ is one. The message from
    $\phi_A$ to $h_2$ is 
    \begin{align}
      \fxmess{\phi_A}{h_2}{1}(h_2=0) & = \sum_{h_1} \phi_A(h_1, h_2=0)\\
      & = 0.3\\
      \fxmess{\phi_A}{h_2}{1}(h_2=1) & = \sum_{h_1} \phi_A(h_1, h_2=1)\\
      & = 0.2
    \end{align}
    
    Since $v_2$ is not observed and $p(v_2|h_2)$ normalised, the message from $p(v_2|h_2)$ to $h_2$ equals one.

    This means that the message from $h_2$ to $p(h_3 | h_2)$, which is
    $\alpha(h_2)$ equals $\fxmess{\phi_A}{h_2}{1}(h_2)$, i.e.\ 
    \begin{align}
      \alpha(h_2=0) & = 0.3\\
      \alpha(h_2=1) & = 0.2
    \end{align}

  \end{solution}
  
\item With $\alpha(h_2)$ defined as above, use message passing to show
  that the predictive probability $p(v_3=1 | v_1=1)$ can be expressed in terms of $\alpha(h_2)$
  as
  \begin{equation}
    p(v_3=1 | v_1=1) = \frac{ x \alpha(h_2=1) + y \alpha(h_2=0)}{\alpha(h_2=1)+\alpha(h_2=0)}
  \end{equation}
  and report the values of $x$ and $y$.  

  \begin{solution}

    Given the definition of $p(h_3| h_2)$, the message $\fxmess{p(h_3|h_2)}{h_3}{1}(h_3)$ is
    \begin{align}
      \fxmess{p(h_3|h_2)}{h_3}{1}(h_3=0) & = \alpha(h_2 =1)\\
      \fxmess{p(h_3|h_2)}{h_3}{1}(h_3=1) & = \alpha(h_2 =0)
  \end{align}
    
  The variable node $h_3$ copies the message so that we have
  \begin{align}
    \fxmess{p(v_3|h_3)}{v_3}{1}(v_3=0) & = \sum_{h_3} p(v_3=0 | h_3)  \fxmess{p(h_3|h_2)}{h_3}{1}(h_3) \\
    & =  p(v_3=0 | h_3=0)  \alpha(h_2 =1) + p(v_3=0 | h_3=1)  \alpha(h_2 =0)\\
    & = 0.6  \alpha(h_2 =1) + 0.4  \alpha(h_2 =0)\\
    \fxmess{p(v_3|h_h3)}{v_3}{1}(v_3=1) & = \sum_{h_3} p(v_3=1 | h_3) ) \fxmess{p(h_3|h_2)}{h_3}{1}(h_3) \\
    & = p(v_3=1 | h_3=0)  \alpha(h_2 =1) + p(v_3=1 | h_3=1)  \alpha(h_2 =0)\\
    & = 0.4 \alpha(h_2=1) + 0.6 \alpha(h_2=0)
  \end{align}

  We thus have
  \begin{align}
    p(v_3 = 1| v_1=1) &=  \frac{0.4 \alpha(h_2=1) + 0.6 \alpha(h_2=0)}{ 0.4 \alpha(h_2=1) + 0.6 \alpha(h_2=0) +  0.6  \alpha(h_2 =1) + 0.4  \alpha(h_2 =0)}\\
    & =  \frac{0.4 \alpha(h_2=1) + 0.6 \alpha(h_2=0)}{\alpha(h_2=1)+\alpha(h_2=0)}
  \end{align}
  The requested $x$ and $y$ are thus: $x =0.4$, $y=0.6$.

  \end{solution}
  
\item Compute the numerical value of $p(v_3 =1 | v_1=1)$.

  \begin{solution}
    
    Inserting the numbers gives $\alpha(h_2=0)+\alpha(h_2=1) = 5/10 = 1/2$ so that
    \begin{align}
      p(v_3 = 1| v_1=1) &= \frac{0.4 \cdot 0.2 + 0.6 \cdot 0.3}{\frac{1}{2}}\\
      &= 2\cdot \left( \frac{4}{10} \cdot \frac{2}{10} + \frac{6}{10} \frac{3}{10}\right)\\
      &=  \frac{4}{10} \cdot \frac{4}{10} + \frac{6}{10} \frac{6}{10}\\
      &= \frac{1}{100} (16+ 36)\\
      &= \frac{1}{100} 52\\
      & = \frac{52}{100} = 0.52
    \end{align}

  \end{solution}
  
\end{exenumerate}


\ex{Hidden Markov models and change of measure}

We take here a change of measure perspective on the alpha-recursion.

Consider the following directed graph for a hidden Markov model
where the $y_i$ correspond to observed (visible) variables and the
$x_i$ to unobserved (hidden/latent) variables.

\begin{center}
  \scalebox{1}{
    \begin{tikzpicture}[dgraph]
      \node[cont] (x1) at (0,2) {$x_1$};
      \node[cont] (x2) at (2,2) {$x_2$};
      \node[cont] (x3) at (4,2) {$x_3$};
      \node       (a) at (6,2) {};
      \node       (b) at (7,2) {$\ldots$};
      \node       (c) at (7,0) {$\ldots$};
      \node       (d) at (8,2) {};
      \node[cont] (x4) at (10,2) {$x_n$};
      \node[cont] (y1) at (0,0) {$y_1$};
      \node[cont] (y2) at (2,0) {$y_2$};
      \node[cont] (y3) at (4,0) {$y_3$};
      \node[cont] (y4) at (10,0) {$y_n$};
      \draw(x1)--(y1);\draw(x2)--(y2);\draw(x3)--(y3);\draw(x4)--(y4);
      \draw(x1)--(x2);\draw(x2)--(x3);\draw(x3)--(a);\draw(d)--(x4);
  \end{tikzpicture}}
\end{center}

The joint model for $\x=(x_1, \ldots, x_n)$ and $\y = (y_1, \ldots,
y_n)$ thus is
 \begin{align}
   p(\x,\y) = p(x_1) \prod_{i=2}^n p(x_i|x_{i-1}) \prod_{i=1}^n p(y_i|x_i).
 \end{align}

 \begin{exenumerate}

 \item Show that
   \begin{equation}
     p(x_1, \ldots, x_n, y_1, \ldots, y_t) =  p(x_1) \prod_{i=2}^n p(x_i|x_{i-1}) \prod_{i=1}^t p(y_i|x_i)
   \end{equation}
   for $t=0, \ldots, n$. We take the case $t=0$ to correspond to $p(x_1, \ldots, x_n)$,
   \begin{equation}
     p(x_1, \ldots, x_n) =  p(x_1) \prod_{i=2}^n p(x_i|x_{i-1}).
   \end{equation}
   
   \begin{solution}
     The result follows by integrating/summing out $y_{t+1} \ldots n$.
     \begin{align}
       p(x_1, \ldots, x_n, y_1, \ldots, y_t) & = \int p(x_1, \ldots, x_n, y_1, \ldots, y_n) \ud y_{t+1} \ldots \ud y_n \\
       & = \int  p(x_1) \prod_{i=2}^n p(x_i|x_{i-1}) \prod_{i=1}^n p(y_i|x_i)\ud y_{t+1} \ldots \ud y_n \\
       & =  p(x_1) \prod_{i=2}^n p(x_i|x_{i-1}) \prod_{i=1}^t p(y_i|x_i) \int  \prod_{i=t+1}^n p(y_i|x_i)\ud y_{t+1} \ldots \ud y_n \\
       &=  p(x_1) \prod_{i=2}^n p(x_i|x_{i-1}) \prod_{i=1}^t p(y_i|x_i)   \prod_{i=t+1}^n \underbrace{\int p(y_i|x_i)\ud y_{i}}_{=1} \\
       & =  p(x_1) \prod_{i=2}^n p(x_i|x_{i-1}) \prod_{i=1}^t p(y_i|x_i)
     \end{align}
     The result for $p(x_1, \ldots, x_n)$ is obtained when we integrate out all
     $y$'s.

   \end{solution}

 \item Show that $p(x_1, \ldots, x_n | y_1, \ldots, y_t)$, $t=0,
   \ldots, n$, factorises as
\begin{equation}
  p(x_1, \ldots, x_n | y_1, \ldots, y_t) \propto  p(x_1) \prod_{i=2}^n p(x_i|x_{i-1}) \prod_{i=1}^t g_i(x_i)
  \label{eq:HMM-conditional-factorisation}
\end{equation}
where $g_i(x_i) = p(y_i|x_i)$ for a fixed value of $y_i$, and that its normalising constant $Z_t$ equals the likelihood $p(y_1, \ldots, y_t)$

\begin{solution}
The result follows from the basic definition of the conditional
\begin{align}
  p(x_1, \ldots, x_n | y_1, \ldots, y_t) & = \frac{p(x_1, \ldots, x_n, y_1, \ldots, y_t)}{p(y_1, \ldots, y_t)}
\end{align}
together with the expression for $p(x_1, \ldots, x_n, y_1, \ldots, y_t)$ when the $y_i$ are kept fixed.
\end{solution}

\item Denote $ p(x_1, \ldots, x_n|y_1, \ldots, y_t)$ by $p_t(x_1,
  \ldots, x_n)$. The index $t \le n$ thus indicates the time of the
  last $y$-variable we are conditioning on. Show the following
  recursion for $1 \le t \le n$:
  \begin{align}
    p_{t-1}(x_1, \ldots, x_t) &=
    \begin{cases}
      p(x_1) & \text{if } t=1\\
      p_{t-1}(x_1, \ldots, x_{t-1}) p(x_t|x_{t-1}) & \text{otherwise}
      \label{eq:HMM-extension}
    \end{cases} && \text{\small (extension)}\\
  p_{t}(x_1, \ldots, x_{t}) &= \frac{1}{Z_t} p_{t-1}(x_1, \ldots, x_t) g_t(x_t) \label{eq:HMM-change-of-measure} && \text{\small (change of measure)}\\
  Z_t & = \int p_{t-1}(x_t) g_t(x_t) \ud x_t 
\end{align}
By iterating from $t=1$ to $t=n$, we can thus recursively compute
$p(x_1, \ldots, x_n|y_1, \ldots, y_n)$, including its normalising
constant $Z_n$, which equals the likelihood $ Z_n= p(y_1, \ldots, y_n)$ 

\begin{solution}
  We start with \eqref{eq:HMM-conditional-factorisation} which shows
  that by definition of $p_t(x_1, \ldots, x_n)$ we have
  \begin{align}
    p_t(x_1, \ldots, x_n) & = p(x_1, \ldots, x_n | y_1, \ldots, y_t)\\
    &\propto p(x_1) \prod_{i=2}^n p(x_i|x_{i-1}) \prod_{i=1}^t g_i(x_i)
    \label{eq:HMM-conditional-factorisation-2} 
  \end{align}
  For $t=1$, we thus have
  \begin{equation}
    p_1(x_1, \ldots, x_n) \propto  p(x_1) \prod_{i=2}^n p(x_i|x_{i-1}) g_1(x_1)
  \end{equation}
  Integrating out $x_2, \ldots, x_n$ gives
  \begin{align}
    p_1(x_1) & = \int p_1(x_1, \ldots, x_n) \ud x_2 \ldots \ud x_n\\
    &\propto \int  p(x_1) \prod_{i=2}^n p(x_i|x_{i-1}) g_1(x_1)\ud x_2 \ldots \ud x_n\\
    &\propto p(x_1) g_1(x_1) \int  \prod_{i=2}^n p(x_i|x_{i-1})\ud x_2 \ldots \ud x_n\\
    &\propto p(x_1) g_1(x_1) \underbrace{\prod_{i=2}^n \int p(x_i|x_{i-1})\ud x_i}_{=1}\\
    &\propto p(x_1) g_1(x_1)
  \end{align}

  The normalising constant is
  \begin{align}
    Z_1 & = \int  p(x_1) g_1(x_1) \ud x_1
  \end{align}
  This establishes the result for $t=1$.

  From \eqref{eq:HMM-conditional-factorisation}, we further have
   \begin{align}
     p_{t-1}(x_1, \ldots, x_n) & = p(x_1, \ldots, x_n | y_1, \ldots, y_{t-1})\\
     &\propto p(x_1) \prod_{i=2}^n p(x_i|x_{i-1}) \prod_{i=1}^{t-1} g_i(x_i)
   \end{align}
   Integrating out $x_{t+1}, \ldots, x_n$ thus gives
   \begin{align}
     p_{t-1}(x_1, \ldots, x_t) & = \int p_{t-1}(x_1, \ldots, x_n) \ud x_{t+1} \ldots \ud x_n\\
     &\propto \int p(x_1) \prod_{i=2}^n p(x_i|x_{i-1}) \prod_{i=1}^{t-1} g_i(x_i)\ud x_{t+1} \ldots \ud x_n\\
     & \propto  p(x_1) \prod_{i=2}^t p(x_i|x_{i-1}) \prod_{i=1}^{t-1} g_i(x_i) \int \prod_{i=t+1}^n p(x_i|x_{i-1}) \ud x_{t+1} \ldots \ud x_n\\
     & \propto  p(x_1) \prod_{i=2}^t p(x_i|x_{i-1}) \prod_{i=1}^{t-1} g_i(x_i) \prod_{i=t+1}^n \int p(x_i|x_{i-1}) \ud x_i\\
     & \propto  p(x_1) \prod_{i=2}^t p(x_i|x_{i-1}) \prod_{i=1}^{t-1} g_i(x_i)
   \end{align}
    Noting that the product over the $g_i$ does not involve $x_t$ and
    that $p(x_t | x_{t-1})$ is a pdf, we have further
    \begin{align}
      p_{t-1}(x_1, \ldots, x_{t-1}) &=\int  p_{t-1}(x_1, \ldots, x_t) \ud x_t\\
      &  \propto p(x_1) \prod_{i=2}^{t-1} p(x_i|x_{i-1}) \prod_{i=1}^{t-1} g_i(x_i)
    \end{align}
    Hence
    \begin{equation}
      p_{t-1}(x_1, \ldots, x_t) =  p_{t-1}(x_1, \ldots, x_{t-1}) p(x_t|x_{t-1})
    \end{equation}
    Note that we can have an equal sign since $p(x_t|x_{t-1})$ is a
    pdf and hence integrates to one. This is sometimes called the
    ``extension'' since the inputs for $p_{t-1}$ are extended from
    $(x_1, \ldots, x_{t-1})$ to $x_1, \ldots, x_t$.
  
    From \eqref{eq:HMM-conditional-factorisation-2}, we further have
    \begin{align}
      p_t(x_1, \ldots, x_n) & \propto  p_{t-1}(x_1, \ldots, x_n) g_t(x_t)
    \end{align}
    Integrating out $x_{t+1}, \ldots, x_n$ thus gives
    \begin{align}
      p_t(x_1, \ldots, x_t) & \propto  p_{t-1}(x_1, \ldots, x_t) g_t(x_t)
    \end{align}
    This is a change of measure from $p_{t-1}(x_1, \ldots, x_t)$ to
    $p_t(x_1, \ldots, x_t)$. Note that $p_{t-1}(x_1, \ldots, x_t)$ only
    involves $g_i$, and hence observations $y_i$, up to index (time) $t-1$. The
    change of measure multiplies-in the additional factor
    $g_t(x_t)=p(y_t|x_t)$, and thereby incorporates the observation at index (time) $t$
    into the model.
  
    The stated recursion is complete by computing the normalising constant $Z_t$ for
    $p_t(x_1, \ldots, x_t)$, which equals
    \begin{align}
      Z_t & = \int  p_{t-1}(x_1, \ldots, x_t) g_t(x_t) \ud x_1, \ldots \ud x_t\\
      & = \int g_t(x_t) \left[\int p_{t-1}(x_1, \ldots, x_t) \ud x_1, \ldots \ud x_{t-1} \right] \ud x_t\\
      & = \int g_t(x_t) p_{t-1}(x_t) \ud x_t
    \end{align}
    
    This recursion, and some slight generalisations, forms the basis
    for what is known as the ``forward recursion'' in particle
    filtering and sequential Monte Carlo. An excellent introduction to
    these topics is book \citep{Chopin2020}.
  
\end{solution}

\item Use the recursion above to derive the following form of the alpha recursion: 
  \begin{align}
    p_{t-1}(x_{t-1}, x_t) &= p_{t-1}(x_{t-1}) p(x_t|x_{t-1}) && \text{\small (extension)} \\
    p_{t-1}(x_t) & = \int  p_{t-1}(x_{t-1}, x_t) \ud x_{t-1} &&\text{\small (marginalisation)}\\
    p_{t}(x_t) &= \frac{1}{Z_t} p_{t-1}(x_t) g_t(x_t)  &&\text{\small (change of measure)}\\
    Z_t & = \int p_{t-1}(x_t) g_t(x_t) \ud x_t
  \end{align}
  with $p_0(x_1) = p(x_1)$.
  
  The term $p_{t}(x_t)$ corresponds to $\alpha(x_t)$ from the
  alpha-recursion after normalisation. Moreover, $p_{t-1}(x_t)$ is the
  predictive distribution for $x_t$ given observations until time
  $t-1$. Multiplying $p_{t-1}(x_t)$ with $g_t(x_t)$ gives the new
  $\alpha(x_t)$. The term $g_t(x_t) = p(y_t|x_t)$ is sometimes called
  the ``correction'' term. We see here that the correction has the
  effect of a change of measure, changing the predictive distribution
  $p_{t-1}(x_t)$ into the filtering distribution $p_t(x_t)$.
  
  \begin{solution}
  Let $t>1$. With \eqref{eq:HMM-extension}, we have
    \begin{align}
      p_{t-1}(x_{t-1}, x_t) & = \int p_{t-1}(x_1, \ldots, x_t) \ud x_1 \ldots \ud x_{t-2}\\
    &=  \int p_{t-1}(x_1, \ldots, x_{t-1}) p(x_t|x_{t-1})  \ud x_1 \ldots \ud x_{t-2}\\
      &=  p(x_t|x_{t-1})  \int p_{t-1}(x_1, \ldots, x_{t-1}) \ud x_1 \ldots \ud x_{t-2}\\
      & =  p(x_t|x_{t-1}) p_{t-1}(x_{t-1})
    \end{align}
    which proves the ``extension''.

    With \eqref{eq:HMM-change-of-measure}, we have
    \begin{align}
      p_t(x_t) & = \int  p_{t}(x_1, \ldots, x_{t}) \ud x_1, \ldots \ud x_{t-1}\\
      &=  \frac{1}{Z_t} \int p_{t-1}(x_1, \ldots, x_t) g_t(x_t) \ud x_1, \ldots \ud x_{t-1}\\
      & = \frac{1}{Z_t} g_t(x_t) \int p_{t-1}(x_1, \ldots, x_t)  \ud x_1, \ldots \ud x_{t-1}\\
      & = \frac{1}{Z_t} g_t(x_t) p_{t-1}(x_t)
    \end{align}
    which proves the ``change of measure''. Moreover, the normalising
    constant $Z_t$ is the same as before. Hence completing the
    iteration until $t=n$ yields the likelihood $p(y_1, \ldots, y_n)=
    Z_n$ as a by-product of the recursion. The initialisation of the
    recursion with $p_0(x_1)=p(x_1)$ is also the same as above.
      
  \end{solution}
  
\end{exenumerate}


\ex{Kalman filtering}
\label{ex:kalman}

We here consider filtering for hidden Markov models with Gaussian
transition and emission distributions. For simplicity, we assume
one-dimensional hidden variables and observables. We denote the
probability density function of a Gaussian random variable $x$ with
mean $\mu$ and variance $\sigma^2$ by $\Gauss(x | \mu, \sigma^2)$,
\begin{equation}
  \Gauss(x | \mu, \sigma^2) = \frac{1}{\sqrt{2 \pi \sigma^2}} \exp\left[ -\frac{(x-\mu)^2}{2\sigma^2} \right].
\end{equation}
The transition and emission distributions are assumed to be
\begin{align}
  p(h_s | h_{s-1}) &= \Gauss(h_s | A_s h_{s-1}, B^2_s) \label{eq:transition}\\
  p(v_s | h_s) & = \Gauss(v_s | C_s h_s, D^2_s). \label{eq:emission}
\end{align}
The distribution $p(h_1)$ is assumed Gaussian with known
parameters. The $A_s, B_s, C_s, D_s$ are also assumed known.

\begin{exenumerate}

\item Show that $h_s$ and $v_s$ as defined in the following update and
observation equations
\begin{align}
  h_s & = A_s h_{s-1} + B_s \xi_s \label{eq:transition-model}\\
  v_s & = C_s h_s  + D_s \eta_s \label{eq:emission-model}
\end{align}
follow the conditional distributions in \eqref{eq:transition} and
\eqref{eq:emission}. The random variables $\xi_s$ and $\eta_s$ are
independent from the other variables in the model and follow a
standard normal Gaussian distribution, e.g. $\xi_s \sim \Gauss(\xi_s |
0, 1)$.\\ Hint: For two constants $c_1$ and $c_2$, $y = c_1 + c_2 x$
is Gaussian if $x$ is Gaussian. In other words, an affine
transformation of a Gaussian is Gaussian.

The equations mean that $h_s$ is obtained by scaling $h_{s-1}$ and by
adding noise with variance $B_s^2$. The observed value $v_s$ is
obtained by scaling the hidden $h_s$ and by corrupting it with
Gaussian observation noise of variance $D_s^2$.

\begin{solution}

By assumption, $\xi_s$ is Gaussian. Since we condition on $h_{s-1}$,
$A_s h_{s-1}$ in \eqref{eq:transition-model} is a constant, and since
$B_s$ is a constant too, $h_s$ is Gaussian. 

What we have to show next is that \eqref{eq:transition-model} defines
the same conditional mean and variance as the conditional Gaussian in
\eqref{eq:transition}: The conditional expectation of $h_s$ given
$h_{s-1}$ is
\begin{align}
  \E(h_s | h_{s-1}) & = A_s h_{s-1} + \E( B_s \xi_s) &&\text{(since we condition on $h_{s-1}$)}\\
  & = A_s h_{s-1} + B_s \E(\xi_s) && \text{(by linearity of expectation)}\\
  & = A_s h_{s-1} && \text{(since $\xi_s$ has zero mean)}
\end{align}
The conditional variance of $h_s$ given $h_{s-1}$ is
\begin{align}
  \var(h_s | h_{s-1}) & =  \var( B_s \xi_s) &&\text{(since we condition on $h_{s-1}$)}\\
  & = B_s^2 \var(\xi_s) && \text{(by properties of the variance)}\\
  & = B_s^2 && \text{(since $\xi_s$ has variance one)}
\end{align}
We see that the conditional mean and variance of $h_s$ given $h_{s-1}$ match those in
\eqref{eq:transition}. And since $h_s$ given $h_{s-1}$ is Gaussian as argued above,
the result follows.

Exactly the same reasoning also applies to the case of
\eqref{eq:emission-model}. Conditional on $h_s$, $v_s$ is Gaussian
because it is an affine transformation of a Gaussian. The conditional mean of $v_s$ given $h_s$ is:
\begin{align}
  \E(v_s | h_s) & = C_s h_s + \E( D_s \eta_s) &&\text{(since we condition on $h_s$)}\\
  & = C_s h_s + D_s \E(\eta_s) && \text{(by linearity of expectation)}\\
  & = C_s h_s && \text{(since $\eta_s$ has zero mean)}
\end{align}
The conditional variance of $v_s$ given $h_s$ is
\begin{align}
  \var(v_s | h_s) & =  \var( D_s \eta_s) &&\text{(since we condition on $h_s$)}\\
  & = D_s^2 \var(\eta_s) && \text{(by properties of the variance)}\\
  & = D_s^2 && \text{(since $\eta_s$ has variance one)}
\end{align}
Hence, conditional on $h_s$, $v_s$ is Gaussian with mean and variance as in \eqref{eq:emission}.

\end{solution}

\item Show that
\begin{align}
  \int \Gauss(x | \mu, \sigma^2) \Gauss(y | Ax, B^2) \ud x & \propto \Gauss(y | A \mu, A^2 \sigma^2 + B^2) \label{eq:Gaussian-int}
\end{align} 
Hint: While this result can be obtained by integration, an approach that avoids this is as follows: First note that $\Gauss(x | \mu, \sigma^2) \Gauss(y | Ax, B^2)$ is proportional to the joint pdf of $x$ and $y$. We can thus consider the integral to correspond to the computation of the marginal of $y$ from the joint. Using the equivalence of Equations \eqref{eq:transition}-\eqref{eq:emission} and \eqref{eq:transition-model}-\eqref{eq:emission-model}, and the fact that the weighted sum of two Gaussian random variables is a Gaussian random variable then allows one to obtain the result.
\begin{solution}
We follow the procedure outlined above. The two Gaussian densities correspond to the equations
    \begin{align}
      x & = \mu + \sigma \xi\\
      y &= A x + B \eta
    \end{align}
    where $\xi$ and $\eta$ are independent standard normal random variables. The mean of $y$ is
    \begin{align}
      \E(y) & = A \E(x) + B \E(\eta)\\
      & = A \mu
    \end{align}
    where we have use the linearity of expectation and $\E(\eta)=0$. The variance of $y$ is
    \begin{align}
      \var(y) & = \var(A x) + \var (B \eta) \quad \quad \text{(since $x$ and $\eta$ are independent)}\\
      & = A^2 \var(x) + B^2 \var(\eta) \quad \quad \text{(by properties of the variance)}\\
      & = A^2 \sigma^2 + B^2
    \end{align}
    Since $y$ is the (weighted) sum of two Gaussians, it is Gaussian
    itself, and hence its distribution is completely defined by its
    mean and variance, so that
    \begin{align}
      y & \sim \Gauss(y | A \mu, A^2 \sigma^2+B^2).
    \end{align}
    Now, the product $\Gauss(x | \mu, \sigma^2) \Gauss(y | Ax, B^2)$ is proportional to the joint pdf of $x$ and $y$, so that the integral can be considered to correspond to the marginalisation of $x$, and hence its result is proportional to the density of $y$, which is $\Gauss(y | A \mu, A^2 \sigma^2+B^2)$.
  \end{solution}
  
\item Show that
  \begin{equation}
    \Gauss(x | m_1, \sigma_1^2) \Gauss(x | m_2, \sigma_2^2) \propto \Gauss(x | m_3, \sigma_3^2) \label{eq:Gaussian-mult}
  \end{equation}
  where
  \begin{align}
    \sigma^2_3 & = \left( \frac{1}{\sigma_1^2} + \frac{1}{\sigma_2^2}\right)^{-1} = \frac{\sigma_1^2 \sigma_2^2}{\sigma_1^2+\sigma_2^2}\\
    m_3 & = \sigma_3^2\left( \frac{m_1}{\sigma_1^2}+ \frac{m_2}{\sigma_2^2}\right) = m_1 + \frac{\sigma_1^2}{\sigma_1^2+\sigma_2^2}(m_2-m_1)
  \end{align}
  \emph{Hint: Work in the negative log domain.}
  
  \begin{solution}
    We show the result using a classical technique called ``completing the square'', see e.g.\ \url{https://en.wikipedia.org/wiki/Completing_the_square}.

    We work in the (negative) log-domain and use that
    \begin{align}
      -\log \left[\Gauss(x | m, \sigma^2)\right] & = \frac{ (x-m)^2 }{2 \sigma^2} + \text{const} \\
      & = \frac{ x^2}{2\sigma^2} - x \frac{m}{\sigma^2} + \frac{m^2}{2\sigma^2} + \text{const} \\
      & = \frac{ x^2}{2\sigma^2} - x \frac{m}{\sigma^2} +  \text{const} \label{eq:expansion}
    \end{align}
    where const indicates terms not depending on $x$. We thus obtain
    \begin{align}
      -\log \left[\Gauss(x | m_1, \sigma_1^2) \Gauss(x | m_2, \sigma_2^2) \right]  & = -\log \left[\Gauss(x | m_1, \sigma_1^2)\right]-\log \left[ \Gauss(x | m_2, \sigma_2^2) \right]\\
      &= \frac{ (x-m_1)^2 }{2 \sigma_1^2} + \frac{ (x-m_2)^2 }{2 \sigma_2^2} + \text{const}\\
      &= \frac{ x^2}{2\sigma_1^2} - x \frac{m_1}{\sigma_1^2} + \frac{ x^2}{2\sigma_2^2} - x \frac{m_2}{\sigma_2^2} + \text{const}\\
      &= \frac{x^2}{2} \left(\frac{1}{\sigma_1^2}+\frac{1}{\sigma_2^2}\right) -x \left( \frac{m_1}{\sigma_1^2}+\frac{m_2}{\sigma_2^2}\right) + \text{const}\\
      &=  \frac{x^2}{2 \sigma_3^2} - \frac{x}{\sigma_3^2} \sigma_3^2\left( \frac{m_1}{\sigma_1^2}+\frac{m_2}{\sigma_2^2}\right) + \text{const},
    \end{align}
    where
    \begin{align}
      \frac{1}{\sigma_3^2} &= \frac{1}{\sigma_1^2}+\frac{1}{\sigma_2^2}.
    \end{align}
    Comparison with \eqref{eq:expansion} shows that we can further write
    \begin{align}
      \frac{x^2}{2 \sigma_3^2} - \frac{x}{\sigma_3^2} \sigma_3^2\left( \frac{m_1}{\sigma_1^2}+\frac{m_2}{\sigma_2^2}\right) &=  \frac{ (x-m_3)^2 }{2 \sigma_3^2} + \text{const}
    \end{align}
    where
    \begin{align}
      m_3 & =  \sigma_3^2\left( \frac{m_1}{\sigma_1^2}+\frac{m_2}{\sigma_2^2}\right)
    \end{align}
    so that
    \begin{align}
     -\log \left[\Gauss(x | m_1, \sigma_1^2) \Gauss(x | m_2, \sigma_2^2) \right]  & = \frac{ (x-m_3)^2 }{2 \sigma_3^2} + \text{const}
    \end{align}
    and hence
    \begin{align}
     \Gauss(x | m_1, \sigma_1^2) \Gauss(x | m_2, \sigma_2^2) & \propto  \Gauss(x | m_3, \sigma_3^2).
    \end{align}
    Note that the identity
    \begin{align}
      m_3 & = \sigma_3^2\left( \frac{m_1}{\sigma_1^2}+ \frac{m_2}{\sigma_2^2}\right) = m_1 + \frac{\sigma_1^2}{\sigma_1^2+\sigma_2^2}(m_2-m_1)
    \end{align}
    is obtained as follows
    \begin{align}
      \sigma_3^2\left( \frac{m_1}{\sigma_1^2}+ \frac{m_2}{\sigma_2^2}\right) & =  \frac{\sigma_1^2 \sigma_2^2}{\sigma_1^2+\sigma_2^2}\left( \frac{m_1}{\sigma_1^2}+ \frac{m_2}{\sigma_2^2}\right)\\
      & =  m_1 \frac{\sigma_2^2}{\sigma_1^2+\sigma_2^2}+ m_2 \frac{\sigma_1^2}{\sigma_1^2+\sigma_2^2}\\
      & = m_1 \left(1- \frac{\sigma_1^2}{\sigma_1^2+\sigma_2^2}\right) + m_2 \frac{\sigma_1^2}{\sigma_1^2+\sigma_2^2}\\
      & = m_1 + \frac{\sigma_1^2}{\sigma_1^2+\sigma_2^2} (m_2-m_1)
    \end{align}
     
  \end{solution}

\item We can use the ``alpha-recursion'' to recursively compute $p(h_t | v_{1:t}) \propto
  \alpha(h_t)$ as follows.
  \begin{align}
    \alpha(h_1) & = p(h_1) \cdot  p(v_1 | h_1) &
    \alpha(h_s) &=  p(v_s | h_s) \sum_{h_{s-1}} p(h_s | h_{s-1}) \alpha(h_{s-1}).
  \end{align}
  For continuous random variables, the sum above becomes an integral so that
  \begin{align}
    \alpha(h_s) &=  p(v_s | h_s) \int p(h_s | h_{s-1}) \alpha(h_{s-1}) \ud h_{s-1}.
    \label{eq:cont-alpha}
  \end{align}

  For reference, let us denote the integral by $I(h_s)$,
  \begin{equation}
    I(h_s) = \int p(h_s | h_{s-1}) \alpha(h_{s-1}) \ud h_{s-1}.
    \label{eq:I-def-kalman}
  \end{equation}
  Note that $I(h_s)$ is proportional to the predictive distribution $p(h_s | v_{1:s-1})$.
  
  For a Gaussian prior distribution for $h_1$ and Gaussian emission
  probability $p(v_1 | h_1)$, $\alpha(h_1) = p(h_1) \cdot p(v_1 | h_1)
  \propto p(h_1 | v_1)$ is proportional to a Gaussian. We denote its
  mean by $\mu_1$ and its variance by $\sigma_1^2$ so that
  \begin{equation}
    \alpha(h_1) \propto \Gauss(h_1 | \mu_1, \sigma_1^2).
  \end{equation}
  Assuming $\alpha(h_{s-1}) \propto \Gauss(h_{s-1} | \mu_{s-1}, \sigma_{s-1}^2)$ (which holds for $s=2$), use Equation \eqref{eq:Gaussian-int} to show that
  \begin{align}
    I(h_s) & \propto \Gauss(h_s | A_s \mu_{s-1}, P_s)
  \end{align}
  where
  \begin{align}
    P_s & = A^2_s \sigma^2_{s-1} + B_s^2.
  \end{align}
  
  \begin{solution}
    We can set $\alpha(h_{s-1}) \propto \Gauss(h_{s-1} | \mu_{s-1}, \sigma_{s-1}^2)$. Since $p(h_s | h_{s-1})$ is Gaussian, see Equation \eqref{eq:transition},  Equation \eqref{eq:I-def-kalman} becomes
    \begin{align}
      I(h_s) &\propto \int  \Gauss(h_s | A_s h_{s-1}, B^2_s)  \Gauss(h_{s-1} | \mu_{s-1}, \sigma_{s-1}^2) \ud h_{s-1}.
    \end{align}
    Equation \eqref{eq:Gaussian-int} with $x \equiv h_{s-1}$ and $y \equiv h_s$ yields the desired result,
    \begin{align}
      I(h_s) &\propto  \Gauss(h_s | A_s \mu_{s-1}, A_s^2 \sigma_{s-1}^2+B_s^2).
      \label{eq:prediction-Gaussian}
    \end{align}
    We can understand the equation as follows: To compute the
    predictive mean of $h_s $ given $v_{1:s-1}$, we forward propagate
    the mean of $h_{s-1} | v_{1:s-1}$ using the update equation
    \eqref{eq:transition-model}. This gives the mean term $A_s
    \mu_{s-1}$. Since $h_{s-1} | v_{1:s-1}$ has variance
    $\sigma_{s-1}^2$, the variance of $h_s | v_{1:s-1}$ is given by
    $A_s^2 \sigma_{s-1}^2$ plus an additional term, $B_s^2$, due to
    the noise in the forward propagation. This gives the variance term
    $A_s^2 \sigma_{s-1}^2+B_s^2$.
    
  \end{solution}

  \item Use Equation \eqref{eq:Gaussian-mult} to show that
  \begin{align}
    \alpha(h_s) & \propto \Gauss\left(h_s | \mu_s , \sigma_s^2\right)
  \end{align}
  where
  \begin{align}
    \mu_s & =  A_s \mu_{s-1} + \frac{P_sC_s}{C_s^2 P_s+ D_s^2}\left(v_s - C_s A_s \mu_{s-1}\right)\\
     \sigma^2_s & =  \frac{P_s D_s^2}{P_s C_s^2+ D_s^2}
  \end{align}

  \begin{solution}
    Having computed $I(h_s)$, the final step in the alpha-recursion is
    \begin{align}
      \alpha(h_s) & = p(v_s | h_s) I(h_s)
    \end{align}
    With Equation \eqref{eq:emission} we obtain
    \begin{align}
      \alpha(h_s) & \propto \Gauss(v_s | C_s h_s, D^2_s)  \Gauss(h_s | A_s \mu_{s-1}, P_s).
    \end{align}
    We further note that
    \begin{align}
      \Gauss(v_s | C_s h_s, D_s^2) \propto \Gauss \left(h_s | C_s^{-1}
      v_s, \frac{D_s^2}{C_s^2}\right)
    \end{align}
    so that we can apply Equation \eqref{eq:Gaussian-mult} (with $m_1=A \mu_{s-1}$, $\sigma_1^2 = P_s$)
    \begin{align}
      \alpha(h_s) & \propto  \Gauss\left(h_s | C_s^{-1} v_s, \frac{D_s^2}{C_s^2}\right) \Gauss(h_s | A_s \mu_{s-1}, P_s)\\
      & \propto \Gauss\left(h_s, \mu_s, \sigma_s^2\right)
    \end{align}
    with
    \begin{align}
      \mu_s & =  A_s \mu_{s-1} + \frac{P_s}{P_s+ \frac{D_s^2}{C_s^2}}\left(C_s^{-1} v_s - A_s \mu_{s-1}\right)\\
            & =  A_s \mu_{s-1} + \frac{P_sC_s^2}{C_s^2 P_s+ D_s^2}\left(C_s^{-1}v_s - A_s \mu_{s-1}\right)\\
      & =  A_s \mu_{s-1} + \frac{P_sC_s}{C_s^2 P_s+ D_s^2}\left(v_s - C_s A_s \mu_{s-1}\right)\\
      \sigma^2_s & = \frac{P_s \frac{D_s^2}{C_s^2}}{P_s+ \frac{D_s^2}{C_s^2}}\\
      & =  \frac{P_s D_s^2}{P_s C_s^2+ D_s^2}\\
    \end{align}
  \end{solution}
  
  \item Show that $\alpha(h_s)$ can be re-written as
     \begin{align}
    \alpha(h_s) & \propto \Gauss\left(h_s | \mu_s , \sigma_s^2\right)
     \end{align}
     where
     \begin{align}
       \mu_s & = A_s \mu_{s-1} + K_s \left(v_s - C_s A_s \mu_{s-1}\right) \label{eq:mu-def}\\
       \sigma_s^2 & = (1-K_sC_s)P_s\\
       K_s & = \frac{P_s C_s}{C_s^2 P_s + D_s^2}
     \end{align}
     These are the Kalman filter equations and $K_s$ is called the Kalman filter gain.

     \begin{solution}
       We start from
       \begin{align}
         \mu_s & = A_s \mu_{s-1} + \frac{P_sC_s}{C_s^2 P_s+ D_s^2}\left(v_s - C_s A_s \mu_{s-1}\right),
       \end{align}
       and see that
       \begin{equation}
         \frac{P_sC_s}{C_s^2 P_s+ D_s^2} = K_s
       \end{equation}
       so that
       \begin{align}
         \mu_s & = A_s \mu_{s-1} + K_s \left(v_s - C_s A_s \mu_{s-1}\right).
       \end{align}
       For the variance $\sigma_s^2$, we have
       \begin{align}
         \sigma^2_s  & =  \frac{P_s D_s^2}{P_s C_s^2+ D_s^2}\\
         & =\frac{D_s^2}{P_s C_s^2+ D_s^2}  P_s \\
         & = \left(1- \frac{P_sC_s^2}{P_s C_s^2+ D_s^2}\right) P_s\\
         & = (1- K_s C_s) P_s,
       \end{align}
       which is the desired result.

       The filtering result generalises to vector valued latents and
       visibles where the transition and emission distributions in
       \eqref{eq:transition} and \eqref{eq:emission} become
       \begin{align}
         p(\h_s | \h_{s-1}) & = \Gauss(\h_s | \A \h_{s-1}, \Sigmab^h),\\
         p(\v_s | \h_s) & = \Gauss( \v_s | \C_s \h_s, \Sigmab^v),
       \end{align}
       where $\Gauss()$ denotes multivariate Gaussian pdfs, e.g.\
       \begin{align}
         \Gauss(\v_s | \C_s \h_s, \Sigmab^v) &= \frac{1}{|\det( 2\pi \Sigmab^v)|^{1/2}}\exp\left( -\frac{1}{2} (\v_s-\C_s \h_s)^\top (\Sigmab^v)^{-1}(\v_s-\C_s \h_s)\right).
       \end{align}
       We then have
       \begin{align}
         p(\h_t | \v_{1:t}) & = \Gauss( \h_t | \mub_t, \Sigmab_t)
       \end{align}
       where the posterior mean and variance are recursively computed as
       \begin{align}
         \mub_s & = \A_s \mub_{s-1} + \K_s ( \v_s-\C_s \A_s \mub_{s-1} )\\
         \Sigmab_s & = (\I - \K_s \C_s) \PP_s\\
         \PP_s & = \A_s \Sigmab_{s-1} \A_s^\top + \Sigmab^h\\
         \K_s & = \PP_s \C_s^\top \left( \C_s \PP_s \C_s^\top + \Sigmab^v \right)^{-1}
       \end{align}
       and initialised with $\mub_1$ and $\Sigmab_1$ equal to the mean
       and variance of $p(\h_1 | \v_1)$. The matrix $\K_s$ is then
       called the Kalman gain matrix.

       The Kalman filter is widely applicable, see
       e.g.\ \url{https://en.wikipedia.org/wiki/Kalman_filter}, and
       has played a role in historic events such as the moon
       landing, see e.g.\ \citep{Grewal2010}.

       An example of the application of the Kalman filter to tracking is shown in Figure \ref{fig:Kalman-tracking}.
     \begin{figure}[h]
       \centering
       \includegraphics[width=0.5 \textwidth]{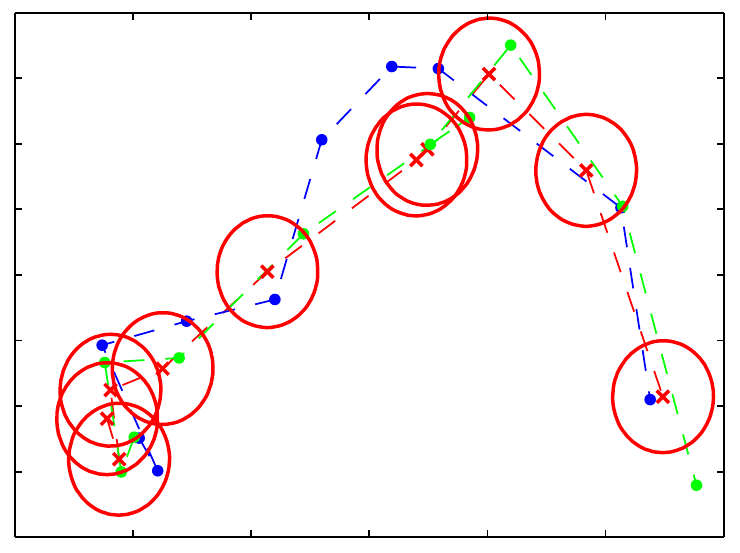}
       \caption{ \label{fig:Kalman-tracking} Kalman filtering for tracking of a moving object. The blue points indicate the true positions of the object in a two-dimensional space at successive time steps, the green points denote noisy measurements of the positions, and the red crosses indicate the means of the inferred posterior distributions of the positions obtained by running the Kalman filtering equations. The covariances of the inferred positions are indicated by the red ellipses, which correspond to contours having one standard deviation. \citep[Figure 13.22]{Bishop2006}}
     \end{figure}

     \end{solution}
     
   \item Explain Equation \eqref{eq:mu-def} in non-technical terms. What happens if the variance $D_s^2$ of the observation noise goes to zero?
     
     \begin{solution}
       We have already seen that $A_s \mu_{s-1}$ is the predictive
       mean of $h_s$ given $v_{1:s-1}$. The term $C_s A_s \mu_{s-1}$
       is thus the predictive mean of $v_s$ given the observations so
       far, $v_{1:s-1}$. The difference $v_s - C_s A_s \mu_{s-1}$ is
       thus the prediction error of the observable. Since $\alpha(h_s)$ is
       proportional to $p(h_s | v_{1:s})$ and $\mu_s$ its mean, we
       thus see that the posterior mean of $h_s | v_{1:s}$ equals the
       posterior mean of $h_s | v_{1:s-1}$, $A_s \mu_{s-1}$, updated
       by the prediction error of the observable weighted by the
       Kalman gain.

       For $D_s^2 \to 0$, $K_s \to C_s^{-1}$ and
        \begin{align}
          \mu_s & = A_s \mu_{s-1} + K_s \left(v_s - C_s A_s \mu_{s-1}\right) \\
          & = A_s \mu_{s-1} + C_s^{-1} \left(v_s - C_s A_s \mu_{s-1}\right) \\
          & = A_s \mu_{s-1} + C_s^{-1} v_s - A_s \mu_{s-1}\\
          & =  C_s^{-1} v_s,
        \end{align}
        so that the posterior mean of $p(h_s | v_{1:s})$ is obtained
        by inverting the observation equation. Moreover, the variance
        $\sigma_s^2$ of $h_s | v_{1:s}$ goes to zero so that the value
        of $h_s$ is known precisely and equals $C_s^{-1} v_s$.
       
     \end{solution}
\end{exenumerate}

\chapter{Model-Based Learning} 
\minitoc

\ex{Maximum likelihood estimation for a Gaussian}
\label{ex:Gauss-MLE}

  The Gaussian pdf parametrised by mean $\mu$ and standard deviation $\sigma$ is given by
  $$ p(x; \thetab) = \frac{1}{\sqrt{2\pi\sigma^2}}\exp\left[-\frac{(x-\mu)^2}{2\sigma^2}\right], \quad \quad \thetab = (\mu, \sigma).$$
  
\begin{exenumerate}
  
\item Given iid data $\data = \{x_1, \ldots, x_n\}$, what is the likelihood function $L(\thetab)$ for the Gaussian model?
  
  \begin{solution}
    For iid data, the likelihood function is
    \begin{align}
      L(\thetab) &= \prod_i^n p(x_i;\thetab) \\
		    &= \prod_i^n \frac{1}{\sqrt{2\pi\sigma^2}}\exp\left[-\frac{(x_i-\mu)^2}{2\sigma^2}\right]\\
		    &= \frac{1}{(2\pi\sigma^2)^{n/2}}\exp\left[-\frac{1}{2\sigma^2}\sum_{i=1}^n (x_i-\mu)^2\right].
    \end{align}
    
  \end{solution}

\item What is the log-likelihood function $\ell(\thetab)$?

\begin{solution}

  Taking the log of the likelihood function gives
  \begin{align}
    \ell(\thetab) & = -\frac{n}{2} \log(2\pi \sigma^2) -\frac{1}{2\sigma^2} \sum_{i=1}^n (x_i-\mu)^2
  \end{align}

\end{solution}
    
\item Show that the maximum likelihood estimates for the mean $\mu$ and standard deviation $\sigma$ are the
  sample mean 
  \begin{equation}
    \bar{x}= \frac{1}{n} \sum_{i=1}^n x_i
    \end{equation}
  and the square root of the sample variance
  \begin{equation}
    S^2 = \frac{1}{n} \sum_{i=1}^n (x_i -\bar{x})^2.
  \end{equation}
    
  \begin{solution}
    Since the logarithm is strictly monotonically increasing, the
    maximiser of the log-likelihood equals the maximiser of the
    likelihood. It is easier to take derivatives for the
    log-likelihood function than for the likelihood function so that
    the maximum likelihood estimate is typically determined using the log-likelihood.

    Given the algebraic expression of $\ell(\thetab)$, it is simpler
    to work with the variance $v=\sigma^2$ rather than the standard
    deviation. Since $\sigma > 0$ the function $v = g(\sigma) = \sigma^2$
    is invertible, and the invariance of the MLE to re-parametrisation
    guarantees that
    $$ \hat{\sigma} = \sqrt{\hat{v}}. $$

    We now thus maximise the function $J(\mu, v)$,
    \begin{align}
      J(\mu, v) & = -\frac{n}{2} \log(2\pi v) -\frac{1}{2v} \sum_{i=1}^n (x_i-\mu)^2
    \end{align}
    with respect to $\mu$ and $v$.

    Taking partial derivatives gives
    \begin{align}
      \frac{\partial J }{\partial\mu} & = \frac{1}{v} \sum_{i=1}^n (x_i-\mu)\\
      &= \frac{1}{v} \sum_{i=1}^n x_i  -\frac{n}{v} \mu\\
      \frac{\partial J}{\partial v} &= -\frac{n}{2} \frac{1}{v} + \frac{1}{2v^2} \sum_{i=1}^n (x_i-\mu)^2
    \end{align}
    A necessary condition for optimality is that the partial derivatives are zero. We thus obtain the conditions
    \begin{align}
      \frac{1}{v} \sum_{i=1}^n (x_i-\mu) &=0\\
      -\frac{n}{2} \frac{1}{v} + \frac{1}{2v^2} \sum_{i=1}^n (x_i-\mu)^2 & = 0
    \end{align}
    From the first condition it follows that
    \begin{align}
      \hat{\mu} &= \frac{1}{n}\sum_{i=1}^n x_i
    \end{align}
    The second condition thus becomes
      \begin{align}
        -\frac{n}{2} \frac{1}{v} + \frac{1}{2v^2} \sum_{i=1}^n (x_i-\hat{\mu})^2 & = 0 \quad \quad (\text{multiply with } v^2 \text{ and rearrange})\\
        \frac{1}{2} \sum_{i=1}^n (x_i-\hat{\mu})^2 & = \frac{n}{2} v,
      \end{align}
      and hence
      \begin{align}
        \hat{v} & = \frac{1}{n} \sum_{i=1}^n (x_i-\hat{\mu})^2,
      \end{align}
      We now check that this solution corresponds to a maximum by computing the Hessian matrix
      \begin{align}
        \H(\mu, v) & = \begin{pmatrix}
          \frac{\partial^2 J }{\partial \mu^2} & \frac{\partial^2 J }{\partial \mu \partial v}\\
          \frac{\partial^2 J }{\partial \mu \partial v} & \frac{\partial^2 J }{\partial v^2}
        \end{pmatrix}
      \end{align}
      If the Hessian negative definite at $(\hat{\mu}, \hat{v})$, the
      point is a (local) maximum. Since we only have one critical
      point, $(\hat{\mu}, \hat{v})$, the local maximum is also a
      global maximum. Taking second derivatives gives
      \begin{align}
        \H(\mu, v) & = \begin{pmatrix}
          -\frac{n}{v} & -\frac{1}{v^2} \sum_{i=1}^n (x_i-\mu)\\
          -\frac{1}{v^2} \sum_{i=1}^n (x_i-\mu) & \frac{n}{2} \frac{1}{v^2} - \frac{1}{v^3} \sum_{i=1}^n (x_i-\mu)^2
        \end{pmatrix}.
      \end{align}
      Substituting the values for $(\hat{\mu}, \hat{v})$ gives
      \begin{align}
        \H(\hat{\mu}, \hat{v}) & = \begin{pmatrix}
          -\frac{n}{\hat{v}} & 0 \\
          0 & -\frac{n}{2} \frac{1}{\hat{v}^2}
        \end{pmatrix},
         \end{align}
      which is negative definite. Note that the the (negative)
      curvature increases with $n$, which means that $J(\mu, v)$, and
      hence the log-likelihood becomes more and more peaked as the
      number of data points $n$ increases.
  \end{solution}
  
\end{exenumerate}

\ex{Posterior of the mean of a Gaussian with known variance} Given iid data $\data = \{x_1, \ldots, x_n\}$, compute $p(\mu | \data, \sigma^2)$ for the
Bayesian model
\begin{align}
  p(x | \mu) &= \frac{1}{\sqrt{2\pi\sigma^2}}\exp\left[-\frac{(x-\mu)^2}{2\sigma^2}\right] &
  p(\mu; \mu_0, \sigma_0^2) &=  \frac{1}{\sqrt{2\pi\sigma_0^2}}\exp\left[-\frac{(\mu-\mu_0)^2}{2\sigma_0^2}\right]
\end{align}
where $\sigma^2$ is a fixed known quantity.\\
Hint: You may use that
\begin{equation}
  \Gauss(x ; m_1, \sigma_1^2) \Gauss(x ; m_2, \sigma_2^2) \propto \Gauss(x ; m_3, \sigma_3^2) \label{eq:Gaussian-mult-model-based-learning}
\end{equation}
where 
\begin{align}
   \Gauss(x ; \mu, \sigma^2) &= \frac{1}{\sqrt{2 \pi \sigma^2}} \exp\left[ -\frac{(x-\mu)^2}{2\sigma^2} \right]\\
  \sigma^2_3 & = \left( \frac{1}{\sigma_1^2} + \frac{1}{\sigma_2^2}\right)^{-1} = \frac{\sigma_1^2 \sigma_2^2}{\sigma_1^2+\sigma_2^2}\\
  m_3 & = \sigma_3^2\left( \frac{m_1}{\sigma_1^2}+ \frac{m_2}{\sigma_2^2}\right) = m_1 + \frac{\sigma_1^2}{\sigma_1^2+\sigma_2^2}(m_2-m_1) \label{eq:mu-mult}
\end{align}

\begin{solution}
We re-use the expression for the likelihood $L(\mu)$ from \exref{ex:Gauss-MLE}.
\begin{equation}
  L(\mu) = \frac{1}{(2\pi\sigma^2)^{n/2}}\exp\left[-\frac{1}{2\sigma^2}\sum_{i=1}^n (x_i-\mu)^2\right],
\end{equation}
which we can write as
\begin{align}
  L(\mu) & \propto \exp\left[-\frac{1}{2\sigma^2}\sum_{i=1}^n (x_i-\mu)^2\right]\\
  & \propto  \exp\left[-\frac{1}{2\sigma^2}\sum_{i=1}^n (x_i^2-2\mu x_i + \mu^2)\right]\\
  & \propto  \exp\left[-\frac{1}{2\sigma^2}\left( -2 \mu \sum_{i=1}^n x_i + n \mu^2 \right)\right]\\
  & \propto  \exp\left[-\frac{1}{2\sigma^2}\left( -2n \mu \xBar + n \mu^2 \right)\right]\\
  & \propto  \exp\left[-\frac{n}{2\sigma^2}(\mu-\xBar)^2 \right]\\
  & \propto  \Gauss(\mu ; \xBar, \sigma^2/n).
\end{align}
The posterior is
\begin{align}
  p(\mu | \data) & \propto L(\theta) p(\mu; \mu_0, \sigma_0^2)\\
  & \propto  \Gauss(\mu ; \xBar, \sigma^2/n)  \Gauss(\mu ; \mu_0, \sigma_0^2)
\end{align}
so that with \eqref{eq:Gaussian-mult-model-based-learning}, we have
\begin{align}
  p(\mu | \data) & \propto \Gauss(\mu ; \mu_n, \sigma_n^2)\\
  \sigma^2_n & =  \left( \frac{1}{\sigma^2/n} + \frac{1}{\sigma_0^2}\right)^{-1} \\
  & =  \frac{ \sigma_0^2 \sigma^2/n}{\sigma_0^2+\sigma^2/n}\\
  \mu_n & =  \sigma_n^2\left( \frac{\xBar}{\sigma^2/n}+ \frac{\mu_0}{\sigma_0^2}\right)\\
  & = \frac{1}{\sigma_0^2+\sigma^2/n}\left(\sigma_0^2 \xBar + (\sigma^2/n) \mu_0\right)\\
  & = \frac{\sigma_0^2}{\sigma_0^2+\sigma^2/n} \xBar + \frac{\sigma^2/n} {\sigma_0^2+\sigma^2/n} \mu_0.
\end{align}
As $n$ increases, $\sigma^2/n$ goes to zero so that $\sigma_n^2 \to 0$
and $\mu_n \to \xBar$. This means that with an increasing amount of
data, the posterior of the mean tends to be concentrated around the
maximum likelihood estimate $\xBar$.

From \eqref{eq:mu-mult}, we also have that
\begin{align}
\mu_n & = \mu_0 + \frac{\sigma_0^2}{\sigma^2/n+\sigma_0^2}(\xBar-\mu_0),
\end{align}
which shows more clearly that the value of $\mu_n$ lies on a line with end-points $\mu_0$ (for $n =0$) and $\xBar$ (for $n \to \infty$). As
the amount of data increases, $\mu_n$ moves form the mean under the prior, $\mu_0$, to the average of the observed sample, that is the MLE $\xBar$.
\end{solution}


\ex{Maximum likelihood estimation of probability tables in fully observed directed graphical models of binary variables}
\label{ex:MLE-DGM}
We assume that we are given a parametrised directed graphical model for variables
$x_1, \ldots, x_d$, 
\begin{align}
  p(\x; \thetab) &= \prod_{i=1}^d p(x_i | \pa_i; \thetab_i) \quad \quad x_i \in \{0,1\} \label{eq:p-dag}
\end{align}
where the conditionals are represented by parametrised probability
tables, For example, if $\pa_3 = \{x_1, x_2\}$, $p(x_3 | \pa_3;
\thetab_3)$ is represented as

\begin{center}
  \begin{tabular}{@{}cll@{}}
      \toprule
      $p(x_3=1 | x_1, x_2; \theta^1_3, \ldots, \theta_3^4))$ & $x_1$ & $x_2$\\
      \midrule
      $\theta^1_3$ & 0 & 0\\
      $\theta^2_3$ & 1 & 0\\
      $\theta^3_3$ & 0 & 1\\
      $\theta^4_3$ & 1 & 1\\
      \bottomrule
  \end{tabular}
\end{center}
with $\thetab_3 = (\theta_3^1,\theta_3^2, \theta_3^3, \theta_3^4)$,
and where the superscripts $j$ of $\theta_3^j$ enumerate the different states
that the parents can be in.

\begin{exenumerate}
  
\item Assuming that $x_i$ has $m_i$ parents, verify that the table parametrisation of $p(x_i | \pa_i; \thetab_i)$ is equivalent to writing $p(x_i | \pa_i; \thetab_i)$ as
  \begin{align}
    p(x_i | \pa_i; \thetab_i) &= \prod_{s=1}^{S_i} (\theta_i^s)^{\ind(x_i=1, \pa_i=s)} (1-\theta_i^s)^{\ind(x_i=0, \pa_i=s)} \label{eq:pcond-fac}
  \end{align}
  where $S_i = 2^{m_i}$ is the total number of states/configurations that the
  parents can be in, and $\ind(x_i=1, \pa_i=s)$ is one if $x_i=1$ and $\pa_i=s$, and zero otherwise.
  
  \begin{solution}
    The number of configurations that $m$ binary parents can be in is given by $S_i$. The questions thus boils down to showing that $p(x_i=1 | \pa_i=k; \thetab_i) = \theta_i^k$ for any state $k \in \{1, \ldots, S_i\}$ of the parents of $x_i$. Since $\ind(x_i=1, \pa_i=s)=0$ unless $s=k$, we have indeed that
     \begin{align}
       p(x_i = 1| \pa_i = k; \thetab_i) &= \left[\prod_{s \neq k} (\theta_i^s)^0 (1-\theta_i^s)^0\right] (\theta_i^k)^{\ind(x_i=1, \pa_i=k)} (1-\theta_i^k)^{\ind(x_i=0, \pa_i=k)}\\
       &= 1 \cdot  (\theta_i^k)^{\ind(x_i=1, \pa_i=k)} (1-\theta_i^k)^0\\
       & = \theta_i^k.
     \end{align}
    
  \end{solution}
  
\item For iid data $\data = \{\x^{(1)}, \ldots, \x^{(n)}\}$ show that the likelihood can be represented as
  \begin{align}
    p(\data; \thetab) & = \prod_{i=1}^d \prod_{s=1}^{S_i}  (\theta_i^s)^{n_{x_i=1}^s} (1-\theta_i^s)^{n_{x_i=0}^s} \label{eq:joint-dag}
  \end{align}
  where $n_{x_i=1}^s$ is the number of times the pattern $(x_i=1, \pa_i=s)$ occurs in the data $\data$, and equivalently for $n_{x_i=0}^s$.
  
  \begin{solution}
    Since the data are iid, we have
    \begin{align}
      p(\data; \thetab) & = \prod_{j=1}^n p(\x^{(j)}; \thetab) \\
    \end{align}
    where each term $p(\x^{(j)}; \thetab)$ factorises as in \eqref{eq:p-dag},
    \begin{align}
      p(\x^{(j)}; \thetab) &= \prod_{i=1}^d p(x_i^{(j)} | \pa_i^{(j)}; \thetab_i)
    \end{align}
    with $x_i^{(j)}$ denoting the $i$-th element of $\x^{(j)}$ and $\pa_i^{(j)}$ the corresponding parents. The conditionals $p(x_i^{(j)} | \pa_i^{(j)}; \thetab_i)$ factorise further according to \eqref{eq:pcond-fac},
 \begin{align}
   p(x_i^{(j)} | \pa_i^{(j)}; \thetab_i) &= \prod_{s=1}^{S_i} (\theta_i^s)^{\ind(x_i^{(j)}=1, \pa_i^{(j)}=s)} (1-\theta_i^s)^{\ind(x_i^{(j)}=0, \pa_i^{(j)}=s)},
  \end{align}
 so that
 \begin{align}
   p(\data; \thetab) & = \prod_{j=1}^n \prod_{i=1}^d p(x_i^{(j)} | \pa_i^{(j)}; \thetab_i)\\
   & =  \prod_{j=1}^n  \prod_{i=1}^d \prod_{s=1}^{S_i} (\theta_i^s)^{\ind(x_i^{(j)}=1, \pa_i^{(j)}=s)} (1-\theta_i^s)^{\ind(x_i^{(j)}=0, \pa_i^{(j)}=s)}
 \end{align}
 Swapping the order of the products so that the product over the data points comes first, we obtain
 \begin{align}
   p(\data; \thetab)  & = \prod_{i=1}^d \prod_{s=1}^{S_i} \prod_{j=1}^n(\theta_i^s)^{\ind(x_i^{(j)}=1, \pa_i^{(j)}=s)} (1-\theta_i^s)^{\ind(x_i^{(j)}=0, \pa_i^{(j)}=s)}
 \end{align}
 We next split the product over $j$ into two products, one for all $j$ where $x_i^{(j)}=1$, and one for all $j$ where $x_i^{(j)}=0$
   \begin{align}
p(\data; \thetab)   &=  \prod_{i=1}^d \prod_{s=1}^{S_i} \prod_{\substack{j: \\x_i^{(j)}=1}}\prod_{\substack{j: \\x_i^{(j)}=0}} (\theta_i^s)^{\ind(x_i^{(j)}=1, \pa_i^{(j)}=s)} (1-\theta_i^s)^{\ind(x_i^{(j)}=0, \pa_i^{(j)}=s)}\\
    &=  \prod_{i=1}^d \prod_{s=1}^{S_i} \prod_{\substack{j: \\x_i^{(j)}=1}} (\theta_i^s)^{\ind(x_i^{(j)}=1, \pa_i^{(j)}=s)} \prod_{\substack{j: \\x_i^{(j)}=0}}(1-\theta_i^s)^{\ind(x_i^{(j)}=0, \pa_i^{(j)}=s)}\\
 & = \prod_{i=1}^d \prod_{s=1}^{S_i} (\theta_i^s)^{ \sum_{j=1}^n \ind(x_i^{(j)}=1, \pa_i^{(j)}=s)} (1-\theta_i^s)^{ \sum_{j=1}^n\ind(x_i^{(j)}=0, \pa_i^{(j)}=s)}\\
 & =  \prod_{i=1}^d \prod_{s=1}^{S_i}  (\theta_i^s)^{n_{x_i=1}^s} (1-\theta_i^s)^{n_{x_i=0}^s}
 \end{align}
 where
 \begin{align}
   n_{x_i=1}^s &=  \sum_{j=1}^n\ind(x_i^{(j)}=1, \pa_i^{(j)}=s) & n_{x_i=0}^s &=  \sum_{j=1}^n\ind(x_i^{(j)}=0, \pa_i^{(j)}=s)
 \end{align}
 is the number of times $x_i=1$ and $x_i=0$, respectively, with its parents being in state $s$.

\end{solution}
 
\item \label{q:loglik-fully-observed-dgm-sum-bernoulli}Show that the log-likelihood decomposes into sums of terms that can be independently optimised, and that each term corresponds to the log-likelihood for a Bernoulli model.
  
  \begin{solution}
    The log-likelihood $\ell(\thetab)$ equals
    \begin{align}
      \ell(\thetab) & = \log p(\data; \thetab) \\
      & = \log \prod_{i=1}^d \prod_{s=1}^{S_i}  (\theta_i^s)^{n_{x_i=1}^s} (1-\theta_i^s)^{n_{x_i=0}^s}\\
      & = \sum_{i=1}^d \sum_{s=1}^{S_i} \log \left[  (\theta_i^s)^{n_{x_i=1}^s} (1-\theta_i^s)^{n_{x_i=0}^s} \right]\\
      & = \sum_{i=1}^d \sum_{s=1}^{S_i} n_{x_i=1}^s \log (\theta_i^s) + n_{x_i=0}^s \log (1-\theta_i^s)
    \end{align}
    Since the parameters $\theta_i^s$ are not coupled in any way, maximising $\ell(\thetab)$ can be achieved by maximising each term $\ell_{is}(\theta_i^s)$ individually,
    \begin{align}
      \ell_{is}(\theta_i^s) & = n_{x_i=1}^s \log (\theta_i^s) + n_{x_i=0}^s \log (1-\theta_i^s).
    \end{align}
    Moreover, $\ell_{is}(\theta_i^s)$ corresponds to the log-likelihood for a Bernoulli model with success probability $\theta_i^s$ and data with $n_{x_i=1}^s$ number of ones and $n_{x_i=0}^s$ number of zeros.
    
  \end{solution}

\item \label{q:mle-bernoulli} Determine the maximum likelihood estimate $\hat{\theta}$ for the
  Bernoulli model
  \begin{align}
    p(x; \theta) &= \theta^{x} (1-\theta)^{1-x},& \theta &\in [0,\;1],&  x &\in\{0,1\} 
  \end{align}
  and iid data $x_1, \ldots, x_n$.
  \begin{solution}
   The log-likelihood function is
  \begin{align}      
    \ell(\theta) & = \sum_{i=1}^n \log p(x_i; \theta) \\
    & = \sum_{i=1}^n x_i \log(\theta) + 1-x_i \log(1-\theta).
  \end{align}
  Since $\log(\theta)$ and $\log(1-\theta)$ do not depend on $i$, we
  can pull them outside the sum and the log-likelihood function can be
  written as
  \begin{equation}
    \ell(\theta) = n_{x=1} \log (\theta) + n_{x=0} \log(1-\theta)
  \end{equation}
  where $n_{x=1} = \sum_{i=1}^n x_i = \sum_{i=1}^n \ind(x_i=1)$ and
  $n_{x=0} = n-n_{x=1}$ are the number of ones and zeros in the
  data. Since $\theta \in [0,\;1]$, we have to solve the constrained
  optimisation problem
  \begin{equation}
    \hat{\theta} = \argmax_{\theta \in [0,\;1]}   \ell(\theta)
  \end{equation}
  There are multiple ways to solve the problem. One option is to
  determine the \emph{unconstrained} optimiser and then check whether
  it satisfies the constraint. The first derivative equals
  \begin{equation}
    \ell'(\theta) = \frac{n_{x=1}}{\theta} - \frac{n_{x=0}}{1-\theta}
  \end{equation}
  and the second derivative is
  \begin{equation}
    \ell''(\theta) = -\frac{n_{x=1}}{\theta^2} - \frac{n_{x=0}}{(1-\theta)^2}
  \end{equation}
  The second derivative is always negative for $\theta \in (0,1)$,
  which means that $\ell(\theta)$ is strictly concave on $(0,1)$
  and that an optimiser that is not on the boundary corresponds to a
  maximum. Setting the first derivative to zero gives the condition
  \begin{equation}
   \frac{n_{x=1}}{\theta} = \frac{n_{x=0}}{1-\theta}
  \end{equation}
  Solving for $\theta$ gives
  \begin{align}
    (1-\theta) n_{x=1} &= n_{x=0} \theta\\
  \end{align}
  so that
  \begin{align}
    n_{x=1} & = \theta( n_{x=0}+ n_{x=1})\\
    & = \theta n
  \end{align}
  Hence, we find
  \begin{equation}
    \hat{\theta} = \frac{n_{x=1}}{n}.
  \end{equation}
  For $n_{x=1}< n$, we have $\hat{\theta} \in (0,1)$ so that the
  constraint is actually not active.
  
  In the derivation, we had to exclude boundary cases where $\theta$
  is 0 or 1. We note that e.g.\ $\hat{\theta}=1$ is obtained when
  $n_{x=1}=n$, i.e.\ when we only observe 1's in the data set. In that
  case, $n_{x=0}=0$ and the log-likelihood function equals
  $n\log(\theta)$, which is strictly increasing and hence attains the
  maximum at $\hat{\theta}=1$. A similar argument shows that if
  $n_{x=1}=0$, the maximum is at $\hat{\theta}=0$.  Hence, the maximum
  likelihood estimate
  \begin{equation}
    \hat{\theta} = \frac{n_{x=1}}{n}
  \end{equation}
  is valid for all $n_{x=1} \in \{0, \ldots, n\}$.

  An alternative approach to deal with the constraint is to
  reparametrise the objective function and work with the log-odds $\eta$,
  \begin{equation}
    \eta = g(\theta) = \log\left[ \frac{\theta}{1-\theta}\right].
  \end{equation}
  The log-odds take values in $\mathbb{R}$ so that $\eta$ is
  unconstrained. The transformation from $\theta$ to $\eta$ is
  invertible and
  \begin{equation}
    \theta = g^{-1}(\eta) = \frac{\exp(\eta)}{1+\exp(\eta)} = \frac{1}{1+\exp(-\eta)}.
  \end{equation}
  The optimisation problem then becomes
  \begin{align*}
    \hat{\eta} &= \argmax_{\eta}  n_{x=1} \eta - n \log(1+\exp(\eta))
  \end{align*}
  Computing the second derivative shows that the objective is concave
  for all $\eta$ and the maximiser $\hat{\eta}$ can be determined by
  setting the first derivative to zero. The maximum likelihood
  estimate of $\theta$ is then given by
  \begin{equation}
    \hat{\theta} = \frac{\exp(\hat{\eta})}{1+\exp(\hat{\eta})}
  \end{equation}
  The reason for this is as follows: Let $J(\eta) = \ell(
  g^{-1}(\eta))$ be the log-likelihood seen as a function of
  $\eta$. Since $g$ and $g^{-1}$ are invertible, we have that
  \begin{align}
    \max_{\theta \in [0,1]} \ell(\theta) &= \max_\eta J(\eta) \\
    \argmax_{\theta \in [0,1]} \ell(\theta) &= g^{-1}\left( \argmax_{\eta} J(\eta) \right).
  \end{align}
  \end{solution}
  
\item Returning to the fully observed directed graphical model, conclude that the maximum likelihood estimates are given by
  \begin{align}
    \hat{\theta}_i^s &= \frac{n_{x_i=1}^s}{n_{x_i=1}^s+n_{x_i=0}^s} = \frac{ \sum_{j=1}^n\ind(x_i^{(j)}=1, \pa_i^{(j)}=s) }{ \sum_{j=1}^n\ind(\pa_i^{(j)}=s) }
  \end{align}
  
  \begin{solution}
    Given the result from question
    \ref{q:loglik-fully-observed-dgm-sum-bernoulli}, we can optimise
    each term $\ell_{is}(\theta_i^s)$ separately. Each term formally
    corresponds to a log-likelihood for a Bernoulli model, so that we
    can use the results from question \ref{q:mle-bernoulli} to obtain
    \begin{align}
      \hat{\theta}_i^s &= \frac{n_{x_i=1}^s}{n_{x_i=1}^s+n_{x_i=0}^s}.
    \end{align}
    Since $n_{x_i=1}^s = \sum_{j=1}^n\ind(x_i^{(j)}=1, \pa_i^{(j)}=s)$ and
    \begin{align}
      n_{x_i=1}^s+n_{x_i=0}^s &=  \sum_{j=1}^n\ind(x_i^{(j)}=1, \pa_i^{(j)}=s) +  \sum_{j=1}^n\ind(x_i^{(j)}=0, \pa_i^{(j)}=s)\\
      & =  \sum_{j=1}^n\ind( \pa_i^{(j)}=s),
    \end{align}
    we further have
    \begin{align}
      \hat{\theta}_i^s &= \frac{ \sum_{j=1}^n\ind(x_i^{(j)}=1, \pa_i^{(j)}=s) }{ \sum_{j=1}^n\ind(\pa_i^{(j)}=s) }.
      \label{eq:dag-table-mle}
    \end{align}
    Hence, to determine $\hat{\theta}_i^s$, we first count the
    number of times the parents of $x_i$ are in state $s$, which
    gives the denominator, and then among them, count the number of
    times $x_i=1$, which gives the numerator.
    
  \end{solution}
  
\end{exenumerate}


\ex{Cancer-asbestos-smoking example: MLE}
\label{ex:cancer-smoking-asbestos-mle} 
Consider the model specified by the DAG

\begin{center}
\begin{tikzpicture}[dgraph]
  \node[cont] (x) at (0,0) {a};
  \node[cont] (y) at (2,0) {s};
  \node[cont] (z) at (1,-1) {c};
  \draw(x) -- (z);
  \draw(y) -- (z);
\end{tikzpicture}
\end{center}

The distribution of $a$ and $s$ are Bernoulli distributions with
parameter (success probability) $\theta_a$ and $\theta_s$,
respectively, i.e.\
\begin{equation}
  p(a; \theta_a) = \theta_a^a(1-\theta_a)^{1-a} \quad \quad p(s; \theta_s) = \theta_s^s(1-\theta_s)^{1-s},
\end{equation}
and the distribution of $c$ given the parents is parametrised as specified in the following table
\begin{center}
  \begin{tabular}{@{}cll@{}}
    \toprule
    $p(c=1 | a,s; \theta^1_c, \ldots, \theta_c^4))$ & $a$ & $s$\\
    \midrule
    $\theta^1_c$ & 0 & 0\\
    $\theta^2_c$ & 1 & 0\\
    $\theta^3_c$ & 0 & 1\\
    $\theta^4_c$ & 1 & 1\\
    \bottomrule
  \end{tabular}
\end{center}
The free parameters of the model are $(\theta_a, \theta_s, \theta^1_c, \ldots, \theta^4_c)$.

Assume we observe the following iid data (each row is a data point).

\begin{center}
  \scalebox{1}{
    \begin{tabular}{lll}
      \toprule
      a & s & c\\
      \midrule
      0 &   1 &   1\\
      0 &   0 &   0\\
      1 &   0 &   1\\
      0 &   0 &   0\\
      0 &   1 &   0\\
      \bottomrule
  \end{tabular}}
\end{center}

\begin{exenumerate}
\item Determine the maximum-likelihood estimates of $\theta_a$ and $\theta_s$
  \begin{solution}
    The maximum likelihood estimate (MLE) $\hat{\theta}_a$ is given by
    the fraction of times that $a$ is 1 in the data set. Hence
    $\hat{\theta}_a = 1/5$. Similarly, the MLE $\hat{\theta}_s$ is
    $2/5$.
    
  \end{solution}

\item \label{q:cancer-smoking-asbestos-mle} Determine the maximum-likelihood estimates of $\theta_c^1, \ldots, \theta_c^4$.
 
  \begin{solution}
    With \eqref{eq:dag-table-mle}, we have
    \begin{center}
      \begin{tabular}{@{}lll@{}}
        \toprule
        $\hat{p}(c=1 | a,s)$ & $a$ & $s$\\
        \midrule
        $\hat{\theta}^1_c = 0$ & 0 & 0\\
        $\hat{\theta}^2_c = 1/1$ & 1 & 0\\
        $\hat{\theta}^3_c = 1/2$ & 0 & 1\\
        $\hat{\theta}^4_c$ not defined & 1 & 1\\
        \bottomrule
      \end{tabular}
    \end{center}
    This because, for example, we have two observations where
    $(a,s)=(0,0)$, and among them, $c=1$ never occurs, so that the MLE for
    $p(c=1 | a,s)$ is zero.

    This example illustrates some issues with maximum likelihood
    estimates: We may get extreme probabilities, zero or one, or if the
    parent configuration does not occur in the observed data, the estimate
    is undefined.

  \end{solution}
  
\end{exenumerate}


\ex{Bayesian inference for the Bernoulli model}
\label{ex:Bayesian-inference-Bernoulli}
Consider the Bayesian model
\begin{align*}
  p(x | \theta) &= \theta^x (1-\theta)^{1-x} & p(\theta ; \alphab_0) = \BetaDist(\theta ; \alpha_0, \beta_0)
\end{align*}
where $x \in \{0,1\}, \ \theta \in [0,1], \alphab_0 = (\alpha_0,\beta_0)$, and
\begin{equation}
  \BetaDist(\theta; \alpha, \beta) \propto \theta^{\alpha-1}(1-\theta)^{\beta-1} \quad \quad \theta \in [0,1]
\end{equation}

\begin{exenumerate}
\item Given iid data $\data = \{x_1, \ldots, x_n\}$ show that the posterior of $\theta$ given $\data$ is
  \begin{align*}
    p(\theta | \data) &= \BetaDist(\theta ; \alpha_n, \beta_n)\\
    \alpha_n &= \alpha_0 + n_{x=1} & \beta_n &= \beta_0 + n_{x=0}
  \end{align*}
  where $n_{x=1}$ denotes the number of ones and $n_{x=0}$ the number of zeros in the data.

  \begin{solution}
    This follows from
    \begin{align}
      p(\theta | \data) \propto L(\theta) p(\theta ; \alphab_0) \label{eq:Bernoulli-posterior-def}
    \end{align}
    and from the expression for the likelihood function of the Bernoulli model, which is
    \begin{align}
      L(\theta) &= \prod_{i=1}^n p(x_i|\theta)\\
      & = \prod_{i=1}^n  \theta^{x_i} (1-\theta)^{1-x_i}\\
      & = \theta^{\sum_{i=1}^n x_i} (1-\theta)^{\sum_{i=1}^n (1-x_i)}\\
      & = \theta^{n_{x=1}}(1-\theta)^{n_{x=0}},
    \end{align}
    where $n_{x=1} = \sum_{i=1}^n x_i$ denotes the number of 1's in
    the data, and $n_{x=0}=\sum_{i=1}^n (1-x_i)=n-n_{x=1}$ the number
    of 0's.

    Inserting the expressions for the likelihood and prior into
    \eqref{eq:Bernoulli-posterior-def} gives
    \begin{align}
      p(\theta | \data) &\propto \theta^{n_{x=1}}(1-\theta)^{n_{x=0}}\theta^{\alpha_0-1}(1-\theta)^{\beta_0-1}\\
      & \propto \theta^{\alpha_0+n_{x=1}-1}(1-\theta)^{\beta_0+n_{x=0}-1}\\
      & \propto \BetaDist(\theta, \alpha_0+n_{x=1}, \beta_0+n_{x=0}),
    \end{align}
    which is the desired result. Since $\alpha_0$ and $\beta_0$ are
    updated by the counts of ones and zeros in the data, these
    hyperparameters are also referred to as
    ``pseudo-counts''. Alternatively, one can think that they are the
    counts that are observed in another iid data set which has been
    previously analysed and used to determine the prior.
  \end{solution}

\item \label{q:mean-beta-var} Compute the mean of a Beta random variable $f$,
  \begin{equation}
    p(f; \alpha, \beta) =  \BetaDist(f ; \alpha, \beta) \quad \quad f \in [0,1],
  \end{equation}
  using that
    \begin{equation}
    \int_0^1 f^{\alpha-1} (1-f)^{\beta-1} \ud f = B(\alpha, \beta) = \frac{\Gamma(\alpha)\Gamma(\beta)}{\Gamma(\alpha+\beta)}
  \end{equation}
  where $ B(\alpha, \beta)$ denotes the Beta function and where the Gamma function $\Gamma(t)$ is defined as
  \begin{equation}
    \Gamma(t) = \int_o^\infty f^{t-1} \exp(-f) \ud f
  \end{equation}
  and satisfies $\Gamma(t+1) = t \Gamma(t)$.\\
  \emph{Hint: It will be useful to represent the partition function in terms of the Beta function.}

  \begin{solution}
    We first write the partition function of $p(f; \alpha, \beta)$ in terms of the Beta function
    \begin{align}
      Z(\alpha, \beta) &= \int_0^1 f^{\alpha-1} (1-f)^{\beta-1}\\
      & =  B(\alpha, \beta).
    \end{align}
    We then have that the mean $\E[f]$ is given by
    \begin{align}
      \E[f] & = \int_0^1 f p(f; \alpha, \beta) \ud f \\
      & = \frac{1}{B(\alpha, \beta)} \int_0^1 f  f^{\alpha-1} (1-f)^{\beta-1} \ud f\\
      & = \frac{1}{B(\alpha, \beta)} \int_0^1 f^{\alpha+1-1} (1-f)^{\beta-1} \ud f\\
      & = \frac{B(\alpha+1, \beta)}{B(\alpha, \beta)} \\
      & = \frac{\Gamma(\alpha+1)\Gamma(\beta)}{\Gamma(\alpha+1+\beta)} \frac{\Gamma(\alpha+\beta)}{\Gamma(\alpha)\Gamma(\beta)}\\
      & = \frac{\alpha \Gamma(\alpha)\Gamma(\beta)}{(\alpha+\beta) \Gamma(\alpha+\beta)} \frac{\Gamma(\alpha+\beta)}{\Gamma(\alpha)\Gamma(\beta)}\\
      & = \frac{\alpha}{\alpha+\beta}
      \label{eq:mean-beta-var}
    \end{align}
    where we have used the definition of the Beta function in terms of the Gamma function and the property $\Gamma(t+1) = t \Gamma(t)$.
  \end{solution}
  
\item  \label{q:bernoulli-posterior-predictive} Show that the predictive posterior probability $p( x=1 | \data )$ for a
  new independently observed data point $x$ equals the posterior mean
  of $p(\theta | \data)$, which in turn is given by
  \begin{align}
    \E( \theta | \data) & = \frac{\alpha_0 + n_{x=1}}{\alpha_0+\beta_0+n}.
    \label{eq:bernoulli-posterior-mean}
  \end{align}
  \begin{solution}
    We obtain
    \begin{align}
      p(x=1 | \data) & = \int_0^1 p(x=1, \theta | \data) \ud \theta \quad \quad \text{(sum rule)}\\
      & = \int_0^1 p(x=1 | \theta, \data) p(\theta | \data) \ud \theta \quad \quad \text{(product rule)} \\
      & = \int_0^1 p(x=1 | \theta) p(\theta | \data) \ud \theta \quad \quad (x \independent \data | \theta) \\
      & = \int_0^1 \theta p(\theta | \data) \ud \theta\\
      & = \E[ \theta | \data]
    \end{align}
    From the previous question we know the mean of a Beta random variable. Since $\theta \sim \BetaDist(\theta ; \alpha_n, \beta_n)$, we obtain
    \begin{align}
      p(x=1 | \data) & =  \E[ \theta | \data]\\
      & = \frac{\alpha_n}{\alpha_n+\beta_n}\\
      & = \frac{\alpha_0 +  n_{x=1}} {\alpha_0+ n_{x=1} + \beta_0 +  n_{x=0}}\\
      & = \frac{\alpha_0 +  n_{x=1}}{\alpha_0+ \beta_0 + n}
    \end{align}
    where the last equation follows from the fact that $n= n_{x=0}+n_{x=1}$. Note that for $n \to \infty$, the posterior mean tends to the MLE
    $n_{x=1}/n$.

  \end{solution}

 \end{exenumerate}


\ex{Bayesian inference of probability tables in fully observed directed graphical models of binary variables}
\label{ex:Bayesian-Inference-DGM}

This is the Bayesian analogue of Exercise \ref{ex:MLE-DGM} and the
notation follows that exercise. We consider the Bayesian model
\begin{align}
  p(\x | \thetab) &= \prod_{i=1}^d p(x_i | \pa_i, \thetab_i) \quad \quad x_i \in \{0,1\} \label{eq:p-dag2}\\
  p(\thetab; \alphab_0, \betab_0) & = \prod_{i=1}^d \prod_{s=1}^{S_i} \BetaDist(\theta_i^s; \alpha_{i,0}^s, \beta_{i,0}^s)
\end{align}
where $p(x_i | \pa_i, \thetab_i)$ is defined via \eqref{eq:pcond-fac}, $\alphab_0$ is a vector of hyperparameters containing all
$\alpha_{i,0}^s$, $\betab_0$ the vector containing all
$\beta_{i,0}^s$, and as before $\BetaDist$ denotes the
Beta distribution. Under the prior, all parameters are independent.

\begin{exenumerate}
\item For iid data $\data = \{\x^{(1)}, \ldots, \x^{(n)}\}$ show that 
  \begin{align}
    p(\thetab | \data) & = \prod_{i=1}^d \prod_{s=1}^{S_i} \BetaDist(\theta_i^s, \alpha_{i,n}^s, \beta_{i,n}^s)
  \end{align}
  where
  \begin{align}
    \alpha_{i,n}^s & = \alpha_{i,0}^s + n_{x_i=1}^s &  \beta_{i,n}^s & = \beta_{i,0}^s + n_{x_i=0}^s
  \end{align}
  and that the parameters are also independent under the posterior.
  
  \begin{solution}
    We start with
    \begin{equation}
      p(\thetab | \data) \propto p(\data | \thetab) p(\thetab; \alphab_0, \betab_0).
    \end{equation}
    Inserting the expression for $p(\data | \thetab)$ given in \eqref{eq:joint-dag} and the assumed form of the prior gives
    \begin{align}
      p(\thetab | \data) &\propto  \prod_{i=1}^d \prod_{s=1}^{S_i}  (\theta_i^s)^{n_{x_i=1}^s} (1-\theta_i^s)^{n_{x_i=0}^s} \prod_{i=1}^d \prod_{s=1}^{S_i} \BetaDist(\theta_i^s; \alpha_{i,0}^s, \beta_{i,0}^s)\\
      & \propto \prod_{i=1}^d \prod_{s=1}^{S_i}   (\theta_i^s)^{n_{x_i=1}^s} (1-\theta_i^s)^{n_{x_i=0}^s}  \BetaDist(\theta_i^s; \alpha_{i,0}^s, \beta_{i,0}^s)\\
      & \propto \prod_{i=1}^d \prod_{s=1}^{S_i}   (\theta_i^s)^{n_{x_i=1}^s} (1-\theta_i^s)^{n_{x_i=0}^s} (\theta_i^s)^{\alpha_{i,0}^s-1} (1-\theta_i^s)^{\beta_{i,0}^s-1}\\
      & \propto \prod_{i=1}^d \prod_{s=1}^{S_i}   (\theta_i^s)^{\alpha_{i,0}^s+n_{x_i=1}^s-1} (1-\theta_i^s)^{\beta_{i,0}^s+n_{x_i=0}^s-1}\\
      & \propto \prod_{i=1}^d \prod_{s=1}^{S_i} \BetaDist(\theta_i^s; \alpha_{i,0}^s+n_{x_i=1}^s, \beta_{i,0}^s+n_{x_i=0}^s)
    \end{align}
    It can be immediately verified that $\BetaDist(\theta_i^s; \alpha_{i,0}^s+n_{x_i=1}^s, \beta_{i,0}^s+n_{x_i=0}^s)$ is proportional to the marginal $p(\theta_i^s | \data)$ so that the parameters are independent under the posterior too.
  \end{solution}

\item \label{q:dgm-posterior-predictive} For a variable $x_i$ with parents $\pa_i$, compute the posterior predictive probability $p(x_i=1|\pa_i, \data)$

  \begin{solution}
    The solution is analogue to the solution for question
    \ref{q:bernoulli-posterior-predictive}, using the sum rule,
    independencies, and properties of beta random variables:
    \begin{align}
      p(x_i=1 |\pa_i=s, \data) &= \int p(x_i=1, \theta_i^s | \pa_i=s,
      \data) \ud \theta_i^s\\ & = \int p(x_i=1 | \theta_i^s, \pa_i=s,
      \data) p(\theta_i^s | \pa_i=s , \data)\\ & = \int p(x_i=1 |
      \theta_i^s, \pa_i=s) p(\theta_i^s | \data)\\ & = \int \theta_i^s
      p(\theta_i^s | \data)\\ & = \E[ \theta_i^s | \data)]\\ &
      \overset{\eqref{eq:mean-beta-var}}{=}
      \frac{\alpha_{i,n}^s}{\alpha_{i,n}^s+\beta_{i,n}^s}\\ & =
      \frac{\alpha_{i,0}^s+n^s_{x_i=1}}{\alpha_{i,0}^s+\beta_{i,0}^s+n^s}
      \label{eq:dgm-posterior-predictive}
    \end{align}
  \end{solution}
    where $n^s=n^s_{x_i=0}+n^s_{x_i=1}$ denotes the number of times
    the parent configuration $s$ occurs in the observed data $\data$.
    
\end{exenumerate}
  
\ex{Cancer-asbestos-smoking example: Bayesian inference}

Consider the model specified by the DAG

\begin{center}
\begin{tikzpicture}[dgraph]
  \node[cont] (x) at (0,0) {a};
  \node[cont] (y) at (2,0) {s};
  \node[cont] (z) at (1,-1) {c};
  \draw(x) -- (z);
  \draw(y) -- (z);
\end{tikzpicture}
\end{center}

The distribution of $a$ and $s$ are Bernoulli distributions with
parameter (success probability) $\theta_a$ and $\theta_s$,
respectively, i.e.\
\begin{equation}
  p(a | \theta_a) = \theta_a^a(1-\theta_a)^{1-a} \quad \quad p(s | \theta_s) = \theta_s^s(1-\theta_s)^{1-s},
\end{equation}
and the distribution of $c$ given the parents is parametrised as specified in the following table
\begin{center}
  \begin{tabular}{@{}rll@{}}
    \toprule
    $p(c=1 | a,s, \theta^1_c, \ldots, \theta_c^4))$ & $a$ & $s$\\
    \midrule
    $\theta^1_c$ & 0 & 0\\
    $\theta^2_c$ & 1 & 0\\
    $\theta^3_c$ & 0 & 1\\
    $\theta^4_c$ & 1 & 1\\
    \bottomrule
  \end{tabular}
\end{center}
We assume that the prior over the parameters of the model, $(\theta_a,
\theta_s, \theta^1_c, \ldots, \theta^4_c)$, factorises and is given by
beta distributions with hyperparameters $\alpha_0 = 1$ and $\beta_0 = 1$
(same for all parameters).

Assume we observe the following iid data (each row is a data point).

\begin{center}
  \scalebox{1}{
    \begin{tabular}{lll}
      \toprule
      a & s & c\\
      \midrule
      0 &   1 &   1\\
      0 &   0 &   0\\
      1 &   0 &   1\\
      0 &   0 &   0\\
      0 &   1 &   0\\
      \bottomrule
  \end{tabular}}
\end{center}

\begin{exenumerate}
\item Determine the posterior predictive probabilities $p(a =1 |\data)$ and $p(s=1|\data)$. 
  \begin{solution}
    With \exref {ex:Bayesian-inference-Bernoulli} question
    \ref{q:bernoulli-posterior-predictive}, we have
    \begin{align}
      p(a =1 |\data) &= \E(\theta^a | \data) = \frac{1 + 1}{1+1+5} = \frac{2}{7}\\
      p(s=1|\data) &=\E(\theta^s | \data) = \frac{1 + 2}{1+1+5} = \frac{3}{7}
    \end{align}
  \end{solution}

\item Determine the posterior predictive probabilities $p(c=1 | \pa,
  \data)$ for all possible parent configurations.
 
  \begin{solution}
    The parents of $c$ are $(a,s)$. With
    \exref{ex:Bayesian-Inference-DGM} question
    \ref{q:dgm-posterior-predictive}, we have
    \begin{center}
      \begin{tabular}{@{}lll@{}}
        \toprule
        $p(c=1 | a,s, \data)$ & $a$ & $s$\\
        \midrule
        $(1+0)/(1+1+2)=1/4 $ & 0 & 0\\
        $(1+1)/(1+1+1)=2/3 $ & 1 & 0\\
        $(1+1)/(1+1+2)=1/2 $ & 0 & 1\\
        $(1+0)/(1+1)=1/2 $ & 1 & 1\\
        \bottomrule
      \end{tabular}
    \end{center}
    Compared to the MLE solution in
    \exref{q:cancer-smoking-asbestos-mle} question
    \ref{q:cancer-smoking-asbestos-mle}, we see that the estimates are
    less extreme. This is because they are a combination of the prior
    knowledge and the observed data. Moreover, when we do not have any
    data, the posterior equals the prior, unlike for the mle where the
    estimate is not defined.

  \end{solution}
  
\end{exenumerate}


\ex{Learning parameters of a directed graphical model}

 We consider the directed graphical model shown below on the left for
 the four binary variables $t,b,s,x$, each being either zero or
 one. Assume that we have observed the data shown in the table on the
 right.\\
 
  \begin{minipage}[t]{0.45 \textwidth}
    \begin{center}
      {\small Model:\\[2ex]}
      \scalebox{1}{ 
        \begin{tikzpicture}[dgraph]
          \node[cont] (t) at (0,0) {$t$};
          \node[cont] (b) at (2,0) {$b$};
          \node[cont] (s) at (1,-1.5) {$s$};
          \node[cont] (x) at (-1,-1.5) {$x$};

          \draw (t) -- (s);
          \draw (b) -- (s);
          \draw (t) -- (x);
      \end{tikzpicture}}
    \end{center}\vspace{2ex}
    \begin{tabular}{l l}
    $t=1$& has tuberculosis\\
    $b=1$& has bronchitis \\
    $s=1$& has shortness of breath\\
      $x=1$& has positive x-ray
    \end{tabular}
  \end{minipage}
  \hspace{2ex}
  \begin{minipage}[t]{0.45\textwidth}
    \begin{center}
      {\small Observed data:\\[2ex]}
     \scalebox{1}{
      \begin{tabular}{llll}
       \toprule
        x & s & t & b\\
        \midrule
        0 &   1 &   0 &   1\\
        0 &   0 &   0 &   0\\
        0 &   1 &   0 &   1\\
        0 &   1 &   0 &   1\\
        0 &   0 &   0 &   0\\
        0 &   0 &   0 &   0\\
        0 &   1 &   0 &   1\\
        0 &   1 &   0 &   1\\
        0 &   0 &   0 &   1\\
        1 &   1 &   1 &   0\\
        \bottomrule
      \end{tabular}}
    \end{center}
  \end{minipage}\\[2ex]
We assume the (conditional) pmf of $s|t,b$ is specified by the following
parametrised probability table:
\begin{center}
  \begin{tabular}{@{}cll@{}}
    \toprule
    $p(s=1 | t, b; \theta^1_s, \ldots, \theta_s^4))$ & $t$ & $b$\\
    \midrule
    $\theta^1_s$ & 0 & 0\\
    $\theta^2_s$ & 1 & 0\\
    $\theta^3_s$ & 0 & 1\\
    $\theta^4_s$ & 1 & 1\\
    \bottomrule
  \end{tabular}
\end{center}

  \begin{exenumerate}
  \item What are the maximum likelihood estimates for $p(s=1 | b=0,
    t=0)$ and $p(s=1 | b=0, t=1)$, i.e.\ the parameters $\theta^1_s$
    and $\theta^3_s$?
    
    \begin{solution}
      The maximum likelihood estimates (MLEs) are equal to the fraction of occurrences of the relevant events.
      \begin{align}
        \hat{\theta}^1_s & =\frac{\sum_{i=1}^n \ind(s_i=1, b_i=0, t_i=0)}{\sum_{i=1}^n \ind(b_i=0, t_i=0)} = \frac{0}{3} = 0\\
        \hat{\theta}^3_s & =\frac{\sum_{i=1}^n \ind(s_i=1, b_i=0, t_i=1)}{\sum_{i=1}^n \ind(b_i=0, t_i=1)} = \frac{1}{1} = 1
      \end{align}

    \end{solution}
    
  \item Assume each parameter in the table for $p(s | t,b)$ has a
    uniform prior on $(0,1)$. Compute the posterior mean of the
    parameters of \mbox{$p(s=1 | b=0, t=0)$} and $p(s=1 | b=0, t=1)$
    and explain the difference to the maximum likelihood estimates.

    \begin{solution}
      A uniform prior corresponds to a Beta distribution with
      hyperparameters $\alpha_0=\beta_0=1$. With
      \exref{ex:Bayesian-Inference-DGM} question
      \ref{q:dgm-posterior-predictive}, we have
      \begin{align}
        \E(\theta_s^1 | \data) & = \frac{\alpha_0+0}{\alpha_0+\beta_0+3} = \frac{1}{5}\\
        \E(\theta_s^3 | \data) & = \frac{\alpha_0+1}{\alpha_0+\beta_0+1} = \frac{2}{3}
      \end{align}
      Compared to the MLE, the posterior mean is less extreme. It can
      be considered a ``smoothed out'' or regularised estimate, where
      $\alpha_0 >0$ and $\beta_0>0$ provides regularisation (see
      \url{https://en.wikipedia.org/wiki/Additive_smoothing}). We can
      see a pull of the parameters towards the prior predictive mean,
      which equals 1/2.

    \end{solution}
    
  \end{exenumerate}
  
%



\ex{Factor analysis}
\label{ex:FA}

A friend proposes to improve the factor analysis model by
working with correlated latent variables. The proposed model is
\begin{align}
  p(\h; \C) &= \Gauss(\h; \zerob, \C) & p(\v | \h; \F, \Psib, \c) = \Gauss(\v; \F \h+\c, \Psib)
\end{align}
where $\C$ is some $H \times H$ covariance matrix, $\F$ is the $D
\times H$ matrix with the factor loadings, $\Psib=\diag(\Psi_1,
\ldots, \Psi_D)$, $\c\in \mathbb{R}^D$ and the dimension of the
latents $H$ is less than the dimension of the visibles
$D$. $\Gauss(\x; \mub, \Sigmab)$ denotes the pdf of a Gaussian with
mean $\mub$ and covariance matrix $\Sigmab$. The standard factor analysis model is obtained when $\C$ is the identity matrix.

\begin{exenumerate}
  
\item What is marginal distribution of the visibles $p(\v; \thetab)$ where $\thetab$ stands for the parameters $\C, \F, \c, \Psib$?
   
  \begin{solution}
    The model specifications are equivalent to the following data generating process:
    \begin{align}
      \h &\sim \Gauss(\h; \zerob, \C) & \epsilonb &\sim \Gauss(\epsilonb; \zerob, \Psib) &  \v &= \F \h + \c + \epsilonb
    \end{align}
    Recall the basic result on the distribution of linear
    transformations of Gaussians: if $\x$ has density $\Gauss(\x;
    \mub_x, \C_x)$, $\z$ density $\Gauss(\z; \mub_z, \C_z)$, and $\x
    \independent \z$ then $\y = \A \x + \z$ has density
    $$\Gauss(\y; \A\mub_x+\mub_z, \A \C_x \A^\top + \C_z).$$
    It thus follows that $\v$ is Gaussian with mean $\mub$ and covariance $\Sigmab$,
    \begin{align}
      \mub &= \F \underbrace{\E[\h]}_{\zerob} + \c + \underbrace{\E[\epsilonb]}_{\zerob}\\
      & = \c\\
      \Sigmab &= \F \Var[\h] \F^\top + \Var[\epsilonb]\\
      & = \F \C \F^\top + \Psib.
    \end{align}

  \end{solution}
  
\item Assume that the singular value decomposition of $\C$ is given by
  \begin{align}
    \C & = \EE \Lambdab \EE^\top
  \end{align}
  where $\Lambdab = \diag(\lambda_1, \ldots, \lambda_D)$ is a diagonal
  matrix containing the eigenvalues, and $\EE$ is a orthonormal matrix
  containing the corresponding eigenvectors. The matrix square root of $\C$ is the matrix $\M$ such that
  \begin{align}
    \M \M = \C,
  \end{align}
  and we denote it by $\C^{1/2}$. Show that the matrix square root of
  $\C$ equals
  \begin{align}
    \C^{1/2} = \EE \diag(\sqrt{\lambda_1}, \ldots, \sqrt{\lambda_D}) \EE^\top.
  \end{align}
  
  \begin{solution}
    We verify that $\C^{1/2}\C^{1/2} = \C$:
    \begin{align}
      \C^{1/2} \C^{1/2} & = \EE \diag(\sqrt{\lambda_1}, \ldots, \sqrt{\lambda_D})\EE^\top \EE \diag(\sqrt{\lambda_1}, \ldots, \sqrt{\lambda_D})\EE^\top\\
      & = \EE \diag(\sqrt{\lambda_1}, \ldots, \sqrt{\lambda_D}) \; \I \; \diag(\sqrt{\lambda_1}, \ldots, \sqrt{\lambda_D})\EE^\top\\
     &= \EE \diag(\sqrt{\lambda_1}, \ldots, \sqrt{\lambda_D}) \diag(\sqrt{\lambda_1}, \ldots, \sqrt{\lambda_D})\EE^\top\\
      &= \EE \diag(\lambda_1, \ldots, \lambda_D) \EE^\top\\
      &= \EE \Lambdab \EE^\top\\
      & =\C
    \end{align}
    
  \end{solution}

\item Show that the proposed factor analysis model is equivalent to the original factor analysis model
  \begin{align}
    p(\h; \I) &= \Gauss(\h; \zerob, \I) & p(\v | \h; \tilde{\F}, \Psib, \c) = \Gauss(\v; \tilde{\F} \h+\c, \Psib)
  \end{align}
with $\tilde{\F} = \F \C^{1/2}$, so that the extra parameters given by
the covariance matrix $\C$ are actually redundant and nothing is gained with the richer parametrisation. 

\begin{solution}
  We verify that the model has the same distribution for the visibles. As before $\E[\v] = \c$, and the covariance matrix is
  \begin{align}
    \Var[\v] & = \tilde{\F} \I \tilde{\F}^\top + \Psib\\
    & =  \F \C^{1/2} \C^{1/2}  \F^\top + \Psib\\
    & =  \F \C  \F^\top + \Psib
  \end{align}
  where we have used that $\C^{1/2}$ is a symmetric matrix. This means
  that the correlation between the $\h$ can be absorbed into the
  factor matrix $\F$ and the set of pdfs defined by the proposed model
  equals the set of pdfs of the original factor analysis model.

  Another way to see the result is to consider the data generating
  process and noting that we can sample $\h$ from $\Gauss(\h; \zerob, \C)$
  by first sampling $\h'$ from $\Gauss(\h'; \zerob, \I)$ and then transforming the sample by $\C^{1/2}$,
  \begin{align}
    \h &\sim \Gauss(\h; \zerob, \C)  & \Longleftrightarrow && \h &= \C^{1/2} \h' \quad \quad \quad \h' \sim \Gauss(\h'; \zerob, \I).
  \end{align}
  This follows again from the basic properties of linear transformations of Gaussians, i.e.
  $$ \Var(\C^{1/2} \h') = \C^{1/2}\Var(\h')(\C^{1/2})^\top =  \C^{1/2}\I \C^{1/2} =  \C$$
  and $\E(\C^{1/2} \h') = \C^{1/2} \E(\h') = \zerob$.

  To generate samples from the proposed factor analysis model, we would thus proceed as follows:
  \begin{align}
  \h' &\sim \Gauss(\h'; \zerob, \I) & \epsilonb &\sim \Gauss(\epsilonb; \zerob, \Psib) &  \v &= \F (\C^{1/2}\h') + \c + \epsilonb
  \end{align}
  But the term
  $$ \v = \F (\C^{1/2}\h') + \c + \epsilonb$$
  can be written as
  $$\v = (\F \C^{1/2} ) \h' + \c + \epsilonb = \tilde{\F} \h' + \c + \epsilonb$$
  and since $\h'$ follows $\Gauss(\h'; \zerob, \I)$, we are back at the original factor analysis model.
  
\end{solution}
\end{exenumerate}

\ex{Independent component analysis}
\label{ex:ex2}

\begin{exenumerate}
  
\item Whitening corresponds to linearly transforming a random variable
  $\x$ (or the corresponding data) so that the resulting random variable $\z$ has an identity covariance matrix, i.e.\
  $$\z = \V \x \quad \text{with} \quad \Var[\x] = \C \quad
  \text{and}\quad \Var[\z] = \I. $$ The matrix $\V$ is called the
  whitening matrix. We do not make a distributional assumption on
  $\x$, in particular $\x$ may or may not be Gaussian.

  Given the eigenvalue decomposition $\C = \EE \Lambdab \EE^\top$,
  show that
  \begin{equation}
    \V = \diag (\lambda_1^{-1/2}, \ldots,  \lambda_d^{-1/2}) \EE^\top
  \end{equation}
  is a whitening matrix.
    
  \begin{solution}
    From $\Var[\z] = \Var[\V\x]= \V \Var[\x] \V^\top$, it follows that
    \begin{align}
      \Var[\z] & =  \V \Var[\x] \V^\top\\
      & =  \V \C \V^\top\\
      & =  \V \EE \Lambdab \EE^\top \V^\top\\
      & =   \diag (\lambda_1^{-1/2}, \ldots,
  \lambda_d^{-1/2}) \EE^\top \EE \Lambdab \EE^\top  \V^\top\\
& = \diag (\lambda_1^{-1/2}, \ldots,
  \lambda_d^{-1/2}) \Lambdab \EE^\top \V^\top
    \end{align}
    where we have used that $\EE^\top \EE = \I$. Since
    $$ \V^\top = \left[  \diag (\lambda_1^{-1/2}, \ldots, \lambda_d^{-1/2}) \EE^\top \right]^\top = \EE  \diag (\lambda_1^{-1/2}, \ldots, \lambda_d^{-1/2})$$
    we further have
    \begin{align}
      \Var[\z] & = \diag (\lambda_1^{-1/2}, \ldots,
      \lambda_d^{-1/2}) \Lambdab \EE^\top  \EE  \diag (\lambda_1^{-1/2}, \ldots, \lambda_d^{-1/2})\\
      & =  \diag (\lambda_1^{-1/2}, \ldots,
      \lambda_d^{-1/2}) \Lambdab \diag (\lambda_1^{-1/2}, \ldots, \lambda_d^{-1/2})\\
      & =  \diag (\lambda_1^{-1/2}, \ldots,
      \lambda_d^{-1/2}) \diag(\lambda_1, \ldots, \lambda_d) \diag (\lambda_1^{-1/2}, \ldots, \lambda_d^{-1/2})\\
      & = \I,
    \end{align}
    so that $\V$ is indeed a valid whitening matrix. Note that whitening matrices are not unique. For example,
   $$ \tilde{\V} = \EE \diag (\lambda_1^{-1/2}, \ldots,
    \lambda_d^{-1/2}) \EE^\top$$ is also a valid whitening
    matrix. More generally, if $\V$ is a whitening matrix, then $\R \V$ is
    also a whitening matrix when $\R$ is an orthonormal
    matrix. This is because
    $$ \Var[ \R \V \x ] = \R \Var[\V \x] \R^\top = \R \I \R^{\top} = \I $$
    where we have used that $\V$ is a whitening matrix so that $\V \x$ has identity covariance matrix.

  \end{solution}
  
\item Consider the ICA model
\begin{align}
      \v &= \A \h, & \h &\sim p_{\h}(\h), &  p_{\h}(\h)&= \prod_{i=1}^D p_h(h_i),
\end{align}
where the matrix $\A$ is invertible and the $h_i$ are independent
random variables of mean zero and variance one. Let $\V$ be a
whitening matrix for $\v$. Show that $\z = \V \v$ follows the ICA
model
\begin{align}
      \z &= \tilde{\A} \h, & \h &\sim p_{\h}(\h), &  p_{\h}(\h)&= \prod_{i=1}^D p_h(h_i),
\end{align}
where $\tilde{\A}$ is an orthonormal matrix.

\begin{solution}
  If $\v$ follows the ICA model, we have
  \begin{align}
    \z & = \V \v\\
    & = \V \A \h\\
    &= \tilde{\A} \h
  \end{align}
  with $\tilde{\A} = \V \A$. By the whitening operation, the covariance matrix of $\z$ is identity, so that
  \begin{align}
    \I = \Var(\z) =  \tilde{\A} \Var(\h)  \tilde{\A}^{\top}.
  \end{align}
  By the ICA model, $\Var(\h) = \I$, so that $\tilde{\A}$ must satisfy
  \begin{equation}
    \I =  \tilde{\A} \tilde{\A}^{\top},
  \end{equation}
  which means that $\tilde{\A}$ is orthonormal.

  In the original ICA model, the number of parameters is given by the
  number of elements of the matrix $\A$, which is $D^2$ if $\v$ is
  D-dimensional. An orthogonal matrix contains $D(D-1)/2$ degrees of
  freedom {\small (see
    e.g. \url{https://en.wikipedia.org/wiki/Orthogonal_matrix}}), so
  that we can think that whitening ``solves half of the ICA
  problem''. Since whitening is a relatively simple standard
  operation, many algorithms \citep[e.g.\ ``fastICA'',][]{Hyvarinen1999}
  first reduce the complexity of the estimation problem by whitening
  the data. Moreover, due to the properties of the orthogonal matrix,
  the log-likelihood for the ICA model also simplifies for whitened
  data: The log-likelihood for ICA model without whitening is
  \begin{equation}
    \ell(\B) = \sum_{i=1}^n  \sum_{j=1}^D\log p_h (\b_j \v_i) + n \log |\det \B|
  \end{equation}
  where $\B = \A^{-1}$. If we first whiten the data, the log-likelihood becomes
  \begin{equation}
    \ell(\tilde{\B}) = \sum_{i=1}^n  \sum_{j=1}^D\log p_h (\tilde{\b}_j \z_i) + n \log |\det \tilde{\B}|
  \end{equation}
  where $\tilde{\B} = \tilde{\A}^{-1} =\tilde{\A}^\top$ since $\A$ is
  an orthogonal matrix. This means $\tilde{\B}^{-1} = \tilde{\A} =
  \tilde{\B}^\top$ and $\tilde{\B}$ is an orthogonal matrix. Hence
  $\det \tilde{\B}=1$, and the $\log \det$ term is zero. Hence, the
  log-likelihood on whitened data simplifies to
   \begin{equation}
    \ell(\tilde{\B}) = \sum_{i=1}^n  \sum_{j=1}^D\log p_h (\tilde{\b}_j \z_i).
  \end{equation}
   While the log-likelihood takes a simpler form, the optimisation
   problem is now a constrained optimisation problem: $\tilde{\B}$ is
   constrained to be orthonormal. For further information, see
   e.g.\ \citep[Chapter 9]{Hyvarinen2001}.  
   
  \end{solution}
  
\end{exenumerate}


\ex{Score matching for the exponential family}
\label{ex:score-matching-exp-family}
The objective function $J(\thetab)$ that is minimised in score matching is
    \begin{align} 
      J(\thetab) &= \frac{1}{n} \sum_{i=1}^n \sum_{j=1}^m \left[ \partial_j
        \psi_j(\x_i;\thetab) + \frac{1}{2} \psi_j(\x_i;\thetab)^2 \right],\label{eq:Jsm-def}
    \end{align}
where $\psi_j$ is the partial derivative of the log model-pdf $\log
p(\x ;\thetab)$ with respect to the $j$-th coordinate (slope) and
$\partial_j \psi_j$ its second partial derivative (curvature). The
observed data are denoted by $\x_1,\ldots,\x_n$ and $\x \in
\mathbb{R}^m$.

The goal of this exercise is to show that for statistical models of the form
\begin{align}
  \log p(\x;\thetab) &= \sum_{k=1}^K \theta_k F_k(\x) -\log Z(\thetab), \quad \quad \quad \x \in \mathbb{R}^m,
  \label{eq:exp-family-def}
\end{align}
the score matching objective function becomes a quadratic form, which can be optimised efficiently \citep[see e.g.\ ][Appendix A.5.3]{Barber2012}.

The set of models above are called the (continuous) exponential
family, or also log-linear models because the models are linear in the
parameters $\theta_k$. Since the exponential family generally includes
probability mass functions as well, the qualifier ``continuous'' may
be used to highlight that we are here considering continuous random
variables only. The functions $F_k(\x)$ are assumed to be known (they
are called the sufficient statistics).

\begin{exenumerate}

\item Denote by $\K(\x)$ the matrix with elements $K_{kj}(\x)$,
  \begin{equation}
    K_{kj}(\x) = \frac{\partial F_k(\x)}{\partial x_j}, \quad \quad \quad k=1 \ldots K, \quad j=1 \ldots m,
  \end{equation}
  and by $\H(\x)$ the matrix with elements $H_{kj}(\x)$,
  \begin{equation}
    H_{kj}(\x) = \frac{\partial^2 F_k(\x)}{\partial x_j^2}, \quad \quad \quad k=1 \ldots K, \quad j=1 \ldots m.
  \end{equation}
  Furthermore, let $\h_j(\x) = (H_{1j}(\x), \ldots, H_{Kj}(\x))^\top$ be the $j$–th column vector of $\H(\x)$.
  
  Show that for the continuous exponential family, the score matching objective in Equation \eqref{eq:Jsm-def} becomes
  \begin{equation}
    J(\thetab) =  \thetab^\top \r +  \frac{1}{2} \thetab^\top \M \thetab,
    \end{equation}
    where
    \begin{align}
      \label{eq:Jsm-sol}
       \r&=\frac{1}{n}\sum_{i=1}^n \sum_{j=1}^m \h_j(\x_i), & \M &= \frac{1}{n}\sum_{i=1}^n \K(\x_i)\K(\x_i)^\top.
    \end{align}
     
  \begin{solution}
    For
    \begin{align}
      \log p(\x;\thetab) &= \sum_{k=1}^K \theta_k F_k(\x) -\log Z(\thetab)
    \end{align}
    the first derivative with respect to $x_j$, the $j$-th element of $\x$, is
    \begin{align}
      \psi_j(\x; \thetab) &= \frac{\partial \log p(\x; \thetab)}{\partial x_j}\\
      & = \sum_{k=1}^K \theta_k \frac{\partial F_k(\x)}{\partial x_j}\\
      & =  \sum_{k=1}^K \theta_k K_{kj}(\x).
    \end{align}
    The second derivative is
    \begin{align}
      \partial_j \psi_j(\x; \thetab) & = \frac{\partial^2 \log p(\x; \thetab)}{\partial x_j^2}\\
      & =  \sum_{k=1}^K \theta_k \frac{\partial^2 F_k(\x)}{\partial x_j^2}\\
      & =  \sum_{k=1}^K \theta_k  H_{kj}(\x),
    \end{align}
    which we can write more compactly as
    \begin{align}
      \partial_j  \psi_j(\x; \thetab) &= \thetab^\top \h_j(\x).
    \end{align}

    The score matching objective in Equation \eqref{eq:Jsm-def}  features the sum $\sum_j \psi_j(\x;\thetab)^2$. The term $\psi_j(\x;\thetab)^2$ equals
    \begin{align}
      \psi_j(\x; \thetab)^2 & = \left[ \sum_{k=1}^K \theta_k K_{kj}(\x)\right]^2\\
      & =  \sum_{k=1}^K \sum_{k'=1}^K  K_{kj}(\x) K_{k'j}(\x) \theta_{k} \theta_{k'},
    \end{align}
    so that
    \begin{align}
      \sum_{j=1}^m \psi_j(\x; \thetab)^2  &= \sum_{j=1}^m  \sum_{k=1}^K \sum_{k'=1}^K  K_{kj}(\x) K_{k'j}(\x) \theta_{k} \theta_{k'}\\
      & = \sum_{k=1}^K \sum_{k'=1}^K \theta_{k} \theta_{k'} \left[ \sum_{j=1}^m K_{kj}(\x) K_{k'j}(\x) \right],
    \end{align}
    which can be more compactly expressed using matrix notation. Noting
    that $$\sum_{j=1}^m K_{kj}(\x_i) K_{k'j}(\x_i)$$ equals the
    $(k,k')$ element of the matrix-matrix product $\K(\x_i)\K(\x_i)^\top$,
    \begin{equation}
      \sum_{j=1}^m K_{kj}(\x_i) K_{k'j}(\x_i) = \left[\K(\x_i)\K(\x_i)^\top\right]_{k,k'},
    \end{equation}
    we can write
    \begin{align}
      \sum_{j=1}^m \psi_j(\x; \thetab)^2 & =  \sum_{k=1}^K \sum_{k'=1}^K \theta_{k} \theta_{k'}  \left[\K(\x_i)\K(\x_i)^\top\right]_{k,k'}\\
      & =  \thetab^\top \K(\x_i)\K(\x_i)^\top \thetab
    \end{align}
    where we have used that for some matrix $\A$
    \begin{equation}
      \thetab^\top \A \thetab = \sum_{k,k'} \theta_k \theta_{k'} [\A]_{k,k'}
    \end{equation}
    where $[\A]_{k,k'}$ is the $(k,k')$ element of the matrix $\A$.
    
    Inserting the expressions into Equation \eqref{eq:Jsm-def} gives
    \begin{align} 
      J(\thetab) &= \frac{1}{n} \sum_{i=1}^n  \sum_{j=1}^m\left[ \partial_j \psi_j(\x_i;\thetab) + \frac{1}{2} \psi_j(\x_i;\thetab)^2 \right] \\
      &= \frac{1}{n} \sum_{i=1}^n  \sum_{j=1}^m \partial_j \psi_j(\x_i;\thetab) + \frac{1}{2}  \frac{1}{n} \sum_{i=1}^n  \sum_{j=1}^m \psi_j(\x_i;\thetab)^2 \\
      & =  \frac{1}{n} \sum_{i=1}^n \sum_{j=1}^m \thetab^\top \h_j(\x_i) + \frac{1}{2} \frac{1}{n} \sum_{i=1}^n \thetab^\top \K(\x_i)\K(\x_i)^\top \thetab\\
        & =  \thetab^\top \left[  \frac{1}{n} \sum_{i=1}^n \sum_{j=1}^m \h_j(\x_i)\right] + \frac{1}{2}  \thetab^\top \left[ \frac{1}{n}\sum_{i=1}^n  \K(\x_i)\K(\x_i)^\top \right] \thetab\\
        & = \thetab^\top \r + \frac{1}{2}\thetab^\top \M \thetab,
    \end{align}
    which is the desired result. 
  \end{solution}

\item The pdf of a zero mean Gaussian parametrised by the variance $\sigma^2$ is 
  \begin{align}
    p(x; \sigma^2) = \frac{1}{\sqrt{2 \pi \sigma^2}}\exp\left(-\frac{x^2}{2\sigma^2} \right), \quad \quad \quad x \in \mathbb{R}.
  \end{align}
  The (multivariate) Gaussian is a member of the exponential
  family. By comparison with Equation \eqref{eq:exp-family-def}, we
  can re-parametrise the statistical model $\{p(x; \sigma^2)\}_{\sigma^2}$ and work with
  \begin{align}
    p(x; \theta) = \frac{1}{Z(\theta)}\exp\left(\theta x^2 \right), \quad \quad \theta <0, \quad \quad x \in \mathbb{R},
  \end{align}
  instead. The two parametrisations are related by $\theta =
  -1/(2 \sigma^2)$. Using the previous result on the (continuous)
  exponential family, determine the score matching estimate
  $\hat{\theta}$, and show that the corresponding $\hat{\sigma}^2$ is
  the same as the maximum likelihood estimate. This result is
  noteworthy because unlike in maximum likelihood estimation, score
  matching does not need the partition function $Z(\theta)$ for the estimation.

  \begin{solution}
    By comparison with Equation \eqref{eq:exp-family-def}, the
    sufficient statistics $F(x)$ is $x^2$.

    We first determine the score matching objective function. For
    that, we need to determine the quantities $\r$ and $\M$ in
    Equation \eqref{eq:Jsm-sol}. Here, both $\r$ and $\M$ are scalars,
    and so are the matrices $\K$ and $\H$ that define $\r$ and $\M$. By
    their definitions, we obtain
    \begin{align}
      K(x) &=  \frac{\partial F(x)}{\partial x} = 2x\\
      H(x) &= \frac{\partial^2 F(x)}{\partial x^2} = 2\\
      r & = 2\\
      M &=  \frac{1}{n}\sum_{i=1}^n K(x_i)^2 \\
      & = 4 m_2
    \end{align}
    where $m_2$ denotes the second empirical moment,
    \begin{equation}
      m_2 = \frac{1}{n} \sum_{i=1}^n x_i^2.
    \end{equation}
    With Equation \eqref{eq:Jsm-def}, the score matching objective thus is
    \begin{align}
      J(\theta) &= 2 \theta + \frac{1}{2} 4 m_2 \theta^2 \\
      & = 2 \theta + 2 m_2 \theta^2
    \end{align}
    A necessary condition for the minimiser to satisfy is
    \begin{align}
      \frac{\partial J(\theta)}{\partial \theta} & = 2 + 4 \theta m_2 \\
      & = 0
    \end{align}
    The only parameter value that satisfies the condition is
    \begin{equation}
      \hat{\theta} = -\frac{1}{2 m_2}.
    \end{equation}
    The second derivative of $J(\theta)$ is
      \begin{align}
      \frac{\partial^2 J(\theta)}{\theta^2} & =  m_2,
    \end{align}
      which is positive (as long as all data points are non-zero). Hence $\hat{\theta}$ is a minimiser.

      From the relation $\theta = -1/(2 \sigma^2)$, we obtain that the score matching estimate of the variance $\sigma^2$ is
      \begin{equation}
        \hat{\sigma}^2 = - \frac{1}{2\hat{\theta}} = m_2.
      \end{equation}
      We can obtain the score matching estimate $\hat{\sigma}^2$ from
      $\hat{\theta}$ in this manner for the same reason that we were
      able to work with transformed parameters in maximum likelihood
      estimation.

      For zero mean Gaussians, the second moment $m_2$ is the maximum likelihood
      estimate of the variance, which shows that the score matching and maximum
      likelihood estimate are here the same. While the two methods generally
      yield different estimates, the result also holds for multivariate
      Gaussians where the score matching estimates also equal the maximum
      likelihood estimates, see the original article on score matching by
      \citet{Hyvarinen2005c}.
      
  \end{solution}
  
\end{exenumerate}


\ex{Maximum likelihood estimation and unnormalised models}
Consider the Ising model for two binary random variables $(x_1,x_2)$,
\begin{align*}
  p(x_1,x_2;\theta) \propto \exp\left(\theta x_1 x_2+x_1+x_2\right), \quad \quad x_i \in \{-1,1\},
  \label{eq:Q1energy}
\end{align*}

\begin{exenumerate}
\item Compute the partition function $Z(\theta)$.

  \begin{solution}
    The definition of the partition function is
    \begin{align}
      Z(\theta) & = \sum_{ \{-1,1\}^2} \exp\left(\theta x_1 x_2+x_1+x_2\right).
    \end{align}
    where have have to sum over $(x_1,x_2) \in  \{-1,1\}^2 = \{ (-1,1),\, (1,1),\, (1,-1)\, (-1-1)\}$. This gives
    \begin{align}
      Z(\theta) & = \exp(-\theta-1+1)+\exp(\theta+2)+\exp(-\theta+1-1)+\exp(\theta-2)\\
      & = 2\exp(-\theta)+\exp(\theta+2)+\exp(\theta-2)
    \end{align}

  \end{solution}
  
\item The figure below shows the graph of $f(\theta)= \frac{\partial \log Z(\theta)}{\partial \theta}$.\vspace{1ex}

  Assume you observe three data points $(x_1, x_2)$ equal to
  $(-1,-1)$, $(-1,1)$, and $(1,-1)$. Using the figure, what is the
  maximum likelihood estimate of $\theta$? Justify your answer. 
    
\begin{center}
  \includegraphics[width=0.4\textwidth]{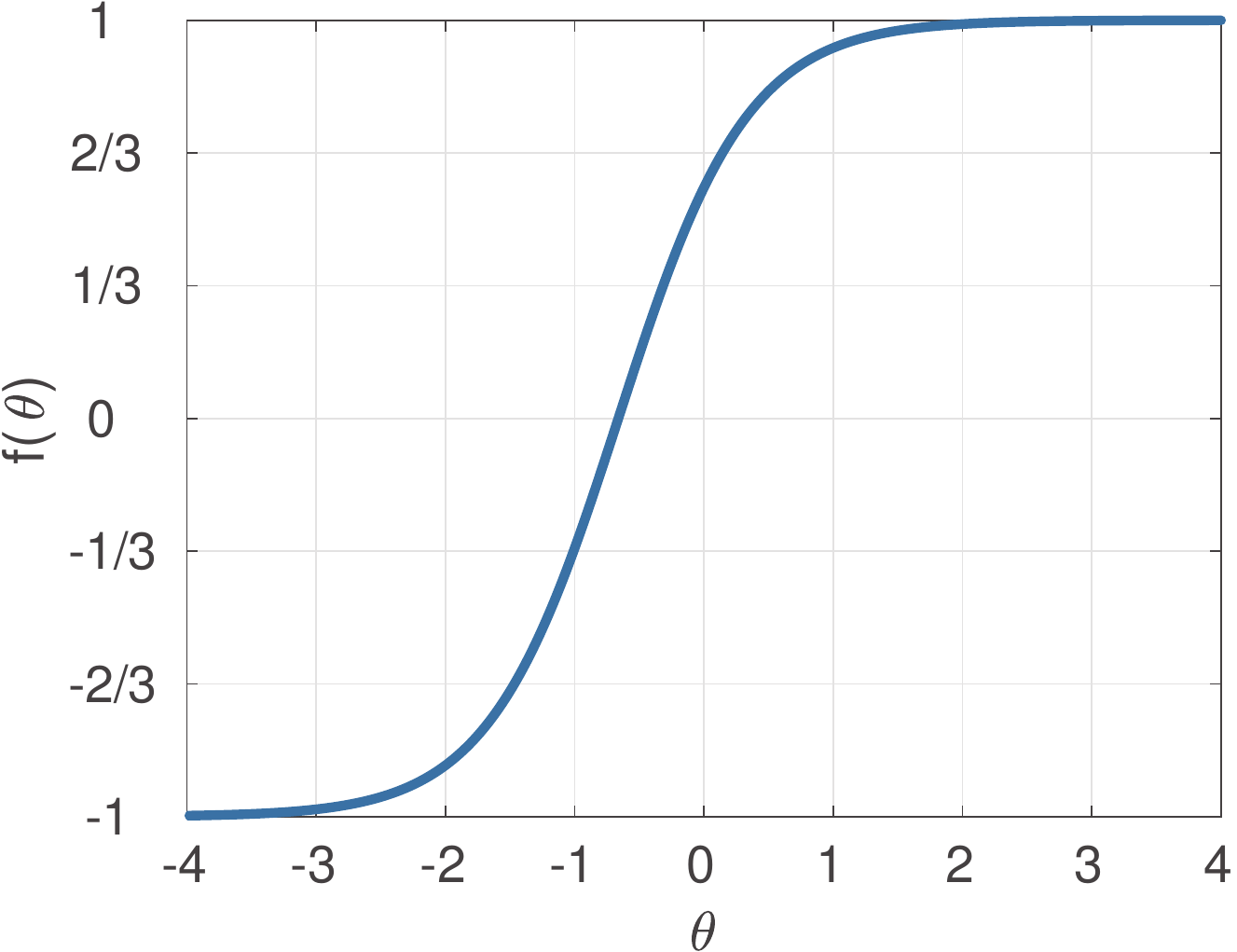}
\end{center}

  \begin{solution}

    Denoting the $i$-th observed data point by $(x_1^i,x_2^i)$,
    the log-likelihood is
    \begin{align}
      \ell(\theta) & = \sum_{i=1}^n \log p(x_1^i,x_2^i; \theta)
    \end{align}
    Inserting the definition of the $p(x_1,x_2;\theta)$ yields
    \begin{align}
      \ell(\theta)  & = \sum_{i=1}^n \left[ \theta x_1^i x_2^i+x_1^i+x_2^i \right] - n \log Z(\theta)\\
      &= \theta \sum_{i=1}^n \left[x_1^i x_2^i\right] + \sum_{i=1}^n \left[x_1^i+x_2^i \right] - n\log Z(\theta)
    \end{align}
    Its derivative with respect to the $\theta$ is
    \begin{align}
      \frac{\partial \ell(\theta)}{\partial \theta} & =  \sum_{i=1}^n \left[x_1^i x_2^i\right] - n  \frac{\partial \log Z(\theta)}{\partial \theta}\\
      & =  \sum_{i=1}^n \left[x_1^i x_2^i\right] - n f(\theta)
    \end{align}
    Setting it to zero yields
    \begin{align}
      \frac{1}{n}  \sum_{i=1}^n \left[x_1^i x_2^i\right] &= f(\theta)
    \end{align}

    An alternative approach is to start with the more general
    relationship that relates the gradient of the partition function
    to the gradient of the log unnormalised model. For example, if
    $$ p(\x,\thetab) = \frac{\phi(\x; \thetab)}{Z(\thetab)}$$
    we have
    \begin{align}
      \ell(\thetab) &= \sum_{i=1}^n \log p(\x_i;\thetab) \\
      & = \sum_{i=1}^n \log \phi(\x_i;\thetab) -n \log Z(\thetab)
    \end{align}
    Setting the derivative to zero gives,
    $$\frac{1}{n} \sum_{i=1}^n \nabla_{\thetab}  \log \phi(\x_i;\thetab) = \nabla_{\thetab} \log Z(\thetab)$$
      
    In either case, numerical evaluation of  $1/n \sum_{i=1}^n x_1^i x_2^i$ gives
    \begin{align}
      \frac{1}{n}  \sum_{i=1}^n \left[x_1^i x_2^i\right] & = \frac{1}{3}\left(1-1-1\right)\\
      &= -\frac{1}{3}
    \end{align}
    From the graph, we see that $f(\theta)$ takes on the value $-1/3$
    for $\theta = -1$, which is the desired MLE. 

  \end{solution}

\end{exenumerate}

\ex{Parameter estimation for unnormalised models}
Let $p(\x; \A) \propto \exp(-\x^\top \A \x)$ be a parametric
statistical model for $\x = (x_1, \ldots, x_{100})$, where the
parameters are the elements of the matrix $\A$. Assume that $\A$ is
symmetric and positive semi-definite, i.e.\ $\A$ satisfies $\x^\top
\A \x \ge 0$ for all values of $\x$.

\begin{exenumerate}
\item For $n$ iid data points $\x_1, \ldots, \x_n$, a friend
  proposes to estimate $\A$ by maximising $J(\A)$,
  \begin{equation}
    J(\A) = \prod_{k=1}^n \exp\left( - \x_k^\top \A \x_k \right).
  \end{equation}
  Explain why this procedure cannot give reasonable parameter estimates. 

  \begin{solution}
    We have that $\x_k^\top \A \x_k \ge 0$ so that $\exp\left( -
    \x_k^\top \A \x_k \right) \le 1$. Hence $\exp\left( - \x_k^\top \A
    \x_k \right)$ is maximal if the elements of $\A$ are zero. This
    means that $J(\A)$ is maximal if $\A = 0$ whatever the observed
    data, which does not correspond to a meaningful estimation
    procedure (estimator).
    
  \end{solution}
  
\item Explain why maximum likelihood estimation is easy when the $x_i$
  are real numbers, i.e. $x_i \in \mathbb{R}$, while typically very
  difficult when the $x_i$ are binary, i.e.\ $x_i \in \{0,1\}$.

  \begin{solution}

    For maximum likelihood estimation, we needed to normalise the
    model by computing the partition function $Z(\thetab)$, which is
    defined as the sum/integral of $\exp(-\x^\top \A \x)$ over the
    domain of $\x$.

    When the $x_i$ are numbers, we can here obtain an analytical
    expression for $Z(\thetab)$.  However, if the $x_i$ are binary, no
    such analytical expression is available and computing $Z(\thetab)$
    is then very costly.
    
  \end{solution}
  
\item Can we use score matching instead of maximum likelihood
  estimation to learn $\A$ if the $x_i$ are binary?
  
  \begin{solution}
    No, score matching cannot be used for binary data.
  \end{solution}

\end{exenumerate}

\chapter{Sampling and Monte Carlo Integration}
\minitoc

\ex{Importance sampling to estimate tail probabilities {\small \citep[based on][Exercise 3.5]{Robert2010}}}
\label{ex:importance-sampling-to-estimate-tail-probabilities}
We would like to use importance sampling to compute the probability
that a standard Gaussian random variable $x$ takes on a value larger
than $5$, i.e
\begin{equation}
  \Pr(x>5) = \int_{5}^\infty \frac{1}{\sqrt{2\pi}} \exp\left(-\frac{x^2}{2}\right) dx
\end{equation}
We know that the probability equals
\begin{align}
  \Pr(x>5) & = 1-\int_{-\infty}^5 \frac{1}{\sqrt{2\pi}} \exp\left(-\frac{x^2}{2}\right) dx\\
  & = 1-\Phi(5)\\
  & \approx 2.87 \cdot 10^{-7}
\end{align}
where $\Phi(.)$ is the cumulative distribution function of a standard
normal random variable.

\begin{exenumerate}
\item With the indicator function $\mathbbm{1}_{x>5}(x)$, which
  equals one if $x$ is larger than $5$ and zero otherwise, we can
  write $\Pr(x>5)$ in form of the expectation
  \begin{align}
    \Pr(x>5) &= \E[ \mathbbm{1}_{x>5}(x) ],
  \end{align}
  where the expectation is taken with respect to the density
  $\mathcal{N}(x; 0, 1)$ of a standard normal random variable,
  \begin{equation}
    \mathcal{N}(x; 0, 1) = \frac{1}{\sqrt{2\pi}} \exp\left(-\frac{x^2}{2}\right).
  \end{equation}
  This suggests that we can approximate $\Pr(x>5)$ by a Monte Carlo average
  \begin{equation}
    \Pr(x>5) \approx \frac{1}{n} \sum_{i=1}^n \mathbbm{1}_{x>5}(x_i), \quad \quad \quad x_i \sim  \mathcal{N}(x; 0, 1).
  \end{equation}
  Explain why this approach does not work well.

  \begin{solution}
    In this  approach, we essentially  count how many times  the $x_i$
    are larger than  5. However, we know that the  chance that $x_i>5$
    is only  $2.87 \cdot 10^{-7}$. That  is, we only  get about one value above 5
    every 20 million simulations! The approach is thus very sample inefficient.

  \end{solution}

\item Another approach is to use importance sampling with an importance distribution $q(x)$ that is zero for $x<5$. We can then write $\Pr(x>5)$ as
  \begin{align}
    \Pr(x>5) & = \int_{5}^\infty \frac{1}{\sqrt{2\pi}} \exp\left(-\frac{x^2}{2}\right) dx\\
    & = \int_{5}^\infty \frac{1}{\sqrt{2\pi}} \exp\left(-\frac{x^2}{2}\right) \frac{q(x)}{q(x)} dx \\
    & = \E_{q(x)} \left[\frac{1}{\sqrt{2\pi}} \exp\left(-\frac{x^2}{2}\right) \frac{1}{q(x)}\right]
  \end{align}
  and estimate $\Pr(x>5)$ as a sample average.
  
  We here use an exponential distribution shifted by $5$ to the right. It has pdf
  \begin{equation}
    q(x) = \begin{cases}
      \exp(- (x-5)) & \text{if } x\ge 5\\
      0 & \text{otherwise}
    \end{cases}
  \end{equation}
  For background on the exponential distribution, see
  e.g.\ \url{https://en.wikipedia.org/wiki/Exponential_distribution}.
  
  Provide a formula that approximates $\Pr(x>5)$ as a sample average
  over $n$ samples $x_i \sim q(x)$.

  \begin{solution}
    The provided equation
    \begin{equation}
      \Pr(x>5)= \E_{q(x)} \left[\frac{1}{\sqrt{2\pi}} \exp\left(-\frac{x^2}{2}\right) \frac{1}{q(x)}\right]
    \end{equation}
    can be approximated as a sample average as follows:
    \begin{align}
      \Pr(x>5) &\approx \frac{1}{n} \sum_{i=1}^n \frac{1}{\sqrt{2\pi}} \exp\left(-\frac{x_i^2}{2}\right) \frac{1}{q(x_i)}\\
      & = \frac{1}{n} \sum_{i=1}^n  \frac{1}{\sqrt{2\pi}} \exp\left(-\frac{x_i^2}{2} + x-5\right)
    \end{align}
    with $x_i \sim q(x)$.
  \end{solution}
  
\item Numerically compute the importance estimate for various sample sizes $n
  \in [0, 1000]$. Plot the estimate against the sample size and
  compare with the ground truth value.

  \begin{solution}
    The following figure shows the importance sampling estimate as a
    function of the sample size (numbers do depend on the random seed
    used). We can see that we can obtain a good estimate with a few
    hundred samples already.
    \begin{center}
      \includegraphics[width=0.75 \textwidth]{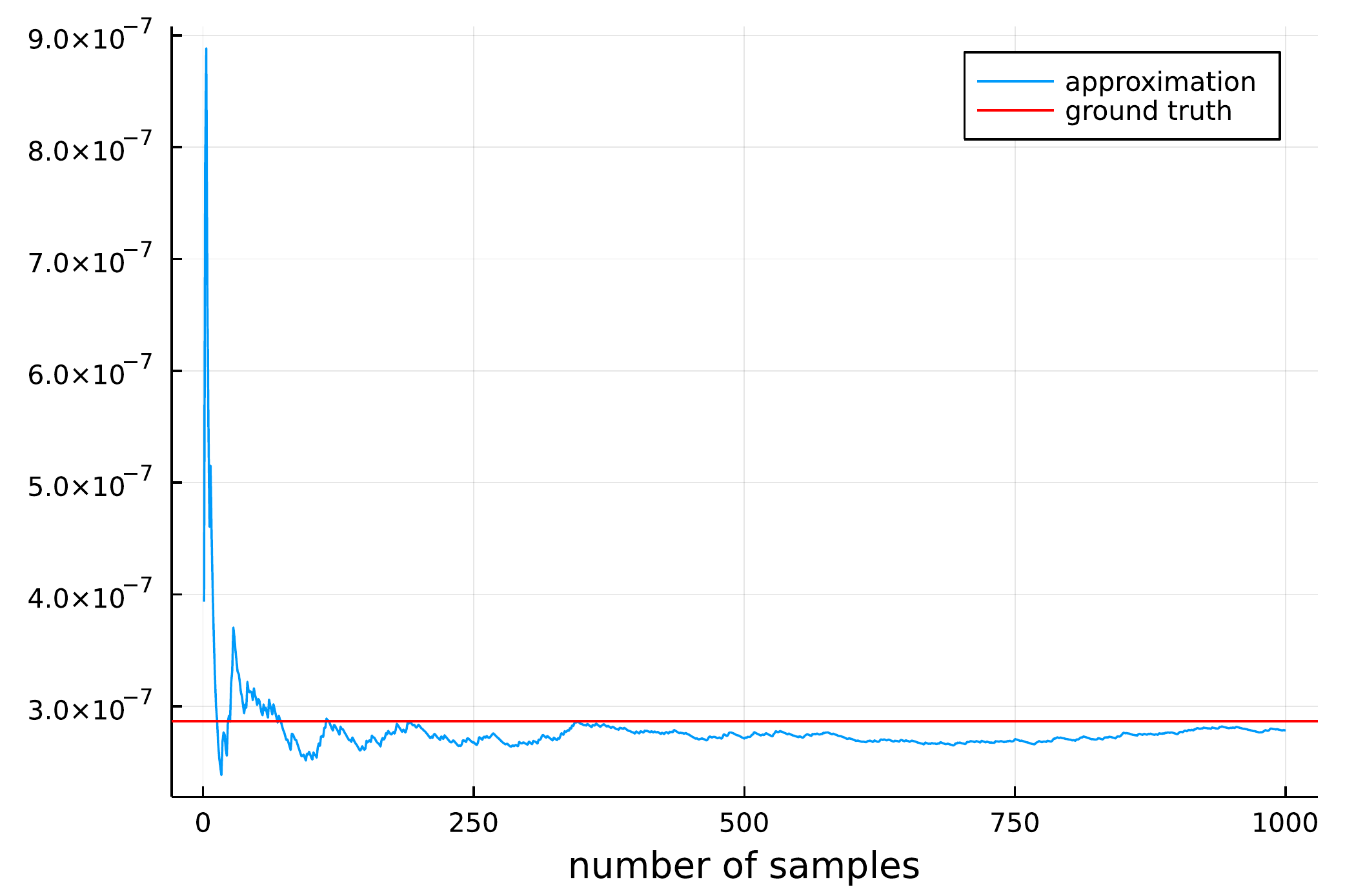}
    \end{center}

    Python code is as follows.
    \begin{lstlisting}
      import numpy as np
      from numpy.random import default_rng
      import matplotlib.pyplot as plt
      from scipy.stats import norm

      n = 1000
      alpha = 5

      # compute the tail probability
      p = 1-norm.cdf(alpha)

      # sample from the importance distribution
      rng = default_rng()
      vals = rng.exponential(scale=1, size=n) + alpha

      # compute average
      def w(x):
      return 1/np.sqrt(2*np.pi)*np.exp(-x**2/2+x-alpha)

      Ihat = np.cumsum(w(vals))/ np.arange(1, n+1)

      # plot
      plt.plot(Ihat)
      plt.axhline(y=p, color="r")
      plt.xlabel("number of samples")
\end{lstlisting}
    
    And code in Julia is:
    \begin{lstlisting}
      using Distributions
      using Plots
      using Statistics

      # compute the tail probability
      phi(x) = cdf(Normal(0,1),x)
      alpha = 5
      p = (1-phi(alpha))

      # sample from the importance distribution
      n = 1000
      exprv = Exponential(1)
      x = rand(exprv, n).+alpha;

      # compute the approximation
      w(x) = 1/sqrt(2*pi)*exp(-x^2/2+x-alpha)
      #w(x) = pdf(Normal(0,1),x)/pdf(exprv, x-alpha);  

      Ihat = zeros(length(x));
      for k in 1:length(x)
         Ihat[k] = mean(w.(x[1:k]));
      end

      # plot
      plt=plot(Ihat, label="approximation");
      hline!([p], color=:red, label="ground truth")
      xlabel!("number of samples")
\end{lstlisting}
    
  \end{solution}
  
\end{exenumerate}

\ex{Monte Carlo integration and importance sampling}
\label{ex:monte-carlo-integration-and-importance-sampling}
A standard Cauchy distribution has the density function (pdf)
\begin{equation}
  \label{eq:cauchy-pdf}
  p(x) = \frac{1}{\pi}\frac{1}{1+x^2}
\end{equation}
with $x \in \mathbb{R}$. A friend would like to verify that $\int p(x)
dx =1$ but doesn't quite know how to solve the integral
analytically. They thus use importance sampling and approximate the integral as
\begin{equation}
  \int p(x) dx \approx \frac{1}{n} \sum_{i=1}^n \frac{p(x_i)}{q(x_i)} \quad \quad x_i \sim q
\end{equation}
where $q$ is the density of the auxiliary/importance
distribution. Your friend chooses a standard normal density for $q$
and produces the following figure:
\begin{center}
  \includegraphics[width=0.65\textwidth]{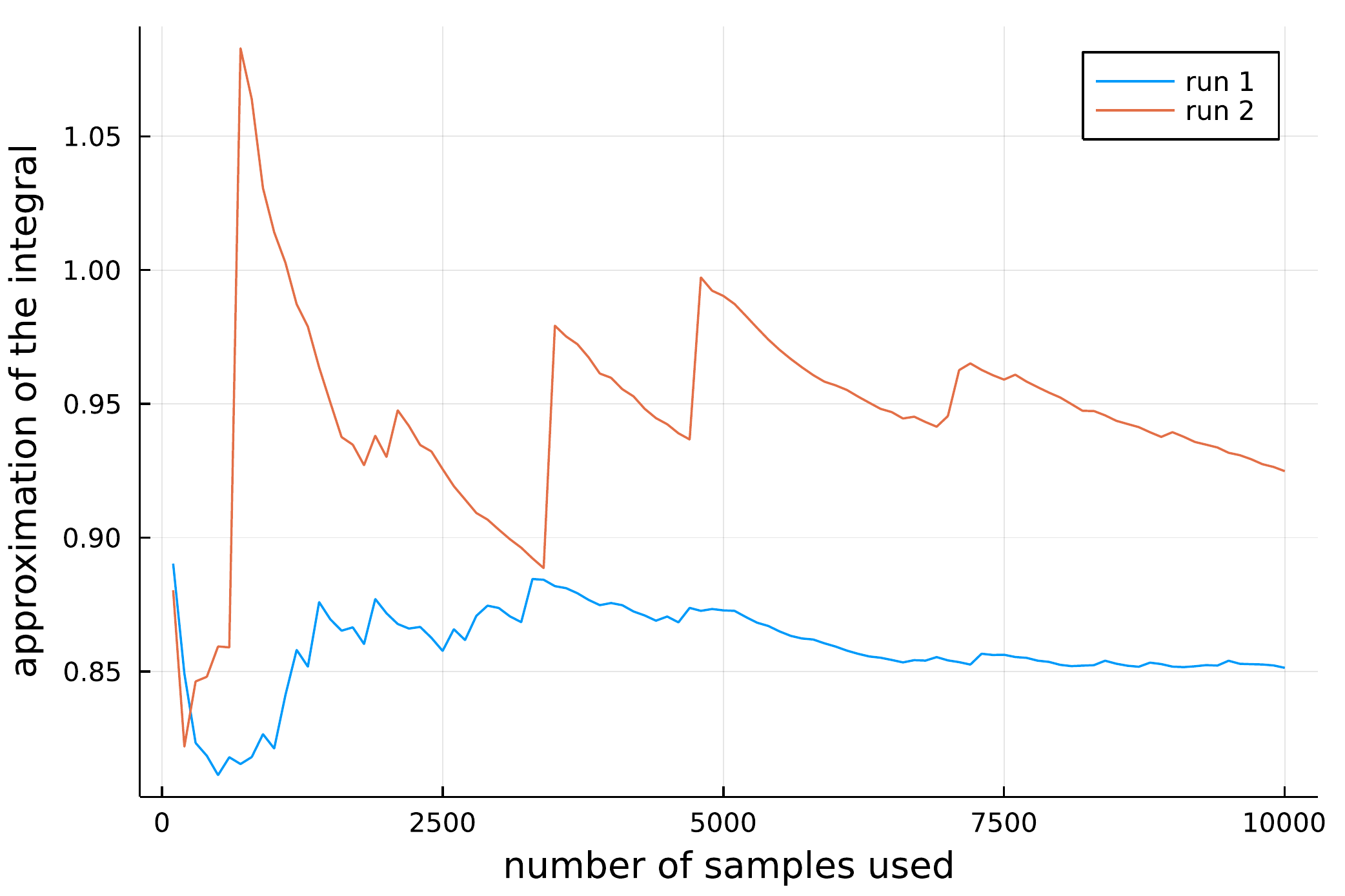}
\end{center}
The figure shows two independent runs. In each run, your friend
computes the approximation with different sample sizes by subsequently
including more and more $x_i$ in the approximation, so that, for example, the
approximation with $n=2000$ shares the first 1000 samples with the
approximation that uses $n=1000$.

Your friend is puzzled that the two runs give rather different results
(which are not equal to one), and also that within each run, the
estimate very much depends on the sample size. Explain these findings.

\begin{solution}
  While the estimate $\hat{I}_n$
  \begin{equation}
    \hat{I}_n =  \frac{1}{n} \sum_{i=1}^n \frac{p(x_i)}{q(x_i)}
  \end{equation}
  is unbiased by construction, we have to check whether its second
  moment is finite. Otherwise, we have an invalid estimator that
  behaves erratically in practice. The ratio $w(x)$ between $p(x)$ and
  $q(x)$ equals
  \begin{align}
    w(x) & = \frac{p(x)}{q(x)}\\
    & = \frac{\frac{1}{\pi}\frac{1}{1+x^2}}{ \frac{1}{\sqrt{2\pi}}\exp(-x^2/2)}
  \end{align}
  which can be simplified to
  \begin{equation}
    w(x)  = \frac{\sqrt{2\pi}\exp(x^2/2)}{\pi(1+x^2)}.
  \end{equation}
  The second moment of $w(x)$ under $q(x)$ thus is
  \begin{align}
    \E_{q(x)}\left[w(x)^2\right] & = \int_{-\infty}^\infty \frac{2\pi}{\pi^2} \frac{\exp(x^2)}{(1+x^2)^2} q(x) dx\\
    & = \int_{-\infty}^\infty \frac{2\pi}{\pi^2} \frac{\exp(x^2)}{(1+x^2)^2} \frac{1}{\sqrt{2\pi}} \exp(-x^2/2) dx\\
    & \propto \int_{-\infty}^\infty \frac{\exp(x^2/2)}{(1+x^2)^2} dx
  \end{align}
  The exponential function grows more quickly than any polynomial so
  that the integral becomes arbitrarily large. Hence, the second
  moment (and the variance) of $\hat{I}_n$ is unbounded, which
  explains the erratic behaviour of the curves in the plot.

  A less formal but quicker way to see that, for this problem, a
  standard normal is a poor choice of an importance distribution is to
  note that its density decays more quickly than the Cauchy pdf in
  \eqref{eq:cauchy-pdf}, which means that the standard normal pdf is
  ``small'' when the Cauchy pdf is still ``large'' (see Figure
  \ref{fig:gauss-cauchy-logpdf}). This leads to large variance of the
  estimate. The overall conclusion is that the integral $\int p(x) dx$
  should not be approximated with importance sampling with a Gaussian
  importance distribution.

  \begin{figure}[h!]
    \centering
    \includegraphics[width=0.6\textwidth]{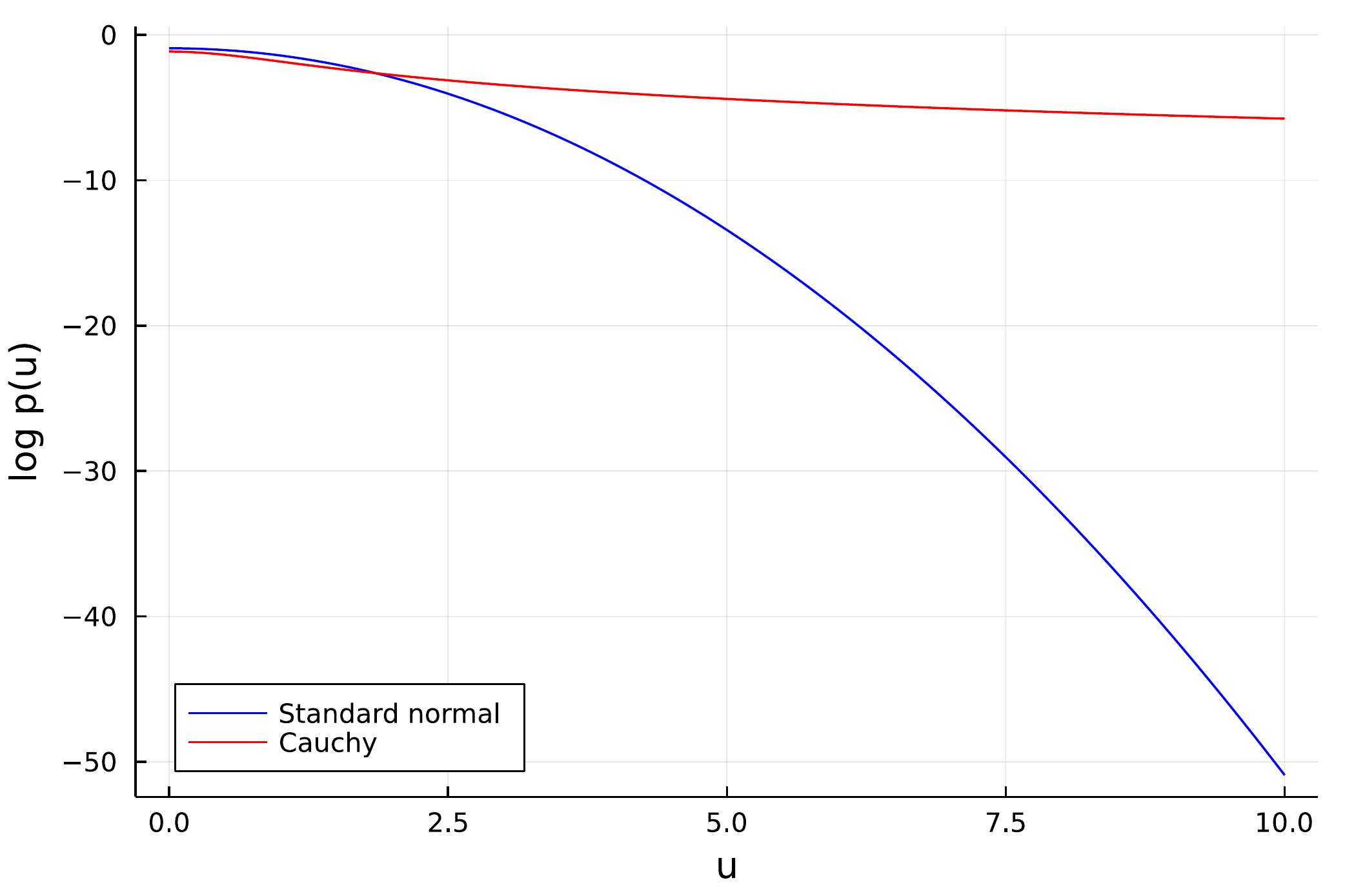}
    \caption{\label{fig:gauss-cauchy-logpdf} Exercise
      \ref{ex:monte-carlo-integration-and-importance-sampling}. Comparison
      of the log pdf of a standard normal (blue) and the Cauchy random
      variable (red) for positive inputs. The Cauchy pdf has much
      heavier tails than a Gaussian so that the Gaussian pdf is
      already ``small'' when the Cauchy pdf is still ``large''.}
  \end{figure}
\end{solution}

\ex{Inverse transform sampling}
\label{ex:ex1}
The cumulative distribution function (cdf) $F_x(\alpha)$ of a (continuous or
discrete) random variable $x$ indicates the probability that $x$ takes
on values smaller or equal to $\alpha$,
\begin{equation}
  F_x(\alpha) = \Pr( x \le \alpha).
\end{equation}
For continuous random variables, the cdf is defined via the integral
\begin{equation}
  F_x(\alpha) = \int_{-\infty}^{\alpha} p_x(u) \ud u,
\end{equation}
where $p_x$ denotes the pdf of the random variable $x$ ($u$ is here a
dummy variable). Note that $F_x$ maps the domain of $x$ to the interval $[0,1]$. For simplicity, we here assume that $F_x$ is invertible.

For a continuous random variable $x$ with cdf $F_x$ show that the random variable $y = F_x(x)$ is uniformly distributed on $[0,1]$. 

Importantly, this implies that for a random variable $y$ which is
uniformly distributed on $[0,1]$, the transformed random variable
$F_x^{-1}(y)$ has cdf $F_x$. This gives rise to a method called
``inverse transform sampling'' to generate $n$ iid samples
of a random variable $x$ with cdf $F_x$.
Given a target cdf $F_x$, the method consists of:
\begin{itemize}
\item calculating the inverse $F_x^{-1}$
\item sampling $n$ iid random variables uniformly distributed on $[0,1]$: $y^{(i)} \sim \mathcal{U}(0,1)$, $i=1, \ldots, n$.
\item transforming each sample by $F_x^{-1}$: $x^{(i)} = F_x^{-1}(y^{(i)})$, $i=1, \ldots, n$.
\end{itemize}
By construction of the method, the $x^{(i)}$ are $n$ iid samples of $x$.
  
  \begin{solution}
    We start with the cumulative distribution function (cdf) $F_y$ for $y$,
    \begin{align}
      F_y(\beta) & = \Pr( y \le \beta).
    \end{align}
    Since $F_x(x)$ maps $x$ to $[0,1]$, $F_y(\beta)$ is zero for $\beta <0$ and one for $\beta>1$. We next consider $\beta \in [0,1]$.

   Let $\alpha$ be the value of $x$ that $F_x$ maps to $\beta$, i.e.\ $F_x(\alpha) = \beta$, which means $\alpha = F_x^{-1}(\beta)$. Since $F_x$ is a non-decreasing function, we have
    \begin{equation}
     F_y(\beta) = \Pr( y \le \beta) = \Pr( F_x(x) \le \beta) = \Pr(x \le F_x^{-1}(\beta)) = \Pr( x \le \alpha) = F_x(\alpha).
    \end{equation}
    Since $\alpha = F_x^{-1}(\beta)$ we obtain
    \begin{equation}
      F_y(\beta)  =  F_x( F_x^{-1}(\beta) ) = \beta
    \end{equation}
    The cdf $F_y$ is thus given by
    \begin{equation}
      F_y(\beta) = \begin{cases}
        0 & \text{if } \beta <0\\
        \beta & \text{if } \beta \in [0,1]\\
        1 & \text{if } \beta >1
      \end{cases}
    \end{equation}
    which is the cdf of a uniform random variable on $[0,1]$. Hence $y=F_x(x)$ is uniformly distributed on $[0,1]$.
    
  \end{solution}

\ex{Sampling from the exponential distribution}
The exponential distribution has the density
\begin{equation}
  p(x; \lambda) = \begin{cases} \lambda \exp(-\lambda x) & x \ge 0\\ 0
    & x<0,
  \end{cases}
  \end{equation}
where $\lambda$ is a parameter of the distribution. Use inverse transform sampling to generate $n$ iid samples from $p(x; \lambda)$.

\begin{solution}
  We first compute the cumulative distribution function.
  \begin{align}
    F_x(\alpha) & = \Pr(x\le \alpha)\\
    & = \int_0^\alpha  \lambda \exp(-\lambda x)\\
    & = -\exp(-\lambda)\big|_0^\alpha\\
    & = 1-\exp(-\lambda \alpha)
  \end{align}
  It's inverse is obtained by solving
  \begin{equation}
    y = 1-\exp(-\lambda x)
  \end{equation}
  for $x$, which gives:
  \begin{align}
    \exp(-\lambda x) &= 1-y\\
    -\lambda x &= \log(1-y)\\
    x &= \frac{-\log(1-y)}{\lambda}
  \end{align}
  To generate samples $x^{(i)}  \sim p(x; \lambda)$, we thus first sample $y^{(i)} \sim U(0,1)$, and then set
  \begin{equation}
    x^{(i)} = \frac{-\log(1-y^{(i)})}{\lambda}.
  \end{equation}
 
  Inverse transform sampling can be used to generate samples
  from many standard distributions. For example, it allows one
  to generate Gaussian random variables from uniformly
  distributed random variables. The method is called the
  Box-Muller transform, see
  e.g. \url{https://en.wikipedia.org/wiki/Box-Muller_transform}. How
  to generate the required samples from the uniform distribution is a
  research field on its own, see
  e.g. \url{https://en.wikipedia.org/wiki/Random_number_generation}
  and \citep[Chapter 3]{Owen2013}.

\end{solution}

\ex{Sampling from a Laplace distribution}
\label{ex:sampling-from-a-Laplace-distribution}
A Laplace random variable $x$ of mean zero and variance one has the density $p(x)$
\begin{equation}
  p(x) = \frac{1}{\sqrt{2}} \exp\left(-\sqrt{2}|x|\right) \quad \quad x \in \mathbb{R}.  
\end{equation}
Use inverse transform sampling to generate $n$ iid samples from $x$.

\begin{solution}
  The main task is to compute the cumulative distribution function (cdf) $F_x$ of $x$ and its inverse. The cdf is by definition
  \begin{align}
    F_x(\alpha) & = \int_{-\infty}^\alpha  \frac{1}{\sqrt{2}} \exp\left(-\sqrt{2}|u|\right) \ud u.
  \end{align}
  We first consider the case where $\alpha \le 0$. Since $-|u| = u$ for $u\le 0$, we have 
    \begin{align}
      F_x(\alpha) & = \int_{-\infty}^\alpha  \frac{1}{\sqrt{2}} \exp\left(\sqrt{2}u\right) \ud u   \\
      & = \frac{1}{2} \exp\left(\sqrt{2}u\right) \bigg |_{-\infty}^\alpha\\
      & =  \frac{1}{2} \exp\left(\sqrt{2} \alpha\right).
    \end{align}
  For $\alpha >0$, we have
   \begin{align}
     F_x(\alpha) & = \int_{-\infty}^\alpha  \frac{1}{\sqrt{2}} \exp\left(-\sqrt{2}|u|\right) \ud u   \\
     &= 1- \int_{\alpha}^\infty  \frac{1}{\sqrt{2}} \exp\left(-\sqrt{2}|u|\right) \ud u        
   \end{align}
  where we have used the fact that the pdf has to integrate to one. For values of $u>0$, $-|u| = -u$, so that
   \begin{align}
     F_x(\alpha) & = 1 - \int_{\alpha}^\infty  \frac{1}{\sqrt{2}} \exp\left(-\sqrt{2} u \right) \ud u  \\
     &= 1 + \frac{1}{2}  \exp\left(-\sqrt{2} u \right) \bigg |_{\alpha}^{\infty}\\
     &= 1 - \frac{1}{2}  \exp\left(-\sqrt{2} \alpha \right).
   \end{align}
  In total, for $\alpha \in \mathbb{R}$, we thus have
   \begin{equation}
     F_x(\alpha) = \begin{cases}
       \frac{1}{2} \exp\left(\sqrt{2} \alpha\right) & \text{if } \alpha \le 0\\
       1 - \frac{1}{2}  \exp\left(-\sqrt{2} \alpha\right) & \text{if } \alpha>0
     \end{cases}
   \end{equation}
  Figure \ref{fig:Fx} visualises $F_x(\alpha)$.
   \begin{figure}[h!]
     \centering
     \includegraphics[width = 0.6 \textwidth]{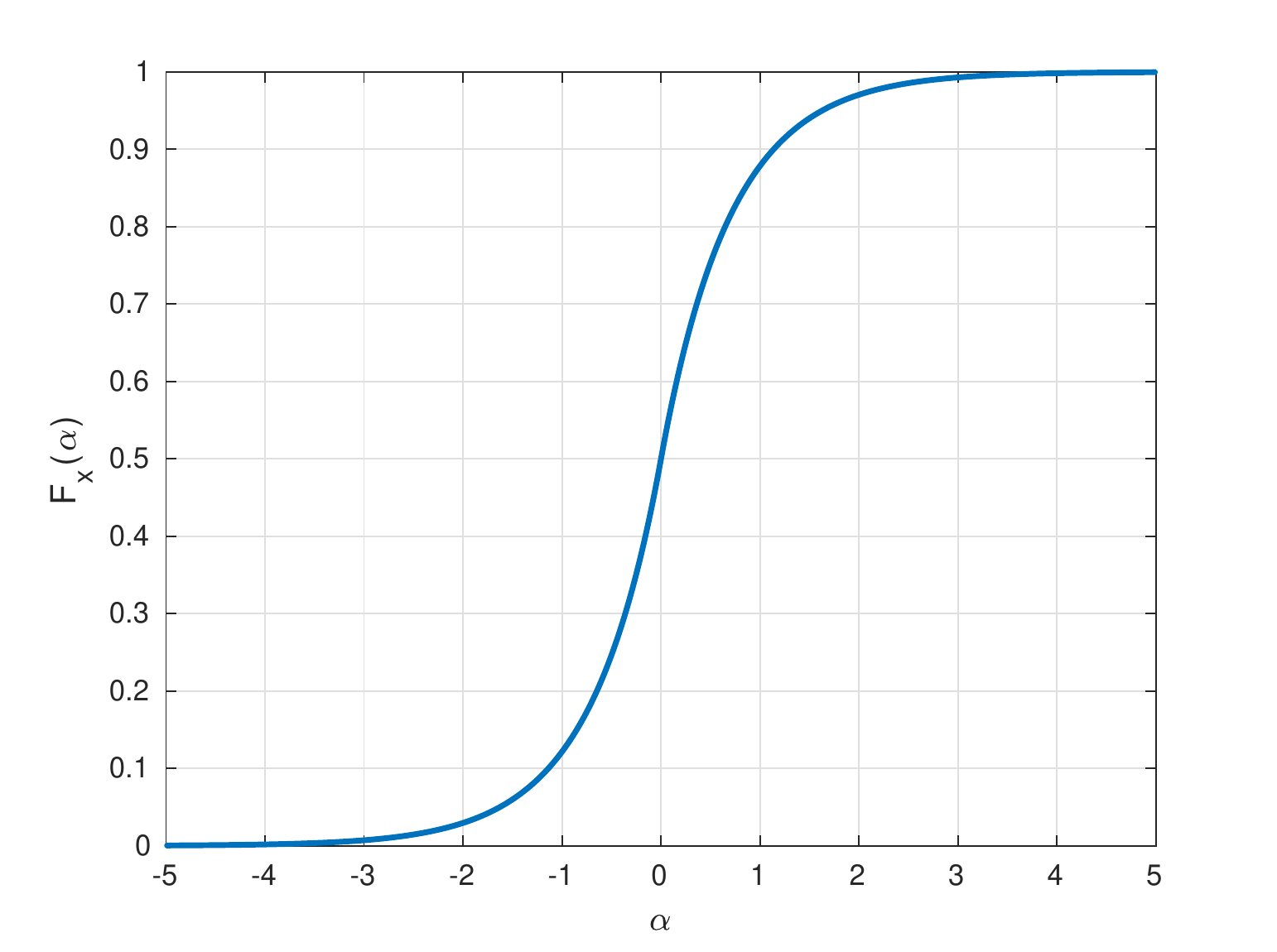}
     \caption{\label{fig:Fx} The cumulative distribution function $F_x(\alpha)$ for a Laplace distributed random variable.}
   \end{figure}
   
  As the figure suggests, there is a unique inverse to $y = F_x(\alpha)$. For $y \le 1/2$, we have
   \begin{align}
     y & =  \frac{1}{2} \exp\left(\sqrt{2} \alpha\right)\\
     \log(2y) & = \sqrt{2} \alpha\\
     \alpha & = \frac{1}{\sqrt{2}} \log (2y)
   \end{align}
  For $y > 1/2$, we have
   \begin{align}
     y & =  1 - \frac{1}{2}  \exp\left(-\sqrt{2} \alpha \right)\\
     -y & =  -1 + \frac{1}{2}  \exp\left(-\sqrt{2} \alpha \right)\\
     1-y &= \frac{1}{2}  \exp\left(-\sqrt{2} \alpha \right) \\
     \log( 2-2y) & = -\sqrt{2} \alpha \\
     \alpha & = -\frac{1}{\sqrt{2}} \log(2-2y)
   \end{align}
  The function $y \mapsto g(y)$ that occurs in the logarithm in both cases is
   \begin{equation}
     g(y) = \begin{cases}
       2 y & \text{if } y\le \frac{1}{2}\\
       2-2y & \text{if } y> \frac{1}{2}
     \end{cases}.
   \end{equation}
  It is shown below and can be written more compactly as $g(y) = 1-2|y-1/2|$.
    \begin{figure}[h]
     \centering
     \includegraphics[width = 0.6 \textwidth]{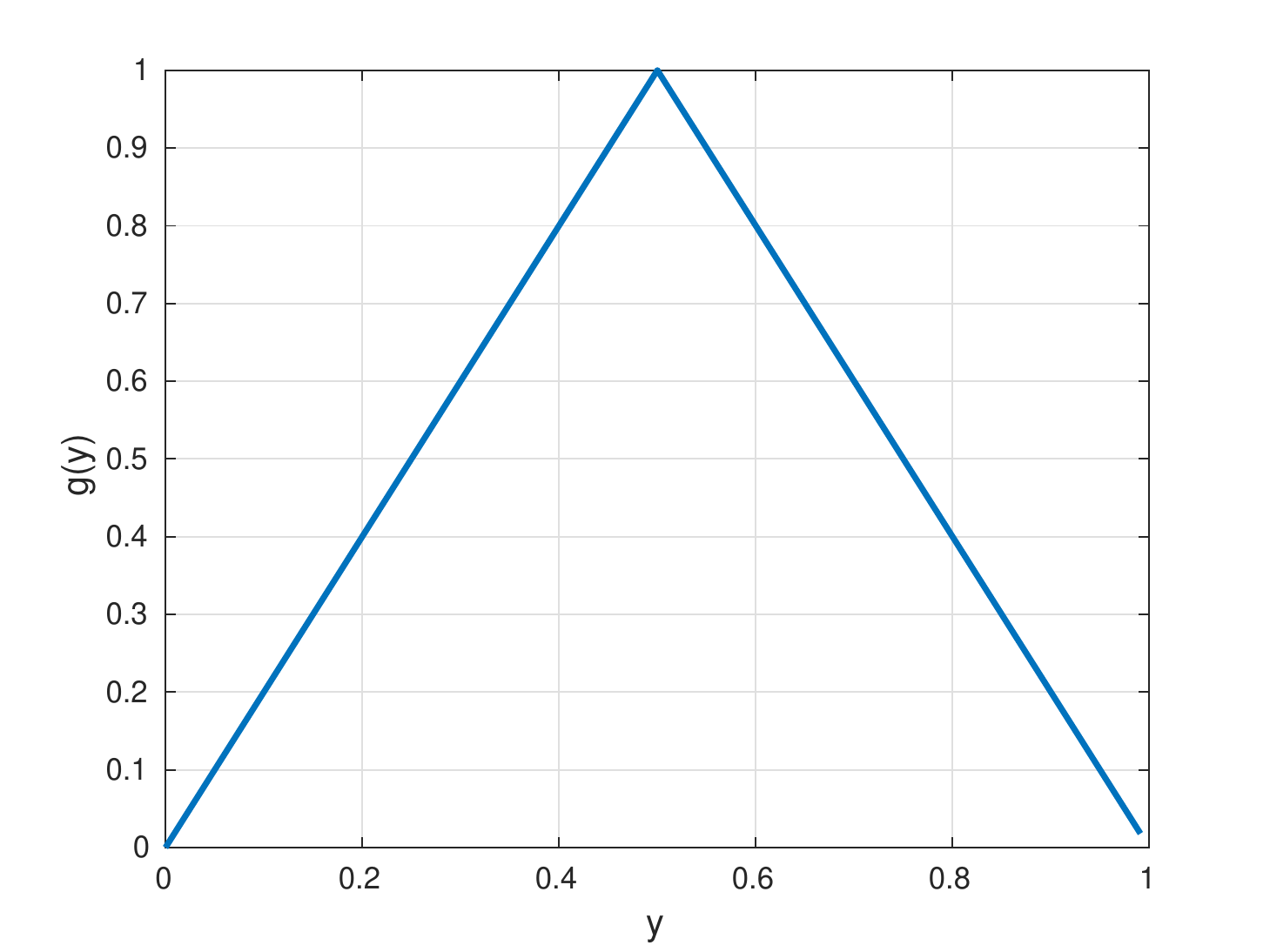}
    \end{figure}

  We thus can write the inverse $F_x^{-1}(y)$ of the cdf $y=F_x(\alpha)$ as
    \begin{equation}
      F_x^{-1}(y) = -\text{sign}\left(y-\frac{1}{2}\right) \frac{1}{\sqrt{2}} \log\left[  1-2\big|y-\frac{1}{2}\big| \right].
    \end{equation}
  To generate $n$ iid samples from $x$, we first generate $n$
  iid samples $y^{(i)}$ that are uniformly distributed on $[0,1]$,
  and then compute for each $F_x^{-1}(y^{(i)})$. The properties of
  inverse transform sampling guarantee that the $x^{(i)}$,
    \begin{equation}
      x^{(i)} = F_x^{-1}(y^{(i)}),
    \end{equation}
  are independent and Laplace distributed.
  \end{solution}

\ex{Rejection sampling {\small \citep[based on][Exercise 2.8]{Robert2010}}}
\label{ex:rejection-sampling}
Most compute environments provide functions to sample from a standard
normal distribution. Popular algorithms include the Box-Muller
transform, see
e.g. \url{https://en.wikipedia.org/wiki/Box-Muller_transform}. We here
use rejection sampling to sample from a standard normal distribution
with density $p(x)$ using a Laplace distribution as our
proposal/auxiliary distribution.

The density $q(x)$ of a zero-mean Laplace distribution with variance
$2b^2$ is
\begin{equation}
  q(x;b) = \frac{1}{2b}\exp\left(-\frac{|x|}{b}\right).
\end{equation}
We can sample from it by sampling a Laplace variable with variance 1
as in Exercise \ref{ex:sampling-from-a-Laplace-distribution} and then
scaling the sample by $\sqrt{2}b$.

Rejection sampling then repeats the following steps:
\begin{itemize}
  \item Generate $x \sim q(x; b)$
  \item Accept $x$ with probability $f(x) = \frac{1}{M}
    \frac{p(x)}{q(x)}$, i.e. generate $u \sim U(0,1)$ and accept $x$
    if $u \le f(x)$.
\end{itemize}

\begin{exenumerate}
\item Compute the ratio $M(b) = \max_x \frac{p(x)}{q(x; b)}$.
  \begin{solution}
    By the definitions of the pdf $p(x)$ of a standard normal and the pdf $q(x; b)$ of the Laplace distribution, we have
    \begin{align}
      \frac{p(x)}{q(x;b)} &= \frac{ \frac{1}{\sqrt{2\pi}} \exp(-x^2/2)}{\frac{1}{2b}\exp(-|x|/b)}\\
          & = \frac{2b}{\sqrt{2\pi}} \exp(-x^2/2 + |x|/b)
    \end{align}
    The ratio is symmetric in $x$. Moreover, since the exponential
    function is strictly increasing, we can find the maximiser of
    $-x^2/2+x/b$ for $x\ge 0$ to determine the maximiser of
    $M(b)$. With $g(x) = -x^2/2+x/b$, we have
    \begin{align}
      g'(x) &= -x + 1/b\\
      g''(x) &= -1
    \end{align}
    The critical point (for which the first derivative is zero) is $x = 1/b$ and since the second derivative is negative for all $x$, the point is a maximum. The maximal ratio $M(b)$ thus is
    \begin{align}
      M(b)  & = \frac{2b}{\sqrt{2\pi}} \exp\left(-x^2/2 + |x|/b\right)\Big|_{x=1/b}\\
      & = \frac{2b}{\sqrt{2\pi}} \exp\left(-1/(2b^2) + 1/b^2\right)\\
      & = \frac{2b}{\sqrt{2\pi}} \exp\left(1/(2b^2)\right)
    \end{align}
  \end{solution}

\item How should you choose $b$ to maximise the probability of acceptance?

  \begin{solution}
    The probability of acceptance is $1/M$. Hence to maximise it, we
    have to choose $b$ such that $M(b)$ is minimal. We compute the derivatives
    \begin{align}
      M'(b) & = \frac{2}{\sqrt{2\pi}} \exp(1/(2b^2)) - \frac{2b}{\sqrt{2\pi}} \exp(1/(2b^2)) b^{-3}\\
      & = \frac{2}{\sqrt{2\pi}} \exp(1/(2b^2)) - \frac{2}{\sqrt{2\pi}} \exp(1/(2b^2)) b^{-2}\\
      & = \frac{2}{\sqrt{2\pi}} \exp(1/(2b^2))(1-b^{-2})\\
      M''(b) & = -b^{-3}\frac{2}{\sqrt{2\pi}} \exp(1/(2b^2))(1-b^{-2})+2b^{-3}\frac{2}{\sqrt{2\pi}} \exp(1/(2b^2))\\
    \end{align}
    Setting the first derivative to zero gives
    \begin{align}
      \frac{2}{\sqrt{2\pi}} \exp(1/(2b^2)) &= \frac{2}{\sqrt{2\pi}} \exp(1/(2b^2)) b^{-2}\\
      1 & = b^{-2}
    \end{align}
    Hence the optimal $b=1$. The second derivative at $b=1$ is
    \begin{align}
      M''(1) & = 2 \frac{2}{\sqrt{2\pi}} \exp(1/2)
    \end{align}
    which is positive so that the $b=1$ is a minimum. The smallest value of $M$ thus is
    \begin{align}
      M(1) & =  \frac{2b}{\sqrt{2\pi}} \exp(1/(2b^2))\Big |_{b=1} \\
      & =  \frac{2}{\sqrt{2\pi}} \exp(1/2)\\
      & =  \sqrt{\frac{2 e}{\pi}}
    \end{align}
    where $e=\exp(1)$. The maximal acceptance probability thus is
    \begin{align}
      \frac{1}{\min_b M(b)} & = \sqrt{\frac{\pi}{2e}}\\
      & \approx 0.76
    \end{align}
    This means for each sample $x$ generated from $q(x; 1)$, there is
    chance of $0.76$ that it gets accepted. In other words, for each
    accepted sample, we need to generate $1/0.76 = 1.32$ samples from
    $q(x; 1)$.

    The variance of the Laplace distribution for $b=1$ equals 2. Hence
    the variance of the auxiliary distribution is larger (twice as
    large) as the variance of the distribution we would like to sample from. 
  \end{solution}
\item Assume you sample from $p(x_1, \ldots, x_d) = \prod_{i=1}^d
  p(x_i)$ using $q(x_1, \ldots, x_d) = \prod_{i=1}^d q(x_i; b)$ as
  auxiliary distribution without exploiting any independencies. How
  does the acceptance probability scale as a function of $d$? You may
  denote the acceptance probability in case of $d=1$ by $A$.

  \begin{solution}
    We have to determine the maximal ratio
    \begin{equation}
      M_d = \max_{x_1, \ldots, x_d} \frac{p(x_1, \ldots, x_d)}{q(x_1, \ldots, x_d)}
    \end{equation}
    Plugging-in the factorisation gives
    \begin{align}
      M_d &= \max_{x_1, \ldots, x_d} \prod_{i=1}^d \frac{p(x_i)}{q(x_i)}\\
      & = \prod_{i=1}^d \underbrace{\max_{x_i} \frac{p(x_i)}{q(x_i)}}_{M_1 = 1/A}\\
      & = \prod_{i=1}^d \frac{1}{A}\\
      & = \frac{1}{A^d}
    \end{align}
    Hence, the acceptance probability is
    \begin{equation}
      \frac{1}{M_d} = A^d
    \end{equation}
    Note that $A\le 1$ since it is a probability. This means that, unless $A=1$, we
    have an acceptance probability that decays exponentially in the
    number of dimensions if the target and auxiliary distributions
    factorise and we do not exploit the independencies.
  \end{solution}
\end{exenumerate}

\ex{Sampling from a restricted Boltzmann machine}

The restricted Boltzmann machine (RBM) is a model for binary variables
$\v =(v_1, \ldots, v_n)^\top$ and $\h=(h_1, \ldots, h_m)^\top$ which
asserts that the joint distribution of $(\v,\h)$ can be described by
the probability mass function
\begin{equation}
  p(\v,\h) \propto \exp\left( \v^\top \W \h + \a^\top\v + \b^\top\h \right),
\end{equation}
where $\W$ is a $n \times m$ matrix, and $\a$ and $\b$ vectors of size
$n$ and $m$, respectively. Both the $v_i$ and $h_i$ take
values in $\{0,1\}$. The $v_i$ are called the ``visibles'' variables
since they are assumed to be observed while the $h_i$ are the hidden
variables since it is assumed that we cannot measure them.

Explain how to use Gibbs sampling to generate samples from the
marginal $p(\v)$,
  \begin{equation}
    p(\v) = \frac{\sum_{\h}  \exp\left( \v^\top \W \h + \a^\top\v + \b^\top\h \right)}{\sum_{\h, \v}  \exp\left( \v^\top \W \h + \a^\top\v + \b^\top\h \right)},
  \end{equation}
for any given values of $\W$, $\a$, and $\b$.

\emph{Hint:} You may use that
\begin{align}
  p(\h | \v) &= \prod_{i=1}^m p(h_i | \v), &  p(h_i=1 | \v) & =  \frac{1}{ 1+  \exp\left(- \sum_j v_j W_{ji} - b_i \right)}, \\
  p(\v | \h) &= \prod_{i=1}^n p(v_i | \h), &  p(v_i = 1 | \h) &= \frac{1}{1+\exp\left(-\sum_j W_{ij} h_j-a_i\right)}.
\end{align}

\begin{solution}
In order to generate samples $\v^{(k)}$ from $p(\v)$ we generate
samples $(\v^{(k)}, \h^{(k)})$ from $p(\v, \h)$ and then ignore the
$\h^{(k)}$.

Gibbs sampling is a MCMC method to produce a sequence of samples
$\x^{(1)}, \x^{(2)}, \x^{(3)}, \ldots$ that follow a pdf/pmf $p(\x)$
(if the chain is run long enough). Assuming that $\x$ is
$d$-dimensional, we generate the next sample $\x^{(k+1)}$ in the
sequence from the previous sample $\x^{(k)}$ by:
\begin{enumerate}
  \item picking (randomly) an index $i \in \{1, \ldots, d \}$
  \item sampling $x_i^{(k+1)}$ from $p(x_i \mid \x^{(k)}_{\backslash
    i})$ where $\x^{(k)}_{\backslash i}$ is vector $\x$ with $x_i$
    removed, i.e. $\x^{(k)}_{\backslash i}
    =({x}_1^{(k)},\ldots,{x}_{i-1}^{(k)},{x}_{i+1}^{(k)},\ldots,{x}_d^{(k)})$
  \item setting $\x^{(k+1)} =({x}_1^{(k)},\ldots,{x}_{i-1}^{(k)},x_i^{(k+1)},{x}_{i+1}^{(k)},\ldots,{x}_d^{(k)})$.
\end{enumerate}
For the RBM, the tuple $(\h,\v)$ corresponds to $\x$ so that a $x_i$
in the above steps can either be a hidden variable or a visible. Hence
\begin{equation}
  p(x_i \mid \x_{\backslash i}) = \begin{cases}
    p(h_i \mid \h_{\backslash i}, \v) & \text{if $x_i$ is a hidden variable $h_i$}\\
    p(v_i \mid \v_{\backslash i}, \h) & \text{if $x_i$ is a visible variable $v_i$}
  \end{cases}
\end{equation}
($\h_{\backslash i}$ denotes the vector $\h$ with element $h_i$
removed, and equivalently for $\v_{\backslash i}$)

To compute the conditionals on the right hand side, we use the hint:
\begin{align}
  p(\h | \v) &= \prod_{i=1}^m p(h_i | \v), &  p(h_i=1 | \v) & =  \frac{1}{ 1+  \exp\left(- \sum_j v_j W_{ji} - b_i \right)}, \\
  p(\v | \h) &= \prod_{i=1}^n p(v_i | \h), &  p(v_i = 1 | \h) &= \frac{1}{1+\exp\left(-\sum_j W_{ij} h_j-a_i\right)}.
\end{align}
Given the independencies between the hiddens given the visibles and vice versa, we have
\begin{align}
  p(h_i \mid \h_{\backslash i}, \v) & = p(h_i \mid \v) &   p(v_i \mid \v_{\backslash i}, \h) & = p(v_i \mid \h)
\end{align}
so that the expressions for $p(h_i=1 | \v)$ and $p(v_i = 1 | \h)$ allow us to implement the Gibbs sampler.

Given the independencies, it makes further sense to sample the $\h$
and $\v$ variables in blocks: first we sample all the $h_i$ given
$\v$, and then all the $v_i$ given the $\h$ (or vice versa). This is
also known as block Gibbs sampling.

In summary, given a sample $(\h^{(k)},\v^{(k)})$, we thus generate the next sample
$(\h^{(k+1)},\v^{(k+1)})$ in the sequence as follows:
\begin{itemize}
\item For all $h_i$, $i=1, \ldots, m$:
  \begin{itemize}
  \item compute $p^h_i = p(h_i=1 | \v^{(k)})$
  \item sample $u_i$ from a uniform distribution on $[0,1]$ and set $h^{(k+1)}_i$ to 1 if $u_i \le p^h_i$.
  \end{itemize}
\item For all $v_i$, $i=1, \ldots, n$:
  \begin{itemize}
  \item compute $p^v_i = p(v_i=1 | \h^{(k+1)})$
  \item sample $u_i$ from a uniform distribution on $[0,1]$ and set $v^{(k+1)}_i$ to 1 if $u_i \le p^v_i$.
  \end{itemize}
\end{itemize}
As final step, after sampling $S$ pairs $(\h^{(k)},\v^{(k)})$, $k=1,
\ldots, S$, the set of visibles $\v^{(k)}$ form samples from the
marginal $p(\v)$.
\end{solution}

\begin{figure}[!h]
	\centering
	\subfloat[True density]{
	  \includegraphics[width=0.45\textwidth]{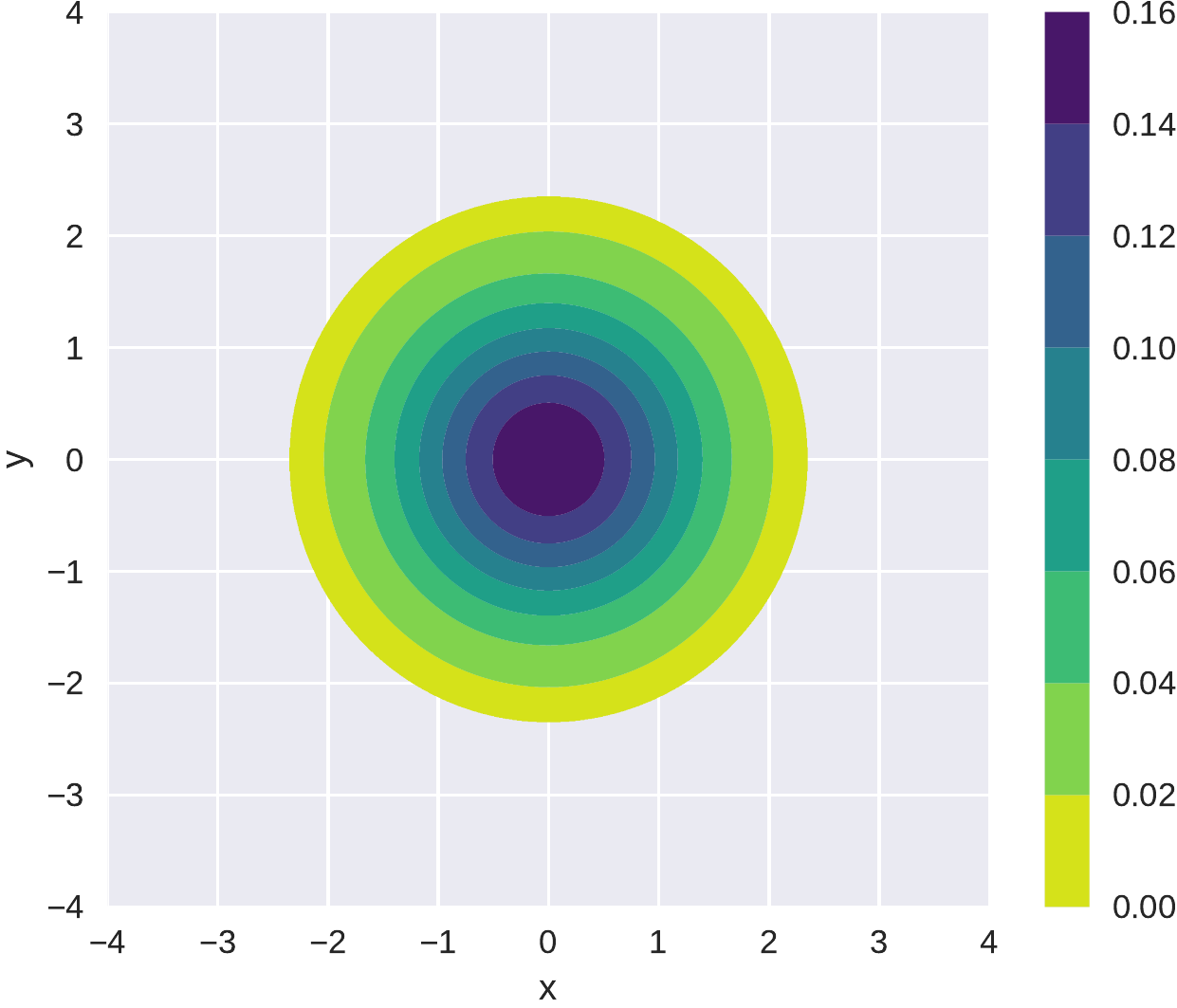}
	  \label{sfig:density}}
	\subfloat[Density represented by $10,000$ samples.]{
	  \includegraphics[width=0.45\textwidth]{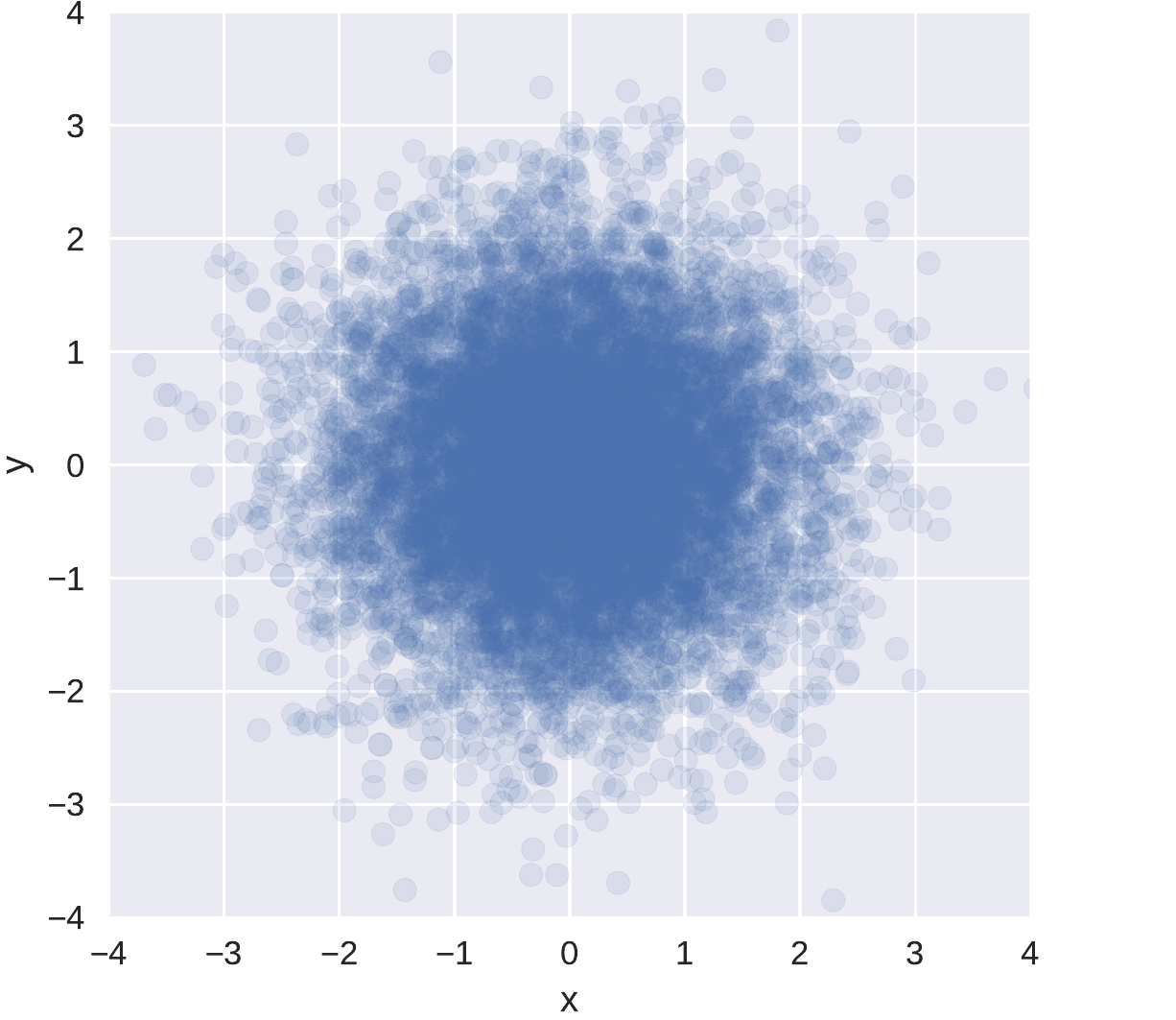}
	  \label{sfig:scatter}
        }
	\caption{Density and samples from $p(x,y) = \normal(x; 0,1)\normal(y;0,1)$.}
	\label{fig:mcmc_approx}
\end{figure}

\ex{Basic Markov chain Monte Carlo inference} \label{q:basic-mcmc}
This exercise is on sampling and approximate inference by Markov
chain Monte Carlo (MCMC). MCMC can be used to obtain
samples from a probability distribution, e.g. a posterior
distribution. The samples approximately represent the distribution, as
illustrated in Figure~\ref{fig:mcmc_approx}, and can be used to
approximate expectations.

We denote the density of a zero mean Gaussian with variance $\sigma^2$ by
$\normal(x; \mu, \sigma^2)$, i.e.\
\begin{equation}
  \normal(x; \mu, \sigma^2) = \frac{1}{\sqrt{2\pi \sigma^2}}\exp\left(-\frac{(x-\mu)^2}{2\sigma^2}\right)
\end{equation}

Consider a vector of $d$ random variables $\thetab = (\theta_1, \dots, \theta_d)$ and some observed data $\data$.
In many cases, we are interested in computing expectations under
the posterior distribution $p(\thetab \mid \data)$, e.g.\
\begin{equation}
  \E_{p(\thetab \mid \data)}\left[g(\thetab)\right] = \int g(\thetab) p(\thetab \mid \data) \mathrm{d} \thetab
\end{equation}
for some function $g(\thetab)$. If $d$ is small, e.g.\ $d \le3$, deterministic
numerical methods can be used to approximate the integral to high accuracy, see
e.g.\ \url{https://en.wikipedia.org/wiki/Numerical_integration}. But for higher
dimensions, these methods are generally not applicable any more. The
expectation, however, can be approximated as a sample average if we have samples
$\thetab^{(i)}$ from $p(\thetab \mid \data)$:
\begin{equation}
\E_{p(\thetab \mid \data)}\left[g(\thetab)\right] \approx \frac{1}{S}\sum_{i=1}^{S} g(\thetab^{(i)})
\end{equation}
Note that in MCMC methods, the samples $\thetab^{(1)}, \ldots, \thetab^{(S)}$ used in the above
approximation are typically not statistically independent.

Metropolis-Hastings is an MCMC algorithm that generates samples from a
distribution $p(\thetab)$, where $p(\thetab)$ can be any distribution
on the parameters (and not only posteriors). The algorithm is iterative
and at iteration $t$, it uses:
\begin{itemize}
\item a proposal distribution $q(\thetab; \thetab^{(t)})$, parametrised by the
	current state of the Markov chain, i.e.\ $\thetab^{(t)}$;
\item a function $p^*(\thetab)$, which is proportional to $p(\thetab)$. In other
  words, $p^*(\thetab)$ is unnormalised\footnote{We here follow the notation of \citet{Barber2012}; $\tilde{p}$ or $\phi$ are often to denote unnormalised models too.} and the normalised density
  $p(\thetab)$ is
  \begin{equation} p(\thetab) = \frac{p^*(\thetab)}{\int p^*(\thetab) \mathrm{d}\thetab}.
  \end{equation}
\end{itemize} 
For all tasks in this exercise, we work with a Gaussian proposal distribution
$q(\thetab; \thetab^{(t)})$, whose mean is the previous sample in the Markov
chain, and whose variance is $\epsilon^2$. That is, at iteration $t$ of our
Metropolis-Hastings algorithm,
\begin{equation}
  q(\thetab; \thetab^{(t-1)}) = \prod_{k=1}^d\normal(\theta_k; \theta_k^{(t-1)}, \epsilon^2).
  \label{eq:gauss-prop}
\end{equation}
When used with this proposal distribution, the algorithm is
called Random Walk Metropolis-Hastings algorithm.

\begin{exenumerate}
\item Read Section 27.4 of \citet{Barber2012} to familiarise yourself with the Metropolis-Hastings algorithm.

\item\label{qpt:mh} Write a function \lstinline|mh| implementing the
  Metropolis Hasting algorithm, as given in Algorithm 27.3 in
\citet{Barber2012}, using the Gaussian proposal distribution in \eqref{eq:gauss-prop} above. The
  function should take as arguments
\begin{itemize}
	\item \lstinline{p_star}: a function on $\thetab$ that is proportional to the density of interest $p(\thetab)$;
	\item \lstinline{param_init}: the initial sample --- a value for $\thetab$ from where the Markov chain starts;
	\item \lstinline{num_samples}: the number $S$ of samples to generate;
	\item \lstinline{vari}: the variance $\epsilon^2$ for the Gaussian proposal distribution $q$;
\end{itemize}
and return $[\thetab^{(1)}, \dots, \thetab^{(S)}]$ --- a list of $S$ samples from $p(\thetab) \propto p^*(\thetab)$. For example:
\begin{lstlisting}
  def mh(p_star, param_init, num_samples=5000, vari=1.0):
      # your code here
      return samples
\end{lstlisting}

\begin{solution}

  Below is a Python implementation.
\begin{lstlisting}
def mh(p_star, param_init, num_samples=5000, vari=1.0):
   x = []

   x_current = param_init
   for n in range(num_samples):

     # proposal
     x_proposed = multivariate_normal.rvs(mean=x_current, cov=vari)

     # MH step
     a = multivariate_normal.pdf(x_current, mean=x_proposed, cov=vari) * p_star(x_proposed)
     a = a / (multivariate_normal.pdf(x_proposed, mean=x_current, cov=vari) * p_star(x_current))

     # accept or not 
     if a >= 1:
       x_next = np.copy(x_proposed)
     elif uniform.rvs(0, 1) < a:
       x_next = np.copy(x_proposed)
     else:
       x_next = np.copy(x_current)

     # keep record
     x.append(x_next)
     x_current = x_next
	
   return x
\end{lstlisting}
As we are using a symmetrical proposal distribution, $q(\thetab \mid \thetab^*) = q(\thetab^* \mid \thetab)$, and one could simplify the algorithm by having $a = \frac{p^*(\thetab^*)}{p^*(\thetab)}$, where $\thetab$ is the current sample and $\thetab^*$ is the proposed sample.

In practice, it is desirable to implement the function in the log domain, to avoid numerical problems. That is, instead of $p^*$, \lstinline|mh| will accept as an argument $\log p^*$, and $a$ will be calculated as: 
$$a = (\log q(\thetab \mid \thetab^*) + \log p^*(\thetab^*)) - (\log q(\thetab^* \mid \thetab) + \log p^*(\thetab))$$
\end{solution}

\item \label{qpt:mh_test} Test your algorithm by sampling $5,000$
  samples from $p(x, y) = \normal(x; 0, 1)\normal(y; 0,
  1)$. Initialise at $(x=0, y=0)$ and use $\epsilon^2=1$.
Generate a scatter plot of the obtained samples. The plot should be
similar to Figure~\ref{sfig:scatter}. Highlight the first 20 samples
only. Do these 20 samples alone adequately approximate the true
density?

Sample another $5,000$ points from $p(x, y) = \normal(x; 0,
1)\normal(y; 0, 1)$ using \lstinline|mh| with $\epsilon^2=1$, but this
time initialise at $(x=7,y=7)$. Generate a scatter plot of the drawn
samples and highlight the first 20 samples. If everything went as
expected, your plot probably shows a ``trail'' of samples, starting at
$(x=7, y=7)$ and slowly approaching the region of space where most of
the probability mass is.

\begin{solution}

  \begin{figure}[!h]
    \captionsetup{type=figure}
    \centering
    \subfloat[\label{sfig:00} Starting the chain at $(0,0)$.]{
      \includegraphics[width=0.45\textwidth]{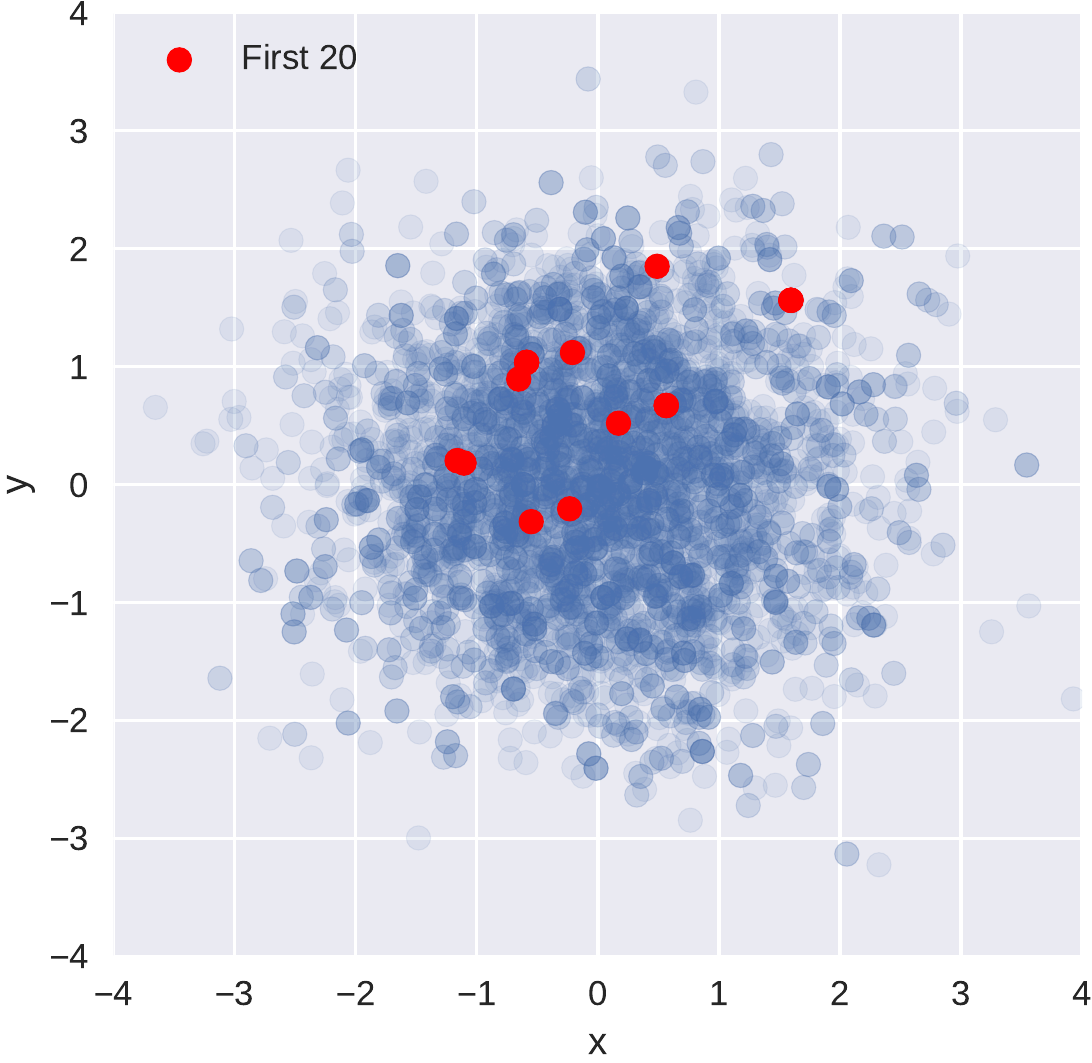}}
    \subfloat[\label{sfig:77}Starting the chain at $(7,7)$]{
      \includegraphics[width=0.45\textwidth]{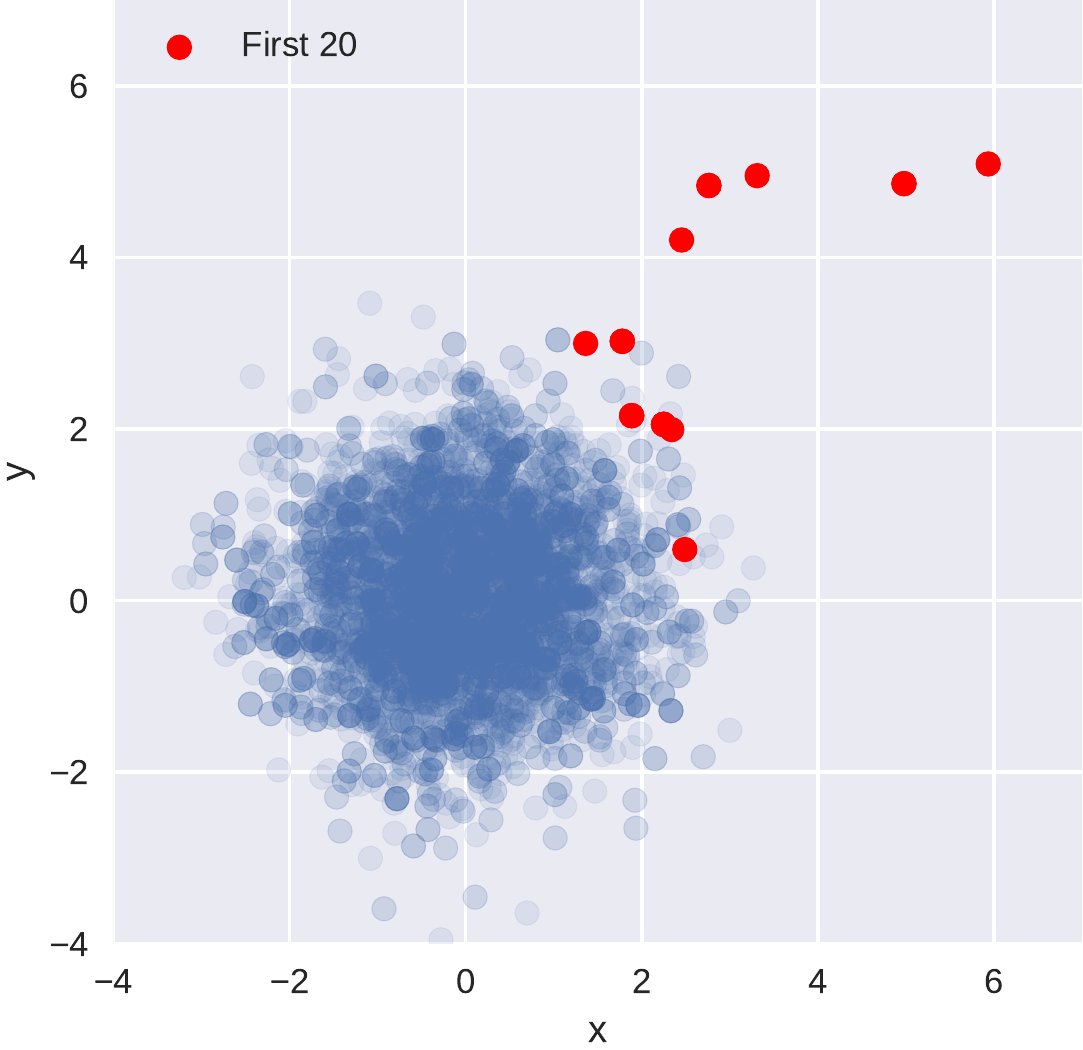}}
    \caption{\label{fig:initial} $5,000$ samples from $\normal(x; 0, 1)\normal(y; 0, 1)$ (blue), with the first $20$ samples highlighted (red). Drawn using Metropolis-Hastings with different starting points.}
  \end{figure}

Figure~\ref{fig:initial} shows the two scatter plots of draws from $\normal(x; 0, 1)\normal(y; 0, 1)$:
\begin{itemize}
	\item Figure~\ref{sfig:00} highlights the first 20 samples
          obtained by the chain when starting at $(x=0, y=0)$. They
          appear to be representative samples from the distribution,
          however, they are not enough to approximate the distribution
          on their own. This would mean that a sample average computed
          with 20 samples only would have high variance, i.e. its
          value would depend strongly on the values of the 20 samples
          used to compute the average.
	\item Figure~\ref{sfig:77} highlights the first 20 samples
          obtained by the chain when starting at $(x=7, y=7)$. One can
          clearly see the ``burn-in'' tail which slowly approaches the
          region where most of the probability mass is.
\end{itemize}

\end{solution}

\item In practice, we don't know where the distribution we wish to
  sample from has high density, so we typically initialise the Markov
  Chain somewhat arbitrarily, or at the maximum a-posterior (MAP) sample if
  available. The samples obtained in the beginning of the chain are
  typically discarded, as they are not considered to be representative
  of the target distribution. This initial period between
  initialisation and starting to collect samples is called
  ``warm-up'', or also ``burn-in''.

Extended your function \lstinline|mh| to include an additional warm-up
argument $W$, which specifies the number of MCMC steps taken before
starting to collect samples. Your function should still return a list
of $S$ samples as in \ref{qpt:mh}.

\begin{solution}

We can extend the \lstinline|mh| function with a warm-up argument by, for example, iterating for \lstinline|num_samples + warmup|
steps, and start recording samples only after the warm-up period:
\begin{lstlisting}
def mh(p_star, param_init, num_samples=5000, vari=1.0, warmup=0):
	x = []
	x_current = param_init
	for n in range(num_samples+warmup):
		...  # body same as before
	
		if n >= warmup: x.append(x_next)
		x_current = x_next
	
	return x
\end{lstlisting}
\end{solution}

\end{exenumerate}

\ex{Bayesian Poisson regression}\label{q:poisson-reg}
  Consider a Bayesian Poisson regression
  model, where outputs $y_n$ are generated from a Poisson distribution
  of rate $\exp(\alpha x_n + \beta)$, where the $x_n$ are the inputs (covariates), and $\alpha$ and $\beta$ the parameters of the regression model for which we assume a broad Gaussian prior:
  \begin{align}
    \alpha &\sim \normal(\alpha; 0, 100) \\
    \beta &\sim \normal(\beta; 0, 100)\\
    y_n &\sim \mathrm{Poisson}(y_n; \exp(\alpha x_n + \beta)) \quad \text{for } n = 1, \dots, N 
\end{align}
$\mathrm{Poisson}(y ;\lambda)$ denotes the probability mass
function of a Poisson random variable with rate $\lambda$,
\begin{equation}
  \mathrm{Poisson}(y; \lambda) = \frac{\lambda^y}{y!}\exp(-\lambda), \quad \quad y \in \{0, 1, 2, \ldots\}, \quad \lambda>0
\end{equation}
Consider $\data = \{(x_n, y_n)\}_{n=1}^N$ where $N=5$ and
\begin{align}
(x_1, \ldots, x_5) &= (-0.50519053, -0.17185719,  0.16147614,  0.49480947,  0.81509851)\\
  (y_1, \ldots, y_5) &= (1, 0, 2, 1, 2)
\end{align}

We are interested in computing the posterior density of the parameters
$(\alpha, \beta)$ given the data $\data$ above.

\begin{exenumerate}

\item Derive an expression for the unnormalised posterior density of
  $\alpha$ and $\beta$ given $\data$, i.e. a function $p^*$ of the
  parameters $\alpha$ and $\beta$ that is proportional to the
  posterior density $p(\alpha, \beta\mid \data)$, and which can thus
  be used as target density in the Metropolis Hastings algorithm.
  
  \begin{solution}
    By the product rule, the joint distribution described by the
    model, with $\data$ plugged in, is proportional to the posterior
    and hence can be taken as $p^*$:
    \begin{align}
      p^*(\alpha, \beta) &= p(\alpha, \beta, \{(x_n,y_n)\}_{n=1}^N ) \\ 
      &= \normal(\alpha; 0, 100)\normal(\beta; 0, 100)\prod_{n=1}^{N}\mathrm{Poisson}(y_n \mid \text{exp}(\alpha x_n + \beta))
    \end{align}
  \end{solution}
  
\item Implement the derived unnormalised posterior density $p^*$. If
  your coding environment provides an implementation of the above
  Poisson pmf, you may use it directly rather than implementing the
  pmf yourself.

  Use the Metropolis Hastings algorithm from Question
  \ref{q:basic-mcmc}\ref{qpt:mh_test} to draw $5,000$ samples from the
  posterior density $p(\alpha, \beta\mid \data)$. Set the
  hyperparameters of the Metropolis-Hastings algorithm to:
    \begin{itemize}
    \item \lstinline{param_init} $=(\alpha_{\mathrm{init}},\beta_{\mathrm{init}}) = (0,0)$,
    \item \lstinline{vari} $= 1$, and 
    \item number of warm-up steps $W = 1000$.
    \end{itemize}
    
    Plot the drawn samples with x-axis $\alpha$ and y-axis $\beta$ and report the posterior mean of $\alpha$ and $\beta$, as well as their correlation coefficient under the posterior.

    \begin{solution}
      A Python implementation is:
      
      \begin{lstlisting}
import numpy as np
from scipy.stats import multivariate_normal, norm, poisson, uniform

xx1 = np.array([-0.5051905265552105, -0.17185719322187715, 0.16147614011145617, 0.49480947344478954, 0.8150985069051909])
yy1 = np.array([1, 0, 2, 1, 2])
N1 = len(xx1)

def poisson_regression(params):
    a = params[0]
    b = params[1]
    # mean zero, standard deviation 10 == variance 100
    p = norm.pdf(a, loc=0, scale=10) * norm.pdf(b, loc=0, scale=10) 
    for n in range(N1):
        p = p * poisson.pmf(yy1[n], np.exp(a * xx1[n] + b)) 

    return p

# sample
S = 5000
samples = np.array(mh(poisson_regression, np.array([0, 0]), num_samples=S, vari=1.0, warmup=1000))
\end{lstlisting}

A scatter plot showing $5,000$ samples from the posterior is shown on
Figure~\ref{fig:easy}. The posterior mean of $\alpha$ is 0.84, the
posterior mean of $\beta$ is -0.2, and posterior correlation
coefficient is -0.63. Note that the numerical values are sample-specific.
\begin{figure}
  \centering
  \includegraphics[width=0.5\textwidth]{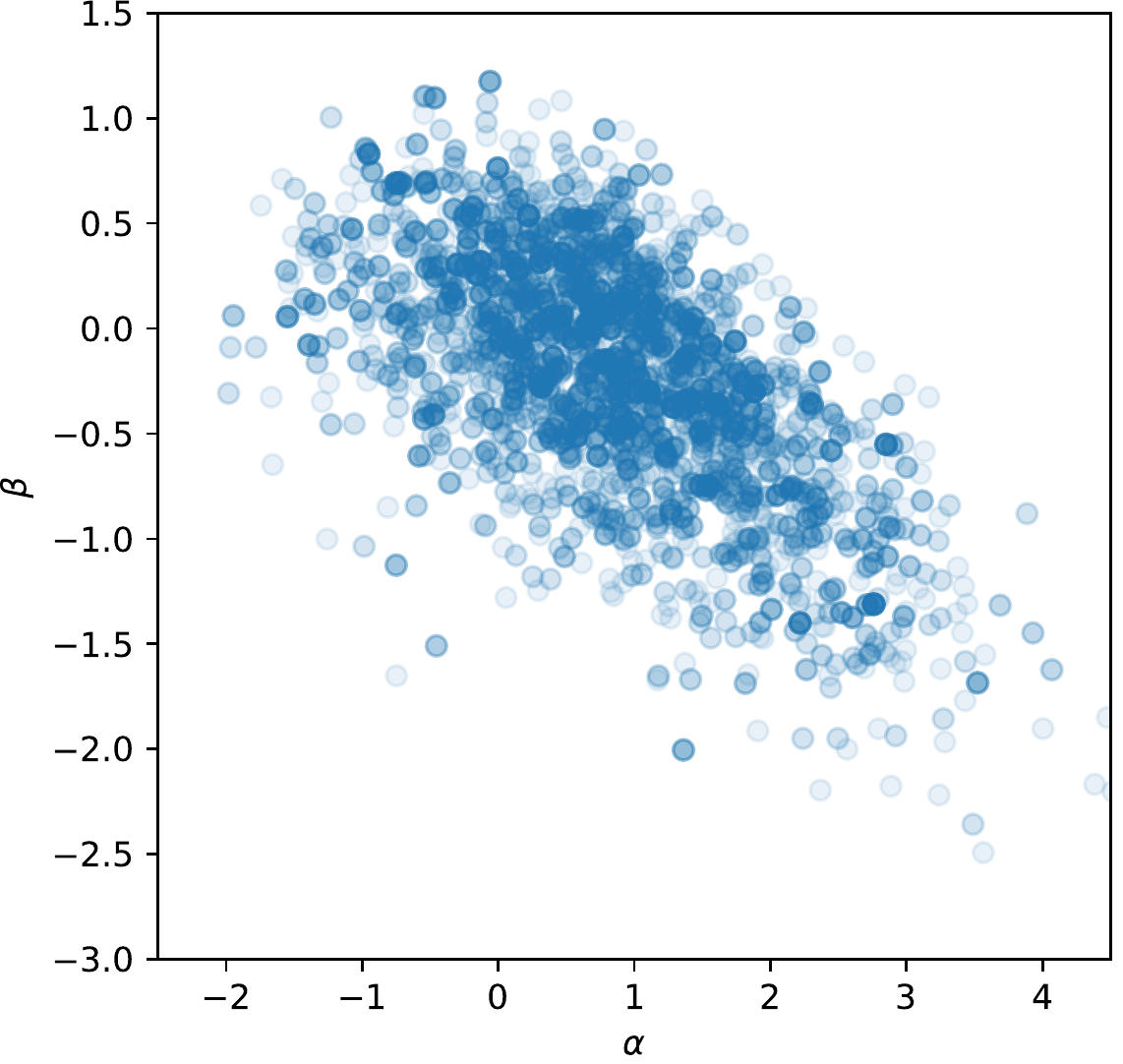}
	\caption{Posterior samples for Poisson regression problem;
          $\thetab_{\mathrm{init}} = (0,0)$.}
	\label{fig:easy}
\end{figure}
      
    \end{solution}

\end{exenumerate}

\ex{Mixing and convergence of Metropolis-Hasting MCMC}\label{q:mh_mixing} Under weak conditions, an MCMC
algorithm is an asymptotically exact inference algorithm,
meaning that if it is run forever, it will generate samples that
correspond to the desired probability distribution. In this case, the
chain is said to converge.

In practice, we want to run the algorithm long enough to be able to
approximate the posterior adequately. How long is long enough for the
chain to converge varies drastically depending on the algorithm, the
hyperparameters (e.g.\ the variance \lstinline{vari}), and the target
posterior distribution. It is impossible to determine exactly whether
the chain has run long enough, but there exist various diagnostics
that can help us determine if we can ``trust'' the sample-based
approximation to the posterior.

A very quick and common way of assessing convergence of the Markov
chain is to visually inspect the \emph{trace plots} for each
parameter. A trace plot shows how the drawn samples evolve through
time, i.e.\ they are a time-series of the samples generated by the
Markov chain. Figure~\ref{fig:traceplots} shows examples of trace
plots obtained by running the Metropolis Hastings algorithm for
different values of the hyperparameters \lstinline{vari} and
\lstinline{param_init}. Ideally, the time series covers the whole
domain of the target distribution and it is hard to ``see'' any
structure in it so that predicting values of future samples from the
current one is difficult. If so, the samples are likely independent from
each other and the chain is said to be well ``mixed''.

\begin{exenumerate}
  \item \label{qpt:mh_trace}  Consider the trace plots in Figure~\ref{fig:traceplots}: Is
    the variance \lstinline{vari} used in Figure
    \ref{sfig:traceplot_many_small} larger or smaller than the value of
    \lstinline{vari} used in Figure \ref{sfig:traceplot_many_good}? Is
    \lstinline{vari} used in Figure \ref{sfig:traceplot_many_big}
    larger or smaller than the value used in Figure
    \ref{sfig:traceplot_many_good}?

    In both cases, explain the behaviour of the trace plots in terms of
    the workings of the Metropolis Hastings algorithm and the effect of
    the variance \lstinline{vari}.

    \begin{figure}[p]
      \setcounter{subfigure}{0}
	\centering
	\subfloat[variance \lstinline{vari}: $1$]{
	  \includegraphics[width=\textwidth]{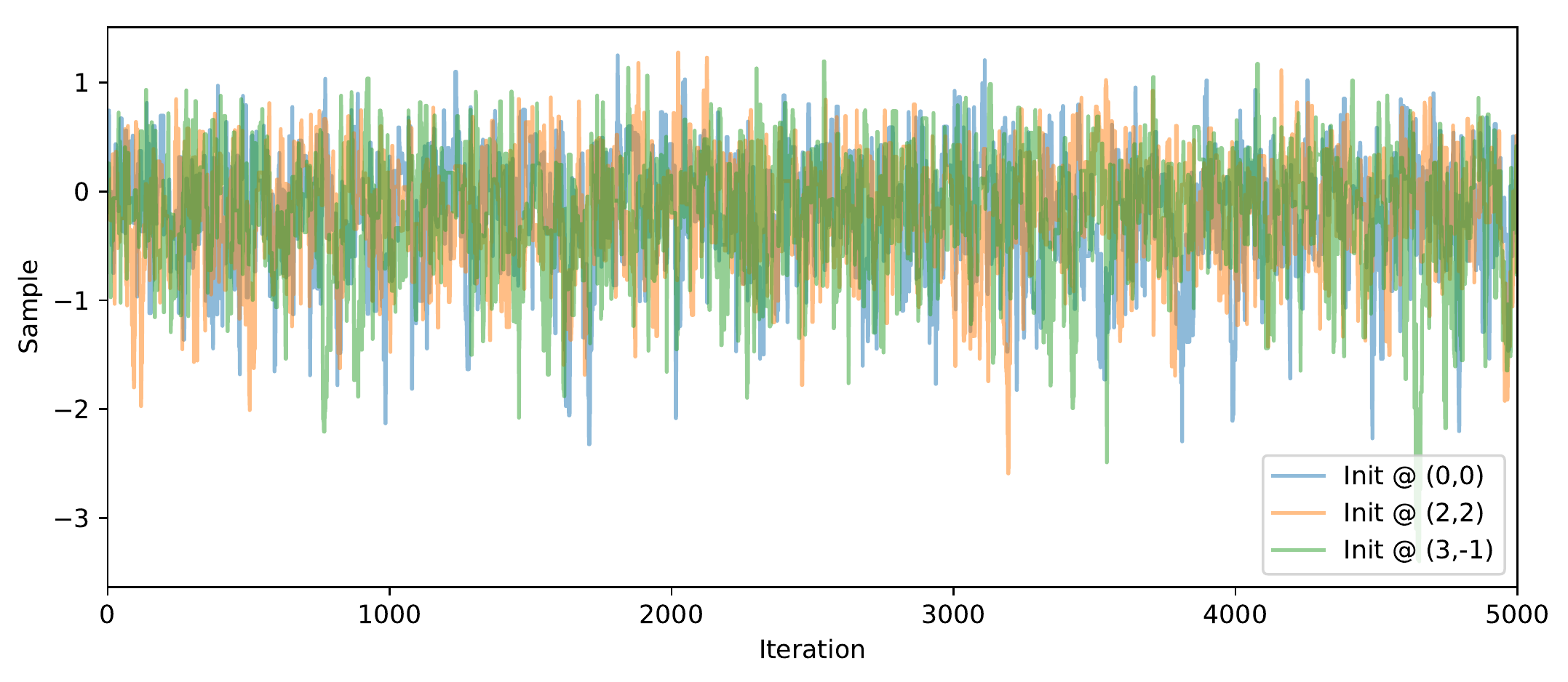}
	  \label{sfig:traceplot_many_good}}\\
	\subfloat[Alternative value of \lstinline{vari}]{
	  \includegraphics[width=\textwidth]{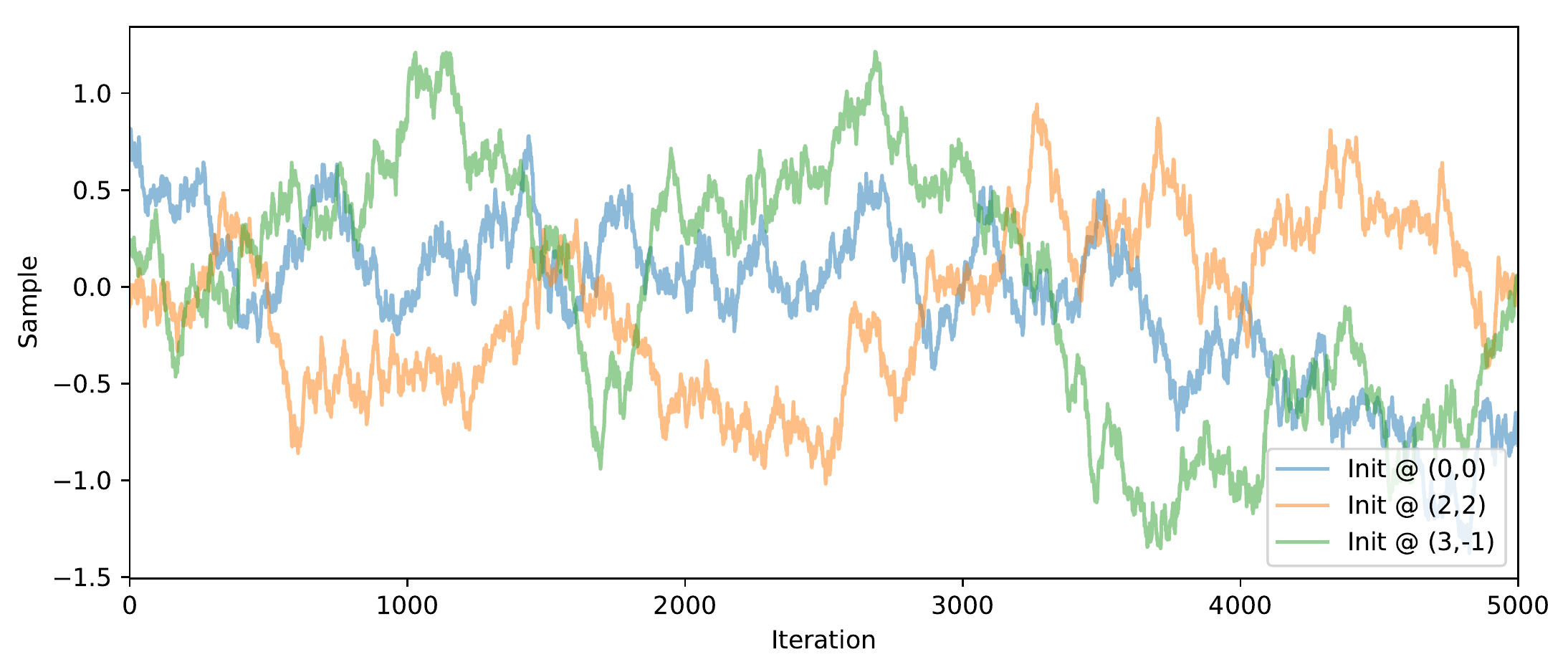}
	  \label{sfig:traceplot_many_small}}\\
        \subfloat[Alternative value of \lstinline{vari}]{
	  \includegraphics[width=\textwidth]{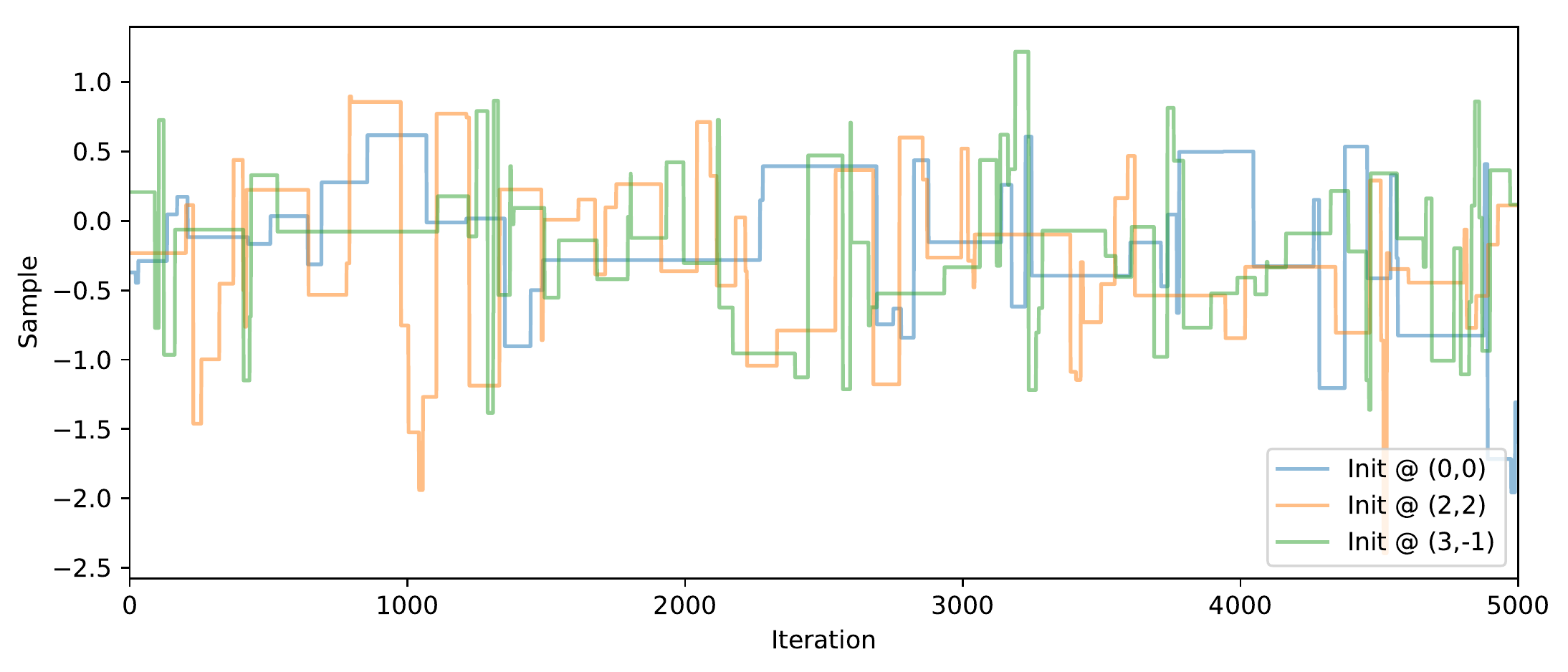}
	  \label{sfig:traceplot_many_big}}
	\caption{For Question \ref{q:mh_mixing}\ref{qpt:mh_trace}: Trace plots of the parameter $\beta$ from Question \ref{q:poisson-reg} drawn using Metropolis-Hastings with different variances of the proposal distribution.}
	\label{fig:traceplots}
\end{figure}

\begin{solution}

  MCMC methods are sensitive to different hyperparameters, and we
  usually need to carefully diagnose the inference results to ensure
  that our algorithm adequately approximates the target posterior
  distribution.

  \begin{enumerate}[label=(\roman*)]
	
  \item Figure~\ref{sfig:traceplot_many_small} uses a \textit{small}
    variance (\lstinline{vari} was set to $0.001$) . The trace plots show
    that the samples for $\beta$ are very highly correlated and evolve
    very slowly through time. This is because the introduced
    randomness is quite small compared to the scale of the
    posterior, thus the proposed sample at each MCMC iteration will be
    very close to the current sample and hence likely accepted.

    More mathematical explanation: for a symmetric proposal
    distribution, the acceptance ratio $a$ becomes
    \begin{equation}
      a =  \frac{p^*(\thetab^*)}{p^*(\thetab)},
    \end{equation}
    where $\thetab$ is the current sample and $\thetab^*$ is the
    proposed sample. For variances that are small compared to the
    (squared) scale of the posterior, $a$ is close to one and the proposed
    sample $\thetab^*$ gets likely accepted. This then gives rise to the
    slowly changing time series shown in
    Figure~\ref{sfig:traceplot_many_small}.
	
  \item In Figure~\ref{sfig:traceplot_many_big}, the variance is
    larger than the reference (\lstinline{vari} was set to $50$) . The trace
    plots suggest that many iterations of the algorithm result in the
    proposed sample being rejected, and thus we end up copying the
    same sample over and over again. This is because if the random
    perturbations are large compared to the scale of the
    posterior, $p^*(\thetab^*)$ may be very different from
    $p^*(\thetab)$ and $a$ may be very small.

  \end{enumerate}

  \end{solution}

\item In Metropolis-Hastings, and MCMC in general, any sample depends
  on the previously generated sample, and hence the algorithm
  generates samples that are generally statistically dependent. The
  \textit{effective sample size} of a sequence of dependent samples is
  the number of independent samples that are, in some sense,
  equivalent to our number of dependent samples. A definition of the
  effective sample size (ESS) is
  \begin{equation}
    \text{ESS} = \frac{S}{1+ 2\sum_{k=1}^{\infty} \rho(k)}
  \end{equation}
  where $S$ is the number of dependent samples drawn and $\rho(k)$ the
  correlation coefficient between two samples in the Markov chain that
  are $k$ time points apart. We can see that if the samples are
  strongly correlated, $\sum_{k=1}^{\infty} \rho(k)$ is large and the
  effective sample size is small. On the other hand, if $\rho(k)=0$
  for all $k$, the effective sample size is $S$.

$\text{ESS}$, as defined above, is the number of independent
samples which are needed to obtain a sample average that has the same
variance as the sample average computed from correlated samples.

To illustrate how correlation between samples is related to a
reduction of sample size, consider two pairs of samples $(\theta_1,
\theta_2)$ and $(\omega_1, \omega_2)$. All variables have variance
$\sigma^2$ and the same mean $\mu$, but $\omega_1$ and $\omega_1$ are
uncorrelated while the covariance matrix for $\theta_1, \theta_2$ is
$\C$,
\begin{equation}
  \C = \sigma^2 \begin{pmatrix}
    1 & \rho \\
    \rho & 1
  \end{pmatrix},
\end{equation}
with $\rho>0$. The variance of the average $\bar{\omega} = 0.5
(\omega_1+\omega_2)$ is
\begin{equation}
  \var \left( \bar{\omega} \right) = \frac{\sigma^2}{2},
\end{equation}
where the $2$ in the denominator is the sample size.

Derive an equation for the variance of $\bar{\theta} =
0.5(\theta_1+\theta_2)$ and compute the reduction $\alpha$ of the
sample size when working with the correlated $(\theta_1,\theta_2)$. In
other words, derive an equation of $\alpha$ in
\begin{equation}
   \var \left(\bar{\theta}\right) = \frac{\sigma^2}{2/\alpha}.
\end{equation}
What is the effective sample size $2/\alpha$ as $\rho \to 1$?

\begin{solution}
  
  Note that $\E(\bar{\theta}) = \mu$. From the definition of
  variance, we then have
  \begin{align}
    \var(\bar{\theta}) & = \E\left( (\bar{\theta} - \mu)^2\right)\\
    & = \E\left( \left(\frac{1}{2}(\theta_1+\theta_2) - \mu\right)^2\right) \\
    & =  \E\left( \left(\frac{1}{2}(\theta_1-\mu+\theta_2-\mu\right)^2\right) \\
    & =  \frac{1}{4} \E\left( (\theta_1-\mu)^2+(\theta_2-\mu)^2 + 2 (\theta_1-\mu) (\theta_2-\mu)\right) \\
    & =  \frac{1}{4} ( \sigma^2 + \sigma^2 + 2 \sigma^2 \rho) \\
    & =  \frac{1}{4} ( 2\sigma^2 + 2 \sigma^2 \rho) \\
    & =  \frac{\sigma^2}{2} ( 1 + \rho) \\
    & =  \frac{\sigma^2}{2/(1+\rho)}
  \end{align}
  Hence: $\alpha = (1+\rho)$, and for $\rho \to 1$, $2/\alpha \to 1$.
  
  Because of the strong correlation, we effectively only have one sample and not two if $\rho \to 1$.
  
\end{solution}

\end{exenumerate}

\chapter{Variational Inference}
\minitoc

\ex{Mean field variational inference I}
\label{ex:MFVI-I}
Let $\ELBOx(q)$ be the evidence lower bound for the marginal $p(\x)$ of a joint pdf/pmf $p(\x,\y)$,
\begin{equation}
  \ELBOx(q) = \E_{q(\y|\x)} \left[\log \frac{p(\x,\y)}{q(\y|\x)}\right].
  \end{equation}
Mean field variational inference assumes that the variational distribution $q(\y|\x)$ fully factorises, i.e.
\begin{equation}
  q(\y | \x) = \prod_{i=1}^d q_i(y_i | \x),
\end{equation}
when $\y$ is $d$-dimensional. An approach to learning the $q_i$ for
each dimension is to update one at a time while keeping the others fixed. We
here derive the corresponding update equations.

\begin{exenumerate}
\item Show that the evidence lower bound $\ELBOx(q)$ can be written as
\begin{equation}
  \ELBOx(q) = \E_{q_1(y_1|\x)} \E_{q(\y_{\setminus 1}|\x)}\left[ \log p(\x,\y)\right] - \sum_{i=1}^d \E_{q_i(y_i|\x)} \left[\log q_i(y_i|\x)\right]
\end{equation}
where $q(\y_{\setminus 1}|\x) = \prod_{i=2}^d q_i(y_i | \x)$ is the variational distribution without $q_1(y_1|\x)$.

\begin{solution}
  This follows directly from the definition of the ELBO and the assumed factorisation of $q(\y|\x)$. We have
  \begin{align}
    \ELBOx(q) &= \E_{q(\y|\x)} \log p(\x,\y) - \E_{q(\y|\x)} \log q(\y|\x) \\
    & = \E_{ \prod_{i=1}^d q_i(y_i | \x)}  \log p(\x,\y) - \E_{ \prod_{i=1}^d q_i(y_i | \x)} \sum_{i=1}^d \log q_i(y_i|\x) \\
    & = \E_{ \prod_{i=1}^d q_i(y_i | \x)}  \log p(\x,\y) - \sum_{i=1}^d  \E_{q_i(y_i | \x)} \log q_i(y_i|\x) \\
    & = \E_{q_1(y_1|\x)} \E_{ \prod_{i=2}^d q_i(y_i | \x)}  \log p(\x,\y) - \sum_{i=1}^d  \E_{q_i(y_i | \x)} \log q_i(y_i|\x)\\
    & = \E_{q_1(y_1|\x)} \E_{q(\y_{\setminus 1}|\x)}\left[ \log p(\x,\y)\right] - \sum_{i=1}^d \E_{q_i(y_i|\x)} \left[\log q_i(y_i|\x)\right]
  \end{align}
  We have here used the linearity of expectation. In case of continuous random variables, for instance, we have
  \begin{align}
 \E_{ \prod_{i=1}^d q_i(y_i | \x)} \sum_{i=1}^d \log q_i(y_i|\x) & = \int  q_1(y_1 | \x)\cdot \ldots\cdot q_d(y_d|\x)  \sum_{i=1}^d \log q_i(y_i|\x) d y_1\ldots d y_d\\
 & =  \sum_{i=1}^d \int q_1(y_1 | \x)\cdot \ldots \cdot q_d(y_d|\x) \log q_i(y_i|\x) d y_1\ldots d y_d\\
 & =  \sum_{i=1}^d \int q_i(y_i | \x)\log q_i(y_i|\x) d y_i \underbrace{\int \prod_{j\neq i} q_j(y_j|\x) d y_j}_{=1}\\
 & =  \sum_{i=1}^d E_{q_i(y_i|\x)} \log q_i(y_i|\x)
\end{align}
For discrete random variables, the integral is replaced with a sum and leads to the same result.

\end{solution}

\item Assume that we would like to update $q_1(y_1 |\x)$ and that the
  variational marginals of the other dimensions are kept fixed. Show that
  \begin{equation}
    \argmax_{q_1(y_1|\x)} \ELBOx(q) = \argmin_{q_1(y_1|\x)} \KL(q_1(y_1|\x) || \bar{p}(y_1|\x))
  \end{equation}
  with
\begin{equation}
  \log \bar{p}(y_1|\x) = \E_{q(\y_{\setminus 1}|\x)}\left[ \log p(\x,\y)\right] + \text{const},
\end{equation}
where $\text{const}$ refers to terms not depending on $y_1$. That is,
\begin{equation}
\bar{p}(y_1|\x) = \frac{1}{Z} \exp\left[ \E_{q(\y_{\setminus
      1}|\x)}\left[ \log p(\x,\y)\right] \right],
\end{equation}
where $Z$ is the normalising constant. Note that variables $y_2, \ldots, y_d$
are marginalised out due to the expectation with respect to $q(\y_{\setminus
  1}|\x)$.

  \begin{solution}
    Starting from 
    \begin{equation}
      \ELBOx(q) = \E_{q_1(y_1|\x)} \E_{q(\y_{\setminus 1}|\x)}\left[ \log p(\x,\y)\right] - \sum_{i=1}^d \E_{q_i(y_i|\x)} \left[\log q_i(y_i|\x)\right]
    \end{equation}
    we drop terms that do not depend on $q_1$. We then obtain
    \begin{align}
      J(q_1) & =  \E_{q_1(y_1|\x)} \E_{q(\y_{\setminus 1}|\x)}\left[ \log p(\x,\y)\right] -  \E_{q_1(y_1|\x)} \left[\log q_1(y_1|\x)\right]\\
      & = \E_{q_1(y_1|\x)} \log \bar{p}(y_1|\x) - \E_{q_1(y_1|\x)} \left[\log q_1(y_1|\x)\right] + \text{const}\\
      & = \E_{q_1(y_1|\x)}\left[ \log \frac{\bar{p}(y_1|\x)}{ q_1(y_1|\x)} \right]\\
      & = -\KL( q_1(y_1|\x) || \bar{p}(y_1|\x) )
    \end{align}
    Hence
    \begin{equation}
      \argmax_{q_1(y_1|\x)} \ELBOx(q) = \argmin_{q_1(y_1|\x)} \KL(q_1(y_1|\x) || \bar{p}(y_1|\x))
    \end{equation}
    
  \end{solution}

\item Conclude that given $q_i(y_i|\x)$, $i=2, \ldots, d$, the optimal $q_1(y_1|\x)$ equals $\bar{p}(y_1|\x)$.

  This then leads to an iterative updating scheme where we cycle
  through the different dimensions, each time updating the corresponding marginal variational distribution according to:
  \begin{align}
    q_i(y_i|\x) &=  \bar{p}(y_i|\x), &  \bar{p}(y_i|\x) &= \frac{1}{Z} \exp\left[ \E_{q(\y_{\setminus i}|\x)}\left[ \log p(\x,\y)\right] \right]
  \end{align}
  where $q(\y_{\setminus i}|\x) = \prod_{j \neq i} q(y_j | \x)$ is the product of all marginals without marginal $q_i(y_i|\x)$.

  \begin{solution}
    This follows immediately from the fact that the KL divergence is
    minimised when $q_1(y_1|\x) = \bar{p}(y_1|\x)$. Side-note: The
    iterative update rule can be considered to be coordinate ascent
    optimisation in function space, where each ``coordinate''
    corresponds to a $q_i(y_i|\x)$.
    
  \end{solution}
\end{exenumerate}


\ex{Mean field variational inference II}
Assume random variables $y_1, y_2, x$ are generated according to the following process
\begin{align}
  y_1 &\sim \Gauss(y_1; 0, 1) &   y_2 &\sim \Gauss(y_2; 0, 1) \\
  n &\sim \Gauss(n; 0, 1)    &    x & = y_1+y_2+n & 
\end{align}
where $y_1, y_2, n$ are statistically independent.

\begin{exenumerate}

\item $y_1, y_2, x$ are jointly Gaussian. Determine their mean and their
  covariance matrix.

  \begin{solution}
    
    The expected value of $y_1$ and $y_2$ is zero. By linearity of
    expectation, the expected value of $x$ is
    \begin{equation}
      \E(x) = \E(y_1) + \E(y_2) + \E(n) = 0
    \end{equation}
    The variance of $y_1$ and $y_2$ is 1. Since $y_1, y_2, n$ are statistically independent,
    \begin{equation}
      \Var(x) = \Var(y_1) + \Var(y_2) + \Var(n) = 1 + 1 + 1 = 3.
    \end{equation}
    The covariance between $y_1$ and $x$ is
    \begin{align}
      \text{cov}(y_1, x) &= \E( (y_1-\E(y_1))(x-\E(x))) = \E( y_1 x) \\
      &= \E( y_1(y_1+y_2+n) ) = \E(y_1^2) + \E(y_1 y_2) + \E(y_1 n)\\
      &= 1 +  \E(y_1)\E(y_2) + \E(y_1)\E(n)\\
      &= 1 + 0 + 0
    \end{align}
    where we have used that $y_1$ and $x$ have zero mean and the independence assumptions.

    The covariance between $y_2$ and $x$ is computed in the same way and equals 1 too.

    We thus obtain the covariance matrix $\Sigmab$,
    \begin{equation}
      \Sigmab = \begin{pmatrix}
        1 & 0 & 1\\
        0 & 1 & 1\\
        1 & 1 & 3
      \end{pmatrix}
    \end{equation}
    
  \end{solution}

\item The conditional $p(y_1, y_2 | x)$ is Gaussian with mean $\m$ and covariance $\C$,
  \begin{align}
    \m &= \frac{x}{3} \begin{pmatrix}
      1\\
      1
    \end{pmatrix}
    &
    \C & = \frac{1}{3} \begin{pmatrix}
      2 &-1 \\
     -1 & 2
    \end{pmatrix}
  \end{align}
  Since $x$ is the sum of three random variables that have the
  same distribution, it makes intuitive sense that the mean assigns
  $1/3$ of the observed value of $x$ to $y_1$ and $y_2$. Moreover,
  $y_1$ and $y_2$ are negatively corrected since an increase in $y_1$
  must be compensated with a decrease in $y_2$.

  Let us now approximate the posterior $p(y_1, y_2 | x)$ with mean
  field variational inference. Determine the optimal variational
  distribution using the method and results from \exref{ex:MFVI-I}. You may use that
  \begin{align}
    p(y_1, y_2, x) &= \Gauss\left( (y_1, y_2, x); \zerob, \Sigmab \right) &  \Sigmab &= \begin{pmatrix} 
      1 & 0 & 1\\
      0 & 1 & 1\\
      0 & 1 & 3
    \end{pmatrix}
    &
    \Sigmab^{-1} &= \begin{pmatrix} 
      2 & 1 & -1\\
      1 & 2 & -1\\
      -1 & -1 & 1
    \end{pmatrix}
  \end{align}
  
  \begin{solution}
    The mean field assumption means that the variational distribution is assumed to factorise as
    \begin{equation}
      q(y_1, y_2 | x) = q_1(y_1 | x) q_2(y_2 | x)
    \end{equation}
    From \exref{ex:MFVI-I}, the optimal $q_1(y_1|x)$ and $q_2(y_2|x)$ satisfy
     \begin{align}
       q_1(y_1|x) &=  \bar{p}(y_1|x), &  \bar{p}(y_1|x) &= \frac{1}{Z} \exp\left[ \E_{q_2(y_2|x)}\left[ \log p(y_1, y_2, x)\right] \right]\\
       q_2(y_2|x) &=  \bar{p}(y_2|x), &  \bar{p}(y_2|x) &= \frac{1}{Z} \exp\left[ \E_{q_1(y_1|x)}\left[ \log p(y_1, y_2, x)\right] \right]
     \end{align}
    Note that these are coupled equations: $q_2$ features in the
    equation for $q_1$ via $\bar{p}(y_1|x)$, and $q_1$ features in
    the equation for $q_2$ via $\bar{p}(y_2|x)$. But we have two
    equations for two unknowns, which for the Gaussian joint model
    $p(x, y_1, y_2)$ can be solved in closed form.

    Given the provided equation for $p(y_1, y_2, x)$, we have that
    \begin{align}
      \log p(y_1, y_2, x) & = -\frac{1}{2} \begin{pmatrix}
        y_1\\
        y_2\\
        x
      \end{pmatrix}^\top
      \begin{pmatrix}
        2 & 1 & -1\\
        1 & 2 & -1\\
        -1 & -1 & 1
      \end{pmatrix}
      \begin{pmatrix}
        y_1\\
        y_2\\
        x
      \end{pmatrix} + \text{const}\\
        & = -\frac{1}{2} \left( 2 y_1^2 + 2 y_2^2 + x^2 + 2 y_1 y_2 - 2 y_1 x - 2y_2 x \right) + \text{const}
    \end{align}
    
    Let us start with the equation for $\bar{p}(y_1|x)$. It is easier to work in the logarithmic domain, where we obtain:
    \begin{align}
      \log  \bar{p}(y_1|x) & =  \E_{q_2(y_2|x)}\left[ \log p(y_1, y_2, x)\right] + \text{const}\\
      & = -\frac{1}{2} \E_{q_2(y_2|x)}\left[  2 y_1^2 + 2 y_2^2 + x^2 + 2 y_1 y_2 - 2 y_1 x - 2y_2 x \right] + \text{const}\\
      & = -\frac{1}{2} \left(2 y_1^2 + 2 y_1 \E_{q_2(y_2|x)}[y_2] - 2 y_1 x \right) + \text{const}\\
      & = -\frac{1}{2} \left( 2 y_1^2 +2y_1 m_2 - 2 y_1 x \right) + \text{const}\\
      & = -\frac{1}{2} \left( 2 y_1^2 - 2 y_1 (x-m_2) \right) + \text{const}
      \label{eq:barp1}
    \end{align}
    where we have absorbed all terms not involving $y_1$ into the constant. Moreover, we set $\E_{q_2(y_2|x)}[y_2]=m_2$.

    Note that an arbitrary Gaussian density $\Gauss(y; m, \sigma^2)$
    with mean $m$ and variance $\sigma^2$ can be written in the
    log-domain as
    \begin{align}
      \log \Gauss(y; m, \sigma^2) & = -\frac{1}{2}\frac{(y-m)^2}{\sigma^2} + \text{const}\\
      & = -\frac{1}{2} \left( \frac{y^2}{\sigma^2} -2y \frac{m}{\sigma^2} \right) + \text{const}
      \label{eq:loggauss}
    \end{align}
    Comparison with \eqref{eq:barp1} shows that $\bar{p}(y_1|x)$, and
    hence $q_1(y_1 | x)$, is Gaussian with variance and mean equal to
    \begin{align}
      \sigma_1^2 &= \frac{1}{2} & m_1 & = \frac{1}{2}(x-m_2)
    \end{align}
    Note that we have not made a Gaussianity assumption on $q_1(y_1 |
    x)$. The optimal $q_1(y_1 | x)$ turns out to be Gaussian because
    the model $p(y_1, y_2, x)$ is Gaussian.
    
    The equation for $\bar{p}(y_2|x)$ gives similarly
    \begin{align}
      \log  \bar{p}(y_2|x) & =  \E_{q_1(y_1|x)}\left[ \log p(y_1, y_2, x)\right] + \text{const}\\
      & = -\frac{1}{2} \E_{q_1(y_1|x)}\left[  2 y_1^2 + 2 y_2^2 + x^2 + 2 y_1 y_2 - 2 y_1 x - 2y_2 x \right] + \text{const}\\
      & = -\frac{1}{2} \left(2 y_2^2 + 2 \E_{q_1(y_1|x)}[y_1] y_2 - 2y_2 x \right) + \text{const}\\
      & = -\frac{1}{2} \left(2 y_2^2 + 2 m_1 y_2 - 2 y_2 x \right) + \text{const}\\
      & = -\frac{1}{2} \left(2 y_2^2 - 2 y_2(x-m_1) \right) + \text{const}
      \label{eq:barp2}
    \end{align}
    where we have absorbed all terms not involving $y_2$ into the
    constant. Moreover, we set $\E_{q_1(y_1|x)}[y_1]=m_1$. With
    \eqref{eq:loggauss}, this is defines a Gaussian distribution with 
    variance and mean equal to
    \begin{align}
      \sigma_2^2 &= \frac{1}{2} & m_2 & = \frac{1}{2}(x-m_1)
    \end{align}
    Hence the optimal marginal variational distributions $q_1(y_1|x)$ and
    $q_2(y_2|x)$ are both Gaussian with variance equal to $1/2$. Their
    means satisfy
    \begin{align}
      m_1 & = \frac{1}{2}(x-m_2) & m_2  & = \frac{1}{2}(x-m_1)
    \end{align}
    These are two equations for two unknowns. We can solve them as follows   
    \begin{align}
      2 m_1 &= x - m_2 \\
      & = x -\frac{1}{2}(x-m_1)\\
      4 m_1 & = 2 x -x + m_1 \\
      3 m_1 & = x\\
      m_1 & = \frac{1}{3} x
    \end{align}
    Hence
    \begin{equation}
     m_2 = \frac{1}{2} x - \frac{1}{6} x = \frac{2}{6} x = \frac{1}{3} x
    \end{equation}
    In summary, we find
    \begin{align}
      q_1(y_1 | x) & = \Gauss\left(y_1; \frac{x}{3}, \frac{1}{2}\right) &  q_2(y_2 | x) & = \Gauss\left(y_2; \frac{x}{3}, \frac{1}{2}\right)
    \end{align}
    and the optimal variational distribution $q(y_1, y_2|x) =
    q_1(y_1|x) q_2(y_2|x)$ is Gaussian. We have made the mean field
    (independence) assumption but not the Gaussianity
    assumption. Gaussianity of the variational distribution is a
    consequence of the Gaussianity of the model $p(y_1, y_2, x)$.

    Comparison with the true posterior shows that the mean field
    variational distribution $q(y_1, y_2 | x)$ has the same mean but
    ignores the correlation and underestimates the marginal
    variances. The true posterior and the mean field approximation are
    shown in Figure \ref{fig:meanfield}.

    \begin{figure}[h!]
      \centering
      \includegraphics[width=0.75\textwidth]{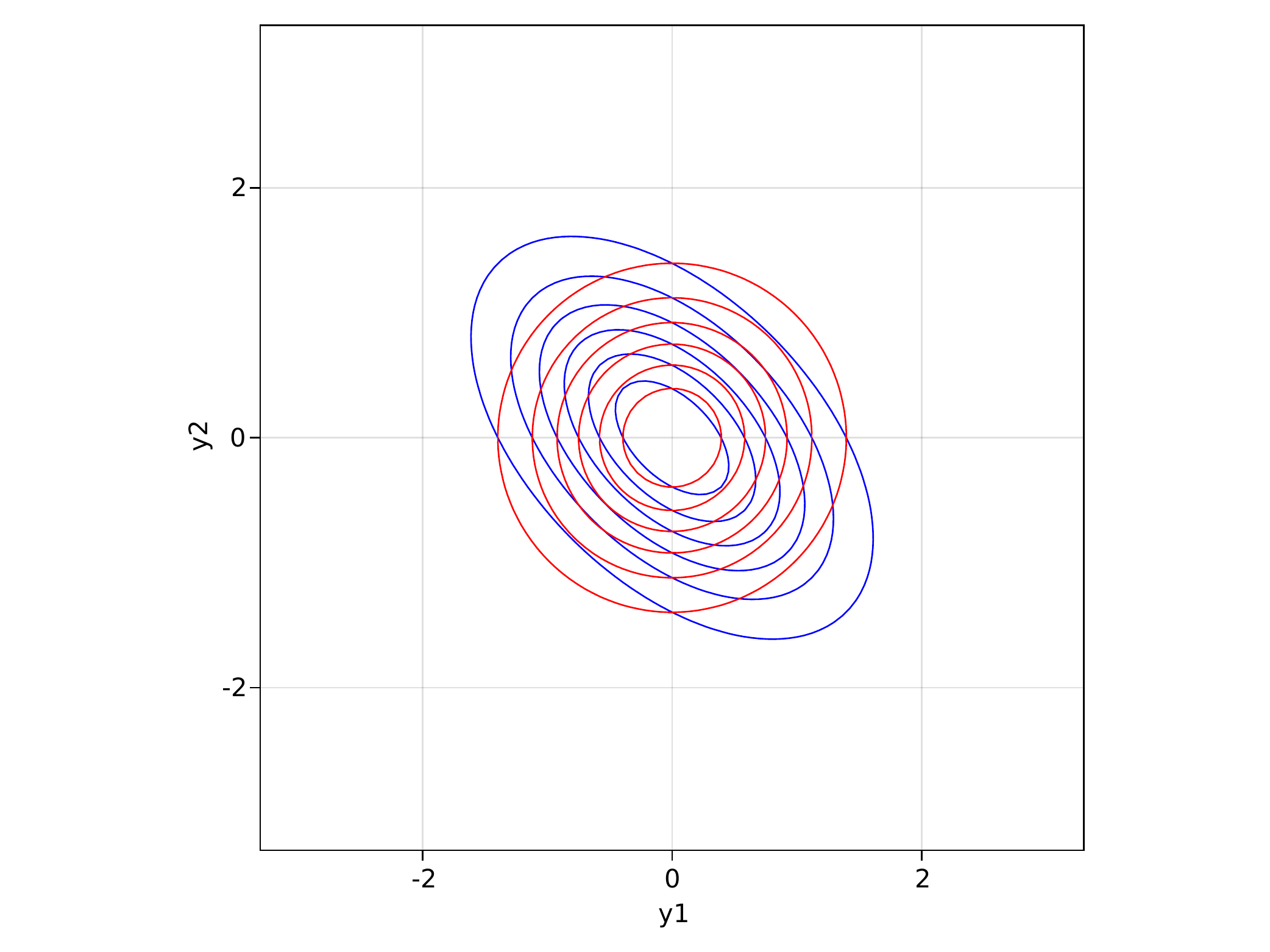}
      \caption{\label{fig:meanfield}In blue: correlated true posterior. In red: mean field approximation.}
    \end{figure}
  \end{solution}

\end{exenumerate}
\ex{Variational posterior approximation I}

We have seen that maximising the evidence lower bound (ELBO) with
respect to the variational distribution $q$ minimises the
Kullback-Leibler divergence to the true posterior $p$. We here assume
that $q$ and $p$ are probability density functions so that the
Kullback-Leibler divergence between them is defined as
\begin{equation}
  \KL(q || p) = \int q(\x) \log \frac{q(\x)}{p(\x)} \ud \x = \E_q \left[\log \frac{q(\x)}{p(\x)}\right].
\end{equation}

\begin{exenumerate}

\item You can here assume that $\x$ is one-dimensional so that $p$
  and $q$ are univariate densities. Consider the case where $p$ is
  a bimodal density but the variational densities $q$ are
  unimodal. Sketch a figure that shows $p$ and a variational
  distribution $q$ that has been learned by minimising $\text{KL}(q
  || p)$. Explain qualitatively why the sketched $q$ minimises
  $\text{KL}(q || p)$.

  \begin{solution}

  A possible sketch is shown in the figure below.

  \begin{figure}[h!]
    \centering
    \includegraphics[width = 0.5 \textwidth]{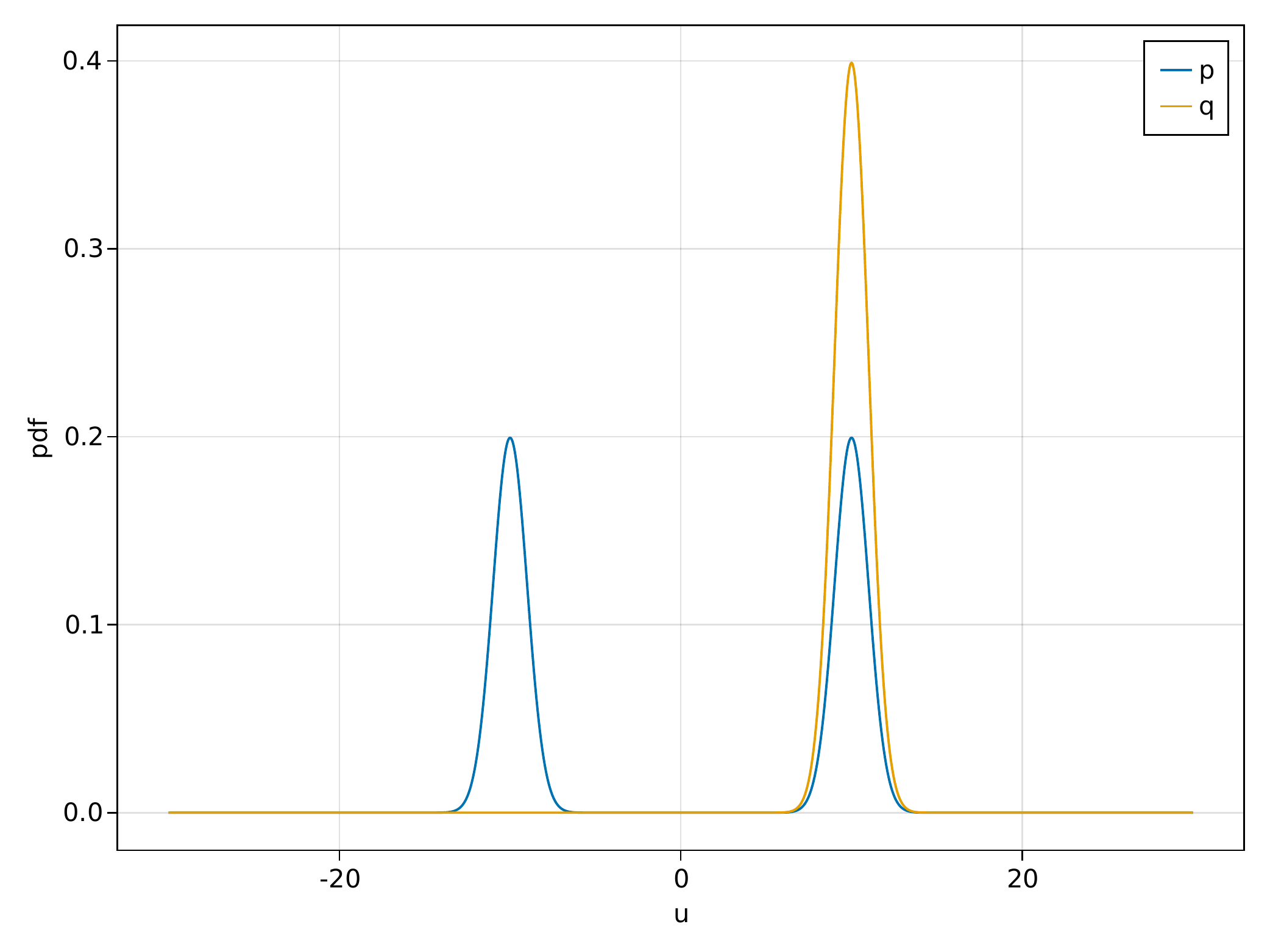}
  \end{figure}

  Explanation: We can divide the domain of $p$ and $q$ into the areas
  where $p$ is small (zero) and those where $p$ has significant
  mass. Since the objective features $q$ in the numerator while $p$ is
  in the denominator, an optimal $q$ needs to be zero where $p$ is
  zero. Otherwise, it would incur a large penalty (division by
  zero). Since we take the expectation with respect to $q$, however,
  regions where $p>0$ do not need to be covered by $q$; cutting them
  out does not incur a penalty. Hence, optimal unimodal $q$ only cover
  one peak of the bimodal $p$.
   
  \end{solution}

\item Assume that the true posterior $p(\x) = p(x_1, x_2)$ factorises into two Gaussians of mean zero and variances $\sigma_1^2$ and $\sigma_2^2$,
  \begin{equation}
    p(x_1,x_2) = \frac{1}{\sqrt{2\pi \sigma_1^2}} \exp\left[-\frac{x_1^2}{2 \sigma_1^2}\right]\frac{1}{\sqrt{2\pi \sigma_2^2}} \exp\left[-\frac{x_2^2}{2 \sigma_2^2}\right].
  \end{equation}
  Assume further that the variational density $q(x_1,x_2; \lambda^2)$ is parametrised as
  \begin{equation}
    q(x_1,x_2;\lambda^2) = \frac{1}{2\pi \lambda^2} \exp\left[-\frac{x_1^2+x_2^2}{2 \lambda^2}\right]
    \end{equation}
  where $\lambda^2$ is the variational parameter that is learned
  by minimising $ \KL(q || p)$. If $\sigma^2_2$ is much
  larger than $\sigma^2_1$, do you expect $\lambda^2$ to be closer to $\sigma_2^2$
  or to $\sigma_1^2$? Provide an explanation.

  \begin{solution}
    The learned variational parameter will be closer to $\sigma_1^2$
    (the smaller of the two $\sigma_i^2$).
      
    Explanation: First note that the $\sigma_i^2$ are the variances
    along the two different axes, and that $\lambda^2$ is the single
    variance for both $x_1$ and $x_2$. The objective penalises $q$ if
    it is non-zero where $p$ is zero (see above). The variational
    parameter $\lambda^2$ thus will get adjusted during learning so
    that the variance of $q$ is close to the smallest of the two
    $\sigma_i^2$.

  \end{solution}

\end{exenumerate}


\ex{Variational posterior approximation II}

We have seen that maximising the evidence lower bound (ELBO) with
respect to the variational distribution minimises the
Kullback-Leibler divergence to the true posterior. We here
investigate the nature of the approximation if the family of
variational distributions does not include the true posterior.

\begin{exenumerate}

\item Assume that the true posterior for $\x = (x_1, x_2)$ is given by
  \begin{equation}
    p(\x) = \normal(x_1 ; \sigma_1^2)\normal(x_2 ; \sigma_2^2) 
  \end{equation}
  and that our variational distribution $q(\x; \lambda^2)$ is
  \begin{equation}
    q(\x; \lambda^2) = \normal(x_1 ; \lambda^2)\normal(x_2 ; \lambda^2),
  \end{equation}
  where $\lambda >0$ is the variational parameter. Provide an
  equation for
  \begin{equation}
    J(\lambda) = \KL(q(\x; \lambda^2) || p(\x)),
  \end{equation}
  where you can omit additive terms that do not depend on
  $\lambda$. 

  \begin{solution}

    We write
    \begin{align}
      \text{KL}(q(\x; \lambda^2) || p(\x)) & = \E_q \left[ \log  \frac{q(\x; \lambda^2)}{p(\x)} \right]\\
      & = \E_q \log q(\x; \lambda^2) - \E_q \log p(\x)\\
      & = \E_q \log  \normal(x_1 ; \lambda^2) + \E_q \log \normal(x_2 ; \lambda^2) \nonumber \\
      & \phantom{=} - \E_q \log \normal(x_1 ; \sigma_1^2) - \E_q \log \normal(x_2 ; \sigma_2^2)
    \end{align}
    
    We further have
    \begin{align}
      \E_q \log  \normal(x_i ; \lambda^2) & = \E_q \log \left[ \frac{1}{\sqrt{2\pi \lambda^2}} \exp \left[-\frac{x_i^2}{2 \lambda^2} \right] \right]\\
      & = \log \left[ \frac{1}{\sqrt{2\pi \lambda^2}}\right] -\E_q \left[\frac{x_i^2}{2 \lambda^2}\right]\\
      & = -\log \lambda - \frac{\lambda^2}{2\lambda^2} + \text{const}\\
      & = -\log \lambda - \frac{1}{2} + \text{const}\\
      & = -\log \lambda + \text{const}
    \end{align}
    where we have used that for zero mean $x_i$, $\E_q [x_i^2] = \var(x_i) = \lambda^2$. 

    We similarly obtain
    \begin{align}
    \E_q \log \normal(x_i ; \sigma_i^2) & = \E_q \log \left[ \frac{1}{\sqrt{2\pi \sigma_i^2}} \exp \left[-\frac{x_i^2}{2 \sigma_i^2} \right] \right]\\
      & = -\log \left[ \frac{1}{\sqrt{2\pi \sigma_i^2}}\right] -\E_q \left[\frac{x_i^2}{2 \sigma_i^2}\right]\\
    & = -\log \sigma_i - \frac{\lambda^2}{2\sigma_i^2} + \text{const}\\
    & = - \frac{\lambda^2}{2\sigma_i^2} + \text{const}
    \end{align}
    
    We thus have
    \begin{align}
      \text{KL}(q(\x; \lambda^2 || p(\x))  & =  -2 \log \lambda + \lambda^2\left(\frac{1}{2\sigma_1^2}+\frac{1}{2\sigma_2^2}\right)+\text{const}
    \end{align}    
    
  \end{solution}

\item Determine the value of $\lambda$ that minimises $J(\lambda) =
  \KL(q(\x; \lambda^2) || p(\x))$. Interpret the result and
  relate it to properties of the Kullback-Leibler
  divergence.

  \begin{solution}

    Taking derivatives of $J(\lambda)$ with respect to $\lambda$ gives
    \begin{align}
      \frac{\partial J(\lambda)}{\partial \lambda} & = -\frac{2}{\lambda} + \lambda \left(\frac{1}{\sigma_1^2}+\frac{1}{\sigma_2^2}\right)
      \end{align}
    Setting it zero yields
    \begin{align}
      \frac{1}{\lambda^2} &= \frac{1}{2}\left(\frac{1}{\sigma_1^2}+\frac{1}{\sigma_2^2}\right)
    \end{align}
    so that
    \begin{align}
      \lambda^2 = 2 \frac{\sigma_1^2 \sigma_2^2}{\sigma_1^2+\sigma_2^2}
    \end{align}
    or
    \begin{equation}
    \lambda = \sqrt{2} \sqrt{\frac{\sigma_1^2 \sigma_2^2}{\sigma_1^2+\sigma_2^2}}
    \end{equation}
    This is a minimum because the second derivative of
    $J(\lambda)$
    \begin{equation}
      \frac{\partial^2 J(\lambda)}{\partial \lambda^2} = \frac{2}{\lambda^2} + \left(\frac{1}{\sigma_1^2}+\frac{1}{\sigma_2^2}\right)
    \end{equation}
    is positive for all $\lambda > 0$.
    
    The result has an intuitive explanation: the optimal variance
    $\lambda^2$ is the harmonic mean of the variances $\sigma_i^2$ of
    the true posterior. In other words, the optimal precision
    $1/\lambda^2$ is given by the average of the precisions
    $1/\sigma_i^2$ of the two dimensions.

    If the variances are not equal, e.g.\ if $\sigma_2^2 >
    \sigma_1^2$, we see that the optimal variance of the variational
    distribution strikes a compromise between two types of penalties
    in the KL-divergence: the penalty of having a bad fit because the
    variational distribution along dimension two is too narrow; and
    along dimension one, the penalty for the variational distribution
    to be nonzero when $p$ is small. 

  \end{solution}
  
\end{exenumerate}

\bibliography{refs}
\addcontentsline{toc}{chapter}{Bibliography}

\end{document}